\documentclass{article}

\usepackage{microtype}
\usepackage{graphicx}
\usepackage{subfigure}
\usepackage{booktabs} 
\usepackage{amsthm}
\usepackage{amssymb}
\usepackage{amsmath}
\usepackage{amsfonts}
\usepackage{placeins}
\usepackage[left=1in,bottom=1in,top=1in,right=1in]{geometry}
\usepackage[colorlinks=true]{hyperref}
\usepackage{thmtools}

\usepackage[utf8]{inputenc} 
\usepackage[T1]{fontenc}    
\usepackage{url}            
\usepackage{nicefrac}       
\usepackage{xcolor}         

\usepackage{enumitem}

\usepackage{url}

\usepackage[disable]{todonotes} 
\presetkeys{todonotes}{}{}

\usepackage{commath} 

\theoremstyle{definition}
\newtheorem{definition}{Definition}[section]

\theoremstyle{remark}
\newtheorem*{remark}{Remark}

\theoremstyle{plain}
\newtheorem{theorem}{Theorem}[section]

\theoremstyle{plain}
\newtheorem{corollary}{Corollary}[theorem]

\theoremstyle{plain}

\theoremstyle{plain}
\newtheorem{lemma}[theorem]{Lemma}
\newtheorem{claim}[theorem]{Claim}
\newtheorem{fact}[theorem]{Fact}

\newcommand{\dout}{d_{\mathrm{out}}}

\newcommand{\relu}{\ensuremath{\mathsf{ReLU}}}

\renewcommand{\tanh}{\ensuremath{\mathsf{tanh}}}

\newcommand{\Reals}{\ensuremath{\mathbb{R}}}

	\newcommand{\bv}{\mathbf{v}}
	\newcommand{\bV}{\mathbf{V}}
	
	\newcommand{\bx}{\mathbf{x}}
	\newcommand{\obx}{\overline{\mathbf{x}}}
	\newcommand{\bw}{\mathbf{w}}
	\newcommand{\bW}{\mathbf{W}}
	\newcommand{\bM}{\mathbf{M}}
	\newcommand{\obW}{\overline{\bW}}
	
	\newcommand{\deriv}{\ensuremath{\,\mathrm{d}}}
	\newcommand{\Ball}{\mathbb{B}}
	\newcommand{\Sphere}{\mathbb{S}}

	\DeclareMathOperator{\tr}{\mathsf{tr}}

	\newtheorem{assumption}{Assumption}

\declaretheoremstyle[%
  spaceabove=0.5\parskip,%
  spacebelow=0.5\parskip,%
  headfont=\normalfont\itshape,%
  postheadspace=1em,%
  qed=\qedsymbol%
]{mystyle} 
\declaretheorem[name={Proof},style=mystyle,unnumbered,
]{prf}

\title{Learning and Generalization in RNNs}

\author{%
	Abhishek Panigrahi\\
	Computer Science, Princeton University\\
	\texttt{ap34@cs.princeton.edu} \\
	\and
	Navin Goyal\\
	Microsoft Research India\\
	\texttt{navingo@microsoft.com}  
}

\begin{document}
\maketitle







\vskip 0.3in




\begin{abstract}
Simple recurrent neural networks (RNNs) and their more advanced cousins LSTMs etc. have been very successful in sequence modeling. Their theoretical understanding, however, is lacking and has not kept pace with the progress for feedforward networks, where a reasonably complete understanding in the special case of highly overparametrized one-hidden-layer networks has emerged. In this paper, we make progress towards remedying this situation by proving that RNNs can learn functions of sequences. In contrast to the previous work that could only deal with functions of sequences that are sums of functions of individual tokens in the sequence, we allow general functions. Conceptually and technically, we introduce new ideas which enable us to extract information from the hidden state of the RNN in our proofs---addressing a crucial weakness in previous work. We illustrate our results on some regular language recognition problems.

\end{abstract}

\section{Introduction}
Simple Recurrent Neural Networks \cite{Elman} also known as Elman RNNs or vanilla RNNs (just RNNs henceforth) along with their more advanced versions such as LSTMs \cite{LSTM} and GRU ~\cite{cho-etal-2014-learning} are among the most successful models for processing sequential data, finding wide-ranging applications including natural language processing, audio processing \cite{jurafsky_martin} and time series classification \cite{lim2020time}. Feedforward networks (FFNs) model functions on inputs of fixed length, such as vectors in $\Reals^d$. In contrast, RNNs model functions whose input consists of sequences of tokens $\bx^{(1)}, \bx^{(2)}, \ldots$, where $\bx^{(i)} \in \Reals^d$ for each $i$. 
 RNNs have a notion of memory; formally it is given by the hidden state vector which is denoted by $\mathbf{h}^{(t)}$ after processing the $t$-th token. RNNs apply a fixed function to $\mathbf{h}^{(t)}$ and $\bx^{(t+1)}$ to compute $\mathbf{h}^{(t+1)}$ and the output. This fixed function is modeled by a neural networks with one hidden-layer. Compared to FFNs, new challenges arise in the analysis of RNNs: for example, the use of memory and the same function at each step introduces dependencies across time and RNN training suffers from vanishing and exploding gradients \cite{pascanu13}. 

Studies aimed at understanding the effectiveness of RNNs have been conducted since their introduction; for some of the early work, see, e.g., \cite{SiegelmannS95, kolen2001field}. These works take the form of experimental probing of the inner workings of these models as well as theoretical studies. 
The theoretical studies are often focused on expressibility, training and generalization questions in isolation rather than all together---the latter needs to be addressed to approach full understanding of RNNs and appears to be far more challenging. While experimental probing has continued apace, e.g., \cite{WeissGY18, BhattamishraAG20}, progress on theoretical front has been slow. It is only recently that training and generalization are starting to be addressed in the wake of progress on the relatively easier case of FFNs as discussed next.

RNNs are closely related to deterministic finite automata \cite{korsky2019computational, WeissGY18} as well as to dynamical systems. With finite precision and ReLU activation, they are equivalent to finite automata \cite{korsky2019computational} in computational power.
In the last few years progress was made on theoretical analysis of overparamterized FFNs with one-hidden-layer, e.g., \cite{JacotNTK, LiLiang2018, du2018gradient, allen2019learning, quanquan, Arora_Generalization, ghorbani2021linearized}. Building upon these techniques,
\cite{allen2019convergence_rnn} proved that RNNs trained with SGD (stochastic gradient descent) achieve small training loss if the number of neurons is sufficiently large polynomial in the number of training datapoints and the maximum sequence length. 

But the gap between our understanding of RNNs and FFNs remains large. \cite{Tao_generalization_RNN, generalization_zhao} provide generalization bounds on RNNs in terms of certain norms of the parameters. While interesting, these bounds shed light on only a part of the picture as they do not consider the training of the networks nor do not preclude the possibility that the norms of the parameters for the trained networks are large leading to poor generalization guarantees.
RNNs can be viewed as dynamical systems and many works have used this viewpoint to study RNNs, e.g., \cite{HardtMR16, MillerH19, pmlr-v99-oymak19a, maheswaranathan2019reverse}. Other related work includes relation to kernel methods, e.g., \cite{Yang19, RNTK, alemohammad2020scalable}, linear RNNs \cite{emami2021implicit}, saturated RNNs~\cite{merrill2019sequential, merrill2020formal, merrill2021formal}, and echo state networks~\cite{grigoryeva2018echo, ozturk2007analysis}. Several other works talk about the expressive power of the novel sequence to sequence models  Transformers~\cite{yun2019transformers, yun2020n}. Due to a large number of works in this area it is not possible to be exhaustive: apart from the references directly relevant to our work we have only been able to include a small subset. 

\cite{allen2019can} gave the first ``end-to-end'' result for RNNs. Very informally, their result is: if the concept class consists of functions that are sums of functions of tokens then overparametrized RNNs trained using SGD with sufficiently small learning rate can learn such a concept class. They introduce new technical ideas, most notably what they call re-randomization which allows one to tackle the dependencies that arise because the same weights are used in RNN across time. 
However, an important shortcoming of their result is limited expressive power of their concept class: 
while this class can be surprisingly useful as noted there, 
it cannot capture problems where the RNN needs to make use of the information in the past tokens when processing a token (in their terminolgy, their concept class can \emph{adapt} to time but not to tokens). Indeed, a key step in their proof shows that RNNs can learn to ignore the hidden state $\mathbf{h}^{(t)}$. (The above concept class comes up because it can be learnt even if $\mathbf{h}^{(t)}$ is ignored.) But the hidden state $\mathbf{h}^{(t)}$ is the hallmark of RNNs and 
is the source of information about the past tokens---in general, not something to be ignored. 
Thus, it is an important question to theoretically analyze RNNs' performance on general concept classes and it was also raised in \cite{allen2019can}. This question is addressed in the present paper. As in previous work, we work with sequences of bounded length $L$. Without loss of generality, we work with token sequences $\bx^{(1)}, \ldots, \bx^{(L)}$ of fixed length as opposed to sequences of length up to $L$.
Informally, our result is:
\vspace{-5pt}
\begin{center}
    \fbox{\begin{minipage}{\columnwidth}
Overparametrized RNNs can efficiently learn concept classes consisting of one-hidden-layer neural networks that take the entire sequence of tokens as input. The training algorithm used is SGD with sufficiently small step size. 
\end{minipage}}
\end{center}
 By the universality theorem for one-hidden-layer networks, 
  such RNNs can approximate all continuous functions of $\bx^{(1)}, \ldots, \bx^{(L)}$---though naturally the more complex the functions in the class the larger the network size required. We note that the above result applies to all three aspects mentioned above: expressive power, training and generalization. 
We illustrate the power of our result by showing that some regular languages such as PARITY can be recognized efficiently by RNNs.

\section{Preliminaries}
\todo{Only keep here notation that's used in the main paper}
Let
$\Sphere^{d-1} := \{\bx \in \Reals^d \mid \norm{\bx}_2 = 1\}$ be the unit sphere in $\Reals^d$.
For positive integer $n$ define $[n]:=\{1, 2, \ldots, n\}$. 
Given a vector $\mathbf{v}$, by $v_i$ we denote its $i$-th component.
Given two vectors $\mathbf{a} \in \mathbb{R}^{d_1}$ and $\mathbf{b} \in \mathbb{R}^{d_2}$, $[\mathbf{a}, \mathbf{b}] \in \mathbb{R}^{d_1 + d_2}$ denotes the concatenation of the two vectors. $\langle \cdot, \cdot \rangle$ denotes the standard dot product. 
Given a matrix $\mathbf{M}$, we will denote its $i$-th row as $\mathbf{m}_i$ and the element in row $i$ and column $j$ as $m_{ij}$. Given two matrices $\mathbf{A} \in \mathbb{R}^{a_1 \times a_2}$ and $\mathbf{B} \in \mathbb{R}^{b_1 \times b_2}$ with $a_1 = b_1$ let $[\mathbf{A}, \mathbf{B}]_r \in \mathbb{R}^{a_1 \times (a_2 + b_2)}$ denote the matrix whose rows are obtained by concatenating the respective rows of $\mathbf{A}$ and $\mathbf{B}$. Similarly, $[\mathbf{A}, \mathbf{B}]_c \in \mathbb{R}^{(a_1 + b_1) \times a_2 }$ (assuming $a_2 = b_2$) denotes the matrix whose columns are obtained by concatenating the columns of $\mathbf{A}$ and $\mathbf{B}$. 

$O(\cdot)$ and $\Omega(\cdot)$ hide absolute constants. Similarly, $\mathrm{poly}(\cdot)$ denotes a polynomial in its arguments with degree and coefficients bounded by absolute constants; different instances of $\mathrm{poly}(\cdot)$ may refer to different polynomials. Writing out explicit constants would lead to unwieldy formulas without any new insights.


Let $\sigma: \Reals \to \Reals$, given by $\sigma(x) := \max\{x, 0\} = x \,\mathbb{I}_{x \ge 0}$, be {\relu} activation function. 
$\relu$ can be extended to act on vectors by coordinate-wise application: $\sigma((x_1, \ldots, x_d)) := (\sigma(x_1), \ldots, \sigma(x_d))$.
Note that 
$\relu$ is a positive homogenous function of degree $1$, that is to say $\sigma(\lambda x) = \lambda \, \sigma(x)$ for all $x$ and all $\lambda \geq 0$.

To be learnable efficiently, the functions in the concept class need to be not too complex. We will quantify this with the following two complexity measures which are weighted norms of the Taylor expansion and intuitively can be thought of as quantifying network size and sample complexities, resp., needed to learn $\phi$ up to error $\epsilon$. 

\begin{definition}[Function complexity \cite{allen2019learning}]\label{def:complexity}
	Suppose that function $\phi: \mathbb{R} \rightarrow \mathbb{R}$ has Taylor expansion $\phi(z)=\sum_{i=0}^{\infty} c_{i} z^{i}$. For $R, \epsilon>0$, define
	$$
	\begin{array}{l}
		\mathfrak{C}_{\varepsilon}(\phi, R) := \sum_{i=0}^{\infty}\left(\left(C^{*} R\right)^{i}+\left(\frac{\sqrt{\log (1 / \varepsilon)}}{\sqrt{i}} C^{*} R\right)^{i}\right)\left|c_{i}\right|, \\
		\mathfrak{C}_{\mathfrak{s}}(\phi, R) := C^{*} \sum_{i=0}^{\infty}(i+1)^{1.75} R^{i}\left|c_{i}\right|,
	\end{array}
	$$ 
	where $C^{*}=10^{4}$. As an example, if $\phi(z) = z^d$ for positive integer $d$, then $\mathfrak{C}_{\mathfrak{s}}(\phi, R) = O(R^d)$ and 
	$\mathfrak{C}_{\varepsilon}(\phi, R)  = O(R^d \log^{d/2}(\frac{1}{\varepsilon}))$. For $\phi(z) = \sin z, \cos z, e^{z}$, we have     
	$\mathfrak{C}_{\mathfrak{s}}(\phi, R) = O(1)$ and $\mathfrak{C}_{\varepsilon}(\phi, R) = \mathrm{poly}(1/\varepsilon)$. 
	We have $\mathfrak{C}_{{s}}(\phi, R) \leq \mathfrak{C}_{\varepsilon}(\phi, R) \leq \mathfrak{C}_{{s}}(\phi, O(R)) \times \mathrm{poly}(1 / \varepsilon)$ for all $\phi$ and for $\phi(z) = \sin z, e^{z}$ or constant degree polynomials, they only differ by $o(1 / \varepsilon)$. 
	See \cite{allen2019learning} for details.
	Note that $\phi$ itself is not a member of our concept class but functions like it will be used to construct members of our concept class. 
\end{definition}

\section{Problem Formulation}    
In our set-up, RNNs output a label after processing the whole input sequence.\footnote{While our set-up has similarity to previous work \cite{allen2019can}, there are also important differences.}
The data are generated from an unknown distribution $\mathcal{D}$ over $ ((\overline{\mathbf{x}}^{(2)}, \ldots, \overline{\mathbf{x}}^{(L-1)}), \mathbf{y}^{\ast}) \in ((\mathbb{S}^{d-2})^{L-2}, \mathcal{Y})$, for some label set $\mathcal{Y} \subset \mathbb{R}^{\dout }$ for some positive integer $\dout $. 
We call $\overline{\bx} := (\overline{\mathbf{x}}^{(2)}, \ldots, \overline{\mathbf{x}}^{(L-1)})$ the \emph{true sequence} and $\mathbf{y}^{\ast}$ the \emph{true label}. Denote by $\mathcal{Z}$ the training dataset containing $N$ i.i.d. samples from $\mathcal{D}$. We preprocess the true sequence to \emph{normalize} it:
\begin{definition}[Normalized Input sequence]\label{def:normalized_seq}
	Let $\overline{\bx} = (\overline{\mathbf{x}}^{(2)}, \ldots, \overline{\mathbf{x}}^{(L-1)})$ be a given true input sequence of length $L-2$, s.t. $\overline{\mathbf{x}}^{(i)} \in  \mathbb{S}^{d-2}$ and $\overline{x}^{(i)}_{d-1} = \frac{1}{2}$, for all $i \in [2, L-1]$. 
	The normalized input sequence $\bx := (\mathbf{x}^{(1)}, \ldots, \mathbf{x}^{(L)})$ is given by
	\begin{align*}
		\mathbf{x}^{(1)} := (\mathbf{0}^{d-1}, 1), \;\;\;
		\mathbf{x}^{(\ell)} := (\varepsilon_x \overline{\mathbf{x}}^{(\ell)}, 0), \quad \forall \ell \in [2, L-1], \;\;\;
		\mathbf{x}^{(L)} := (\mathbf{0}^{d-1}, 1), 
	\end{align*}
	where we set $\varepsilon_x > 0$ later in Theorem~\ref{thm:main_theorem_paper}. 
\end{definition}	
	We use normalized sequence in place of the true sequence as input to RNNs, as it helps in proofs, e.g., with bounds on the changes in the activation patterns at each RNN cell, when the input sequences change and also with inversion of RNNs (defined later). Our method can be applied without normalization too, but in that case our error bound has exponential dependence on the length of the input sequence. The extra dimension in the normalized sequence serves as bias which we do not use explicitly to simplify notation. 	

\subsection{RNNs}
\begin{definition}[Recurrent Neural Networks]\label{def:RNN}
	We assume that the input sequences are of length $L$ for some given $L >0$ and are of the form $\mathbf{x}^{(1)}, \ldots, \mathbf{x}^{(L)}$ with $\mathbf{x}^{(\ell)} \in \mathbb{R}^{d}$ for all $\ell \in [L]$. An RNN is specified by three matrices $\mathbf{W}_{\mathrm{rnn}} \in \mathbb{R}^{m \times m}$, $\mathbf{A}_{\mathrm{rnn}} \in \mathbb{R}^{m \times d}$ and $\mathbf{B}_{\mathrm{rnn}}  \in \mathbb{R}^{\dout  \times m}$, where $m$ is the dimension of the hidden state and $\dout $ is the dimension of the output. The hidden states of the RNN are given by $\mathbf{h}^{(0)}_{\mathrm{rnn}} = \mathbf{0} \in \Reals^m$ and 
	\begin{align} \label{eqn:RNN_def}
		\mathbf{h}^{(\ell)}_{\mathrm{rnn}} &:=  \sigma(\mathbf{A}_{\mathrm{rnn}} \mathbf{x}^{(\ell)} + \mathbf{W}_{\mathrm{rnn}} \mathbf{h}^{(\ell-1)}_{\mathrm{rnn}} ) \quad \text{for } \ell \in [L].
	\end{align}
	The output at each step $\ell \in [L]$ is given by 
	$ \mathbf{y}^{(\ell)}_{\mathrm{rnn}} = \mathbf{B}_{\mathrm{rnn}} \mathbf{h}^{(\ell)}_{\mathrm{rnn}}$. By \emph{RNN cell} we mean the underlying FFN in \eqref{eqn:RNN_def}.
	The $m$ rows of $\mathbf{W}_{\mathrm{rnn}}$ and $\mathbf{A}_{\mathrm{rnn}}$ correspond to the $m$ neurons in the RNN.

	Pick the matrices $ \mathbf{W} \in \mathbb{R}^{m \times m}$ and $\mathbf{A} \in \mathbb{R}^{m \times d}$ by sampling entries i.i.d. from $N(0, \frac{2}{m})$, and pick $\mathbf{B}$ by sampling entries i.i.d. from $N(0, \frac{2}{\dout })$. When $\mathbf{W}_{\mathrm{rnn}} = \mathbf{W} $ and $\mathbf{A}_{\mathrm{rnn}} = \mathbf{A} $, the RNN is said to be at random initialization. 
	We will denote the parameters of an RNN at initialization by dropping the subscript ``rnn'', thus the hidden states are $\{\mathbf{h}^{(\ell)}\}_{\ell \in [L]}$. 
	In the following theorems, we will keep $\mathbf{B}_{\mathrm{rnn}}$ at initialization $\mathbf{B}$ and train only $\mathbf{A}_{\mathrm{rnn}}$ and $\mathbf{W}_{\mathrm{rnn}}$. 

	We write $F_{\mathrm{rnn}}^{(\ell)}(\bx; \mathbf{W}_{\mathrm{rnn}}, \mathbf{A}_{\mathrm{rnn}})=\mathbf{y}^{(\ell)}_{\mathrm{rnn}}=\mathbf{B} \mathbf{h}^{(\ell)}_{\mathrm{rnn}}$ for the output of the $\ell$-th step. Our goal is to use $\mathbf{y}^{(L)}_{\mathrm{rnn}} \in \mathbb{R}^{\dout }$ to fit the true label $\mathbf{y}^{\ast} \in \mathcal{Y}$ using some loss function $G: \mathbb{R}^{\dout } \times \mathcal{Y} \rightarrow \mathbb{R}$. We assume that for every $\mathbf{y}^{\ast} \in \mathcal{Y}, G\left(0^{k}, \mathbf{y}^{\ast}\right) \in[-1,1]$ is bounded, and
	$G\left(\cdot, \mathbf{y}^{\ast}\right)$ is convex and 1-Lipschitz continuous in its first variable. This includes, for instance, the cross-entropy loss and $\ell_{2}$ -regression loss (for bounded arguments).
\end{definition}

\subsection{Concept Class}
We now define our target concept class, which we will show to be learnable by RNNs using SGD. 

\begin{definition}[Concept Class]\label{def:concept_class}
	Our concept class consists of functions $F: \mathbb{R}^{(L-2) \cdot (d-1)} \rightarrow \mathbb{R}^{\dout }$ defined as follows.
	Let $\Phi$ denote a set of smooth functions with Taylor expansions with finite complexity as in Def.~\ref{def:complexity}. To define a function $F$, we choose a subset $\{\Phi_{r, s}: \mathbb{R} \rightarrow \mathbb{R}\}_{r \in [p], s \in [\dout ]}$ from $\Phi$,  $\{\mathbf{w}_{r, s}^{\dagger} \in \mathbb{S}^{(L-2)(d-1)-1}\}_{r \in[p], s \in[\dout ]}$, a set of weight vectors, and $\{b_{r, s}^{\dagger} \in \mathbb{R}\}_{r \in[p], s \in[\dout ]}$, a set of output coefficients with $\abs[0]{b_{r, s}^\dagger} \leq 1$. Then, we define $F: \mathbb{R}^{(L-2) \cdot (d-1)} \rightarrow \mathbb{R}^{\dout }$, where for each output dimension $s \in [\dout ]$ we define the $s$-th coordinate $F_s$ of $F=(F_1, \ldots, F_{\dout })$ by
	\begin{equation}\label{eq:concept_class}
		F_s(\overline{\mathbf{x}}) := \sum_{r \in [p]} b_{r, s}^{\dagger} \Phi_{r, s} \left(\langle  \mathbf{w}_{r, s}^{\dagger}, [\overline{\mathbf{x}}^{(2)}, \ldots, \overline{\mathbf{x}}^{(L-1)}]\rangle\right).
	\end{equation}
	To simplify formulas, we assume $\Phi_{r, s}(0)=0$ for all $r$ and $s$. 
	We denote the complexity of the concept class by
	\begin{align*}
		\mathfrak{C}_{\varepsilon}(\Phi, R):=\max_{\phi \in \Phi}\{\mathfrak{C}_{\varepsilon}(\phi, R)\}, \;\;
		\mathfrak{C}_{\mathfrak{s}}(\Phi, R):=\max _{\phi \in \Phi}\{\mathfrak{C}_{\mathfrak{s}}(\phi, R)\}.
	\end{align*}
\end{definition}
\vspace{-2mm}
	Let $F^{\ast}$ be a function in the concept class with smallest possible population loss which we denote by $\mathrm{OPT}$. Hence, we are in an agnostic learning setting where our aim is to learn a function with population objective $\mathrm{OPT} + \varepsilon$. 
As one can observe, functions in the concept class are given by a one hidden layer network with $p$ neurons and smooth activations. We will show that the complexity of the functions $\Phi_{r, s}$ determines the number of neurons and the number of training samples necessary to train the recurrent neural network that has $\mathrm{OPT} + \varepsilon$ population loss. 

While we have defined $F^*$ as a function of $\overline{\mathbf{x}}$, since there's a one-to-one correspondence between $\overline{\mathbf{x}}$ and $\mathbf{x}$, it will occasionally be convenient to talk about $F^*$ as being a function of $\mathbf{x}$---and this should cause no confusion. And similarly for other functions like $F_{\mathrm{rnn}}^{(\ell)}(\bx; \mathbf{W}, \mathbf{A})$.

\subsection{Objective Function and the Learning Algorithm}
We assume that there exists a function $F^{\ast}$ in the concept class that can achieve a population loss $\mathrm{OPT}$, i.e. $\underset{\left(\obx, \mathbf{y}^{\ast}\right) \sim \mathcal{D}}{\mathbb{E}} G(F^{\ast}(\obx), \mathbf{y}^{\ast}) \le \mathrm{OPT}$. The following loss function is used for gradient descent: 
\begin{align*}
    &\mathrm{Obj}(\mathbf{W}', \mathbf{A}') = \underset{\left(\obx, \mathbf{y}^{\ast}\right) \sim \mathcal{Z}}{\mathbb{E}} \mathrm{Obj}(\obx, \mathbf{y}^{\ast};  \mathbf{W}',  \mathbf{A}'), \text{  where} \\&
    \mathrm{Obj}(\obx, \mathbf{y}^{\ast};  \mathbf{W}',  \mathbf{A}') = G(\lambda F_{\mathrm{rnn}}^{(L)}(\bx;  \mathbf{W}+\mathbf{W}', \mathbf{A} + \mathbf{A}'), \mathbf{y}^{\ast} ).
\end{align*}
Parameter $\lambda$ whose value is set in the main Theorem~\ref{thm:main_theorem_paper} is a scaling factor needed for technical reasons discussed later. We consider vanilla stochastic gradient updates with $\mathbf{W}_t, \mathbf{A}_t$ denoting the matrices after $t$-steps of sgd. $\mathbf{W}_t$ and $\mathbf{A}_t$ are given by
\begin{align*}
	& \mathbf{W}_t = \mathbf{W}_{t-1} - \eta \, \nabla_{\mathbf{W}_{t-1}} \mathrm{Obj}(\obx, \mathbf{y}^{\ast};  \mathbf{W}_{t-1},  \mathbf{A}_{t-1}),
	\\&\mathbf{A}_t = \mathbf{A}_{t-1} - \eta \, \nabla_{\mathbf{A}_{t-1}} \mathrm{Obj}(\obx, \mathbf{y}^{\ast};  \mathbf{W}_{t-1},  \mathbf{A}_{t-1}),
\end{align*} \todo{try to save space}
where $(\obx, \mathbf{y}^{\ast})$ is a random sample from $\mathcal{Z}$ and $\bx$ is its normalized form. It should be noted that \cite{allen2019can} train only $\mathbf{W}$.

\emph{Remark.} We made two assumptions in our set-up: (1) input sequences are of fixed length, and (2) the output is only considered at the last step. These assumptions are without loss of generality and allow us to keep already quite complex formulas manageable without affecting the essential ideas. The main change needed to drop these assumptions is a change in the objective function, which will now include terms not just for how well the output fits the target at step $L$ but also for the earlier steps. The objective function for each step behaves in the same way as that for step $L$, and so the sum can be analyzed similarly. Intuitively speaking, considering the output at the end is the ``hardest'' training regime for RNNs as it uses the ``minimal'' amount of label information.

\subsection{RNNs learn the concept class}
We are now ready to state our main theorem. We use $    \rho :=100 L \dout  \log m$ in the following.  Recall that a set $\Phi$ of smooth functions induces a concept class as in Def.~\ref{def:concept_class}.
\begin{theorem}[Main, restated in the appendix as Theorem~\ref{thm:main_theorem}]\label{thm:main_theorem_paper}
	Let $\Phi$ be a set of smooth functions.
	  For  $\epsilon_x := \frac{1}{\operatorname{poly}(\rho)}$ and $\varepsilon \in \left(0, \frac{1}{p \cdot \operatorname{poly}(\rho) \cdot \mathfrak{C}_{\mathfrak{s}}(\Phi, \mathcal{O}(\epsilon_x^{-1}))}\right)$, define complexity $C:=\mathfrak{C}_{\varepsilon}(\Phi, \mathcal{O}(\epsilon_x^{-1}))$
	and $\lambda:=\frac{\varepsilon}{10 L \rho}$. Assume that the number of neurons $m \geq \operatorname{poly}\left(C, p, L, \dout , \varepsilon^{-1}\right)$ and the number of samples $N \geq \operatorname{poly}\left(C, p, L, \dout , \varepsilon^{-1}\right)$. 
	Then with parameter choices $\eta:=\Theta\left(\frac{1}{\varepsilon \rho^{2} m}\right)$ and $T:=\Theta(p^{2} C^{2} \operatorname{poly}(\rho)\varepsilon^{-2})$
	with probability at least $1-e^{-\Omega\left(\rho^{2}\right)}$ over the random initialization, SGD satisfies
	\vspace{-2mm}
	\begin{align} \label{eqn:main-thm}
		\underset{\mathrm { sgd }}{\mathbb{E}}\Big[\frac{1}{T} \sum_{t=0}^{T-1} \underset{\left(\obx, \mathbf{y}^{\ast}\right) \sim \mathcal{D}}{\mathbb{E}} \mathrm{Obj}\Big(\obx, \mathbf{y}^{\ast};  \mathbf{W}_{t}, \mathbf{A}_t\Big)\Big] 
		\leq  \mathrm{OPT}+\varepsilon+ 1/\operatorname{poly}(\rho).
	\end{align}
\end{theorem}
\vspace{-2mm}
Informally, the above theorem states that by SGD training of overparametrized RNNs with sufficiently small learning rate and appropriate preprocessing of the input sequence, we can efficiently find an RNN that has population objective nearly as small as $\mathrm{OPT}$ as $\varepsilon+ 1/\operatorname{poly}(\rho)$ is small.
The required number of neurons and the number of training samples have polynomial dependence on the function complexity of the concept class, 
the length of the input sequence, the output dimension, and the additional prediction error $\varepsilon$.  
\todo{can this be shortened a bit?}

\section{Proof Sketch}
While the full proof is highly technical, in this section we will sketch the proof focusing on the conceptual aspects while minimizing the technical aspects to the essentials; full proofs are in the appendix. The high-level outline of our proof is as follows.
\begin{enumerate}[leftmargin=*, itemsep=0.1pt, topsep=0pt]
	\item \emph{Overparamtrization simplifies the neural network behavior.} The function $F_{\mathrm{rnn}}^{(L)}(\bx; \mathbf{W}+\mathbf{W}', \mathbf{A} + \mathbf{A}')$ computed by the RNN is a function of the parameters $\mathbf{W}', \mathbf{A}'$ as well as of the input $\overline{\mathbf{x}}$. It is a highly non-linear and non-convex function in both the parameters and in the input. The objective function inherits these properties and its direct analysis is difficult. However, it has been realized in the last few years---predominantly for the FFN setting---that when the network is overparametrized (i.e., as the number of neurons $m$ becomes large compared to other paramters of the problem such as the complexity of the concept class), the network behavior simplifies in a certain sense. The general idea carries over to RNNs as well:
	in \eqref{eqn:linear-approximation} below we write the first-order Taylor approximation of $F_{\mathrm{rnn}}^{(L)}(\bx; \mathbf{W}+\mathbf{W}', \mathbf{A} + \mathbf{A}')$
	at $\mathbf{W}$ and $\mathbf{A}$ as a linear function of  $\mathbf{W}'$ and $\mathbf{A}'$; it is still a non-linear function of the input sequence. As in \cite{allen2019can} we call this function \emph{pseudo-network}, though our notion is more general as we vary both the parameters $\mathbf{W}'$ and $\mathbf{A}'$. Pseudo-network is a good approximation of the target network as a function of $\overline{\mathbf{x}}$ for all $\overline{\mathbf{x}}$. \todo{This statement needs more work and also pointers to the formal statements.}
	\item \emph{Existence of a good RNN.} In order to show that the RNN training successfully learns, we first show that there are parameters values for RNN so that as a function of $\overline{\mathbf{x}}$ it is a good approximation of $F^\ast$. Instead of doing this directly, we show that the pseudo-network can approximate $F^\ast$; this suffices as we know that the RNN and the pseudo-network remain close. This is done by constructing paramters $\mathbf{W}^{\ast}$ and $\mathbf{A}^{\ast}$ so that the resulting pseudo-network approximates the target function in the concept class (Section~\ref{sec:fitting-target}) for all $\overline{\mathbf{x}}$. 
	\item \emph{Optimization.} SGD makes progress because the loss function is convex in terms of the pseudo-network which stays close to the RNN as a function of $\mathbf{x}$. Thus, SGD finds parameters with training loss close to that achieved by $\mathbf{W}^{\ast}, \mathbf{A}^{\ast}$.
	\item \emph{Generalization.} Apply a Rademacher complexity-based argument to show that SGD has low population loss.
\end{enumerate}
Step 2 is the main novel contribution of our paper and we will give more details of this step in the rest of this section.\footnote{The above outline is similar to prior work, e.g., \cite{allen2019can}. Details can be quite different though, e.g., they only train $\mathbf{W}$ and keep $\mathbf{A}$ fixed to its initial value. 
Their contribution was also mainly in Step 2 and the other steps were similar to prior work.} 
\subsection{Pseudo-network}\label{sec:pseudo-network}
We define the pseudo-network here. Suppose $\mathbf{W}, \mathbf{A}, \mathbf{B}$ are at random initialization.
 The linear term in the first-order Taylor approximation is given by the pseudo-network
\begin{align} 
	F^{(L)}(\bx; \mathbf{W}', \mathbf{A}') 
	&:= \sum_{i=1}^{L}  \mathbf{Back}_{i \rightarrow L} \mathbf{D}^{(i)} \left(\mathbf{W'} \mathbf{h}^{(i-1)} + \mathbf{A'} \mathbf{x}^{(i)}\right) \label{eqn:linear-approximation}\\&
	\approx F_{\mathrm{rnn}}^{(L)}(\bx; \mathbf{W}+\mathbf{W}', \mathbf{A}+\mathbf{A}') - F_{\mathrm{rnn}}^{(L)}(\bx; \mathbf{W}, \mathbf{A}) \tag{using Lemma~\ref{lemma:perturb_NTK_small_output}}.
\end{align}
 This function approximates the change in the output of the RNN, when $(\mathbf{W}, \mathbf{A})$ changes to $(\mathbf{W}+\mathbf{W}^{'}, \mathbf{A}+\mathbf{A}^{'})$. 
 The parameter $\lambda$, that we defined in the objective function, will be used to make the contribution of  $F_{\mathrm{rnn}}^{(L)}$ at initialization small thus making pseudo-network a good approximation of RNN. Hence, we can observe that the pseudo network is a good approximation of the RNN, provided the weights stay close to the initialization.
 
 To complete the above definition of pseudo-network we define the two new notations in the above formula. For each $\ell \in [L]$, define $\mathbf{D}^{(\ell)} \in \mathbb{R}^{m \times m}$ as a diagonal matrix, with diagonal entries 
\begin{align}\label{eqn:diagonal_main}
	d_{rr}^{(\ell)} := \mathbb{I}[\mathbf{w}_r^{\top} \mathbf{h}^{(\ell - 1)} + \mathbf{a}_r^{\top} \mathbf{x}^{(\ell)} \ge 0], \quad \forall r \in [m]. 
\end{align}
In words, the diagonal of matrix $\mathbf{D}^{(\ell)}$ represents the activation pattern for the RNN cell at step $\ell$ at initialization.

Define $\mathbf{Back}_{i \to j} \in \mathbb{R}^{\dout  \times m}$ for each $1 \le i \le j \le L$ by
\begin{align*}
	\mathbf{Back}_{i \to j} := \mathbf{B} \mathbf{D}^{(j)} \mathbf{W} \ldots \mathbf{D}^{(i+1)} \mathbf{W}, 
\end{align*}
with $\mathbf{Back}_{i \to i} := \mathbf{B}$ for each $i \in [L]$. Matrices $\mathbf{Back}_{i \to j}$ in Eq. \eqref{eqn:linear-approximation} arise naturally in the 
computation of the first-order Taylor approximation (equivalently, gradients w.r.t. the parameters) using standard matrix calculus.
Very roughly, one can think of $\mathbf{Back}_{i \to j}$ as related to the backpropagation signal from the output at step $j$ to the parameters at step $i$.

\subsection{Existence of good pseudo-network}\label{sec:fitting-target}
Our goal is to construct $\mathbf{W}^{\ast}$ and $\mathbf{A}^{\ast}$ such that for any true input sequence $\overline{\mathbf{x}} = (\overline{\mathbf{x}}^{(2)}, \ldots, \overline{\mathbf{x}}^{(L-1)})$, if we define the normalized sequence $\mathbf{x} = (\mathbf{x}^{(1)}, \ldots, \mathbf{x}^{(L)})$, then with high probability we have
\begin{align}\label{eq:simplifiedFstar}
    F^{(L)}(\bx; \mathbf{W}^{\ast}, \mathbf{A}^{\ast}) \approx   F^{\ast}(\overline{\mathbf{x}}).
\end{align}
To simplify the presentation, in this sketch we will assume that $p$, the number of neurons in the concept class, and the output dimension $\dout $ are both equal to $1$. 
Also, let the output weight $b^{\dagger}:=1$. 
These assumptions retain the main proof ideas while simplifying equations. 
Overall, we assume that the target function $F^{\ast} : \Reals^{(L-2) \cdot (d-1)} \to \Reals$ on a given sequence is given by
\begin{align}\label{eqn:Fstar-Phi}
    F^{\ast}(\overline{\mathbf{x}}) = \Phi^{\ast}(\langle \mathbf{w}^{\dagger}, [\overline{\mathbf{x}}^{(2)}, \cdots, \overline{\mathbf{x}}^{(L-1)}]\rangle),
\end{align}
where $\Phi^{\ast}: \Reals \to \Reals$ is a smooth function and $\mathbf{w}^{\dagger} \in \mathbb{S}^{(L-2) \cdot (d-1) - 1}$. 


First, we state Lemma 6.2 in \cite{allen2019learning}, which is useful for our construction of the matrices $\mathbf{W}^{\ast}$ and $\mathbf{A}^{\ast}$.
Consider a smooth function $\phi: [-1, 1] \to \Reals$. It can be approximated as a linear combination of step functions (derivatives of ReLU) for all $u \in (-1, 1)$, i.e., there exists a ``weight function'' $H: \Reals^2 \to \Reals$ such that
$	 \phi\left(u\right) \approx \mathbb{E}_{\alpha_1, \beta_1, b_0} [{H\left(\alpha_{1}, b_0\right)} \, \mathbb{I}_{\alpha_{1} u + \beta_{1} \sqrt{1 - u^{2}} + b_0   \geq 0}]$
where $\alpha_1, \beta_1 \sim \mathcal{N}\left(0, 1\right) \text{ and } b_0 \sim \mathcal{N}\left(0, 1\right)$ are independent random variables (we omitted some technical details).

The above statement can be straightforwardly extended to the following slightly more general version: 
\begin{lemma}\label{lemma:Expressfunction_fNN}
    For every smooth function $\phi$, any $\overline{\mathbf{w}} \in \mathbb{S}^{d-1}$, and any $\varepsilon \in \left(0, \frac{1}{\mathfrak{C}_{s}\left(\phi, 1 \right)}\right)$ there exists a $H:\mathbb{R}^2 \to \left(- 
	\mathfrak{C}_\varepsilon\left(\phi,  1\right),  \mathfrak{C}_\varepsilon\left(\phi,  1\right)\right)$, which is $ \mathfrak{C}_\varepsilon\left(\phi, 1\right)$-Lipschitz continuous and for all $\mathbf{u} \in  \mathbb{S}^{d-1}$, we have
	$$
	\begin{array}{l}
		\Big|\phi(\overline{\mathbf{w}}^{\top} \mathbf{u}) - \mathbb{E}_{\mathbf{w} \sim \mathcal{N}(\mathbf{0}, \mathbf{I}), b_0 \sim \mathcal{N}(0, 1)} [{H(\mathbf{w}^{\top} \overline{\mathbf{w}}, b_0)} \,\mathbb{I}_{\mathbf{w}^{\top} \mathbf{u} + b_0 \geq 0} ]  \Big| \leq \varepsilon.
	\end{array}
	$$
\end{lemma} 
Very informally, this lemma states that the activation pattern of a one-layer $\relu$ network (given by $\mathbb{I}_{\mathbf{w}^{\top} \mathbf{u} \geq 0}$) at initialization can be used to express a smooth function of the dot product of the input vector with a fixed vector. While the above statement involves an expectation, one can easily replace it by an empirical average with slight increase in error. 
This statement formed the basis for FFN and RNN results in \cite{allen2019learning, allen2019can}. 
Can we use it for RNNs for our general concept class? An attempt to do so is the following lemma showing that the pseudo-network can express any smooth function of the hidden state $\mathbf{h}^{(L-1)}$ and $\mathbf{x}^{(L)}$. 

\begin{lemma}[Informal]\label{lemma:Expressfunction_RNN}
	For a given smooth function $\phi$, a vector $\overline{\mathbf{w}} \in \mathbb{S}^{(m+d-1)}$, and any $\varepsilon \in \left(0, \frac{1}{\mathfrak{C}_{s}\left(\phi, 1 \right)}\right)$, there exist matrices $\mathbf{W}^{\ast}$ and $\mathbf{A}^{\ast}$ such that for every normalized input sequence $\bx = (\bx^{(1)}, \ldots, \bx^{(L)})$ formed from a sequence $\obx$, we have with high probability,
	\begin{align*}
	    \abs{F^{(L)}(\bx; \mathbf{W}^{\ast}, \mathbf{A}^{\ast}) - \phi({\langle  \overline{\mathbf{w}}, [\mathbf{h}^{(L-1)}, \bx^{(L)}_{:d-1}]\rangle})} \le \varepsilon,
	\end{align*}
	provided $m = \mathrm{poly}(\frac{1}{\varepsilon}, L, \mathfrak{C}_{\epsilon}(\phi, \mathcal{O}(1)))$. Vector $\bx^{(L)}_{:d-1}$ is 
	$\bx^{(L)}$ without the last coordinate, the bias term appended to each input.
\end{lemma} 
The reason $\mathbf{h}^{(L-1)}$ and $\mathbf{x}^{(L)}$ come up is because they serve as inputs to the RNN cell when processing the $L$-th input. The proof sketched below uses the fact that RNNs are one-layer FFNs unrolled over time. Hence, we could try to apply the result of Lemma~\ref{lemma:Expressfunction_fNN} to the RNN cell at step $L$. However, a difficulty arises in carrying out this plan because the contributions of previous times steps also come up (as seen in the equations below) and it can be difficult to disentangle the contribution of step $L$. This is addressed in the proof: 

\begin{prf}
    Recall that $\mathbf{W}, \mathbf{A} \sim \mathcal{N}(\mathbf{0}, \frac{2}{m}\mathbf{I})$. Also, recall that we have assumed for simplicity $\dout  = 1$. Hence, $\mathbf{B}$ and $\mathbf{Back}_{i \to L}$ are row and column vectors respectively. For typographical simplicty, denote by $b_r$ and $\mathbf{Back}_{i \to L, r}$ the respective $r$-th components of these vectors.

We set $\mathbf{W}^{\ast} := \mathbf{0}$ and for every $r \in [m]$, $\mathbf{a}_r^{\ast} :=\frac{1}{m} b_r H( \sqrt{m/2} (\langle [\mathbf{w}_{r}, \mathbf{a}_{r, :d-1}],  \overline{\mathbf{w}} \rangle), \sqrt{m/2} a_{r, d}) \mathbf{e}_d$, for a function $H$ that we will describe below. With these choices we have
    \begin{align*}
        &F^{(L)}(\bx; \mathbf{W}^{\ast}, \mathbf{A}^{\ast})  = \sum_{i=1}^{L}  \mathbf{Back}_{i \rightarrow L} \mathbf{D}^{(i)} \left(\mathbf{W}^{\ast} \mathbf{h}^{(i-1)} + \mathbf{A}^{\ast} \mathbf{x}^{(i)}\right) 
        \\&= \frac{1}{m} \sum_{i=1}^{L}   \sum_{r \in [m]}  b_{r}  \mathbf{Back}_{i \to L, r}  H( \sqrt{m/2} (\langle [\mathbf{w}_r, \mathbf{a}_{r, :d-1}],  \overline{\mathbf{w}} \rangle), \sqrt{m/2} a_{r, d})    \cdot \mathbb{I}_{\mathbf{w}_r^{\top} \mathbf{h}^{(i-1)} + \mathbf{a}_r^{\top} \mathbf{x}^{(i)} \ge 0}. 
	\end{align*}
	In the last step, we have simplified the formula using sum over neurons.  
	The first $L-1$ summands in the outer sum above nearly vanish due to small correlation between $\mathbf{B}$ and $\mathbf{Back}_{i \to L}$ for $i < L$ (see  Lemma~\ref{lemma:backward_correlation}). Recall that $\mathbf{Back}_{L \to L} = \mathbf{B}$ and thus the correlation is not small for $i=L$. This  gives
	\begin{align*}
		F^{(L)}(\bx; \mathbf{W}^{\ast}, \mathbf{A}^{\ast}) &\approx   \frac{1}{m} \sum_{r \in [m]}  b_{r}^2 H( \sqrt{m/2} (\langle [\mathbf{w}_r, \mathbf{a}_{r, :d-1}],  \overline{\mathbf{w}} \rangle), \sqrt{m/2} a_{r, d})  \cdot \mathbb{I}_{\mathbf{w}_r^{\top} \mathbf{h}^{(i-1)} + \mathbf{a}_r^{\top} \mathbf{x}^{(i)} \ge 0}, 
	\end{align*}
    Now, this resembles a discretized version of Lemma~\ref{lemma:Expressfunction_fNN}. We can substitute $\mathbf{u}$ as $[\mathbf{h}^{(L-1)}, \mathbf{x}^{(L)}]$ in Lemma~\ref{lemma:Expressfunction_fNN} and use concentration bounds with respect to the randomness of weights $\mathbf{W}$ and $\mathbf{A}$ to complete the proof. 
\end{prf}
More generally, with much more technical work, it might be possible to prove an extension of the above lemma asserting the existence of a pseudo-network approximating a sum of functions of type $\sum_{i \in [L]}\phi_i({\langle  \overline{\mathbf{w}}_i, [\mathbf{h}^{(i-1)}, \bx^{(i)}]\rangle})$. However, even so it is not at all clear what class of functions of $\overline{\mathbf{x}}$ this represents because of the presence of the hidden state vectors.

Thus, the major challenge in constructing $\mathbf{W}^{\ast}$ and $\mathbf{A}^{\ast}$ to express the functions from the desired concept class is to use the information contained in $\mathbf{h}^{(\ell)}$. The construction of $\mathbf{W}^\ast$ in \cite{allen2019can} is not able to use this information and ignores it by treating it as noise (which is also non-trivial). The idea underlying our construction is that $\mathbf{h}^{(\ell)}$ in fact contains information about all the inputs $\bx^{(1)}, \ldots, \bx^{(\ell)}$ up until step $\ell$. Furthermore and crucially, this information can be recovered approximately by a linear transformation (Theorem~\ref{thm:Invertibility_ESN_outline} below). 
This enables us to show: \todo{Quantify }
\begin{theorem}[Existence of pseudo-network approximation for target function; abridged statement of Theorem~\ref{thm:existence_pseudo} in the appendix] \label{thm:existence_pseudo_outline}
	For every target function $F^{\ast}$ of the form Eq.~\eqref{eqn:Fstar-Phi}, there exist matrices $\mathbf{W}^{*}$ and $\mathbf{A}^{\ast}$ such that with probability at least $1-e^{-\Omega\left(\rho^{2}\right)}$ over $\mathbf{W}, \mathbf{A}, \mathbf{B}$, we have for every normalized input sequence $\mathbf{x} = (\mathbf{x}^{(1)}, \ldots, \mathbf{x}^{(L)})$ formed from a true sequence $\overline{\bx}$,  
	\begin{align*}
		\abs{F^{(L)}(\bx; \mathbf{W}^{\ast}, \mathbf{A}^{\ast}) - \Phi^{\ast} \left(\langle  \mathbf{w}^{\dagger}, [\overline{\mathbf{x}}^{(2)}, \ldots, \overline{\mathbf{x}}^{(L-2)}]\rangle\right)}  \le \varepsilon + \frac{1}{\mathrm{poly}(\rho)},   
	\end{align*}
	provided $m \ge \mathrm{poly}(\rho, L, \varepsilon^{-1}, \mathfrak{C}_{\varepsilon}(\Phi, \mathcal{O}(\varepsilon_x^{-1})))$ and $\epsilon_x \le \frac{1}{\mathrm{poly}(\rho)}$.
\end{theorem}

\begin{prf}[Proof sketch]
    By Theorem~\ref{thm:Invertibility_ESN_outline} there exists a matrix $\overline{\mathbf{W}}^{[L]}$ such that $\overline{\mathbf{W}}^{[L] \top} \mathbf{h}^{(L-1)} \approx [\bx^{(1)}, \ldots, \bx^{(L)}]$ for all input sequences $[\bx^{(1)}, \ldots, \bx^{(L)}]$. We can modify $\overline{\mathbf{W}}^{[L]}$ to get $[\obx^{(2)}, \ldots, \obx^{(L-1)}]$. Hence, by using  $[\overline{\mathbf{W}}^{[L]} \mathbf{w}^{\dagger}, \mathbf{0}]$ as $\overline{\mathbf{w}}$ and  $\Phi^{\ast}$ as $\phi$ in Lemma~\ref{lemma:Expressfunction_RNN}, we can have $F^{(L)}(\bx; \mathbf{W}^{\ast}, \mathbf{A}^{\ast}) \approx \Phi^{\ast}(\langle \mathbf{w}^{\dagger}, \overline{\mathbf{W}}^{[L]\top} \mathbf{h}^{(L-1)}\rangle) \approx F^{\ast}(\obx)$, implying $F^{(L)}$ and $F^{\ast}$ are close.   Accounting for all the errors in inversion and approximation of function, we get the final bound.
\end{prf}

\emph{Re-randomization.} In the proof sketches of Lemmas~\ref{lemma:Expressfunction_RNN} and Theorem~\ref{thm:existence_pseudo_outline} above we swept a technical but critical consideration under the rug: the random variables $\{\mathbf{w}_r, \mathbf{a}_r\}_{r \in [m]}$, $\obW^{(L)}$, $\{\mathbf{Back}_{i \to L}\}_{i \in [L]}$ and $\{\mathbf{h}^{(i)}\}_{i \in [L]}$ are not independent. This invalidates application of standard concentration inequalities w.r.t. the randomness of $\mathbf{W}$ and $\mathbf{A}$---this application is required in the proofs. Here our new variation of the re-randomization technique from \cite{allen2019can} comes in handy. The basic idea is the following: whenever we want to apply concentration bounds w.r.t. the randomness of $\mathbf{W}$ and $\mathbf{A}$, we divide the set of rows into disjoint sets of equal sizes. For each set, we will re-randomize the rows of the matrix $[\mathbf{W}, \mathbf{A}]_r$, show that the matrices $\obW^{[L]}$, $\{\mathbf{Back}_{i \to L}\}_{i \in [L]}$ and $\{\mathbf{h}^{(i)}\}_{i \in [L]}$ don't change a lot and then apply concentration bounds w.r.t. the new weights in the set. Finally, we account for the error from each set.

\subsection{The rest of the proof}
Having shown that there exists a pseudo-network approximation of the RNN that can also approximate the concept class,
we will complete the proof by showing that SGD can find matrices with performance similar to $\mathbf{W}^{\ast}$ and $\mathbf{A}^{\ast}$ on the population objective $\mathrm{Obj(\cdot)}$. Lemma~\ref{lem:trainloss}  shows that the training loss decreases with time. The basic idea is to use the fact that within small radius of perturbation, overparametrized RNNs behave as a linear network and hence the training can be analyzed via convex optimization. Then, we show using Lemma~\ref{lem:radcomp} that the Rademacher complexity for overparametrized RNNs is bounded. Again, the main idea here is that overparametrized  RNNs behave as pseudo-networks in our overparametrized regime and hence their Rademacher complexity can be approximated by the Rademacher complexity of pseudo-networks. Finally, using generalization bounds on the Rademacher complexity, we get the final population-level objective in  Theorem~\ref{thm:main_theorem}.


\subsection{Invertibility of RNNs at initialization}\label{sec:invertibility}
In this section, we describe how to get back $\bx^{(1)}, \ldots, \bx^{(L)}$ from the hidden state $\mathbf{h}^{(L)}$. 
The following lemma states that any linear function can be represented by a one-hidden layer FFN with activation function $\relu$,with a small approximation error of the order $\frac{1}{\sqrt{m}}$:
\begin{lemma}\label{lem:normal_linearestimate_appr_outline} [a simpler continuous version can be found in Lemma~\ref{lem:normal_linearestimate} in the appendix]
	For any $\bv \in \Reals^d$, the linear function taking $\bx$ to $\bv^\top \bx$ for $\bx \in \Reals^d$, can be represented as
	\begin{equation}
		\abs{\bv^\top \bx -  \mathbf{p}^{\top} \sigma(\mathbf{T} \mathbf{x})} \le \frac{\norm{\bv} \cdot \norm{\bx}}{\sqrt{m}},
	\end{equation}
	with 
	$\mathbf{p} \;=\;  2 \, \mathbf{T} \mathbf{v},$ where $\mathbf{T} \in \mathbb{R}^{m \times d}$ is a matrix with elements i.i.d. sampled from $\mathcal{N}(0, 1)$. 
\end{lemma}
Using the above lemma,\footnote{This lemma is from a paper that will appear soon; apart from the above lemma, 
	this work is very different from the present paper. We have reproduced the proof in full in the appendix.} we will show that the hidden state $\mathbf{h}^{(L)}$ can be inverted using a matrix $\obW^{[L]}$ to get back the input sequence $\mathbf{x}^{(1)}, \ldots, \mathbf{x}^{(L)}$.


\todo{We need a bound on the norm of the $x$ or it should show up in the RHS}
\begin{theorem}\label{thm:Invertibility_ESN_outline}[informal version of Theorem~\ref{thm:Invertibility_ESN}]
	There exists a set of matrices $\{ \obW^{[\ell]} \}_{\ell \in [L]}$, which can possibly depend on $\mathbf{W}$ and $\mathbf{A}$, such that for any $\varepsilon_x < \frac{1}{L}$ and any given normalized sequence $\bx^{(1)}, \ldots, \bx^{(L)}$, 
	with probability at least $1 - e^{-\Omega(\rho^2)}$ we have
	$$
	\begin{array}{l}
		\norm[1]{[\bx^{(1)}, \ldots, \bx^{(L)}] - \obW^{[L]\top} \mathbf{h}^{(L)}}_\infty  \leq \mathrm{poly}(L, \rho, m^{-1}, \varepsilon_x).
	\end{array}
	$$
\end{theorem}
Very roughly, the above result is obtained by repeated application of Lemma~\ref{lem:normal_linearestimate_appr_outline} to go from $\mathbf{h}^{(\ell)}$ to $(\mathbf{h}^{(\ell-1)}, \mathbf{x}^{(\ell)})$ starting with $\ell = L$. This uses the fact that the RNN cell is a one-hidden layer neural network and hence Lemma~\ref{lem:normal_linearestimate_appr_outline} is applicable. Several difficulties need to be overcome to carry out this plan. One difficulty is that a naive application of Lemma~\ref{lem:normal_linearestimate_appr_outline}
results in exponential blowup of error with $L$. We defer the technical details of this resolution to the full proof in the appendix.
Secondly, we apply re-randomization to tackle the dependence between $\mathbf{W}$, $\mathbf{A}$, $\{\mathbf{h}^{(\ell)}\}_{\ell \in [L]}$ and $\{\obW^{[\ell]}\}_{\ell \in [L]}$.
We performed few toy experiments on the ability of invertibility for  RNNs at initialization (Sec.~\ref{sec:expts}). We observed, as predicted by our theorem above, that the error involved in inversion decreases with the number of neurons and increases with the length of the sequence (Fig.~\ref{fig:RNN_inver}).






\section{On concept classes}

It is apparent that our concept class is very general as it allows arbitrary dependencies across tokens. To concretely illustrate the generality of our concept class, and to compare with previous work, we show that our result implies that  RNNs can recognize a simple formal language $D_{L_1}$. Here we are working in the discrete setting where each input token comes from $\{0, 1\}$ possibly represented as a vector when fed to the RNN. For a sequence $\mathbf{z} \in \{0, 1\}^L$, we define $D_{L_1}(\mathbf{z})$ to be $1$ if the number of $1$'s in $\mathbf{z}$ is exactly $1$ and define it to be $0$ otherwise. We can show that $D_{L_1}$ is not representable in the concept class of \cite{allen2019can} (see Theorem~\ref{thm:allencantdl1} in the appendix).  However, we can show that the language $D_{L_1}$ can be recognized with a one-layer FFN with one neuron and quadratic activation. The idea is that we can simply calculate the number of $1$'s in the input string, which is doable using a single neuron. This implies that our concept class can represent language $D_{L_1}$ with low complexity. More generally, we can show that our concept class can efficiently represent pattern matching problems, where strings belong to a language only if they contain given strings as substrings. In general, we can show that our concept class can express general regular languages. However, the complexity of the concept class may depend super-polynomially on the length of the input sequence, depending on the regular language (more discussion in sec.~\ref{sec:diff}). Some regular languages allow special treatment though. For example, consider the language $\mathsf{PARITY}$. $\mathsf{PARITY}$ is the language over alphabet $\{0, 1\}$ with a string $w = (w_1, \ldots, w_j) \in \mathsf{PARITY}$ iff
$w_1+\ldots +w_j = 1 \,\mathrm{mod}\, 2$, for $j \geq 1$. We can show in sec.~\ref{sec:diff} that $\mathsf{PARITY}$ is easily expressible by our concept class with small complexity. RNNs perform well on regular language recognition task in our experiments in Sec.~\ref{sec:expts}. Figuring out which regular languages can be efficiently expressed by our concept class remains an interesting open problem.




\section{Limitations and Conclusions} \label{sec:conclusion}

We proved the first result on the training and generalization of RNNs when the functions in the concept class are allowed to be essentially arbitrary continuous functions of the token sequence. Conceptually the main new idea was to show that the hidden state of the RNN contains information about the whole input sequence and this can be recovered via a linear transformation. We believe our techniques can be used to prove similar results for echo state networks.  

Two main limitations of the present work are: (1) Our overparametrized setting requires the number of neurons to be large in terms of the problem parameters including the sequence length---and it is often qualitatively different from the practical setting. Theoretical analysis of practical parameter setting remains an outstanding challenge---even for one-hidden layer FFNs. 
(2) We did not consider generalization to sequences longer than those in the training data. Such a result would be very interesting but it appears that it would require stronger assumptions than our very general assumptions about the data distribution. Our techniques might be a useful starting point to that end: for example, if we knew that the distributions of the hidden states are similar at different times steps and the output is the same as the hidden state (i.e. $\mathbf{B}$ is the identity) then our results might easily generalize to higher lengths. We note that to our knowledge the limitation noted here holds for all works dealing with generalization for RNNs.
(3) Understanding LSTMs remains open.

\bibliographystyle{unsrt}
\bibliography{references}


\newpage
\appendix
\tableofcontents
\vspace{10 mm}
The appendix has been structured as follows.   We discuss few more notations and basic facts in section~\ref{sec:further_prelims}. We prove few basic properties of the recurrent neural network at initialization in section~\ref{sec:basic_prop}. In section~\ref{sec:singlecell_RNN}, we prove in lemma~\ref{lemma:singlecell_ESN} that any linear function on $[\mathbf{h}^{(\ell-1)}, \mathbf{x}^{(\ell)}]$ at RNN cell $\ell$ can be expressed as a linear transformation of the hidden state $\mathbf{h}^{(\ell)}$. In section~\ref{sec:Invert_RNN}, we use the above lemma to show in theorem~\ref{thm:Invertibility_ESN} that from hidden state $\mathbf{h}^{[L]}$, one can get $[\bx^{(1)}, \cdots, \bx^{[L]}]$ using a linear transformation. We show in section~\ref{sec:existence} that a linear approximation of the recurrent neural networks exist at initialization that can approximate the target function in our concept class. We finish the proof in section~\ref{sec:optim_general}, where we show that RNNs can find a network with population risk close to the target function. We discuss about the experiments in section~\ref{sec:expts}.
\todo{Review this nearer to the submissino}

\section{Further Preliminaries}\label{sec:further_prelims}
\subsection{Notations}
Let $\Ball^d := \{\bx \in \Reals^d \mid \norm{\bx}_2 \leq 1\}$ be the unit $L_2$-ball in $\Reals^d$, and let
$\Sphere^{d-1} := \{\bx \in \Reals^d \mid \norm{\bx}_2 = 1\}$ be the unit $L_2$-sphere in $\Reals^d$. Let $V_d := \frac{\pi^{d/2}}{\Gamma((d+1)/2)}$ 
be the $d$-dimensional volume of $\Ball^d$ and let $\omega_{d-1} := \frac{2\pi^{d/2}}{\Gamma(d/2)}$ be the surface area (i.e. the $(d-1)$-dimensional volume) of $\Sphere^{d-1}$. Given a matrix $\mathbf{T} \in \mathbb{R}^{d_1 \times d_2}$ and a set $S \subset [d_1]$, we denote $\mathbf{T}_{S}$ as the matrix $\mathbb{R}^{\abs{S} \times d_2}$ that contains the rows of $\mathbf{T}$ whose indices are in the set $S$. We will denote a diagonal matrix $\mathbf{D}_{S}$ for a given set $S \subset [n]$ as $d_{ii} = 1$ for $i \in S$ and is $0$ elsewhere.

\todo{Repetitive with prelims in the main paper}
For positive integer $n$ define $[n]:=\{1, 2, \ldots, n\}$. For a matrix $\bM \in \Reals^{m\times n}$, set
$\norm{\bM}_{2, \infty} := \norm{(\norm{\mathbf{m}_1}_2, \ldots \norm{\mathbf{m}_m}_2)}_\infty$, where $\mathbf{m}_1^T, \mathbf{m}_2^T, \ldots$ are the rows of $\bM$. 
Let $\mu_d^{\beta}$ denote the Gaussian measure on $\Reals^d$ associated with the Gaussian probability distribution $\mathcal{N}(\mathbf{0},  \beta^2\mathbf{I})$.
Let $\mu_d := \mu_d^{1}$ denote the standard Gaussian measure on $\Reals^d$.


\todo{Specify what $O(\cdot)$ and $\Omega(\cdot)$ mean: do they hide absolute constants or do they hide constants depending on other parameters. Similarly, clarify the role of poly.}

For simplicity of notation, we will use
\begin{align*}
    &\varrho :=\frac{100 L \dout  p \cdot \mathfrak{C}_{\varepsilon}(\Phi, \mathcal{O}(\varepsilon_x^{-1})) \cdot \log m}{\varepsilon}. \\&
    \rho :=100 L \dout  \log m.
\end{align*}


\subsection{Extra set of Notations for RNNs}
We denote by $\mathbf{A}_{[d-1]}  \in \mathbb{R}^{m \times (d-1)}$ the matrix containing the first $d-1$ columns of the matrix $\mathbf{A}$. Then, we define an alternate fixed sequence as follows: $\bx_{(0)} := (\mathbf{x}^{(1)}_{(0)}, \ldots, \mathbf{x}^{(L)}_{(0)})$, where
	\begin{align*}
		\mathbf{x}^{(1)}_{(0)} = (\mathbf{0}^{d-1}, 1), \;\;\;
		\mathbf{x}^{(\ell)}_{(0)} = (\mathbf{0}^{d-1}, \varepsilon_x), \quad \forall \ell \in [2, L-1], \;\;\;
		\mathbf{x}^{(L)}_{(0)} = (\mathbf{0}^{d-1}, 1).
	\end{align*}
	We will heavily use this fixed sequence to build our model later on. \todo{Can you give better motivation for defining this? Is this definition necessary in the main paper?} There is a small difference in our definition of normalized sequence and the definition in \cite{allen2019can}. The difference is in the definition of $\bx^{(L)}$; our choice gives a better and simpler error bound. This difference leads to only minor changes in the theorems that we take from \cite{allen2019can} and we will account for those changes.

We re-introduce two more notations here for RNNs in def.~\ref{def:RNN}. For each $\ell \in [L]$, define $\mathbf{D}^{(\ell)} \in \mathbb{R}^{m \times m}$ as a diagonal matrix, with diagonal entries \todo{why do we need to use $d_{rr}^{(\ell)}$ why not $\mathbf{D}^{(\ell)}_{rr}$}
\begin{align}\label{eqn:diagonal}
	d_{rr}^{(\ell)} := \mathbb{I}[\mathbf{w}_r^{\top} \mathbf{h}^{(\ell - 1)} + \mathbf{a}_r^{\top} \mathbf{x}^{(\ell)} \ge 0], \quad \forall r \in [m]. 
\end{align}
Hence, $\mathbf{h}^{(\ell)} = \mathbf{D}^{(\ell)} (\mathbf{W} \mathbf{h}^{(\ell-1)} + \mathbf{A} \mathbf{x}^{(\ell)})$. Also, define $\mathbf{Back}_{i \to j} \in \mathbb{R}^{\dout  \times m}$ for each $1 \le i \le j \le L$ by
\begin{align*}
	\mathbf{Back}_{i \to j} := \mathbf{B} \mathbf{D}^{(j)} \mathbf{W} \ldots \mathbf{D}^{(i+1)} \mathbf{W}, 
\end{align*}
with $\mathbf{Back}_{i \to i} := \mathbf{B}$. Matrices $\mathbf{Back}_{i \to j}$ arise naturally in Eq. \eqref{eqn:linear-approximation} in the first-order Taylor approximation in terms of the parameters of the function $F_{\mathrm{rnn}}^{(\ell)}(\overline{\mathbf{x}} ; \mathbf{W}, \mathbf{A})$. Very roughly, one can interpret $\mathbf{Back}_{i \to j}$ as related to the backpropagation signal from the output at step $j$ to the parameters at step $i$. 

For the fixed base sequence $\mathbf{x}^{(1)}_{(0)}, \ldots, \mathbf{x}^{(L)}_{(0)}$, we will denote the hidden states by $\mathbf{h}^{(\ell)}_{(0)}$ and the diagonal matrices by $\mathbf{D}^{(\ell)}_{(0)}$ for $\ell \le L$.


\subsection{Redefine Concept Class}
In this section, we re-define the concept class introduced in the main paper. We introduce additional symbols related to the lipschitz constant and the absolute bounds over the functions, that are necessary in the proof of the main theorem.
\begin{definition}[Concept Class]
		Our concept class consists of functions $F: \mathbb{R}^{(L-2) \cdot (d-1)} \rightarrow \mathbb{R}^{\dout }$ defined as follows.
	Let $\Phi$ denote a set of smooth functions with Taylor expansions with finite complexity as in Def.~\ref{def:complexity}. To define a function $F$, we choose a subset $\{\Phi_{r, s}: \mathbb{R} \rightarrow \mathbb{R}\}_{r \in [p], s \in [\dout ]}$ from $\Phi$,  $\{\mathbf{w}_{r, s}^{\dagger} \in \mathbb{S}^{(L-2)(d-1)-1}\}_{r \in[p], s \in[\dout ]}$, a set of weight vectors, and $\{b_{r, s}^{\dagger} \in \mathbb{R}\}_{r \in[p], s \in[\dout ]}$, a set of output coefficients with $\abs[0]{b_{r, s}^\dagger} \leq 1$. Then, we define $F: \mathbb{R}^{(L-2) \cdot (d-1)} \rightarrow \mathbb{R}^{\dout }$, where for each output dimension $s \in [\dout ]$ we define the $s$-th coordinate $F_s$ of $F=(F_1, \ldots, F_{\dout })$ by
	\begin{equation}\label{eq:concept_class_appendix}
		F_s(\overline{\mathbf{x}}) := \sum_{r \in [p]} b_{r, s}^{\dagger} \Phi_{r, s} \left(\langle  \mathbf{w}_{r, s}^{\dagger}, [\overline{\mathbf{x}}^{(2)}, \ldots, \overline{\mathbf{x}}^{(L-1)}]\rangle\right).
	\end{equation}
	To simplify formulas, we assume $\Phi_{r, s}(0)=0$ for all $r$ and $s$. 
	We denote the complexity of the concept class by
	\begin{align*}
		\mathfrak{C}_{\varepsilon}(\Phi, R):=\max_{\phi \in \Phi}\{\mathfrak{C}_{\varepsilon}(\phi, R)\}, \;\;
		\mathfrak{C}_{\mathfrak{s}}(\Phi, R):=\max _{\phi \in \Phi}\{\mathfrak{C}_{\mathfrak{s}}(\phi, R)\}.
	\end{align*}
	 Let $L_{\phi}$ denote the Lipschitz constant of function $\phi$ in the range $(-\sqrt{L}, \sqrt{L})$ and let $L_{\Phi} := \max_{\phi \in \Phi} L_{\phi}$. Also, $C_{\phi}$ denote the upper bound on the absolute value of $\phi$ in the range $(-\sqrt{L}, \sqrt{L})$ and let $C_{\Phi} := \max_{\phi \in \Phi} C_{\phi}$. We only focus on the properties of the functions in the above range, since the argument to the functions $\langle  \mathbf{w}_{r, s}^{\dagger}, [\overline{\mathbf{x}}^{(2)}, \ldots, \overline{\mathbf{x}}^{(L-1)}]\rangle$ can be shown to lie in the above range. Using the definition of $\mathfrak{C}$ from def.~\ref{def:complexity}, one can show that
	$    C_{\Phi}, L_{\Phi} \le \mathfrak{C}_s(\Phi, \sqrt{2L})$.
	We assume that there exists some function $F^{\ast}$ in the concept class that achieves population loss $\mathrm{OPT}$. Hence, our aim is to learn a function with population loss $\mathrm{OPT} + \varepsilon$. 
\end{definition}

\subsection{Important facts}

We will need the following well-known results. 
\begin{fact}[e.g. Cor.~5.35 in \cite{vershynin2010introduction}]\label{thm:norm_W}
	Let $\mathbf{A}$ be an $N \times n$ matrix whose entries are independent standard normal random variables. Then for every $t \geq 0,$ with probability at least $1-2 \exp \left(-t^{2} / 2\right)$ one has
	\[
	\sqrt{N}-\sqrt{n}-t \leq s_{\min }(\mathbf{A}) \leq s_{\max }(\mathbf{A}) \leq \sqrt{N}+\sqrt{n}+t.
	\]
\end{fact}

\begin{fact}[e.g. Thm.~1.1 in \cite{vershynin2011spectral}]\label{thm:norm_BA}
	Let $\varepsilon \in(0,1)$ and let $m, n, N$ be positive integers. Consider a random $m \times n$ matrix $\mathbf{W}=\mathbf{B} \mathbf{A},$ where $\mathbf{A}$ is an $N \times n$ random matrix whose
	entries are independent random variables with mean zero and $(4+\varepsilon)$-th moment bounded by 1, and $\mathbf{B}$ is an $m \times N$ non-random matrix. Then w.p. exceeding $1 - 2 \exp{(-t^2/2)}$
	$$ 
	\|\mathbf{W}\| \leq C(\varepsilon) \norm{\mathbf{B}}(\sqrt{n}+\sqrt{m} + t), \mbox{ and }
	$$
	$$ 
	s_{\min}(\mathbf{W}) \geq C(\varepsilon) \norm{\mathbf{B}}(\sqrt{m} - \sqrt{n-1} - t),
	$$
	where $C(\varepsilon)$ is a constant that depends only on $\varepsilon$. 
\end{fact}
\todo{I didn't see the lower bound result in Vershynin's paper cited above, certainly not in the statement of Theorem 1.}

\begin{fact}[Example 2.11 in \cite{wainwright_2019_book}]\label{lem:chi-squared}
	Let $Z_1, Z_2, \ldots$ be i.i.d. one-dimensional standard Gaussian random variables. Then
	\begin{align*}
		\Pr\left[\abs{\frac{1}{n}\sum_{i=1}^n Z_i^2 - 1} \geq t\right] \leq 2 e^{-nt^2/8}, \quad \text{for all } t \in (0, 1).
	\end{align*}
\end{fact}

\begin{fact}[Maximum of Gaussians, see e.g. \cite{boucheron2013concentration}.]\label{fact:max_gauss} 
	Let $x_{1}, x_{2}, \ldots, x_{n}$ be n Gaussians following $\mathcal{N}\left(0, \sigma^{2}\right) .$ Then
	for any $\rho>0$ 
	$$
	\operatorname{Pr}\left\{\max _{i \in[n]}\left|x_{i}\right| \leq \sqrt{2}\rho \sigma \right\} \geq 1-2ne^{-\rho^2}.
	$$
\end{fact}

\begin{fact}[Hoeffding's inequality]\label{fact:hoeffding}
	Let $x_1, \cdots, x_n$ be $n$ independent random variables, with each $x_i$ strictly bounded in the interval $[a_i, b_i]$. Let $\overline{x} = \frac{1}{n} \sum_i x_i$. Then for any $\rho > 0$,
	\begin{align*}
		\Pr\left[ \abs{\overline{x} - \mathbb{E}_x \overline{x}} \ge \frac{\rho}{n} \sqrt{\sum_i (a_i - b_i)^2}\right] \le e^{-2\rho^2},
	\end{align*}
\end{fact}

\begin{definition}[$\epsilon$-net on the sphere] A set $\mathcal{N} \subset \Sphere^{d-1}$ is called an $\epsilon$-net of $\Sphere^{d-1}$ if every point in 
	$\Sphere^{d-1}$ is within Euclidean distance $\epsilon$ of some point in $\mathcal{N}$. In other words, for every $\bx \in \Sphere^{d-1}$ there is a point 
	$\tilde{\bx} \in \mathcal{N}$ such that $\norm{\bx - \tilde{\bx}} \leq \epsilon$.
\end{definition}

\begin{fact}[see the proof of Cor. 4.2.13 in \cite{vershynin_2018_book}]\label{fact:eps-net}
	$\Sphere^{d-1}$ has an $\epsilon$-net of size at most $(3/\epsilon)^d$. 
\end{fact}

Let $\mathcal{F}:\mathbb{R}^{d} \to \mathbb{R}$ be a class of function and $\mathcal{Z} = (\bx_1, \ldots, \bx_N)$ be a set of training examples in $\mathbb{R}^{d}$. The empirical rademacher complexity is given by
\begin{align*}
	\hat{R}(\mathcal{F}, \mathcal{Z}) := \sup_{f \in \mathcal{F}} \mathbb{E}_{\zeta \in \{\pm 1\}^N} \left[\frac{1}{N} \sum_{q \in [N]} \zeta_q f(\bx_q) \right]
\end{align*}

\begin{fact}[Generalization through rademacher complexity, \cite{shalev2014understanding}]\label{fact:genRad}
	If for every function $f \in \mathcal{F}$, $\abs{f} \le b$, then with probability at least $1 - \delta$ for any $\delta \ge 0$,
	\begin{align*}
		\sup_{f \in \mathcal{F}} \left[\mathbb{E}_{\bx \in \mathcal{D}} f(\bx) - \mathbb{E}_{\bx \in \mathcal{Z}} f(\bx) \right] \le 2 \hat{R}(\mathcal{F}, \mathcal{Z})  + \mathcal{O}(\frac{b \sqrt{\log (1/\delta)}}{\sqrt{N}}).
	\end{align*}
\end{fact}

\begin{fact}[Rademacher complexity of linear networks, \cite{shalev2014understanding}]\label{fact:radcomp_linear}
	Suppose $\norm[0]{\bx}_2 = 1$ for all $\bx \in \mathcal{X}$. The class $\mathcal{F}=\left\{\bx \mapsto\langle \bw, \bx\rangle \mid\|\bw\|_{2}<B\right\}$ has rademacher complexity
	$$
	\widehat{R}(\mathcal{F}, \mathcal{Z}) \leq O\left(\frac{B}{\sqrt{\mathrm{N}}}\right).
	$$
\end{fact}

\section{Some basic properties of recurrent neural networks at initialization}\label{sec:basic_prop}
The following lemma shows some basic properties of the recurrent neural network at initialization. They are a result of the concentration bounds that can be applied for gaussian weight matrices $\bW$ and $\mathbf{A}$.
\begin{lemma}\label{lemma:norm_ESN}
	For any $\epsilon_x \in (0, \frac{1}{L})$ and any normalized input sequence $(\mathbf{x}^{(1)}, \mathbf{x}^{(2)}, ..., \mathbf{x}^{(L)})$, for all $\ell \in [L]$ with probability at least $1-e^{-\Omega\left(\rho^2\right)}$ we have
	\todo{Specify the value of $\rho$.}
	\begin{enumerate}
		\item $\abs{\norm[0]{\mathbf{h}^{(\ell)}} - \sqrt{2 + (\ell - 2)\epsilon_x^2}}  \le  \frac{\rho^2}{\sqrt{m}}$.
		\item $\norm{\mathbf{W}\mathbf{h}^{(\ell)}}_{\infty}, \norm{\mathbf{A}\mathbf{x}^{(\ell)}}_{\infty} \le \mathcal{O}(\frac{\rho}{\sqrt{m}})$, for all $1 \le 
		\ell \le L$.
		\item $\norm{\mathbf{h}^{(\ell)} - \mathbf{h}^{(\ell)}_{(0)}} \le \sqrt{L} \epsilon_x$, for all $1 \le 
		\ell \le L$.
		\item $\norm{\mathbf{\mathbf{W} (\mathbf{h}^{(\ell)} - \mathbf{h}_{(0)}^{(\ell)})}}_{\infty}, \norm{\mathbf{\mathbf{A} (\mathbf{x}^{(\ell)} - \mathbf{x}_{(0)}^{(\ell)})}}_{\infty} \le \mathcal{O}(\frac{\rho \sqrt{L} \epsilon_x}{\sqrt{m}})$, for all $1 \le 
		\ell \le L$. 
		\item $(1 - \frac{1}{100L})^{j-i+1} \norm{\mathbf{u}} \le \norm{\mathbf{D}^{(j)} \mathbf{W} \mathbf{D}^{(j-1)} \mathbf{W} \cdots \mathbf{D}^{(i+1)} \mathbf{W} \mathbf{D}^{(i)} \mathbf{W} \mathbf{u}} \le (1 + \frac{1}{100L})^{j-i+1} \norm{\mathbf{u}}$ for all $1 \le i \le j \le L$ and any fixed vector $\mathbf{u}$.
		\item $(1 - \frac{1}{100L})^{j-i+1} \norm{\mathbf{u}} \le \norm{\mathbf{D}^{(j)} \mathbf{W} \mathbf{D}^{(j-1)} \mathbf{W} \cdots \mathbf{D}^{(i+2)} \mathbf{W} \mathbf{D}^{(i+1)} \mathbf{W} \mathbf{D}^{(i)} \mathbf{A} \mathbf{u} } \le (1 + \frac{1}{100L})^{j-i+1} \norm{\mathbf{u}}$ for all $1 \le i \le j \le L$ and all vectors $\mathbf{u} \in \mathbb{R}^{d}$.
		\item $\norm{\mathbf{D}^{(j)} \mathbf{W} \mathbf{D}^{(j-1)} \mathbf{W} \cdots \mathbf{D}^{(i+1)} \mathbf{W} \mathbf{D}^{(i)} \mathbf{W}} \le \mathcal{O}(L^3)$ for all $1 \le i \le j \le L$.
		\item $\norm{\mathbf{D}^{(\ell)} - \mathbf{D}_{(0)}^{(\ell)}}_0 \le \mathcal{O}(L^{1/3} \epsilon_x^{2/3}m)$ for all $1 \le 
		\ell \le L$.
		\item $\abs{\mathbf{u}^{\top} \mathbf{W} \mathbf{D}^{(j)} \mathbf{W} \mathbf{D}^{(j-1)} \mathbf{W} \cdots \mathbf{D}^{(i+1)} \mathbf{W} \mathbf{D}^{(i)} \mathbf{W} \mathbf{v}} \le \mathcal{O}(\frac{\sqrt{s_1} \rho}{\sqrt{m}}) \cdot \norm{\mathbf{u}}\norm{\mathbf{v}}$, for all $1 \le i \le j \le L$ and for all $s_1$-sparse vectors $\mathbf{u}$ and $s_2$-sparse vectors $\mathbf{v}$ with $s_1, s_2 \le \frac{m}{\rho^3}$.
		\item  $\abs{\mathbf{u}^{\top} \mathbf{W} \mathbf{D}^{(j)} \mathbf{W} \mathbf{D}^{(j-1)} \mathbf{W} \cdots \mathbf{D}^{(i+1)} \mathbf{W} \mathbf{D}^{(i)} \mathbf{W} \mathbf{v}} \le \mathcal{O}(\frac{\sqrt{s_1} \rho}{\sqrt{m}}) \cdot \norm{\mathbf{u}}\norm{\mathbf{v}}$, for all $1 \le i \le j \le L$ and for all $s_1$-sparse vectors $\mathbf{u}$,  with $s_1 \le \frac{m}{\rho^3}$, and a fixed vector $\mathbf{v}$.
		\item $\norm{\mathbf{W} \mathbf{D}^{(j)} \mathbf{W} \mathbf{D}^{(j-1)} \mathbf{W} \cdots \mathbf{D}^{(i+1)} \mathbf{W}  \mathbf{D}^{(i)} \mathbf{A}}_{\infty, \infty} \le \mathcal{O}(\frac{\rho}{\sqrt{m}})$, for all $1 \le i \le j \le L$.
	\end{enumerate}
\end{lemma}
\begin{proof}
	All of the properties except 4, 6, 9, 10 and 11 have been taken directly from Lemma B.1 and Lemma D.1 in \cite{allen2019can}. 
	\begin{enumerate}
		\item[4] The proof will follow from the proof of property 3. We outline the proof here. We have $\mathbf{W} (\mathbf{h}^{(\ell)} - \mathbf{h}^{(\ell)}_{(0)}) = \mathbf{W} \mathbf{U} \mathbf{U}^{\top} (\mathbf{h}^{(\ell)} - \mathbf{h}^{(\ell)}_{(0)})$ where $\mathbf{U}=G S\left(\mathbf{h}^{(1)}, \cdots, \mathbf{h}^{(L)}, \mathbf{h}^{(1)}_{(0)}, \cdots, \mathbf{h}^{(L)}_{(0)}\right) .$ Each entry of $\mathbf{W} \mathbf{U}$ is i.i.d. from $\mathcal{N}\left(0, \frac{2}{m}\right) .$ For any fixed $\mathbf{z}$ we have $\|\mathbf{W} \mathbf{U} \mathbf{z}\|_{\infty} \leq O(\sqrt{\rho} / \sqrt{m})$ with probability at least $1-e^{-\Omega\left(\rho^{2}\right)}$
		Taking $\epsilon$ -net over $\mathbf{z}$ and using $\left\|\mathbf{h}^{(\ell)} - \mathbf{h}^{(\ell)}_{(0)}\right\| \leq \sqrt{L}\epsilon_x$ from property 3 gives the desired bound. \footnote{GS denotes Gram-schmidt orthonormalization.}
		\item[6] The proof will follow from property 5. We will give the brief outline here. For a fixed vector $\mathbf{u}$, property 5 shows that
		\begin{equation*}
			(1 - \frac{1}{100L})^{j-i} \norm{\mathbf{D}^{(i)} \mathbf{A} \mathbf{u}} \le \norm{\mathbf{D}^{(j)} \mathbf{W} \mathbf{D}^{(j-1)} \mathbf{W} \cdots \mathbf{D}^{(i+2)} \mathbf{W} \mathbf{D}^{(i+1)} \mathbf{W} \mathbf{D}^{(i)} \mathbf{A} \mathbf{u} } \le (1 + \frac{1}{100L})^{j-i} \norm{\mathbf{D}^{(i)} \mathbf{A} \mathbf{u}}.
		\end{equation*}
		Following the proof technique of Lemma 7.1 in \cite{allen2018convergence}, we can show that with probability $1-e^{-\Omega(\rho^2)}$,
		\begin{align*}
			(1 - \mathcal{O}(\frac{\rho}{\sqrt{m}})) \norm{\mathbf{u}} \le \norm{\mathbf{D}^{(i)} \mathbf{A} \mathbf{u}} \le (1 + \mathcal{O}(\frac{\rho}{\sqrt{m}})) \norm{\mathbf{u}}. 
		\end{align*}
		Thus, assuming $m \ge \mathcal{O}(\rho^2 L^2)$ so that $\frac{\rho}{\sqrt{m}} = \mathcal{O}(\frac{1}{L})$,
		\begin{equation*}
			(1 - \frac{1}{100L})^{j-i + 1} \norm{\mathbf{u}} \le \norm{\mathbf{D}^{(j)} \mathbf{W} \mathbf{D}^{(j-1)} \mathbf{W} \cdots \mathbf{D}^{(i+2)} \mathbf{W} \mathbf{D}^{(i+1)} \mathbf{W} \mathbf{D}^{(i)} \mathbf{A} \mathbf{u}} \le (1 + \frac{1}{100L})^{j-i + 1} \norm{\mathbf{u}}.
		\end{equation*}
		The proof will follow from using an $\epsilon$-net over $\mathbb{R}^{d}$ to quantify for all vectors $\mathbf{u}$.
		\item[9] The proof will follow from Lemma B.14 in \cite{allen2019convergence_rnn}. We will give a brief overview here. Let $\mathbf{v}$ be a fixed $s_2$-sparse vector in $\mathbb{R}^{m}$.  Then, letting $\mathbf{z} =  \mathbf{D}^{(j)} \mathbf{W}  \cdots  \mathbf{D}^{(i)} \mathbf{W} \mathbf{v}$, we have w.p. $1-e^{-\Omega(m/L^2)}$ from Lemma B.12 of \cite{allen2019convergence_rnn}, $(1 - \frac{1}{100L})^{j-i + 1} \norm{\mathbf{v}} \le \norm{\mathbf{z}} \le (1 + \frac{1}{100L})^{j-i + 1} \norm{\mathbf{v}}$. 
		
		Let $\mathbf{z}^{(\ell)} =  \mathbf{D}^{(\ell)} \mathbf{W}  \cdots  \mathbf{D}^{(i)} \mathbf{W} \mathbf{v}$ for $i \le \ell \le j$. Also let $\mathbf{U}=G S\left(\mathbf{h}^{(1)}, \cdots, \mathbf{h}^{(L)}, \mathbf{z}^{(1)}, \cdots, \mathbf{z}^{(L)}\right)$.
		Each entry of $\mathbf{W} \mathbf{U}$ is i.i.d. from $\mathcal{N}\left(0, \frac{2}{m}\right).$ The dimension of $\mathbf{W}\mathbf{U}$ is $(m, j - i + 1 + L)$. Using Fact~\ref{lem:chi-squared}, we can show that for a $s_1$-sparse fixed vector $\mathbf{u}$, w.p. at least $1 - e^{-\Omega(Lt^2)}$,
		\begin{align*}
			\norm{\left(\mathbf{W}\mathbf{U}\right)^{\top} \mathbf{u}} \le \mathcal{O}(\frac{\sqrt{L}t}{\sqrt{m}} )\norm{\mathbf{u}}.
		\end{align*}
		Hence,
		\begin{align*}
			\abs{\mathbf{u}^{\top} \mathbf{W} \mathbf{D}^{(j)} \mathbf{W} \mathbf{D}^{(j-1)} \mathbf{W} \cdots \mathbf{D}^{(i+1)} \mathbf{W} \mathbf{D}^{(i)} \mathbf{W} \mathbf{v}} 
			&=  \abs{\mathbf{u}^{\top} \mathbf{W} \mathbf{U} \mathbf{U}^{\top} \mathbf{D}^{(j)} \mathbf{W} \mathbf{D}^{(j-1)} \mathbf{W} \cdots \mathbf{D}^{(i+1)} \mathbf{W} \mathbf{D}^{(i)} \mathbf{W} \mathbf{v}} 
			\\&\le \norm{\left(\mathbf{W}\mathbf{U}\right)^{\top} \mathbf{u}} \norm{\mathbf{U}^{\top} \mathbf{D}^{(j)} \mathbf{W}  \cdots  \mathbf{D}^{(i)} \mathbf{W} \mathbf{v}} \\&
			= \norm{\left(\mathbf{W}\mathbf{U}\right)^{\top} \mathbf{u}} \norm{ \mathbf{D}^{(j)} \mathbf{W}  \cdots  \mathbf{D}^{(i)} \mathbf{W} \mathbf{v}}\\&
			\le \mathcal{O}(\frac{\sqrt{L}t}{\sqrt{m}}) \norm{\mathbf{u}} \norm{\mathbf{v}}.
		\end{align*}
		
		The proof follows from setting $t = \rho \sqrt{s_1}$ and taking an $\epsilon$-net bound over all $s_2$-sparse vectors $\mathbf{v}$ and $s_1$-sparse vectors $\mathbf{u}$, that amounts to an error probability at least $1-e^{\Omega(s_2 \log m)}e^{-\Omega(m/L^2)}- e^{\Omega(s_1 \log m)}e^{-\Omega(s_1 \rho^2)}$ , which simplifies to atleast $1-e^{-\Omega(\rho^2)}$, since $s_1, s_2 \le \frac{m}{\rho^3}$.
		\item[10] The proof will follow the same technique used for property 9. The only difference is that $\mathbf{v}$ will be fixed and hence, no $\epsilon$-net is necessary for the vector $\mathbf{v}$.
		\item[11] The proof will follow the same technique used for property 9. $\mathbf{u}$ will be chosen from $\mathbf{e}_1, \cdots, \mathbf{e}_m$ and $\mathbf{v}$ will be chosen from the set of vectors $\left\{ \mathbf{D}^{(i)} \mathbf{A}\mathbf{e}_1, \cdots, \mathbf{D}^{(i)} \mathbf{A}\mathbf{e}_d \right\}$. Thus, the union bound over $\mathbf{u}$ and $\mathbf{v}$ needs to consider only $m$ and $d$ vectors respectively, in place of the $\epsilon$-net over $s_1$ and $s_2$ sparse vectors.  
	\end{enumerate}
	
\end{proof}
The following lemma shows that the hidden states at initialization are resilient to re-randomization of few rows of the gaussian matrices $\bW$ and $\mathbf{A}$. The proof again follows from applying concentration bounds w.r.t. the new set of weights. This lemma is used multiple times later to break the correlations among different functions of  $\bW$ and $\mathbf{A}$.

\begin{lemma}[Stability after re-randomization, Lemma E.1 in \cite{allen2019can} ]\label{lemma:rerandESN}
	Consider a fixed set $\mathcal{K} \subseteq[m]$ with cardinality $N=\abs{\mathcal{K}}$. Consider the following matrices.
	\begin{itemize}
		\item $\widetilde{\mathbf{W}} \in \mathbb{R}^{m \times m} \text { where } \widetilde{\mathbf{w}}_{k}=\mathbf{w}_{k} \text { for } k \in [m] \backslash \mathcal{K} \text { but } \widetilde{\mathbf{w}}_{k} \sim \mathcal{N}\left(0, \frac{2 \mathbf{I}}{m}\right) \text { is i.i.d. for } k \in \mathcal{K}$
		\item $\widetilde{\mathbf{A}} \in \mathbb{R}^{m \times d} \text { where } \widetilde{\mathbf{a}}_{k}=\mathbf{a}_{k} \text { for } k \in[m] \backslash \mathcal{K} \text { but } \widetilde{\mathbf{a}}_{k} \sim \mathcal{N}\left(0, \frac{2 \mathbf{I}}{m}\right) \text { is i.i.d. for } k \in \mathcal{K}$
	\end{itemize}
	
	For any \todo{ALL use "normalized" sequences which are different. We need to clearly point out the differences in the statement and proof because of this.} normalized input sequence $\mathbf{x}^{(1)}, \ldots, \mathbf{x}^{(L)} \in \mathbb{S}^{d-1},$ we consider the following two executions of ESNs under $\mathbf{W}$ and $\widetilde{\mathbf{W}}$ respectively:
	\[
	\begin{array}{lrl}
		\mathbf{g}^{(0)}=\mathbf{h}^{(0)}=0 & \mathbf{g}^{(0)\prime}=\mathbf{h}^{(0)\prime}=0 \\
		\mathbf{g}^{(\ell)}=\mathbf{W} \mathbf{h}^{(\ell-1)}+\mathbf{A} \mathbf{x}^{(\ell)} & \tilde{\mathbf{g}}^{(\ell)} = \mathbf{g}^{(\ell)}+\mathbf{g}^{(\ell)\prime}=\widetilde{\mathbf{W}}\left(\mathbf{h}^{(\ell-1)}+\mathbf{h}^{(\ell-1)\prime}\right)+\widetilde{\mathbf{A}} \mathbf{x}^{(\ell)} \\
		\mathbf{h}^{(\ell)}=\sigma\left(\mathbf{W} \mathbf{h}^{(\ell-1)}+\mathbf{A} \mathbf{x}^{(\ell)}\right) & \tilde{\mathbf{h}}^{(\ell)} = \mathbf{h}^{(\ell)}+\mathbf{h}^{(\ell)\prime}=\sigma\left(\widetilde{\mathbf{W}}\left(\mathbf{h}^{(\ell-1)}+\mathbf{h}^{(\ell-1)\prime}\right)+\widetilde{\mathbf{A}} \mathbf{x}^{(\ell)}\right) & \text { for } \ell \in [L]
	\end{array}
	\]
	and define diagonal sign matrices $\mathbf{D}^{(\ell)} \in\{0,1\}^{m \times m}$ and $\widetilde{\mathbf{D}}^{(\ell)} = \mathbf{D}^{(\ell)}+\mathbf{D}^{(\ell)\prime} \in\{0,1\}^{m \times m}$ by letting
	$$
	d^{(\ell)}_{k, k}=\mathbb{I}_{g^{(\ell)}_{k} \geq 0} \text { and }\widetilde{d}^{(\ell)}_{k, k}=\mathbb{I}_{\widetilde{g}^{(\ell)}_{k} \geq 0}
	$$
	
	Let $N=\abs{\mathcal{K}} \leq m / \rho^{23}$. Fix any normalized input sequence $\mathbf{x}^{(1)}, \ldots, \mathbf{x}^{(L)}$. We
	have, with probability at least $1-e^{-\Omega\left(\rho^{2}\right)}$ over the randomness of $\mathbf{W}$, $\widetilde{\mathbf{W}}$, $\mathbf{A}$, $\widetilde{\mathbf{A}}$,
	\begin{enumerate}
		\item $\left\|\mathbf{g}^{(\ell)\prime}\right\|,\left\|\mathbf{h}^{(\ell)\prime}\right\| \leq \mathcal{O}\left(\rho^{5} \sqrt{N / m}\right) \quad$ for every $\ell \in[L].$
		\item $\left|\left\langle \mathbf{w}_{k}, \mathbf{h}^{(\ell)\prime}\right\rangle\right| \leq \mathcal{O}\left(\rho^{5} N^{2 / 3} m^{-2 / 3}\right) \quad \text { for every } k \in [m], \ell \in[L].$
		\item $\norm{ \mathbf{W}_{\mathcal{K}} \mathbf{h}^{(\ell)\prime}}\leq \mathcal{O}\left(\rho^{5} N^{2 / 3} m^{-2 / 3}\right) \quad \text { for every }  \ell \in[L].$
		\item $\norm{\mathbf{D}^{(\ell)} - \widetilde{\mathbf{D}}^{(\ell)}}_0 \le \mathcal{O}(\rho^4 N^{1/3} m^{2/3})$, for all $\ell \in [L]$.
		\item $\norm{ \left(\prod_{i \le \ell' \le j} \widetilde{\mathbf{D}}^{(k - \ell' + 1)} \widetilde{\mathbf{W}}  - \prod_{i \le \ell' \le j} \mathbf{D}^{(k - \ell' + 1)} \mathbf{W} \right) \mathbf{v}}_{2} \le  \mathcal{O}(\rho^5 (N/m)^{1/6}) \norm{\mathbf{v}}$, for all $1 \le i \le j \le L$ and for a fixed vector $\mathbf{v}$.
		\item $\norm{\mathbf{W}_{\mathcal{K}} \left(\prod_{i \le \ell' \le j} \widetilde{\mathbf{D}}^{(k - \ell' + 1)} \widetilde{\mathbf{W}}  - \prod_{i \le \ell' \le j} \mathbf{D}^{(k - \ell' + 1)} \mathbf{W}\right) \mathbf{v}}_2 \le  \mathcal{O}(\rho^6 (N/m)^{2/3})$, for all $1 \le i \le j \le L$  and for a fixed vector $\mathbf{v}$.
	\end{enumerate}
\end{lemma}


\begin{proof}
	All the properties except 3, 5 and 6 follow from Lemma E.1 in \cite{allen2019can}. 
	\begin{enumerate}
		\item[3] The proof will follow the same technique as used for property 2. We give a brief overview here. We follow the same technique to expand the desired term
		\begin{align*}
			\mathbf{W}_{\mathcal{K}} \mathbf{h}^{(\ell)\prime} = \mathbf{D}_{\mathcal{K}} \mathbf{W} \mathbf{D}^{(\ell)\prime} (\mathbf{g}^{(\ell)} + \mathbf{g}^{(\ell)\prime}) + \mathbf{D}_{\mathcal{K}} \mathbf{W} \mathbf{D}^{(\ell)}  \mathbf{g}^{(\ell)\prime}
		\end{align*}
		We bound both the terms using the same technique with the following difference: 
		we use property 9 of Lemma~\ref{lemma:norm_ESN} to bound the terms $\norm{\mathbf{D}_{\mathcal{K}} \mathbf{W} \mathbf{D}^{(\ell)\prime}}$ and $\norm{\mathbf{D}_{\mathcal{K}} \mathbf{W} \mathbf{D}^{(\ell)}  \mathbf{g}^{(\ell)\prime}}$.
		
		\item[5] The proof follows from the proof of Lemma E.1(4) in \cite{allen2019can}. In the proof of lemma E.1(4), the term that has been bounded is
		\begin{align*}
			\norm{ \left(\prod_{i \le \ell' \le j} \widetilde{\mathbf{D}}^{(k - \ell' + 1)} \widetilde{\mathbf{W}}  - \prod_{i \le \ell' \le j} \mathbf{D}^{(k - \ell' + 1)} \mathbf{W} \right) \mathbf{e}_k}_{2}, \text{where } k \in [m].
		\end{align*}
		The important property of the vectors $\mathbf{e}_k$ that is used to bound the term above is the $1$-sparsity of the vectors, which is necessary for using a property similar to property 9 of lemma~\ref{lemma:norm_ESN}.
		However, we can show that the same bound holds for a fixed vector $\mathbf{v}$ by bounding the terms that contain $\mathbf{v}$ using property 10 of lemma~\ref{lemma:norm_ESN}.

		\item[6] The proof will follow the same technique as used for property 5. We give a brief overview here. 
		
		For property 5, the term under consideration, $\left(\prod_{i \le \ell' \le j} \widetilde{\mathbf{D}}^{(k - \ell' + 1)} \widetilde{\mathbf{W}}  - \prod_{i \le \ell' \le j} \mathbf{D}^{(k - \ell' + 1)} \mathbf{W} \right) \mathbf{v}$, was expanded into all the (exponentially many) difference
		terms, which were bounded separately. Denote the difference terms as $\mathbf{T}_1, \mathbf{T}_2, \cdots$. 
		
		For each term $\mathbf{T}_i$, the product $\mathbf{W}_{\mathcal{K}} \mathbf{T}_i$ can be written as a product of $\mathbf{D}_{\mathcal{K}} \mathbf{W} (\prod_{\ell_1 \le \ell \le \ell_2 } \mathbf{D}^{(\ell)} \mathbf{W}) \overline{\mathbf{D}}$ and a term $\overline{\mathbf{T}}_i$, for some $i \le \ell_1, \ell_2 \le j$ and $\overline{\mathbf{D}}$ is either $\mathbf{D}_{\mathcal{K}}$ or $\mathbf{D}^{(\ell)} - \mathbf{D}^{(\ell)}_{(0)}$. The term $\overline{\mathbf{T}}_i$ will be bounded in a similar manner as has been done in the proof of property 5. However, the extra term that appears will be the bound of the norm of $\mathbf{D}_{\mathcal{K}} \mathbf{W} (\prod_{\ell_1 \le \ell \le \ell_2 } \mathbf{D}^{(\ell)} \mathbf{W}) \overline{\mathbf{D}}$, which is bounded by $\mathcal{O}(\rho \sqrt{N/m})$ using property 9 of lemma~\ref{lemma:norm_ESN}. 
		
		We will give an example for a term $\mathbf{T}_i$. Few terms will be of the form \begin{equation*}
			(\prod_{\ell_1 \le \ell \le \ell_2 } \mathbf{D}^{(\ell)} \mathbf{W}) \cdot \mathbf{D}^{(\ell_2)\prime} \mathbf{W}  \cdot (\prod_{\ell_2 < \ell \le \ell_3 } \mathbf{D}^{(\ell)} \mathbf{W}) \mathbf{v}, 
		\end{equation*}
		for some $\ell_1, \ell_2, \ell_3$. We break its product with $\mathbf{W}_{\mathcal{K}}$ as
		\begin{align*}
			&\mathbf{W}_{\mathcal{K}} \cdot (\prod_{\ell_1 \le \ell \le \ell_2 } \mathbf{D}^{(\ell)} \mathbf{W}) \cdot \mathbf{D}^{(\ell_2)\prime} \mathbf{W}  \cdot (\prod_{\ell_2 < \ell \le \ell_3 } \mathbf{D}^{(\ell)} \mathbf{W}) \mathbf{v} \\&=
			\underbrace{\left(\mathcal{D}_{\mathcal{K}} \mathbf{W} \cdot (\prod_{\ell_1 \le \ell \le \ell_2 } \mathbf{D}^{(\ell)} \mathbf{W}) \cdot \mathbf{D}^{(\ell_2)\prime} \right)}_{\text{Term 1}} \cdot \underbrace{\left(\mathbf{D}^{(\ell_2)\prime} \cdot (\prod_{\ell_2 < \ell \le \ell_3 } \mathbf{D}^{(\ell)} \mathbf{W}) \mathbf{v}\right)}_{\text{Term 2}}. 
		\end{align*}
		Term 2 appears in the proof of property 5. Term 1 is the extra term that needs to be bounded and we can use property 9 of lemma~\ref{lemma:norm_ESN} to bound its norm by $\mathcal{O}(\rho \sqrt{N/m})$. 
	\end{enumerate}
\end{proof}

\section{Invertibility at a single step}\label{sec:singlecell_RNN}
The section has been structured as follows: we first prove that a linear transformation of a random $\relu$ network can give back a linear function of the input in lemma~\ref{lem:normal_linearestimate}. We then explain why a simple application of the above lemma doesn't give a similar lemma for random RNNs which is, we need to make sure we break the correlations among input, the output vector and the weight matrices. We show that such correlations can be broken using the arguments in Claims \ref{claim:ff}, \ref{claim:fg}, \ref{claim:ffv} and \ref{claim:gg}. This then helps us to prove lemma~\ref{lemma:singlecell_ESN} using an application of lemma~\ref{lem:normal_linearestimate}.

\subsection{Invertibility of one layer $\relu$ networks}
The following lemma is from a companion paper; we reproduce its proof here for completeness. 
\begin{lemma}\label{lem:normal_linearestimate}
	For any $\bv \in \Reals^d$, the linear function taking $\bx$ to $\bv^\top \bx$ for $\bx \in \Reals^d$, can be represented as
	\begin{equation}\label{eqn:single_layer_inversion_exact}
		\bv^\top \bx = \int_{\Reals^d} p(\bw) \, \sigma(\mathbf{w}^{\top} \bx ) \,\deriv \mu_d (\mathbf{w}),
	\end{equation}
	with 
	$$p\left(\mathbf{w}\right) \;=\;  2 \, \mathbf{w}^{\top} \mathbf{v}.$$ 
\end{lemma}
\begin{remark}
	A similar statement can be gleaned from the proof of Proposition 4 of \cite{bach2017breaking} which gives a similar representation except that 
	$\bw$ is uniformly distributed on $\Sphere^{d-1}$ instead of being Gaussian. The proof there makes use of spherical harmonics and does not seem to immediately apply to the Gaussian case. The proof below is elementary and can be easily adapted to any spherically-symmetric distribution. 
\end{remark}

\begin{proof}
	In the following, we re-parametrize $\mathbf{w}$ as $r \overline{\mathbf{w}}$ for some $r \ge 0, \overline{\mathbf{w}} \in \mathbb{S}^{d-1}$. 
	\begingroup \allowdisplaybreaks
	\begin{align}
		\frac{1}{2}
		\int_{\Reals^d} p(\bw) \,\sigma(\mathbf{w}^{\top} \bx ) \deriv \mu_d (\mathbf{w}) 
		&= \int_{\mathbf{w} \in \mathbb{R}^{d}}  \mathbf{v}^{\top} \mathbf{w}  \left(\mathbf{w}^{\top} \mathbf{x} \right) 
		\mathbb{I}_{\left(\mathbf{w}^{\top} \mathbf{x} \right) \ge 0} \deriv \mu \left(\mathbf{w}\right) \nonumber
		\\&=  \mathbf{v}^{\top} \left(\int_{\mathbf{w} \in \mathbb{R}^{d}} \mathbf{w} \left(\mathbf{w}^{\top} \mathbf{x} \right) \mathbb{I}_{\left(\mathbf{w}^{\top} \mathbf{x} \right) \ge 0} \deriv\mu \left(\mathbf{w}\right) \right) \nonumber
		\\& = \mathbf{v}^{\top} \left( \frac{1}{(\sqrt{2\pi})^{d}} \int_{\overline{\mathbf{w}} \in \mathbb{S}^{d-1}} \int_{r=0}^{\infty}  r \overline{\mathbf{w}} \left(r  \overline{\mathbf{w}}^{\top} \mathbf{x} \right) \mathbb{I}_{\left(r \overline{\mathbf{w}}^{\top} \mathbf{x} \right) \ge 0} r^{d-1}e^{-r^2/2} \deriv r \deriv \overline{\mathbf{w}} \right) \nonumber
		\\& = \left(\frac{1}{(\sqrt{2\pi})^{d}} \int_{r=0}^{\infty} r^{d+1} e^{-r^2/2} \deriv r \right)  \mathbf{v}^{\top} 
		\left(   \int_{\overline{\mathbf{w}} \in \mathbb{S}^{d-1}}  \overline{\mathbf{w}} \left(\overline{\mathbf{w}}^{\top} \mathbf{x} \right)  \mathbb{I}_{\left(\overline{\mathbf{w}}^{\top} \mathbf{x} \right) \ge 0}  \deriv \overline{\mathbf{w}} \right) \nonumber
		\\& = \left(\frac{1}{(\sqrt{2\pi})^{d}} 2^{d/2}\, \Gamma(d/2 +1) \right)  \mathbf{v}^{\top} 
		\left(   \int_{\overline{\mathbf{w}} \in \mathbb{S}^{d-1}}  \overline{\mathbf{w}} \left(\overline{\mathbf{w}}^{\top} \mathbf{x} \right)  \mathbb{I}_{\left(\overline{\mathbf{w}}^{\top} \mathbf{x} \right) \ge 0}  \deriv \overline{\mathbf{w}} \right) \nonumber
		\\&=  \frac{d}{\abs[0]{\Sphere^{d-1}}} \mathbf{v}^{\top} \left(   \int_{\overline{\mathbf{w}} \in \mathbb{S}^{d-1}} \overline{\mathbf{w}} \overline{\mathbf{w}}^{\top} \mathbb{I}_{\left(\overline{\mathbf{w}}^{\top} \mathbf{x} \right) \ge 0}  \deriv \overline{\mathbf{w}} \right) \mathbf{x} \nonumber
		\\& =  \frac{d}{\abs[0]{\Sphere^{d-1}}}  \mathbf{v}^{\top} \mathbf{C}_\bx \mathbf{x},   \label{eqn:pw_computation}
	\end{align}
	\endgroup
	where $\mathbf{C}_\bx :=  \int_{\overline{\mathbf{w}} \in \mathbb{S}^{d-1}} \overline{\mathbf{w}} \overline{\mathbf{w}}^{\top} \mathbb{I}_{\left(\overline{\mathbf{w}}^{\top} \mathbf{x} \right) \ge 0}  \deriv \overline{\mathbf{w}}$.
	Let the orthogonal matrix $\mathbf{U}_\bx$ be such that $\mathbf{U}_\bx^\top \mathbf{x} = \mathbf{e}_1$ (the choice is not unique; we choose one arbitrarily).
	Then 
	\begin{align}
		\mathbf{C}_\bx 
		&=  \int_{\overline{\mathbf{w}} \in \mathbb{S}^{d-1}} \overline{\mathbf{w}} \overline{\mathbf{w}}^{\top} \mathbb{I}_{\left(\overline{\mathbf{w}}^{\top} \mathbf{x} \right) \ge 0}  \deriv \overline{\mathbf{w}} \nonumber
		\\&= \int_{\overline{\mathbf{w}} \in \mathbb{S}^{d-1}} \mathbf{U}_\bx \overline{\mathbf{w}} (\mathbf{U}_\bx\overline{\mathbf{w}})^\top \mathbb{I}_{(\mathbf{U}_\bx \overline{\mathbf{w}})^{\top} \mathbf{x}  \ge 0} \deriv \overline{\mathbf{w}} \nonumber
		\\&= \mathbf{U}_\bx \left( \int_{\overline{\mathbf{w}} \in \mathbb{S}^{d-1}} \overline{\mathbf{w}} \overline{\mathbf{w}}^\top \mathbb{I}_{\overline{\mathbf{w}}^{\top} (\mathbf{U}_\bx^\top \mathbf{x}) \ge 0}  \deriv \overline{\mathbf{w}} \right) \mathbf{U}_\bx^\top \nonumber
		\\&= \mathbf{U}_\bx \left( \int_{\overline{\mathbf{w}} \in \mathbb{S}^{d-1}} \overline{\mathbf{w}} \overline{\mathbf{w}}^\top 
		\mathbb{I}_{\overline{\mathbf{w}}_1 \ge 0}  \deriv \overline{\mathbf{w}} \right) \mathbf{U}_\bx^\top \nonumber
		\\&= \mathbf{U}_\bx \mathbf{C}_{\mathbf{e}_1} \mathbf{U}_\bx^\top.  \label{eqn:Cx}
	\end{align}
	Using the symmetry of $\Sphere^{d-1}$ we claim
	\begin{claim}\label{claim:Kd}
		We have $\mathbf{C}_{\mathbf{e}_1} = K_d \mathbf{I}$, for a constant $K_d$ (evaluated below). 
	\end{claim}
	\begin{proof}
		Let's first restate the claim:      
		\begin{align*}
			[\mathbf{C}_{\mathbf{e}_1}]_{i,j} = \int_{\overline{\mathbf{w}} \in \mathbb{S}^{d-1}} \overline{w}_i \overline{w}_j
			\,\mathbb{I}_{\overline{w}_1 \ge 0}  \deriv \overline{\mathbf{w}} 
			= \begin{cases} 
				0, & \text{if } i \neq j, \\
				K_d, & \text{otherwise.}
			\end{cases}
		\end{align*}
		To prove this, note that 
		\begin{align*}
			\int_{\overline{\mathbf{w}} \in \mathbb{S}^{d-1}} \overline{w}_1^2 \,\mathbb{I}_{\overline{w}_1 \ge 0}  \deriv \overline{\mathbf{w}}
			= \frac{1}{2}\int_{\overline{\mathbf{w}} \in \mathbb{S}^{d-1}} \overline{w}_1^2   \deriv \overline{\mathbf{w}},
		\end{align*}
		because $\overline{w}_1^2$ takes on the same value on $(\overline{w}_1, \overline{w}_2, \overline{w}_3, \ldots)$ and on $(-\overline{w}_1, \overline{w}_2, \overline{w}_3, \ldots)$.
		Similarly
		\begin{align*}
			\int_{\overline{\mathbf{w}} \in \mathbb{S}^{d-1}} \overline{w}_2^2 \,\mathbb{I}_{\overline{w}_1 \ge 0}  \deriv \overline{\mathbf{w}}
			= \frac{1}{2}\int_{\overline{\mathbf{w}} \in \mathbb{S}^{d-1}} \overline{w}_2^2   \deriv \overline{\mathbf{w}},
		\end{align*}
		because $\overline{w}_2^2$ takes on the same value on $(\overline{w}_1, \overline{w}_2, \overline{w}_3, \ldots)$ and on $(-\overline{w}_1, \overline{w}_2, \overline{w}_3, \ldots)$.
		Now clearly 
		\begin{align*}
			\int_{\overline{\mathbf{w}} \in \mathbb{S}^{d-1}} \overline{w}_1^2   \deriv \overline{\mathbf{w}}= 
			\int_{\overline{\mathbf{w}} \in \mathbb{S}^{d-1}} \overline{w}_2^2   \deriv \overline{\mathbf{w}}.
		\end{align*}
		Thus we have shown that $\int_{\overline{\mathbf{w}} \in \mathbb{S}^{d-1}} \overline{w}_i^2 \,\mathbb{I}_{\overline{w}_1 \ge 0}  \deriv \overline{\mathbf{w}}$ does not depend on $i$. 
		Now notice that  
		\begin{align*}
			\int_{\overline{\mathbf{w}} \in \mathbb{S}^{d-1}} \overline{w}_i \overline{w}_j \,\mathbb{I}_{\overline{w}_1 \ge 0}  \deriv \overline{\mathbf{w}} = 0
		\end{align*}
		because for each point $(\overline{w}_1, \overline{w}_2, \overline{w}_3, \ldots)$ there's a corresponding point $(\overline{w}_1, -\overline{w}_2, \overline{w}_3, \ldots)$ 
		with the integrands taking on opposite values (or both are $0$). A similar argument shows more generally that for all $i \neq j$ we have $\int_{\overline{\mathbf{w}} \in \mathbb{S}^{d-1}} \overline{w}_i \overline{w}_j
		\,\mathbb{I}_{\overline{w}_1 \ge 0}  \deriv \overline{\mathbf{w}} = 0$. 
	\end{proof}
	
	Now we evaluate $K_d$. Using Claim~\ref{claim:Kd}, we can write
	\begin{align*}
		K_d = \frac{1}{d} \tr \mathbf{C}_{\mathbf{e}_1}
		= \frac{1}{d} \int_{\overline{\mathbf{w}} \in \mathbb{S}^{d-1}} (\tr \overline{\mathbf{w}} \overline{\mathbf{w}}^{\top})\, \mathbb{I}_{\overline{w}_1 \ge 0}  \deriv \overline{\mathbf{w}} 
		= \frac{1}{d} \int_{\overline{\mathbf{w}} \in \mathbb{S}^{d-1}} \mathbb{I}_{\overline{w}_1 \ge 0}  \deriv \overline{\mathbf{w}} 
		= \frac{\abs[0]{\Sphere^{d-1}}}{2d}. 
	\end{align*}
	Using the orthogonality of $\mathbf{U}_\bx$ and Claim~\ref{claim:Kd} in \eqref{eqn:Cx} it follows that 
	\begin{align*}
		\mathbf{C}_\bx = \mathbf{U}_\bx \mathbf{C}_{\mathbf{e}_1} \mathbf{U}_\bx^\top = K_d \mathbf{U}_\bx \mathbf{I} \mathbf{U}_\bx^\top = K_d \mathbf{I}. 
	\end{align*}
	Continuing from where we left off in \eqref{eqn:pw_computation} we have
	\begin{align*}
		\frac{1}{2} \int_{\Reals^d} p(\bw) \,\sigma(\mathbf{w}^{\top} \bx ) \deriv \mu_d (\mathbf{w}) 
		= \frac{d}{\abs{\Sphere^{d-1}}}  \mathbf{v}^{\top} \mathbf{C}_\bx \mathbf{x} 
		=  \frac{d K_d}{\abs{\Sphere^{d-1}}} \mathbf{v}^{\top} \mathbf{x} 
		= \frac{1}{2} \mathbf{v}^{\top} \mathbf{x}.
	\end{align*}
	Thus, 
	\begin{equation*}
		p\left(\mathbf{w}\right) = 2 \,\mathbf{w}^{\top} \mathbf{v}. 
	\end{equation*}
\end{proof}
Lemma~\ref{lem:normal_linearestimate} can be extended to gaussian distribution over $\mathbf{w}$ with variance $\beta \mathbf{I}$, for any $\beta > 0$.
\begin{corollary}\label{cor:linear_estimate}
	For any $\bv \in \Reals^d$, the linear function taking $\bx$ to $\bv^\top \bx$ for $\bx \in \Reals^d$, can be represented as
	\begin{equation}\label{eqn:single_layer_inversion_exact_cor}
		\frac{\beta^2}{2} \, \bv^\top \bx = \int_{\Reals^d} p(\bw) \, \sigma(\mathbf{w}^{\top} \bx ) \,\deriv \mu_d^{\beta} (\mathbf{w}),
	\end{equation}
	with 
	$$p\left(\mathbf{w}\right) \;=\;   \mathbf{w}^{\top} \mathbf{v}.$$ 
\end{corollary}

Lemma~\ref{lem:normal_linearestimate} can be discretized so that instead of the integral in \eqref{eqn:single_layer_inversion_exact}, we use an empirical average.
This comes at the expense of making the resulting version of \eqref{eqn:single_layer_inversion_exact} approximate. Furthermore, we can generalize the lemma so that instead of taking us from $\mathbf{h}^{(1)} = \sigma(\mathbf{W}\mathbf{x}^{(1)})$ to $\bv^\top \bx^{(1)}$ it takes us from $\mathbf{h}^{(\ell)}$ to  $\bv^\top [\mathbf{h}^{(\ell-1)}, \bx^{(\ell)}]$ for every $\ell \in [L]$. The following lemma does both of these.


\subsection{Invertibility at a single step of RNN}
\begin{lemma}\label{lemma:singlecell_ESN}
	We have an RNN at random initialization as defined in Def.~\ref{def:RNN}. Fix any $\ell \in\{0,1, \ldots, L-$ 1\} and $\zeta \in(0,1)$. Let $g: \mathbb{R}^{m+d } \times \mathbb{R}^{m+d}  \rightarrow \mathbb{R}$ be given by $g(\mathbf{v}, [\mathbf{h}, \mathbf{x}]) = \mathbf{v}^{\top} [\mathbf{h}, \mathbf{x}]$. Consider a vector $\mathbf{v} \in \mathbb{R}^{m+d}$ which is stable against re-randomization, as specified later in Assumption~\ref{ass:variant_v} with constants $(\kappa, \zeta)$.
	Let $f\left(\mathbf{v}, \mathbf{h}^{(\ell-1)}, \mathbf{x}^{(\ell)}\right) = \sum_{i=1}^{m} u_i \sigma\left(\mathbf{w}_i^{\top} \mathbf{h}^{(\ell-1)} +  \mathbf{a}_i^{\top} \mathbf{x}^{(\ell)} \right),$ where 
	$$ u_i = [\mathbf{w}_i, \mathbf{a}_i]^{\top} \mathbf{v}.$$
	
	Then for a given normalized sequence $\mathbf{x}^{(1)}, \cdots, \mathbf{x}^{(L)}$, with $\mathbf{x}^{(\ell)} \in \mathbb{R}^{d}$ for each $\ell \in [L]$, and for any constant $\rho > 0$, we have 
	\begin{equation*} \abs[0]{g({\mathbf{v}}, [\mathbf{h}^{(\ell-1)}, \mathbf{x}^{(\ell)}]) - f(\mathbf{v}, \mathbf{h}^{(\ell-1)}, \mathbf{x}^{(\ell)})} \le \mathcal{O}\left(\rho^{5 + \kappa} m^{-1/12} + \rho^{1+\kappa} m^{-\zeta/2} + \rho^{1+\kappa} m^{-1/4} + \rho^{5+\kappa} m^{-1/4}  \right) \cdot \norm{\mathbf{v}},
	\end{equation*}
	with probability at least $1 -  e^{-\Omega(\rho^2)}$.
	
\end{lemma}

\begin{proof}
	The major issue in using a discrete version of lemma~\ref{lem:normal_linearestimate} directly for input $[\mathbf{h}^{(\ell-1)}, \mathbf{x}]$ is that there is a coupling between the randomness of the weights $\mathbf{W}, \mathbf{A}$ and the hidden vector $\mathbf{h}^{(\ell-1)}$. This coupling can be understood as the dependence of $\mathbf{h}^{(\ell-1)}$ on the choice of weight vectors in $\mathbf{W}$ and $\mathbf{A}$. There may also be a coupling between  the randomness of $\mathbf{v}$ and  the weights $\mathbf{W}, \mathbf{A}$, for which we take some assumption later. To decouple this randomness, we use the fact that ESNs are stable to re-randomization of few rows of the weight matrices and follow the proof technique of Lemma G.3 \cite{allen2019can}.

	Choose a random subset $\mathcal{K} \subset[m]$ of size $|\mathcal{K}|=N$. Define the function $f_{\mathcal{K}}$ as
	\begin{equation*}
		f_{\mathcal{K}}(\mathbf{v}, \mathbf{h}^{(\ell-1)}, \mathbf{x}^{(\ell)}) = \sum_{k \in \mathcal{K}} u_{k} \sigma(\mathbf{w}_k^{\top} \mathbf{h}^{(\ell-1)} + \mathbf{a}_k^{\top} \mathbf{x}^{(\ell)}).  
	\end{equation*}
	Replace the rows $\left\{\mathbf{w}_{k}, \mathbf{a}_{k}\right\}_{k \in \mathcal{K}}$ of $\mathbf{W}$ and $\mathbf{A}$ with freshly new i.i.d. samples $\widetilde{\mathbf{w}}_{k}, \widetilde{\mathbf{a}}_{k} \sim \mathcal{N}\left(0, \frac{2}{m} \mathbf{I}\right).$ to form new matrices $\widetilde{\mathbf{W}}$ and $\widetilde{\mathbf{A}}$. For the given sequence, we follow the notation of Lemma~\ref{lemma:rerandESN} to denote the hidden states corresponding to the old and the new weight matrices. We will assume one property for $\mathbf{v}$. Let say $\mathbf{v}$ depends on the matrices $\mathbf{W}$ and $\mathbf{A}$ and becomes $\widetilde{\mathbf{v}}$, with the new matrices $\widetilde{\mathbf{W}}$ and $\widetilde{\mathbf{A}}$. Then, we assume that the norm difference of $\mathbf{v}$ and $\widetilde{\mathbf{v}}$ is small with high probability.
	\begin{assumption}\label{ass:variant_v}
		With probability at least $1-e^{-\Omega(\rho^2)}$, \todo{clarify which random variables are used for this probability} there exists constants $\kappa \ge 0$ and $\zeta < 1$ such that
		\begin{align*}
			&\norm[2]{\mathbf{v} - \widetilde{\mathbf{v}}} \le \mathcal{O}(\rho^{\kappa} (N/m)^{\zeta} \norm{\mathbf{v}}) \\&
			\norm{[\mathbf{W}_{\mathcal{K}}, \mathbf{A}_{\mathcal{K}}]_r \left(\mathbf{v} - \widetilde{\mathbf{v}}\right)} \le \mathcal{O}(\rho^{\kappa} (N/m)^{0.5 + \zeta} \norm{\mathbf{v}}), \quad \forall k \in [m].
		\end{align*}
	\end{assumption}
	We will show later that the vector $\mathbf{v}$ that we need for inversion satisfies the above assumption with constants $(\kappa, \zeta) = (6, 1/6)$. Also, if $\mathbf{v}$ is independent of  $\mathbf{W}$ and $\mathbf{A}$, then the constants needed in the assumption are $(\kappa, \zeta) = (0, 0)$.

	The following claim shows that under the assumption~\ref{ass:variant_v}, function $f_{\mathcal{K}}$ and $\frac{N}{m}g$ are close to each other with high probability.
	
	\begin{claim}\label{claim:singlesubset}
		For the given sequence $\mathbf{x}^{(1)}, \cdots, \mathbf{x}^{(L)}$,
		\begin{align*}
			&\abs{f_{\mathcal{K}}(\mathbf{v}, \mathbf{h}^{(\ell-1)}, \mathbf{x}^{(\ell)}) - \frac{N}{m}g(\mathbf{v}, [\mathbf{h}^{(\ell-1)}, \mathbf{x}^{(\ell)}])} \\&\le  \mathcal{O}\left(\rho^{5 + \kappa} N^{7/6} m^{-7/6} + \rho^{1 + \kappa} (N/m)^{1+\zeta} + \rho^{1+\kappa} N^{1/2} m^{-1}  + \rho^{5 + \kappa} (N/m)^{3/2}\right) \cdot \norm{\mathbf{v}},
		\end{align*}
		with probability exceeding $1 - e^{-\Omega(\rho^2)}$.
	\end{claim}
	The above claim has been restated and proven in claim~\ref{claim:singlesubset_proof}.
	
	To complete the proof, we divide the set of neurons into $m/N$ disjoint sets $\mathcal{K}_1, \cdots, \mathcal{K}_{m/N}$, each set is of size $N$. We apply the Claim~\ref{claim:singlesubset} to each subset $\mathcal{K}_i$ and then add up the errors from each subset. That is, with probability at least $1 - \frac{m}{N}e^{-\Omega(\rho^2)}$,
	\begin{align*}
		f(\mathbf{h}^{(\ell-1)}, \mathbf{x}^{(\ell)}) &=  \sum_{i=1}^{m/N}  f_{\mathcal{K}_i}(\mathbf{h}^{(\ell-1)}, \mathbf{x}^{(\ell)})\\
		&= \sum_{i=1}^{m/N} \frac{N}{m} g([\mathbf{h}^{(\ell-1)}, \mathbf{x}^{(\ell)}]) + error_{\mathcal{K}_i}\\
		&= g([\mathbf{h}^{(\ell-1)}, \mathbf{x}^{(\ell)}]) + \sum_{i=1}^{m/N} error_{\mathcal{K}_i}, 
	\end{align*}
	where by Claim~\ref{claim:singlesubset},
	\begin{align*}
		\abs{error_{\mathcal{K}_i}} \le \mathcal{O}\left(\rho^{5 + \kappa} N^{7/6} m^{-7/6} + \rho^{1 + \kappa} (N/m)^{1+\zeta} + \rho^{1+\kappa} N^{1/2} m^{-1}  + \rho^{5 + \kappa} (N/m)^{3/2}\right) \cdot \norm{\mathbf{v}}.
	\end{align*}
	
	Thus,
	\begin{equation*}
		\abs{ f(\mathbf{h}^{(\ell-1)}, \mathbf{x}^{(\ell)}) - g([\mathbf{h}^{(\ell-1)}, \mathbf{x}^{(\ell)}])} \le \mathcal{O}\left(\rho^{6 + \kappa} N^{1/6} m^{-1/6} + \rho^{2 + \kappa} (N/m)^{\zeta} + \rho^{1+\kappa} N^{-1/2} + \rho^{5 + \kappa} (N/m)^{1/2}\right) \cdot \norm{\mathbf{v}},
	\end{equation*}
	with probability at least $1 - \frac{m}{N}e^{-\Omega(\rho^2)}$.
	
	Choosing $N = m^{1/2}$, we have
	\begin{equation*}
		\abs{ f(\mathbf{h}^{(\ell-1)}, \mathbf{x}^{(\ell)}) - g([\mathbf{h}^{(\ell-1)}, \mathbf{x}^{(\ell)}])} \le \mathcal{O}\left(\rho^{5 + \kappa} m^{-1/12} + \rho^{1+\kappa} m^{-\zeta/2} + \rho^{1+\kappa} m^{-1/4} + \rho^{5+\kappa} m^{-1/4}  \right) \cdot \norm{\mathbf{v}},
	\end{equation*}
	with probability at least $1 -  \sqrt{m} e^{-\Omega(\rho^2)} \ge 1 - e^{-\Omega(\rho^2)}$.
	For Lemma~\ref{lemma:rerandESN} to hold true, we need $N \le \frac{m}{\rho^{23}}$. Hence, we require $\sqrt{m} \le \frac{m}{\rho^{23}}$, which translates to $m \ge \rho^{46}$.

\end{proof}

\subsection{Proof of Claim~\ref{claim:singlesubset}}
\begin{claim}[Restating claim~\ref{claim:singlesubset}]\label{claim:singlesubset_proof}
	For the given sequence $\mathbf{x}^{(1)}, \cdots, \mathbf{x}^{(L)}$,
	\begin{align*}
		&\abs{f_{\mathcal{K}}(\mathbf{v}, \mathbf{h}^{(\ell-1)}, \mathbf{x}^{(\ell)}) - \frac{N}{m}g(\mathbf{v}, [\mathbf{h}^{(\ell-1)}, \mathbf{x}^{(\ell)}])} \\&\le  \mathcal{O}\left(\rho^{5 + \kappa} N^{7/6} m^{-7/6} + \rho^{1 + \kappa} (N/m)^{1+\zeta} + \rho^{1+\kappa} N^{1/2} m^{-1}  + \rho^{5 + \kappa} (N/m)^{3/2}\right) \cdot \norm{\mathbf{v}},
	\end{align*}
	with probability exceeding $1 - e^{-\Omega(\rho^2)}$.
\end{claim}

\begin{proof}
	We will need $\left\{\mathbf{w}_{k}\right\}_{k \in \mathcal{K}}$ to satisfy several conditions. We will lower-bound the probability of each of these events and finally lower bound the probability of their intersection via the union bound. For the sake of clarity we will explicitly label these events $E_{1}, E_{2}, E_{3}, \text{ and } E_{4}$.

	The following claim shows that since $\mathbf{h}^{(\ell - 1)}$ doesn't change much with re-randomization (from lemma~\ref{lemma:rerandESN}), the function $f$ doesn't change much if we change the argument from $\mathbf{h}^{(\ell-1)}$ to $\widetilde{\mathbf{h}}^{(\ell - 1)}$. 
	
	\begin{claim}\label{claim:ff}
		\begin{equation*}
			\abs{f_{\mathcal{K}}(\mathbf{v}, \mathbf{h}^{(\ell-1)}, \mathbf{x}^{(\ell)}) - f_{\mathcal{K}}(\mathbf{v}, \widetilde{\mathbf{h}}^{(\ell-1)}, \mathbf{x}^{(\ell)}) } \le \mathcal{O}\left(\rho^{5 + \kappa} N^{7/6} m^{-7/6} \norm{\mathbf{v}} \right),
		\end{equation*}
		with probability at least $1 - e^{-\Omega(\rho^2)}$.
	\end{claim}
	
	\begin{proof}

		We have,
		\begingroup \allowdisplaybreaks
		\begin{align*}
			\abs{f_{\mathcal{K}}(\mathbf{v}, \mathbf{h}^{(\ell-1)}, \mathbf{x}^{(\ell)}) - f_{\mathcal{K}}(\mathbf{v}, \widetilde{\mathbf{h}}^{(\ell-1)}, \mathbf{x}^{(\ell)}) }  
			&=  \abs{\sum_{k \in \mathcal{K}} u_k \left(\sigma(\mathbf{w}_k^{\top} \mathbf{h}^{(\ell-1)} + \mathbf{a}_k^{\top} \mathbf{x}^{(\ell)}) - \sigma(\mathbf{w}_k^{\top} \widetilde{\mathbf{h}}^{(\ell-1)} + \mathbf{a}_k^{\top} \mathbf{x}^{(\ell)}) \right)} \\  
			&=  \abs{\sum_{k \in \mathcal{K}}  u_k \left(\sigma(\mathbf{w}_k^{\top} \mathbf{h}^{(\ell-1)} + \mathbf{a}_k^{\top} \mathbf{x}^{(\ell)}) - \sigma(\mathbf{w}_k^{\top} \widetilde{\mathbf{h}}^{(\ell-1)} + \mathbf{a}_k^{\top} \mathbf{x}^{(\ell)}) \right) } \\
			&\le \norm{\mathbf{u}_{\mathcal{K}}} \sqrt{\sum_{k \in \mathcal{K}} \left(\sigma(\mathbf{w}_k^{\top} \mathbf{h}^{(\ell-1)} + \mathbf{a}_k^{\top} \mathbf{x}^{(\ell)}) - \sigma(\mathbf{w}_k^{\top} \widetilde{\mathbf{h}}^{(\ell-1)} + \mathbf{a}_k^{\top} \mathbf{x}^{(\ell)}) \right)^2} \\
			&\le  \norm{\mathbf{u}_{\mathcal{K}}} \sqrt{\sum_{k \in \mathcal{K}} \left(\left(\mathbf{w}_k^{\top} \mathbf{h}^{(\ell-1)} + \mathbf{a}_k^{\top} \mathbf{x}^{(\ell)}\right) - \left(\mathbf{w}_k^{\top} \widetilde{\mathbf{h}}^{(\ell-1)} - \mathbf{a}_k^{\top} \mathbf{x}^{(\ell)}\right) \right)^2} \\
			&=  \norm{\mathbf{u}_{\mathcal{K}}} \sqrt{\sum_{k \in \mathcal{K}} (\mathbf{w}_k^{\top} (\mathbf{h}^{(\ell-1)} - \widetilde{\mathbf{h}}^{(\ell-1)}))^2}, 
		\end{align*}
		\endgroup
		where we use cauchy schwartz inequality in the third step and $1$-lipschitzness of the activation function $\relu$ in the pre-final step.
		We will bound the two factors above separately: 
		
		We have,
		\begin{align}
			\norm{\mathbf{u}_{\mathcal{K}}} = \norm{[\mathbf{W}_{\mathcal{K}}, \mathbf{A}_{\mathcal{K}}]_r \mathbf{v}} = \norm{[\mathbf{W}_{\mathcal{K}}, \mathbf{A}_{\mathcal{K}}]_r \widetilde{\mathbf{v}} + [\mathbf{W}_{\mathcal{K}}, \mathbf{A}_{\mathcal{K}}]_r \left(\mathbf{v} - \widetilde{\mathbf{v}}\right)} \le \norm{[\mathbf{W}_{\mathcal{K}}, \mathbf{A}_{\mathcal{K}}]_r \widetilde{\mathbf{v}}} + \norm{[\mathbf{W}_{\mathcal{K}}, \mathbf{A}_{\mathcal{K}}]_r. \left(\mathbf{v} - \widetilde{\mathbf{v}}\right)} .\label{eq:uK}
		\end{align}
		We break $\mathbf{u}_{\mathcal{K}}$ into two terms, since we need to handle the correlation between $[\mathbf{W}_{\mathcal{K}}, \mathbf{A}_{\mathcal{K}}]_r$ and $\mathbf{v}$ and we will do that using assumption~\ref{ass:variant_v}.
		
		Since, $[\mathbf{W}_{\mathcal{K}}, \mathbf{A}_{\mathcal{K}}]_r$ and $\widetilde{\mathbf{v}}$ are independent, we can use the concentration inequality for chi-squared distributions (Fact~\ref{lem:chi-squared}) to get
		\begin{align*}
			\norm{[\mathbf{W}_{\mathcal{K}}, \mathbf{A}_{\mathcal{K}}]_r \widetilde{\mathbf{v}}} \leq \sqrt{\frac{2N}{m} + \frac{2\rho \sqrt{8N}}{m}} \norm{\widetilde{\mathbf{v}}},
		\end{align*}
		with probability at least $1-2e^{-\rho^2}$. It can be further simplified into 
		\begin{equation*}
			\norm{[\mathbf{W}_{\mathcal{K}}, \mathbf{A}_{\mathcal{K}}]_r \widetilde{\mathbf{v}}} \leq \sqrt{\frac{2N}{m}} \norm{\widetilde{\mathbf{v}}} + \frac{2\rho \sqrt{2N}}{m} \norm{\widetilde{\mathbf{v}}},
		\end{equation*}
		using the fact that $\sqrt{1 + y} \le 1 + y/2$ for any variable $y > 0$. 
		Also, from assumption~\ref{ass:variant_v}, we have with probability $1-e^{-\Omega(\rho^2)}$,
		\begin{align*}
			\norm{[\mathbf{W}_{\mathcal{K}}, \mathbf{A}_{\mathcal{K}}]_r \left(\mathbf{v} - \widetilde{\mathbf{v}}\right)} \le \mathcal{O}(\rho^{\kappa} (N/m)^{0.5 + \zeta} \norm{\mathbf{v}}).
		\end{align*}
		Hence, with probability $1 - e^{-\Omega(\rho^2)}$,
		\begin{align*}
			\norm{[\mathbf{W}_{\mathcal{K}}, \mathbf{A}_{\mathcal{K}}]_r \mathbf{v}} \le \sqrt{\frac{2N}{m}} \norm{\widetilde{\mathbf{v}}} + \frac{2\rho \sqrt{2N}}{m} \norm{\widetilde{\mathbf{v}}} + \mathcal{O}(\rho^{ \kappa} (N/m)^{0.5 + \zeta} \norm{v}).
		\end{align*}
		
		Finally going back to eq.~\ref{eq:uK}, with repeated utilization of assumption~\ref{ass:variant_v}, we have
		\begingroup\allowdisplaybreaks
		\begin{align*}
			\norm{[\mathbf{W}_{\mathcal{K}}, \mathbf{A}_{\mathcal{K}}]_r \mathbf{v}} &\le \sqrt{\frac{2N}{m}} \norm{\widetilde{\mathbf{v}}} + \frac{2\rho \sqrt{2N}}{m} \norm{\widetilde{\mathbf{v}}} + \mathcal{O}(\rho^{\kappa} (N/m)^{0.5 + \zeta} \norm{\mathbf{v}}) \\&
			\le \sqrt{\frac{2N}{m}} \norm{\mathbf{v}} + \frac{2\rho \sqrt{2N}}{m} \norm{\mathbf{v}} + \sqrt{\frac{2N}{m}} \norm{\widetilde{\mathbf{v}} - \mathbf{v}} + \frac{2\rho \sqrt{2N}}{m} \norm{\widetilde{\mathbf{v}} - \mathbf{v}} +  \mathcal{O}(\rho^{\kappa} (N/m)^{0.5 + \zeta} \norm{\mathbf{v}}) \\&
			\le \sqrt{\frac{2N}{m}} \norm{\mathbf{v}} + \frac{2\rho \sqrt{2N}}{m} \norm{\mathbf{v}} + \sqrt{\frac{2N}{m}} \norm{\widetilde{\mathbf{v}} - \mathbf{v}} + \frac{2\rho \sqrt{2N}}{m} \norm{\widetilde{\mathbf{v}} - \mathbf{v}} +  \mathcal{O}(\rho^{\kappa} (N/m)^{0.5 + \zeta} \norm{\mathbf{v}}) \\&
			\le \sqrt{\frac{2N}{m}} \norm{\mathbf{v}} + \frac{2\rho \sqrt{2N}}{m} \norm{\mathbf{v}} + \sqrt{\frac{2N}{m}} \norm{\widetilde{\mathbf{v}} - \mathbf{v}} + \frac{2\rho \sqrt{2N}}{m} \norm{\widetilde{\mathbf{v}} - \mathbf{v}} +  \mathcal{O}(\rho^{\kappa} (N/m)^{0.5 + \zeta} \norm{\mathbf{v}})
			\\&
			\le \sqrt{\frac{2N}{m}} \norm{\mathbf{v}} + \frac{2\rho \sqrt{2N}}{m} \norm{\mathbf{v}} + \sqrt{\frac{2N}{m}} \cdot \mathcal{O}(\rho^{\kappa} (N/m)^{\zeta} \norm{\mathbf{v}}) \\& \quad \quad \quad \quad + \frac{2\rho \sqrt{2N}}{m} \cdot \mathcal{O}(\rho^{\kappa} (N/m)^{\zeta} \norm{\mathbf{v}}) +  \mathcal{O}(\rho^ {\kappa} (N/m)^{0.5 + \zeta} \norm{\mathbf{v}}) \\&
			\le \mathcal{O}\left( \rho^{\kappa} \sqrt{N/m} \cdot \norm{\mathbf{v}}\right),
		\end{align*}
		\endgroup
		giving us a bound on norm of $\mathbf{u}_{\mathcal{K}}$. Let us call this event $E_1$.
		\todo{mabe give some more detail here.}
		
		Lemma~\ref{lemma:rerandESN} shows that with probability at least $1 - e^{-\Omega(\rho^2)}$,
		\begin{align}
			& \norm{\mathbf{h}^{(\ell-1)} - \widetilde{\mathbf{h}}^{(\ell-1)}} \le \mathcal{O}\left(\rho^{5} N^{1/2} m^{-1/2}\right)\\&
			\norm{ \mathbf{W}_{\mathcal{K}} \left(\mathbf{h}^{(\ell-1)} - \widetilde{\mathbf{h}}^{(\ell-1)}\right) } \leq \mathcal{O}\left(\rho^{5} N^{2 / 3} m^{-2 / 3}\right), \quad \forall k \in [m]
		\end{align}
		Let us call this event $E_2$.
		
		Combining the two bounds we get 
		\begin{align*}
			\abs{f_{\mathcal{K}}(\mathbf{v}, \mathbf{h}^{(\ell-1)}, \mathbf{x}^{(\ell)}) - f_{\mathcal{K}}(\mathbf{v}, \widetilde{\mathbf{h}}^{(\ell-1)}, \mathbf{x}^{(\ell)}) }  &\le \norm{\mathbf{u}_{\mathcal{K}}} \sqrt{\sum_{k \in \mathcal{K}} (\mathbf{w}_k^{\top} (\mathbf{h}^{(\ell-1)} - \widetilde{\mathbf{h}}^{(\ell-1)}))^2}
			\\& \leq \left(\rho^{\kappa} \sqrt{N/m} \right) \norm{\mathbf{v}}  \cdot
			\sqrt{\sum_{k \in \mathcal{K}} (\mathbf{w}_k^{\top} (\mathbf{h}^{(\ell-1)} - \widetilde{\mathbf{h}}^{(\ell-1)}))^2} \\&
			= \left(\rho^{\kappa} \sqrt{N/m} \right) \norm{\mathbf{v}}  \cdot
			\norm{ \mathbf{W}_{\mathcal{K}} \left(\mathbf{h}^{(\ell-1)} - \widetilde{\mathbf{h}}^{(\ell-1)}\right) }
			\\& \leq \left(\rho^{\kappa} \sqrt{N/m} \right) \norm{\mathbf{v}} \cdot  \mathcal{O}\left(\rho^{5} N^{2 / 3} m^{-2 / 3}\right) \\&\le \mathcal{O}\left(\rho^{5 + \kappa} N^{7/6} m^{-7/6} \norm{\mathbf{v}} \right),
		\end{align*}
		with probability at least $\Pr[E_1 \cap E_2] \geq 1 - e^{-\Omega(\rho^2)} - 2e^{-\rho^2} \ge 1 - e^{-\Omega(\rho^2)}$.
	\end{proof}
	
	The following claim shows that since $\mathbf{v}$ doesn't change much with re-randomization (from assumption~\ref{ass:variant_v}), the function $f$ doesn't change much if we change the argument from $\mathbf{v}$ to $\widetilde{\mathbf{v}}$.  
	\begin{claim}\label{claim:ffv}
		\begin{equation*}
			\abs{f_{\mathcal{K}}(\mathbf{v}, \widetilde{\mathbf{h}}^{(\ell-1)}, \mathbf{x}^{(\ell)}) - f_{\mathcal{K}}(\widetilde{\mathbf{v}}, \widetilde{\mathbf{h}}^{(\ell-1)}, \mathbf{x}^{(\ell)}) } \le \mathcal{O}(\rho^{1 + \kappa} (N/m)^{1+\zeta} \norm{\mathbf{v}}),
		\end{equation*}
		with probability at least $1 - e^{-\Omega(\rho^2)}$.
	\end{claim}
	
	\begin{proof}
		Let $\mathbf{u} = [\mathbf{W}, \mathbf{A}]_r \widetilde{\mathbf{v}}$.  We have,
		\begin{align*}
			\abs{f_{\mathcal{K}}(\mathbf{v}, \widetilde{\mathbf{h}}^{(\ell-1)}, \mathbf{x}^{(\ell)}) - f_{\mathcal{K}}(\widetilde{\mathbf{v}}, \widetilde{\mathbf{h}}^{(\ell-1)}, \mathbf{x}^{(\ell)})}
			&=  \abs{\sum_{k \in \mathcal{K}} \left(u_k - \widetilde{u}_k\right)  \sigma(\mathbf{w}_k^{\top} \widetilde{\mathbf{h}}^{(\ell-1)} + \mathbf{a}_k^{\top} \mathbf{x}^{(\ell)})} \\  
			&\le \sqrt{\sum_{k \in \mathcal{K}}\sigma(\mathbf{w}_k^{\top} \widetilde{\mathbf{h}}^{(\ell-1)} + \mathbf{a}_k^{\top} \mathbf{x}^{(\ell)})^2 }  \sqrt{\sum_{k \in \mathcal{K}} \left( u_k - \widetilde{u}_k \right)^2} \\&
			\le \sqrt{\sum_{k \in \mathcal{K}} (\mathbf{w}_k^{\top} \widetilde{\mathbf{h}}^{(\ell-1)} + \mathbf{a}_k^{\top} \mathbf{x}^{(\ell)})^2 }  \sqrt{\sum_{k \in \mathcal{K}} \left( u_k - \widetilde{u}_k \right)^2}\\
			&= \sqrt{\sum_{k \in \mathcal{K}}(\mathbf{w}_k^{\top} \widetilde{\mathbf{h}}^{(\ell-1)} + \mathbf{a}_k^{\top} \mathbf{x}^{(\ell)})^2} \sqrt{\sum_{k \in \mathcal{K}} \langle [\mathbf{w}_k, \mathbf{a}_k], (\mathbf{v} - \widetilde{\mathbf{v}}) \rangle^2},
		\end{align*}
		where we use cauchy schwartz inequality in the second step and $1$-lipschitzness of $\relu$ in the pre-final step.
		
		We will bound the two factors above separately: 
		Since, $[\mathbf{W}_{\mathcal{K}}, \mathbf{A}_{\mathcal{K}}]_r$ and $\widetilde{\mathbf{h}}$ are independent, we can use the concentration inequality for chi-squared distributions (Fact~\ref{lem:chi-squared}) to get
		\begin{align*}
			\sqrt{\sum_{k \in \mathcal{K}}(\mathbf{w}_k^{\top} \widetilde{\mathbf{h}}^{(\ell-1)} + \mathbf{a}_k^{\top} \mathbf{x}^{(\ell)})^2} \leq \sqrt{\frac{2N}{m} + \frac{2\rho \sqrt{8N}}{m}} \norm{[\widetilde{\mathbf{h}}^{(\ell-1)}, \mathbf{x}^{(\ell)}]},
		\end{align*}
		with probability at least $1-2e^{-\rho^2}$. It can be further simplified into 
		\begin{equation*}
			\sqrt{\sum_{k \in \mathcal{K}}(\mathbf{w}_k^{\top} \widetilde{\mathbf{h}}^{(\ell-1)} + \mathbf{a}_k^{\top} \mathbf{x}^{(\ell)})^2} \leq \sqrt{\frac{2N}{m}} \norm{[\widetilde{\mathbf{h}}^{(\ell-1)}, \mathbf{x}^{(\ell)}]} + \frac{2\rho \sqrt{2N}}{m} \norm{[\widetilde{\mathbf{h}}^{(\ell-1)}, \mathbf{x}^{(\ell)}]},
		\end{equation*}
		using the fact that $\sqrt{1 + y} \le 1 + y/2$ for any variable $y > 0$. Let's call this event $E_3$.
		
		We assume that our deep random neural network satisfies $\norm[0]{\widetilde{\mathbf{h}}^{(\ell-1)}} \in (\sqrt{2 + (\ell - 2)\epsilon_x^2} - \frac{\rho^2}{\sqrt{m}},$\\$ \sqrt{2 + (\ell - 2)\epsilon_x^2} + 
		\frac{\rho^2}{\sqrt{m}})$ for all $\ell \in [L]$. This happens with probability at least $1-e^{-\Omega(\rho^2)}$ w.r.t. the matrices $\widetilde{\mathbf{W}}$ and $\widetilde{\mathbf{A}}$ from Lemma~\ref{lemma:norm_ESN}. Thus, provided $m \ge \Omega(\rho^4)$ and $\epsilon_x \le \frac{1}{L}$, $\norm[0]{\widetilde{\mathbf{h}}^{(\ell-1)}} \in \left( \sqrt{2}, \sqrt{3}\right)$. Let's call this event $E_4$. Also, since the sequence is assumed to be input normalized, $\norm{\mathbf{x}^{(\ell)}} \le 1$. Hence,
		\begin{equation*}
			\sqrt{\sum_{k \in \mathcal{K}}(\mathbf{w}_k^{\top} \widetilde{\mathbf{h}}^{(\ell-1)} + \mathbf{a}_k^{\top} \mathbf{x}^{(\ell)})^2} \leq 2\sqrt{\frac{2N}{m}}  + \frac{4\rho \sqrt{2N}}{m},
		\end{equation*}
		Again, from assumption~\ref{ass:variant_v}, we have with probability $1-e^{-\Omega(\rho^2)}$,
		\begin{align*}
			\norm{[\mathbf{W}_{\mathcal{K}}, \mathbf{A}_{\mathcal{K}}]_r \left(\mathbf{v} - \widetilde{\mathbf{v}}\right)} \le \mathcal{O}(\rho^{\kappa} (N/m)^{0.5 + \zeta} \norm{\mathbf{v}}).
		\end{align*}
		Let us call this event $E_5$.
		
		Combining the two bounds we get 
		\begin{align*}
			\abs{f_{\mathcal{K}}(\mathbf{v}, \widetilde{\mathbf{h}}^{(\ell-1)}, \mathbf{x}^{(\ell)}) - f_{\mathcal{K}}(\widetilde{\mathbf{v}}, \widetilde{\mathbf{h}}^{(\ell-1)}, \mathbf{x}^{(\ell)})}
			&\le \sqrt{\sum_{k \in \mathcal{K}}(\mathbf{w}_k^{\top} \widetilde{\mathbf{h}}^{(\ell-1)} + \mathbf{a}_k^{\top} \mathbf{x}^{(\ell)})^2} \sqrt{\sum_{k \in \mathcal{K}} \langle [\mathbf{w}_k, \mathbf{a}_k], (\mathbf{v} - \widetilde{\mathbf{v}}) \rangle^2} \\&
			\le \left(2\sqrt{\frac{2N}{m}}  + \frac{4\rho \sqrt{2N}}{m}\right) \cdot \mathcal{O}(\rho^{ \kappa} (N/m)^{0.5 + \zeta} \norm{\mathbf{v}}), \\&
			\le \mathcal{O}(\rho^{1 + \kappa} (N/m)^{1+\zeta} \norm{\mathbf{v}}),
		\end{align*}
		with probability at least $\Pr[E_3 \cap E_4 \cap E_5] \geq 1 - 3e^{-\Omega(\rho^2)} \ge 1 - e^{-\Omega(\rho^2)}$.
	\end{proof}

	The next claim shows that the functions $\frac{N}{m} g$ and $f$ are close to each other, using concentration bounds w.r.t. $\{\bw_r\}_{r \in \mathcal{K}}$ and $\{\mathbf{a}_r\}_{r \in \mathcal{K}}$.
	
	\begin{claim}\label{claim:fg}
		\begin{equation*}
			\abs{f_{\mathcal{K}}(\widetilde{\mathbf{v}}, \widetilde{\mathbf{h}}^{(\ell-1)}, \mathbf{x}^{(\ell)}) - \frac{N}{m} g(\widetilde{\mathbf{v}}, [\widetilde{\mathbf{h}}^{(\ell-1)}, \mathbf{x}^{(\ell)}])} \le \mathcal{O}( \rho^{1+\kappa} N^{1/2} m^{-1} \norm{\mathbf{v}}),  
		\end{equation*}
		with probability at least $1 - e^{-\Omega(\rho^2)}$.    
	\end{claim}
	
	\begin{proof}
		Since $\widetilde{\mathbf{v}}$ and $\widetilde{\mathbf{h}}^{(\ell-1)}$ doesn't depend on $\left\{\mathbf{w}_k\right\}_{k \in \mathcal{K}}$, we can use corollary~\ref{cor:linear_estimate} directly to get 
		\begin{equation*}
			\mathbb{E}_{\left\{\mathbf{w}_k\right\}_{k \in \mathcal{K}}} f_{\mathcal{K}}(\widetilde{\mathbf{h}}^{(\ell-1)}, \mathbf{x}^{(\ell)}) = \frac{1}{2} \cdot \frac{2}{m} \cdot g([\widetilde{\mathbf{h}}^{(\ell-1)}, \mathbf{x}^{(\ell)}]) = \frac{1}{m}g([\widetilde{\mathbf{h}}^{(\ell-1)}, \mathbf{x}^{(\ell)}]) .
		\end{equation*}
		Let $\overline{\widetilde{\mathbf{v}}} := \widetilde{\mathbf{v}}/\norm{\widetilde{\mathbf{v}}}$ and $\overline{[\widetilde{\mathbf{h}}^{(\ell-1)}, \mathbf{x}^{(\ell)}]} = \frac{[\widetilde{\mathbf{h}}^{(\ell-1)}, \mathbf{x}^{(\ell)}]}{\norm{[\widetilde{\mathbf{h}}^{(\ell-1)}, \mathbf{x}^{(\ell)}]}}$.

		For the given sequence $\mathbf{x}^{(1)}, \cdots, \mathbf{x}^{(\ell)}$, we have
		
		\begingroup
		\allowdisplaybreaks
		\begin{align}
			& \abs{\frac{1}{N}f_{\mathcal{K}}(\widetilde{\mathbf{v}}, \widetilde{\mathbf{h}}^{(\ell-1)}, \mathbf{x}^{(\ell)}) - \frac{1}{m}g(\widetilde{\mathbf{v}}, [\widetilde{\mathbf{h}}^{(\ell-1)}, \mathbf{x}^{(\ell)}])} \nonumber
			\\& = \abs{\frac{1}{N}  \sum_{k \in \mathcal{K}} [\mathbf{w}_{k}, \mathbf{a}_{k}]^{\top}  \widetilde{\mathbf{v}} \sigma(\mathbf{w}_k^{\top} \widetilde{\mathbf{h}}^{(\ell-1)} + \mathbf{a}_k^{\top} \mathbf{x}^{(\ell)}) -   \mathbb{E}_{\mathbf{w} \sim \mathcal{N}\left(0, \frac{2}{m}\mathbf{I}_{m+d}\right)} \mathbf{w}^{\top} \widetilde{\mathbf{v}} \sigma(\mathbf{w}^{\top} [\widetilde{\mathbf{h}}^{(\ell-1)},  \mathbf{x}^{(\ell)}])} \nonumber\\
			&= \abs{\widetilde{\mathbf{v}}^{\top} \left(\frac{1}{N} \sum_{k \in \mathcal{K}} [\mathbf{w}_{k}, \mathbf{a}_{k}] [\mathbf{w}_{k}, \mathbf{a}_{k}]^{\top}  \mathbb{I}_{[\mathbf{w}_{k}, \mathbf{a}_{k}]^{\top} [\widetilde{\mathbf{h}}^{(\ell-1)}, \mathbf{x}^{(\ell)}] \ge 0} - \mathbb{E}_{\mathbf{w} \sim \mathcal{N}\left(0, \frac{2}{m}\mathbf{I}_{m+d}\right)}  \mathbf{w}\mathbf{w}^{\top}  \mathbb{I}_{\mathbf{w}^{\top} [\widetilde{\mathbf{h}}^{(\ell-1)}, \mathbf{x}^{(\ell)}] \ge 0}\right) [\widetilde{\mathbf{h}}^{(\ell-1)}, \mathbf{x}^{(\ell)}]}  \nonumber\\
			&= \norm{\widetilde{\mathbf{v}}} \norm[0]{[\widetilde{\mathbf{h}}^{(\ell-1)},  \mathbf{x}^{(\ell)}]}\Big|\overline{\widetilde{\mathbf{v}}}^{\top} \Big(\frac{1}{N} \sum_{k \in \mathcal{K}} [\mathbf{w}_{k}, \mathbf{a}_{k}] [\mathbf{w}_{k}, \mathbf{a}_{k}]^{\top}  \mathbb{I}_{[\mathbf{w}_{k}, \mathbf{a}_{k}]^{\top} \overline{[\widetilde{\mathbf{h}}^{(\ell-1)},  \mathbf{x}^{(\ell)}]} \ge 0} 
			\nonumber\\ & \quad\quad\quad\quad\quad\quad\quad\quad\quad\quad\quad\quad\quad
			- \mathbb{E}_{\mathbf{w} \sim \mathcal{N}\left(0, \frac{2}{m}\mathbf{I}_{m+d}\right)}  \mathbf{w}\mathbf{w}^{\top}  \mathbb{I}_{\mathbf{w}^{\top} \overline{[\widetilde{\mathbf{h}}^{(\ell-1)},  \mathbf{x}^{(\ell)}]} \ge 0}\Big) \overline{[\widetilde{\mathbf{h}}^{(\ell-1)},  \mathbf{x}^{(\ell)}]}\Big|  \nonumber\\
			&\le  \frac{\norm{\widetilde{\mathbf{v}}} \norm[0]{[\widetilde{\mathbf{h}}^{(\ell-1)},  \mathbf{x}^{(\ell)}]}}{2}  \Big|\frac{1}{N}\sum_{k \in \mathcal{K}} \left(\left(\overline{[\widetilde{\mathbf{h}}^{(\ell-1)},  \mathbf{x}^{(\ell)}]} + \overline{\widetilde{\mathbf{v}}}\right)^{\top} [\mathbf{w}_{k}, \mathbf{a}_{k}] \mathbb{I}_{[\mathbf{w}_{k}, \mathbf{a}_{k}]^{\top} \overline{[\widetilde{\mathbf{h}}^{(\ell-1)},  \mathbf{x}^{(\ell)}]} \ge 0} \right)^2 \nonumber\\
			&\quad\quad\quad\quad\quad\quad\quad\quad\quad\quad\quad\quad\quad - \mathbb{E}_{\mathbf{w} \sim \mathcal{N}\left(0, \frac{2}{m}\mathbf{I}\right)} \left(\left(\overline{[\widetilde{\mathbf{h}}^{(\ell-1)},  \mathbf{x}^{(\ell)}]} + \overline{\widetilde{\mathbf{v}}}\right)^{\top} \mathbf{w} \mathbb{I}_{\mathbf{w}^{\top} \overline{[\widetilde{\mathbf{h}}^{(\ell-1)},  \mathbf{x}^{(\ell)}]} \ge 0} \right)^2\Big| \label{eqn:bernstein_xl_v}
			\\&  + \frac{\norm{\widetilde{\mathbf{v}}} \norm[0]{[\widetilde{\mathbf{h}}^{(\ell-1)},  \mathbf{x}^{(\ell)}]}}{2}\Big| \frac{1}{N}\sum_{k \in \mathcal{K}} \left(\overline{[\widetilde{\mathbf{h}}^{(\ell-1)},  \mathbf{x}^{(\ell)}]}^{\top} [\mathbf{w}_{k}, \mathbf{a}_{k}] \mathbb{I}_{[\mathbf{w}_{k}, \mathbf{a}_{k}]^{\top} \overline{[\widetilde{\mathbf{h}}^{(\ell-1)},  \mathbf{x}^{(\ell)}]} \ge 0} \right)^2  \nonumber \\&\quad\quad\quad\quad\quad\quad\quad\quad\quad\quad\quad\quad\quad - \mathbb{E}_{\mathbf{w} \sim \mathcal{N}\left(0, \frac{2}{m}\mathbf{I}\right)} \left(\overline{[\widetilde{\mathbf{h}}^{(\ell-1)},  \mathbf{x}^{(\ell)}]}^{\top} \mathbf{w} \mathbb{I}_{\mathbf{w}^{\top} \overline{[\widetilde{\mathbf{h}}^{(\ell-1)},  \mathbf{x}^{(\ell)}]} \ge 0} \right)^2\Big| \label{eqn:bernstein_xl}\\
			&  + \frac{\norm{\widetilde{\mathbf{v}}} \norm[0]{[\widetilde{\mathbf{h}}^{(\ell-1)},  \mathbf{x}^{(\ell)}]}}{2} \abs{\frac{1}{N}\sum_{k \in \mathcal{K}}  \left(\overline{\widetilde{\mathbf{v}}}^{\top} [\mathbf{w}_{k}, \mathbf{a}_{k}] \mathbb{I}_{[\mathbf{w}_{k}, \mathbf{a}_{k}]^{\top} \overline{[\widetilde{\mathbf{h}}^{(\ell-1)},  \mathbf{x}^{(\ell)}]} \ge 0} \right)^2 - \mathbb{E}_{\mathbf{w} \sim \mathcal{N}\left(0, \frac{2}{m}\mathbf{I}\right)} \left(\overline{\widetilde{\mathbf{v}}}^{\top} \mathbf{w} \mathbb{I}_{\mathbf{w}^{\top} \overline{[\widetilde{\mathbf{h}}^{(\ell-1)},  \mathbf{x}^{(\ell)}]} \ge 0} \right)^2} \label{eqn:bernstein_v}
		\end{align}
		\endgroup

		The three terms above in \eqref{eqn:bernstein_xl_v}, \eqref{eqn:bernstein_xl} and \eqref{eqn:bernstein_v} correspond to the large deviation bounds for random variables $\left(\overline{[\widetilde{\mathbf{h}}^{(\ell-1)},  \mathbf{x}^{(\ell)}]} + \overline{\widetilde{\mathbf{v}}}\right)^{\top} [\mathbf{w}_{k}, \mathbf{a}_{k}] \mathbb{I}_{[\mathbf{w}_{k}, \mathbf{a}_{k}]^{\top} \overline{[\widetilde{\mathbf{h}}^{(\ell-1)},  \mathbf{x}^{(\ell)}]} \ge 0}$, $\overline{[\widetilde{\mathbf{h}}^{(\ell-1)},  \mathbf{x}^{(\ell)}]}^{\top} [\mathbf{w}_{k}, \mathbf{a}_{k}] \mathbb{I}_{[\mathbf{w}_{k}, \mathbf{a}_{k}]^{\top} \overline{[\widetilde{\mathbf{h}}^{(\ell-1)},  \mathbf{x}^{(\ell)}]} \ge 0}$ and $\overline{\widetilde{\mathbf{v}}}^{\top} [\mathbf{w}_{k}, \mathbf{a}_{k}] \mathbb{I}_{[\mathbf{w}_{k}, \mathbf{a}_{k}]^{\top} \overline{[\widetilde{\mathbf{h}}^{(\ell-1)},  \mathbf{x}^{(\ell)}]} \ge 0}$ respectively. Each of these random variables is
		sub-exponential as it is bounded above by a squared Gaussian random variable using the fact that $\mathbf{w} \sim \mathcal{N}\left(0, \frac{2}{m}\mathbf{I}\right)$:
		\todo{We need to explain that our r.v.s are dominated by the corresponding Gaussians.}
		\begin{itemize}
			\item $((\overline{[\widetilde{\mathbf{h}}^{(\ell-1)},  \mathbf{x}^{(\ell)}]} + \overline{\widetilde{\mathbf{v}}})^{\top} [\mathbf{w}_{k}, \mathbf{a}_{k}] \mathbb{I}_{[\mathbf{w}_{k}, \mathbf{a}_{k}]^{\top} \overline{[\widetilde{\mathbf{h}}^{(\ell-1)},  \mathbf{x}^{(\ell)}]} \ge 0} )^2 \leq
			((\overline{[\widetilde{\mathbf{h}}^{(\ell-1)},  \mathbf{x}^{(\ell)}]} + \overline{\widetilde{\mathbf{v}}})^{\top} [\mathbf{w}_{k}, \mathbf{a}_{k}] )^2$ 
			and $(\overline{[\widetilde{\mathbf{h}}^{(\ell-1)},  \mathbf{x}^{(\ell)}]} + \overline{\widetilde{\mathbf{v}}})^{\top} \mathbf{w} \sim N(0, \frac{2}{m}\norm[0]{\overline{[\widetilde{\mathbf{h}}^{(\ell-1)},  \mathbf{x}^{(\ell)}]} + \overline{\widetilde{\mathbf{v}}}}^2)$,

			\item $((\overline{[\widetilde{\mathbf{h}}^{(\ell-1)},  \mathbf{x}^{(\ell)}]})^{\top} [\mathbf{w}_{k}, \mathbf{a}_{k}] \mathbb{I}_{[\mathbf{w}_{k}, \mathbf{a}_{k}]^{\top} \overline{[\widetilde{\mathbf{h}}^{(\ell-1)},  \mathbf{x}^{(\ell)}]} \ge 0} )^2 \leq
			((\overline{[\widetilde{\mathbf{h}}^{(\ell-1)},  \mathbf{x}^{(\ell)}]})^{\top} [\mathbf{w}_{k}, \mathbf{a}_{k}] )^2$ 
			and $(\overline{[\widetilde{\mathbf{h}}^{(\ell-1)},  \mathbf{x}^{(\ell)}]})^{\top} \mathbf{w} \sim N(0, \frac{2}{m}\norm[0]{\overline{[\widetilde{\mathbf{h}}^{(\ell-1)},  \mathbf{x}^{(\ell)}]}}^2)$
			
			\item $((\overline{\widetilde{\mathbf{v}}})^{\top} [\mathbf{w}_{k}, \mathbf{a}_{k}] \mathbb{I}_{[\mathbf{w}_{k}, \mathbf{a}_{k}]^{\top} \overline{[\widetilde{\mathbf{h}}^{(\ell-1)},  \mathbf{x}^{(\ell)}]} \ge 0} )^2 \leq
			((\overline{\widetilde{\mathbf{v}}})^{\top} [\mathbf{w}_{k}, \mathbf{a}_{k}] )^2$ 
			and $(\overline{\widetilde{\mathbf{v}}})^{\top} \mathbf{w} \sim N(0, \frac{2}{m}\norm[0]{\overline{\widetilde{\mathbf{v}}}}^2)$.
		\end{itemize}

		For typographical convenience, denote the expressions in \eqref{eqn:bernstein_xl_v}, \eqref{eqn:bernstein_xl} and \eqref{eqn:bernstein_v} by 
		$P(\{\bx^{(\ell)}\}), Q(\{\bx^{(\ell)}\})$ and $R(\{\bx^{(\ell)}\})$, respectively. We assume that our deep random neural network satisfies $\norm[0]{\widetilde{\mathbf{h}}^{(\ell-1)}} \in \left(\sqrt{2 + (\ell - 2)\epsilon_x^2} - \frac{\rho^2}{\sqrt{m}}, \sqrt{2 + (\ell - 2)\epsilon_x^2} + 
		\frac{\rho^2}{\sqrt{m}}\right)$ for all $\ell \in [L]$, with probability at least $1-e^{-\Omega(\rho^2)}$. Thus, provided $m \ge \Omega(\rho^4)$ and $\epsilon_x \le \frac{1}{L}$, $\norm[0]{\widetilde{\mathbf{h}}^{(\ell-1)}} \in \left(\sqrt{2}, \sqrt{3}\right)$.  
		This event was taken care before in event $E_4$. Using this and concentration of chi-squared random variables
		(Fact~\ref{lem:chi-squared}) we get 
		\begin{itemize}
			\item $\Pr\left[\frac{2P(\{\bx^{(\ell)}\})}{\norm{\widetilde{\mathbf{v}}} \norm{[\widetilde{\mathbf{h}}^{(\ell-1)},  \mathbf{x}^{(\ell)}]}} \leq \frac{2\sqrt{2} \rho}{ \sqrt{N} } \cdot \frac{2}{m}\norm[0]{\overline{[\widetilde{\mathbf{h}}^{(\ell-1)},  \mathbf{x}^{(\ell)}]} + \overline{\widetilde{\mathbf{v}}}}^2\right] \leq e^{-\rho^2}$,
			\item $\Pr\left[\frac{2Q(\{\bx^{(\ell)}\})}{\norm{\widetilde{\mathbf{v}}} \norm{[\widetilde{\mathbf{h}}^{(\ell-1)},  \mathbf{x}^{(\ell)}]}} \leq \frac{2\sqrt{2} \rho}{ \sqrt{N} } \cdot \frac{2}{m}\norm[0]{\overline{[\widetilde{\mathbf{h}}^{(\ell-1)},  \mathbf{x}^{(\ell)}]}}^2\right] \leq e^{-\rho^2}$,
			\item $\Pr\left[\frac{2R(\{\bx^{(\ell)}\})}{\norm{\widetilde{\mathbf{v}}} \norm{[\widetilde{\mathbf{h}}^{(\ell-1)},  \mathbf{x}^{(\ell)}]}} \leq \frac{2\sqrt{2} \rho}{ \sqrt{N} } \cdot \frac{2}{m}\norm[0]{\overline{\widetilde{\mathbf{v}}}}^2\right] \leq e^{-\rho^2}$.
		\end{itemize}
		
		Define the following event 
		
		\begingroup\allowdisplaybreaks
		\begin{align*}
			E_6(\{\bx^{(\ell)}\}) := &\left(\frac{2P(\{\bx^{(\ell)}\})}{\norm{\widetilde{\mathbf{v}}} \norm{[\widetilde{\mathbf{h}}^{(\ell-1)},  \mathbf{x}^{(\ell)}]}} \leq \frac{2\sqrt{2} \rho}{ \sqrt{N} } \cdot \frac{2}{m}\norm[0]{\overline{[\widetilde{\mathbf{h}}^{(\ell-1)},  \mathbf{x}^{(\ell)}]} + \overline{\widetilde{\mathbf{v}}}}^2 \right) \\&
			\cap \left(\frac{2Q(\{\bx^{(\ell)}\})}{\norm{\widetilde{\mathbf{v}}} \norm{[\widetilde{\mathbf{h}}^{(\ell-1)},  \mathbf{x}^{(\ell)}]}} \leq \frac{2\sqrt{2} \rho}{ \sqrt{N} } \cdot  \frac{2}{m}\norm[0]{\overline{[\widetilde{\mathbf{h}}^{(\ell-1)},  \mathbf{x}^{(\ell)}]}}^2 \right)  \\&
			\cap \left(\frac{2R(\{\bx^{(\ell)}\})}{\norm{\widetilde{\mathbf{v}}} \norm{[\widetilde{\mathbf{h}}^{(\ell-1)},  \mathbf{x}^{(\ell)}]}} \leq \frac{2\sqrt{2} \rho}{ \sqrt{N} } \cdot \frac{2}{m}\norm[0]{\overline{\widetilde{\mathbf{v}}}}^2 \right)          \end{align*}
		\endgroup
		We have thus shown that for the given sequence $\bx^{(1)}, \bx^{(2)}, \cdots, \bx^{(L)}$ with probability at least $1 - 3e^{-\rho^2} - e^{-\Omega(\rho^2)}$ the event $E_6(\bx) \cap E_3$ occurs, which implies
		\begin{align}\label{eqn:fg-enet}
			&\abs[0]{\frac{1}{N}f_{\mathcal{K}}(\widetilde{\mathbf{v}}, \widetilde{\mathbf{h}}^{(\ell-1)}, \mathbf{x}^{(\ell)}) - \frac{1}{m}g([\widetilde{\mathbf{v}}, \widetilde{\mathbf{h}}^{(\ell-1)}, \mathbf{x}^{(\ell)}])} \nonumber\\& \quad\quad\quad\quad\leq \frac{\norm{\widetilde{\mathbf{v}}} \norm{[\widetilde{\mathbf{h}}^{(\ell-1)},  \mathbf{x}^{(\ell)}]}}{2} \cdot \frac{2\sqrt{2} \rho}{ \sqrt{N} } \cdot \frac{2}{m} (\norm[0]{\overline{[\widetilde{\mathbf{h}}^{(\ell-1)},  \mathbf{x}^{(\ell)}]} + \overline{\widetilde{\mathbf{v}}}}^2 + \norm[0]{\overline{[\widetilde{\mathbf{h}}^{(\ell-1)},  \mathbf{x}^{(\ell)}]}}^2 + \norm[0]{\overline{\widetilde{\mathbf{v}}}}^2) \\ \nonumber
			& \quad\quad\quad\quad\le \frac{16\sqrt{2} \rho}{ \sqrt{N} m} \norm{\widetilde{\mathbf{v}}} \norm{[\widetilde{\mathbf{h}}^{(\ell-1)},  \mathbf{x}^{(\ell)}]} \le \frac{32\sqrt{2} \rho}{ \sqrt{N} m} \norm{\widetilde{\mathbf{v}}}.
		\end{align} 
		\todo{Might need to add some more information above.}
		Hence with probabilty $1-e^{-\Omega(\rho^2)}$,
		\begin{equation*}
			\abs[0]{f_{\mathcal{K}}(\widetilde{\mathbf{v}}, \widetilde{\mathbf{h}}^{(\ell-1)}, \mathbf{x}^{(\ell)}) - \frac{N}{m}g(\widetilde{\mathbf{v}}, [\widetilde{\mathbf{h}}^{(\ell-1)}, \mathbf{x}^{(\ell)}])} \le \frac{32\sqrt{2N} \rho}{ m} \norm{\widetilde{\mathbf{v}}}.
		\end{equation*}
		We further use assumption~\ref{ass:variant_v} to get
		\begingroup \allowdisplaybreaks
		\begin{align*}
			\abs[0]{f_{\mathcal{K}}(\widetilde{\mathbf{v}}, \widetilde{\mathbf{h}}^{(\ell-1)}, \mathbf{x}^{(\ell)}) - \frac{N}{m}g(\widetilde{\mathbf{v}}, [\widetilde{\mathbf{h}}^{(\ell-1)}, \mathbf{x}^{(\ell)}])} &\le \frac{32\sqrt{2} \rho}{ \sqrt{N} m} \norm{\widetilde{\mathbf{v}}} \\&
			\le \frac{32\sqrt{2} \rho \sqrt{N}}{  m} \left(\norm{\mathbf{v}} + \norm{\mathbf{v} - \widetilde{\mathbf{v}}} \right)\\&
			\le \frac{32\sqrt{2} \sqrt{N}\rho}{  m} \left(\norm{\mathbf{v}} + \mathcal{O}\left(\rho^{\kappa} (N/m)^{\zeta}\right)\norm{\mathbf{v}}\right) \\&\le \mathcal{O}( \rho^{1+\kappa} N^{1/2} m^{-1} \norm{\mathbf{v}}),
		\end{align*}
		\endgroup
		with probability exceeding $1 - e^{-\Omega(\rho^2)}$.
	\end{proof}

	The following claim again uses the property that $\mathbf{v}$ and $\mathbf{h}^{(\ell - 1)}$ doesn't change much with re-randomization to show that function $g$ is also stable to re-randomization.
	\begin{claim}\label{claim:gg}
		\begin{equation*}
			\abs{\frac{N}{m}g([\widetilde{\mathbf{v}}, \widetilde{\mathbf{h}}^{(\ell-1)},  \mathbf{x}^{(\ell)}]) - \frac{N}{m}g(\mathbf{v}, [\mathbf{h}^{(\ell-1)},  \mathbf{x}^{(\ell)}])} \le \mathcal{O}(\rho^{5 + \kappa} (N/m)^{3/2}) \cdot \norm{\mathbf{v}},
		\end{equation*}
		with probability exceeding $1 - e^{-\Omega(\rho^2)}$.
	\end{claim}
	
	\begin{proof}
		\begingroup \allowdisplaybreaks
		\begin{align*}
			&\abs{g(\widetilde{\mathbf{v}}, [\widetilde{\mathbf{h}}^{(\ell-1)},  \mathbf{x}^{(\ell)}]) - g(\mathbf{v}, [\mathbf{h}^{(\ell-1)},  \mathbf{x}^{(\ell)}])}\\ &= \abs{\mathbf{v}^{\top} [\mathbf{h}^{(\ell-1)},  \mathbf{x}^{(\ell)}] - \widetilde{\mathbf{v}}^{\top} [\widetilde{\mathbf{h}}^{(\ell-1)},  \mathbf{x}^{(\ell)}]} \\&
			= \abs{\mathbf{v}^{\top} [\mathbf{h}^{(\ell-1)},  \mathbf{x}^{(\ell)}] - \widetilde{\mathbf{v}}^{\top} [\mathbf{h}^{(\ell-1)},  \mathbf{x}^{(\ell)}] + \widetilde{\mathbf{v}}^{\top} [\widetilde{\mathbf{h}}^{(\ell-1)},  \mathbf{x}^{(\ell)}] - \widetilde{\mathbf{v}}^{\top} [\mathbf{h}^{(\ell-1)},  \mathbf{x}^{(\ell)}]} \\&
			\le\abs{\left(\mathbf{v} - \widetilde{\mathbf{v}}\right)^{\top} [\mathbf{h}^{(\ell-1)},  \mathbf{x}^{(\ell)}] - \widetilde{\mathbf{v}}^{\top} [\mathbf{h}^{(\ell-1)},  \mathbf{x}^{(\ell)}} + \abs{\widetilde{\mathbf{v}}^{\top} [\widetilde{\mathbf{h}}^{(\ell-1)},  \mathbf{x}^{(\ell)}] - \widetilde{\mathbf{v}}^{\top} [\mathbf{h}^{(\ell-1)},  \mathbf{x}^{(\ell)}]} 
			\\&\le \norm{\mathbf{v} - \widetilde{\mathbf{v}}} \norm{[\mathbf{h}^{(\ell-1)},  \mathbf{x}^{(\ell)}]} + \norm{\widetilde{\mathbf{v}}} \norm{\widetilde{\mathbf{h}}^{(\ell-1)} - \mathbf{h}^{(\ell-1)}}
			\\&
			\le \norm{\mathbf{v} - \widetilde{\mathbf{v}}} \norm{[\mathbf{h}^{(\ell-1)},  \mathbf{x}^{(\ell)}]} + \norm{\widetilde{\mathbf{v}} - \mathbf{v}} \norm{\widetilde{\mathbf{h}}^{(\ell-1)} - \mathbf{h}^{(\ell-1)}} + \norm{\mathbf{v}}\norm{\widetilde{\mathbf{h}}^{(\ell-1)} - \mathbf{h}^{(\ell-1)}}.
		\end{align*}
		\endgroup

		We assume that our deep random neural network satisfies $\norm[0]{\mathbf{h}^{(\ell-1)}} \in (\sqrt{2 + (\ell - 2)\epsilon_x^2} - \frac{\rho^2}{\sqrt{m}},$\\$ \sqrt{2 + (\ell - 2)\epsilon_x^2} + 
		\frac{\rho^2}{\sqrt{m}})$ for all $\ell \in [L]$. This happens with probability at least $1-e^{-\Omega(\rho^2)}$ w.r.t. the matrices $\mathbf{W}$ and $\mathbf{A}$ from Lemma~\ref{lemma:norm_ESN}. Thus, provided $m \ge \Omega(\rho^4)$ and $\epsilon_x \le \frac{1}{L}$, $\norm[0]{\mathbf{h}^{(\ell-1)}} \in \left(\sqrt{2}, \sqrt{3}\right)$. Let's call this event $E_7$.
		Using assumption~\ref{ass:variant_v}, event $E_2$ and event $E_7$, we have with probability $1-e^{-\Omega(\rho^2)}$,
		\begin{align*}
			&\abs{g(\widetilde{\mathbf{v}}, [\widetilde{\mathbf{h}}^{(\ell-1)},  \mathbf{x}^{(\ell)}]) - g(\mathbf{v}, [\mathbf{h}^{(\ell-1)},  \mathbf{x}^{(\ell)}])}\\ &\le 
			\norm{\mathbf{v} - \widetilde{\mathbf{v}}} \norm{[\mathbf{h}^{(\ell-1)},  \mathbf{x}^{(\ell)}]} + \norm{\widetilde{\mathbf{v}} - \mathbf{v}} \norm{\widetilde{\mathbf{h}}^{(\ell-1)} - \mathbf{h}^{(\ell-1)}} + \norm{\mathbf{v}}\norm{\widetilde{\mathbf{h}}^{(\ell-1)} - \mathbf{h}^{(\ell-1)}} \\&
			\le \mathcal{O}(\rho^\kappa (N/m)^{\zeta} (2 + \mathcal{O}(\rho^5 (N/m)^{1/2}))) \cdot \norm{\mathbf{v}} + \mathcal{O}(\rho^5 (N/m)^{1/2}) \cdot \norm{\mathbf{v}} \\&
			\le \mathcal{O}(\rho^{5 + \kappa} (N/m)^{1/2}) \cdot \norm{\mathbf{v}}.
		\end{align*}
	\end{proof}
	
	Claims \ref{claim:ff}, \ref{claim:fg}, \ref{claim:ffv} and \ref{claim:gg} hold if the event $E_1 \cap E_2 \cap E_3 \cap E_4 \cap E_5 \cap E_6 \cap E_7$ occurs. This has probability at least $1 - e^{-\Omega(\rho^2)}$. Thus, we have
	\begin{align*}
		&\abs{f_{\mathcal{K}}(\mathbf{v}, \mathbf{h}^{(\ell-1)}, \mathbf{x}^{(\ell)}) - \frac{N}{m}g(\mathbf{v}, [\mathbf{h}^{(\ell-1)}, \mathbf{x}^{(\ell)}])} \\
		&\le  \abs{f_{\mathcal{K}}(\mathbf{v}, \mathbf{h}^{(\ell-1)}, \mathbf{x}^{(\ell)}) - f_{\mathcal{K}}(\mathbf{v}, \widetilde{\mathbf{h}}^{(\ell-1)}, \mathbf{x}^{(\ell)}) } + \abs{f_{\mathcal{K}}(\mathbf{v}, \widetilde{\mathbf{h}}^{(\ell-1)}, \mathbf{x}^{(\ell)}) - f_{\mathcal{K}}(\widetilde{\mathbf{v}}, \widetilde{\mathbf{h}}^{(\ell-1)}, \mathbf{x}^{(\ell)}) }
		\\& \quad  + \abs{f_{\mathcal{K}}(\widetilde{\mathbf{v}}, \widetilde{\mathbf{h}}^{(\ell-1)}, \mathbf{x}^{(\ell)}) - \frac{N}{m}g(\widetilde{\mathbf{v}}, [\widetilde{\mathbf{h}}^{(\ell-1)}, \mathbf{x}^{(\ell)}])}  + \abs{\frac{N}{m}g([\widetilde{\mathbf{v}}, \widetilde{\mathbf{h}}^{(\ell-1)},  \mathbf{x}^{(\ell)}]) - \frac{N}{m}g(\mathbf{v}, [\mathbf{h}^{(\ell-1)},  \mathbf{x}^{(\ell)}])}\\
		&\le  \mathcal{O}\left(\rho^{5 + \kappa} N^{5/3} m^{-7/6} + \rho^{1 + \kappa} (N/m)^{1+\zeta} + \rho^{1+\kappa} N^{1/2} m^{-1}  + \rho^{5 + \kappa} (N/m)^{3/2}\right) \cdot \norm{\mathbf{v}}.
	\end{align*}
\end{proof}

\section{Generalization bounds of Recurrent neural networks}\label{sec:maingensec}
The proof has been structured as follows: In section~\ref{sec:Invert_RNN}, we prove thm.~\ref{thm:Invertibility_ESN} where we show that a linear transformation of $\mathbf{h}^{(L)}$ can give back $[\bx^{(1)}, \ldots, \bx^{(L)}]$. The proof follows from a direct application of lemma~\ref{lemma:singlecell_ESN}. Claim~\ref{clam:stabizable_V} shows that the linear matrix at each induction step satisfies a property of stability necessary for the inductive application of lemma~\ref{lemma:singlecell_ESN}.  

In section~\ref{sec:existence}, we first define a pseudo recurrent neural network that stays close to the over parameterized RNN at initialization throughout SGD. We then show in thm.~\ref{thm:existence_pseudo} that there exists a pseudo network which can approximate the target function in concept class. The proof involves breaking correlations among the hidden states and the weight matrices and then we show that the pseudo network concentrates on the desired signal. The above two steps have been divided among the four intermediate claims: ~\ref{claim:simplifybig}, ~\ref{claim:diffftildef}, ~\ref{claim:difftildefphi} and ~\ref{claim:fbacktildeback}.

In section~\ref{sec:optim_general}, we prove theorem~\ref{thm:main_theorem} which shows that RNNs can attain a population risk similar to the target function in the concept class using SGD. First, we show that the pseudo neural network stays close to RNN with small perturbation around initialization in lemmas~\ref{lemma:perturb_NTK_small_output} and~\ref{lemma:perturb_NTK_small}. We then show that there exists a RNN close to random RNN that can approximate the target function in lemma~\ref{lemma:perturb_small_target}. We complete the argument by showing that the SGD can find matrices with training loss close to the optimal in lemma~\ref{lem:trainloss} and then bounding the Rademacher complexity of RNNs with bounds on the movement in the weight matrices in lemma~\ref{lem:radcomp}.


    \subsection{Invertibility of RNNs at initialization}\label{sec:Invert_RNN}    
    Let $\mathbf{W}^{(k_{b}, k_{e})} = \prod_{k_{b} \ge \ell \ge k_{e}} \mathbf{D}_{(0)}^{(\ell)} \mathbf{W}$, if $k_b \ge k_c$. Otherwise, $\mathbf{W}^{(k_{b}, k_{e})} = \mathbf{I}$. 
Define $\obW^{[\ell]}$ inductively as follows:
\begin{equation*}
	\obW^{[\ell]} = \left[\mathbf{D}^{(\ell)}_{(0)} \mathbf{W} \obW^{[\ell-1]},\text{ } \mathbf{D}_{(0)}^{(\ell)}\mathbf{A}\right]_r, \quad \text{ for } 2 \le \ell \le L,
\end{equation*}
with $\obW^{[1]} = \mathbf{D}^{(1)}_{(0)} \mathbf{A}$.  We can show that $\obW^{[\ell]} = [\mathbf{W}^{(\ell, 2)} \mathbf{D}_{(0)}^{(1)} \mathbf{A}, \mathbf{W}^{(\ell, 3)} \mathbf{D}_{(0)}^{(2)} \mathbf{A}, \cdots, \mathbf{D}_{(0)}^{(\ell)} \mathbf{A}]_r$ for $\ell \ge 2$, which will be helpful for presentation later on.
\begin{theorem}\label{thm:Invertibility_ESN}
	For any $\epsilon_x < \frac{1}{L}$ and a given normalized sequence $\bx^{(1)}, \cdots, \bx^{(L)}$, 
	\begin{align*}
		&\norm{[\bx^{(1)}, \cdots, \bx^{(L)}] - \obW^{[L]\top} \mathbf{h}^{(L)}}_\infty \\&\quad \leq \mathcal{O}\left(L^4 \cdot (\rho^{11} m^{-1/12} + \rho^{7} m^{-1/12} + \rho^{7} m^{-1/4} + \rho^{11} m^{-1/4})  \right) + \mathcal{O}(\rho^2 L^{11/6} \epsilon_x^{5/3})
	\end{align*}
	with probability at least $1 - e^{-\Omega(\rho^2)}$. 
\end{theorem}
\begin{proof}
	The theorem has been restated and proven in theorem~\ref{thm:Invertibility_ESN_proof}.
\end{proof}

\begin{corollary}\label{cor:Invertibility_ESN}
	For a given normalized sequence $\bx$ and any $\varepsilon_x < \frac{1}{L}$, with probability at least $1 - e^{-\Omega(\rho^2)}$ w.r.t. the weights $\mathbf{W}$ and $\mathbf{A}$,
	\begin{align*} 
	&\abs{\overline{\mathbf{W}}^{[L]\top} h^{(L-1)} - \varepsilon_x[\overline{\mathbf{x}}^{(2)}, 
			\cdots, \overline{\mathbf{x}}^{(L-1)}]}  \\&\le \mathcal{O}\left(L^4 \cdot (\rho^{11} m^{-1/12} + \rho^{7} m^{-1/12} + \rho^{7} m^{-1/4} + \rho^{11} m^{-1/4})  \right) + \mathcal{O}(L^{4/3} \varepsilon_x^{2/3}) \\&
		\le  \mathcal{O}(L^4 \rho^{11} m^{-1/12} + \rho^2 L^{11/6} \varepsilon_x^{5/3}), 
	\end{align*}
	where $\overline{{\mathbf{W}}}^{[\ell]}$ is slightly redefined as
	\begin{align*}
		&\obW^{[\ell]} = [\mathbf{D}^{(\ell-1)}_{(0)} \mathbf{W} \obW^{[\ell - 2]},  \mathbf{D}_{(0)}^{(\ell-1)}\mathbf{A}_{[d-1]}]_r, \quad \text{ for all } \ell \ge 4,
	\end{align*}
	with $\obW^{[2]} = \mathbf{D}^{(2)}_{(0)} \mathbf{A}_{[d-1]}$
	and $\mathbf{A}_{[d-1]}  \in \mathbb{R}^{m \times (d-1)}$ denotes the matrix which contains the first $d-1$ columns of the matrix $\mathbf{A}$.
\end{corollary}
\begin{proof}
	The difference from Thm.~\ref{thm:Invertibility_ESN} is that here we attempt to get the first $d-1$ dimensions of the vectors $\mathbf{x}^{(2)}, \cdots, \mathbf{x}^{(L-1)}$. This leads to a small change in the inversion matrix.
\end{proof}

Note that in the above corollary, $\obW^{[L]} = [\mathbf{W}^{(L, 3)} \mathbf{D}_{(0)}^{(2)}\mathbf{A}_{[d-1]}, \mathbf{W}^{(L, 4)} \mathbf{D}_{(0)}^{(3)}\mathbf{A}_{[d-1]}, \cdots, \mathbf{W}^{(L, L)} 
	\mathbf{D}_{(0)}^{(L-1)}\mathbf{A}_{[d-1]}]_r$, where $\mathbf{W}^{(k_b, k_e)} = \prod_{k_b \ge \ell > k_e} \mathbf{D}_{(0)}^{(\ell)} \mathbf{W}$. We are going to use this definition in the following theorems.

    \subsection{Existence of good pseudo network}\label{sec:existence}
    We first define a pseudo RNN model, which is shown later to stay close to the RNN model during the gradient descent dynamics. 

\begin{definition}[Pseudo Recurrent Neural Network]
	Given two matrices $\mathbf{W}^{\ast} \in \mathbb{R}^{m \times m}$ and $\mathbf{A}^{\ast} \in \mathbb{R}^{m \times d}$, the output for a pseudo RNN with activation function $\relu$ for a given sequence $\bx$ are given by
	\begin{equation*}
		F_s^{(\ell)}(\bx; \mathbf{W}^{\ast}, \mathbf{A}^{\ast})   =  \sum_{i \le \ell} \mathbf{Back}_{i \to \ell, s} \mathbf{D}^{(i)} \left(\mathbf{W}^{\ast} \mathbf{h}^{(i-1)} + \mathbf{A}^{\ast} \mathbf{x}^{(i)}\right) \quad \forall 1 \le \ell \le L,  s \in [\dout ],
	\end{equation*}
	where $\mathbf{Back}_{i \to \ell, s} = \mathbf{b}_s^{\top} \mathbf{D}^{(\ell)} \mathbf{W} \cdots \mathbf{D}^{(i+1)} \mathbf{W}$. For typographical simplicity, we will denote $F_s^{(\ell)}(\bx; \mathbf{W}^{\ast}, \mathbf{A}^{\ast})$ as $F_s^{(\ell)}$.
\end{definition}


Now, we show that there exist two matrices $\mathbf{W}^{\ast}$ and $\mathbf{A}^{\ast}$, defined below, such that the pseudo network is close to the concept class under consideration.
\begin{definition}\label{def:existence}
	Define $\mathbf{W}^{\ast}$ and $\mathbf{A}^{\ast}$ as follows.
	\begin{align*}
		\mathbf{W}^{\ast} &= 0 \\
		\mathbf{a}^{*}_{r} &= \frac{\dout }{m} \sum_{s \in [\dout ]} \sum_{r' \in [p]} b_{r, s} b_{r', s}^{\dagger} H_{r', s} \left(\theta_{r', s} \left(\langle \mathbf{w}_{r}, \overline{\mathbf{W}}^{[L]} \mathbf{w}_{r', s}^{\dagger}\rangle\right), \sqrt{m/2} a_{r, d}\right) \mathbf{e}_d, \quad \forall r \in [m],
	\end{align*}
	where
	\begin{equation*} 
		\theta_{r', s} = \frac{\sqrt{m/2}}{\norm[1]{ \overline{\mathbf{W}}^{[L]} \mathbf{w}_{r', s}^{\dagger}}},
	\end{equation*}
	and $\obW^{[L]} = [\mathbf{W}^{(L, 3)} \mathbf{D}_{(0)}^{(2)}\mathbf{A}_{[d-1]}, \mathbf{W}^{(L, 3)} \mathbf{D}_{(0)}^{(2)}\mathbf{A}_{[d-1]}, \cdots, \mathbf{W}^{(L, L)} 
	\mathbf{D}_{(0)}^{(L-1)}\mathbf{A}_{[d-1]}]_r$, where $\mathbf{W}^{(k_b, k_e)} = \prod_{k_b \ge \ell > k_e} \mathbf{D}_{(0)}^{(\ell)} \mathbf{W}$.
\end{definition}

In the following theorem, we show that the pseudo RNN can approximate the target concept class, using the weight $\mathbf{W}^{*}$ and $\mathbf{A}^{\ast}$ define above.
\begin{theorem}[Existence of Good Pseudo Network]\label{thm:existence_pseudo}
	The construction of $\mathbf{W}^{*}$ and $\mathbf{A}^{\ast}$ in Definition~\ref{def:existence} satisfies the following. For every normalized input sequence $\mathbf{x}^{(1)}, \cdots, \mathbf{x}^{(L)}$, we have with probability at least $1-e^{-\Omega\left(\rho^{2}\right)}$ over $\mathbf{W}, \mathbf{A}, \mathbf{B},$ it holds for every $s \in [\dout ]$.
	$$
	\begin{array}{l}
		F_{s}^{(L)} \stackrel{\text { def }}{=} \sum_{i=1}^{L} \mathbf{e}_{s}^{\top} \mathbf{Back}_{i \rightarrow L} D^{(i)} \left(\mathbf{W}^{\ast} \mathbf{h}^{(i-1)} + \mathbf{A}^{\ast} \mathbf{x}^{(i)}\right) \\
		= \sum_{r \in[p]} b_{r, s}^{\dagger} \Phi_{r, s} \left(\left\langle  \mathbf{w}_{r, s}^{\dagger}, [\overline{\mathbf{x}}^{(2)}, \cdots, \overline{\mathbf{x}}^{(L-2)}]\right\rangle\right)\\ \pm \mathcal{O}(\dout Lp\rho^2 \varepsilon + \dout L^{17/6} p \rho^4 L_{\Phi} \varepsilon_x^{2/3} + \dout ^{3/2}L^5 p \rho^{11} L_{\Phi} C_{\Phi}  \mathfrak{C}_{\varepsilon}(\Phi, \mathcal{O}(\varepsilon_x^{-1}))  m^{-1/30} ).
	\end{array}
	$$
\end{theorem}

\begin{proof}
	The theorem has been restated and proven in theorem~\ref{thm:existence_pseudo_proof}.
\end{proof}

\subsection{Optimization and Generalization}\label{sec:optim_general}

First, we show that the training loss decreases with gradient descent. The main component of the proof is to show that overparametrized RNN stays close to its pseudo network throughout training. Since, the pseudo network is a linear network, it is easier to check the trajectory of the pseudo network during gradient descent. Since we have already shown that there exists a pseudo network that can approximate the true function, we can show that gradient descent can find some pseudo network that performs as well as the constructed pseudo network. 
\begin{lemma}[Decrease in training loss]\label{lem:trainloss}
   For a constant $\varepsilon_x = \frac{1}{\operatorname{poly}(\rho)}$ and for every constant $\varepsilon \in \left(0, \frac{1}{p \cdot \operatorname{poly}(\rho) \cdot \mathfrak{C}_{\mathfrak{s}}(\Phi, \mathcal{O}(\varepsilon_x^{-1}))}\right),$ there exists $C^{\prime}=\mathfrak{C}_{\varepsilon}(\Phi, \mathcal{O}(\varepsilon_x^{-1}))$, $C_{\Phi} = \mathfrak{C}_{s}(\Phi, \sqrt{2L})$, and a parameter $\lambda=\Theta\left(\frac{\varepsilon}{L \rho}\right)$
so that, as long as $m \geq \operatorname{poly}\left(C', p, L, \dout , \varepsilon^{-1}\right)$ and $N \geq \Omega\left(\frac{\rho^{3} p C_{\Phi}^2}{\varepsilon^2}\right),$ setting learning rate $\eta=\Theta\left(\frac{1}{\varepsilon \rho^{2} m}\right)$ and
$T=\Theta\left(\frac{p^{2}  C'^2 \mathrm{poly}(\rho)}{\varepsilon^{2}}\right),$ we have
\begin{align*}
\underset{\mathrm{sgd}}{\mathbb{E}}\Big[\frac{1}{T} \sum_{t=0}^{T-1}  \underset{(\obx, \mathbf{y}^{\ast}) \sim \mathcal{Z}}{\mathbb{E}} \mathrm{Obj}(\obx, \mathbf{y}^{\ast};  \mathbf{W}+\mathbf{W}_{t}, \mathbf{A} + \mathbf{A}_t) \Big] \leq \mathrm{OPT} + \frac{\varepsilon}{2} + \frac{1}{\mathrm{poly}(\rho)},
\end{align*}
and $\left\|W_{t}\right\|_{F} \leq \frac{\Delta}{\sqrt{m}}$ for $\Delta=\frac{C'^{2} p^{2} \mathrm { poly }(\rho)}{\varepsilon^{2}}$.
\end{lemma}

\begin{proof}
	The lemma has been restated and proven in lemma~\ref{lem:trainloss_proof}.
\end{proof}

Now, we bound the rademacher complexity of overparametrized RNNs.  The main component of the proof is to use the fact that overparametrized RNN stays close to its pseudo network throughout training. Since, the pseudo network is a linear network, it is easier to compute the rademacher complexity of pseudo network.
\begin{lemma}[Rademacher Complexity of RNNs]\label{lem:radcomp}
   For every $s \in [\dout ]$, we have 
   \begin{align*}
      \underset{\zeta \in \{\pm 1\}^{N}}{\mathbb{E}} &\Big[\sup_{\norm{\mathbf{W}'}_F, \norm{\mathbf{A}'}_F \le \frac{\Delta}{\sqrt{m}}} \frac{1}{N} \sum_{q=1}^{N}  \zeta_q F^{(L)}_{\mathrm{rnn}, s} (\obx_q; \mathbf{W}+\mathbf{W}', \mathbf{A} + \mathbf{A}' ) \Big]  \le \mathcal{O}(\frac{\rho^7 \Delta^{4/3}}{m^{1/6}} + \frac{\rho^2 \Delta}{\sqrt{N}}),
   \end{align*}
   where $\overline{\bx}_1, \ldots, \overline{\bx}_N$ denote the training samples in $\mathcal{D}$.
\end{lemma}
\begin{proof}
    The proof follows the same outline as lemma 8.1 in \cite{allen2019can}. We give the outline here for completeness. From lemma~\ref{lemma:perturb_NTK_small_output}, we have w.p. at least $1 - e^{-\Omega(\rho^2)}$ for all $q \in [N]$, $s \in [\dout ]$ and for any $\mathbf{W}', \mathbf{A}'$ with $\abs{\mathbf{W}'}, \abs{\mathbf{A}'} \le \frac{\Delta}{\sqrt{m}}$,
    \begin{align*}
        \abs{F^{(L)}_{\mathrm{rnn}, s} (\obx_q; \mathbf{W}+\mathbf{W}', \mathbf{A} + \mathbf{A}') -  F^{(L)}_{s} (\obx_q; \mathbf{W}', \mathbf{A}') } \le \mathcal{O}(\rho^7 \Delta^{4/3} m^{-1/6}).
    \end{align*}
    Hence, $F_{\mathrm{rnn}}^{(L)}$ is close to $F^{(L)}$ and thus, its rademacher complexity will be close to that of $F^{(L)}$.  Since, $F^{(L)}$ is a linear network, we can apply the rademacher complexity for linear networks (fact~\ref{fact:radcomp_linear}) to get the final bound.
\end{proof}

Now, we can combine both the theorems above to get the following theorem.
\begin{theorem}\label{thm:main_theorem}
   For a constant $\epsilon_x = \frac{1}{\operatorname{poly}(\rho)}$ and for every constant $\varepsilon \in \left(0, \frac{1}{p \cdot \operatorname{poly}(\rho) \cdot \mathfrak{C}_{\mathfrak{s}}(\Phi, \mathcal{O}(\epsilon_x^{-1}))}\right),$ define complexity $C=\mathfrak{C}_{\varepsilon}(\Phi, \mathcal{O}(\epsilon_x^{-1}))$
and $\lambda=\frac{\varepsilon}{10 L \rho},$ if the number of neurons $m \geq \operatorname{poly}\left(C, p, L, \dout , \varepsilon^{-1}\right)$ and the number of samples is $N \geq \operatorname{poly}\left(C, p, L, \dout , \varepsilon^{-1}\right),$ then $S G D$ with $\eta=\Theta\left(\frac{1}{\varepsilon \rho^{2} m}\right)$ and $T=\Theta(p^{2} C^{2} \operatorname{poly}(\rho)\varepsilon^{-2})$
satisfies that, with probability at least $1-e^{-\Omega\left(\rho^{2}\right)}$ over the random initialization
\begin{align*} 
\underset{\mathrm { sgd }}{\mathbb{E}}\Big[\frac{1}{T} &\sum_{t=0}^{T-1} \underset{\left(\obx, y^{\ast}\right) \sim \mathcal{D}}{\mathbb{E}}\Big[ \mathrm{Obj}\Big(\obx, y^{\ast};  \mathbf{W}_{t},  \mathbf{A}_t\Big)\Big]\Big] \leq \mathrm{OPT}+\varepsilon+\frac{1}{\operatorname{poly}(\rho)}.
\end{align*}
\end{theorem}

\begin{proof}
    The proof follows from Fact~\ref{fact:genRad} using Lemma~\ref{lem:trainloss} and Lemma~\ref{lem:radcomp}.
\end{proof}
\newpage
\section{Invertibility of RNNs at initialization: proofs}
\subsection{Proof of theorem~\ref{thm:Invertibility_ESN}}
\begin{theorem}[Restating theorem~\ref{thm:Invertibility_ESN}]\label{thm:Invertibility_ESN_proof}
	For any $\epsilon_x < \frac{1}{L}$ and a given sequence $\bx^{(1)}, \cdots, \bx^{(L)}$, 
	\begin{align*}
		&\norm{[\bx^{(1)}, \cdots, \bx^{(L)}] - \obW^{[L]\top} \mathbf{h}^{(L)}}_\infty \\&\quad \leq \mathcal{O}\left(L^4 \cdot (\rho^{11} m^{-1/12} + \rho^{7} m^{-1/12} + \rho^{7} m^{-1/4} + \rho^{11} m^{-1/4})  \right) + \mathcal{O}(\rho^2 L^{11/6} \epsilon_x^{5/3})
	\end{align*}
	with probability at least $1 - e^{-\Omega(\rho^2)}$. 
\end{theorem}

\begin{proof}
	Define $\mathbf{V}^{(\ell)}$ inductively as follows:
	\begin{align*}
		\bV^{(1)} &= \left[\mathbf{0}_{d \times m}, \text{ } \mathbf{I}_{d \times d}\right]_r \text{ }  \left[\mathbf{W}, \text{ } \mathbf{A}\right]_r^{\top} \text{ }  \mathbf{D}^{(1)}_{(0)} \\
		\bV^{(\ell)} &= \left[\text{ }\left[\bV^{(\ell - 1)}, \text{ } \mathbf{0}_{(\ell - 1)d \times d}\right]_r,\text{ } \left[\mathbf{0}_{d \times m},\text{ } \mathbf{I}_{d \times d}\right]_r\text{ }\right]_c  \text{ } \left[\mathbf{W}, \text{ } \mathbf{A}\right]_r^{\top} \text{ }  \mathbf{D}^{(\ell)}_{(0)}
	\end{align*}
	
	Now, we show three claims that help us to get the desired inequality.
	\begin{claim}\label{claim:Vhx}
		With probability at least $1 - e^{-\Omega(\rho^2)}$,
		\begin{align*}
			&\norm{[\bx^{(1)}, \cdots, \bx^{(\ell)}] - V^{(\ell)} \mathbf{h}^{(\ell)}}_\infty \\&\leq \mathcal{O}\left(\rho^{11} m^{-1/12} + \rho^{7} m^{-1/12} + \rho^{7} m^{-1/4} + \rho^{11} m^{-1/4}  \right)  \cdot \sum_{k 
				< \ell} \norm{\bV^{(k)}}_{2, \infty} + \mathcal{O}(\ell \rho^2 L^{5/6} \epsilon_x^{5/3}),  
		\end{align*}
		$\text{ for all } \ell \in [L].$
	\end{claim}
	
	\begin{claim}\label{claim:induct_VW}
		\begin{equation*}
			\mathbf{V}^{(\ell)} = \obW^{[\ell]\top}, \quad \text{ for all } \ell \in [L].
		\end{equation*}
	\end{claim}

	\begin{claim}\label{claim:normWbracketedL}
		\begin{equation*}
			\norm{\obW^{[\ell]\top}}_{2, \infty} \le \mathcal{O}(L^3), \quad \text{ for all } \ell \in [L],
		\end{equation*}
		with probability at least $1 - 2 e^{-\rho^2}$.
	\end{claim}

	The above claims have been restated and proven in claims~\ref{claim:Vhx_proof},~\ref{claim:induct_VW_proof} and~\ref{claim:normWbracketedL_proof} respectively. Hence, using claims~\ref{claim:induct_VW},~\ref{claim:Vhx} and ~\ref{claim:normWbracketedL}, we have
	\begingroup \allowdisplaybreaks
	\begin{align*} 
		&\norm{ \obW^{[L]\top}  \mathbf{h}^{(L)} - [\mathbf{x}^{(1)}, \cdots,  \mathbf{x}^{(L)}]}_\infty =  \norm{ \bV^{(L)}  \mathbf{h}^{(L)} - [\mathbf{x}^{(1)}, \cdots,  \mathbf{x}^{(L)}]}_\infty \\& 
		\le \mathcal{O}\left(\rho^{11} m^{-1/12} + \rho^{7} m^{-1/12} + \rho^{7} m^{-1/4} + \rho^{11} m^{-1/4}  \right)  \left( \sum_{\ell < L} \norm{\bV^{(\ell)}}_{2, \infty}\right) + \mathcal{O}(\rho^2 L^{11/6} \epsilon_x^{5/3}) \\& 
		= \mathcal{O}\left(\rho^{11} m^{-1/12} + \rho^{7} m^{-1/12} + \rho^{7} m^{-1/4} + \rho^{11} m^{-1/4}  \right)  \left( \sum_{\ell < L} \norm{\obW^{[\ell]\top}}_{2, \infty}\right) + \mathcal{O}(\rho^2 L^{11/6} \epsilon_x^{5/3})\\& 
		\le \mathcal{O}\left(\rho^{11} m^{-1/12} + \rho^{7} m^{-1/12} + \rho^{7} m^{-1/4} + \rho^{11} m^{-1/4}  \right)  \left( \sum_{\ell < L} \mathcal{O}(L^3)\right) + \mathcal{O}(\rho^2 L^{11/6} \epsilon_x^{5/3}) \\& = \mathcal{O}\left(L^4 \cdot (\rho^{11} m^{-1/12} + \rho^{7} m^{-1/12} + \rho^{7} m^{-1/4} + \rho^{11} m^{-1/4} ) \right)  + \mathcal{O}(\rho^2 L^{11/6} \epsilon_x^{5/3}).
	\end{align*}
	\endgroup
	
\end{proof}

\subsection{Proofs of the helping claims}
The following restatement of Lemma~\ref{lemma:singlecell_ESN} in matrix notation will be useful in the sequel. 
\begin{lemma} \label{cor:inversion_Vx}
	Let $\bV \in \Reals^{k \times (m + d)}$ for $k \geq 1$, such that each row of $\bV$ satisfies assumption~\ref{ass:variant_v} with constants $(\kappa, \zeta)$. For all $\ell \in \{0, 1, \ldots, L-1\}$ and for a given sequence $\bx^{(1)}, \cdots, \bx^{(\ell)}$ we have
	\begin{align*}
		&\norm{\bV [\mathbf{h}^{(\ell-1)}, \mathbf{x}^{(\ell)}] - \bV ([\mathbf{W}, \mathbf{A}])^\top \mathbf{h}^{(\ell)}}_\infty \\&\leq \mathcal{O}\left(\rho^{5 + \kappa} m^{-1/12} + \rho^{1+\kappa} m^{-\zeta/2} + \rho^{1+\kappa} m^{-1/4} + \rho^{5+\kappa} m^{-1/4}  \right) \cdot \norm{\bV}_{2, \infty}, 
	\end{align*}
	with probability at least $1 - ke^{-\Omega(\rho^2)}$.
\end{lemma}


\begin{claim}[Restating claim~\ref{claim:Vhx}]\label{claim:Vhx_proof}
	With probability at least $1 - e^{-\Omega(\rho^2)}$,
	\begin{align*}
		&\norm{[\bx^{(1)}, \cdots, \bx^{(\ell)}] - V^{(\ell)} \mathbf{h}^{(\ell)}}_\infty \\&\leq \mathcal{O}\left(\rho^{11} m^{-1/12} + \rho^{7} m^{-1/12} + \rho^{7} m^{-1/4} + \rho^{11} m^{-1/4}  \right)  \cdot \sum_{k 
			< \ell} \norm{\bV^{(k)}}_{2, \infty} + \mathcal{O}(\ell \rho^2 L^{5/6} \epsilon_x^{5/3}),  
	\end{align*}
	$\text{ for all } \ell \in [L].$
\end{claim}

\begin{proof}
	We prove the claim by induction. For $\ell = 1$, we have 
	\begin{align*}
		\norm{ \mathbf{x}^{(1)} - \bV^{(1)}  \mathbf{h}^{(1)}}_\infty &= \norm{ \left[\mathbf{0}_{d \times m}, \text{ } \mathbf{I}_{d \times d}\right]_r [\mathbf{h}^{(0)}, \mathbf{x}^{(1)}] - \bV^{(1)}  \mathbf{h}^{(1)}}_\infty  
		\\& = \norm{ \left[\mathbf{0}_{d \times m}, \text{ } \mathbf{I}_{d \times d}\right]_r [\mathbf{h}^{(0)}, \mathbf{x}^{(1)}] -  \left[\mathbf{0}_{d \times m}, \text{ } \mathbf{I}_{d \times d}\right]_r  \text{ } \left[\mathbf{W}, \text{ } \mathbf{A}\right]_r^{\top} \text{ } \mathbf{D}^{(1)}_{(0)} \text{ } \mathbf{h}^{(1)}}_\infty
		\\& \leq \norm{ \left[\mathbf{0}_{d \times m}, \text{ } \mathbf{I}_{d \times d}\right]_r [\mathbf{h}^{(0)}, \mathbf{x}^{(1)}] -  \left[\mathbf{0}_{d \times m}, \text{ } \mathbf{I}_{d \times d}\right]_r  \text{ } \left[\mathbf{W}, \text{ } \mathbf{A}\right]_r^{\top} \text{ } \mathbf{D}^{(1)} \text{ } \mathbf{h}^{(1)} }_\infty \\&
		\quad \quad  + \norm{\left[\mathbf{0}_{d \times m}, \text{ } \mathbf{I}_{d \times d}\right]_r  \text{ } \left[\mathbf{W}, \text{ } \mathbf{A}\right]_r^{\top} \text{ } (\mathbf{D}^{(1)}_{(0)} - \mathbf{D}^{(1)}) \text{ } \mathbf{h}^{(1)}}_\infty.
	\end{align*}
	Since by the definition of $\mathbf{D}^{(1)}$, $\mathbf{D}^{(1)} \mathbf{h}^{(1)} = \mathbf{h}^{(1)}$, we have
	\begin{align}
		\norm{ \mathbf{x}^{(1)} - \bV^{(1)}  \mathbf{h}^{(1)}}_\infty &\leq \norm{ \left[\mathbf{0}_{d \times m}, \text{ } \mathbf{I}_{d \times d}\right]_r [\mathbf{h}^{(0)}, \mathbf{x}^{(1)}] -  \left[\mathbf{0}_{d \times m}, \text{ } \mathbf{I}_{d \times d}\right]_r  \text{ } \left[\mathbf{W}, \text{ } \mathbf{A}\right]_r^{\top} \text{ } \mathbf{D}^{(1)} \text{ } \mathbf{h}^{(1)} }_\infty \nonumber\\&
		\quad \quad  + \norm{\left[\mathbf{0}_{d \times m}, \text{ } \mathbf{I}_{d \times d}\right]_r  \text{ } \left[\mathbf{W}, \text{ } \mathbf{A}\right]_r^{\top} \text{ } (\mathbf{D}^{(1)}_{(0)} - \mathbf{D}^{(1)}) \text{ } \mathbf{h}^{(1)}}_\infty \nonumber\\&
		= \norm{ \left[\mathbf{0}_{d \times m}, \text{ } \mathbf{I}_{d \times d}\right]_r [\mathbf{h}^{(0)}, \mathbf{x}^{(1)}] -  \left[\mathbf{0}_{d \times m}, \text{ } \mathbf{I}_{d \times d}\right]_r  \text{ } \left[\mathbf{W}, \text{ } \mathbf{A}\right]_r^{\top} \text{ } \text{ } \mathbf{h}^{(1)} }_\infty \label{eq:induction_V1h1_eq1}\\&
		\quad \quad  + \norm{\left[\mathbf{0}_{d \times m}, \text{ } \mathbf{I}_{d \times d}\right]_r  \text{ } \left[\mathbf{W}, \text{ } \mathbf{A}\right]_r^{\top} \text{ } (\mathbf{D}^{(1)}_{(0)} - \mathbf{D}^{(1)}) \text{ } \mathbf{h}^{(1)}}_\infty. \label{eq:induction_V1h1_eq2}
	\end{align}
	Now, using Lemma~\ref{cor:inversion_Vx}, we can show that Eq.~\ref{eq:induction_V1h1_eq1} is small, i.e. 
	\begin{align*}
		&\norm{ \left[\mathbf{0}_{d \times m}, \text{ } \mathbf{I}_{d \times d}\right]_r [\mathbf{h}^{(0)}, \mathbf{x}^{(1)}] -  \left[\mathbf{0}_{d \times m}, \text{ } \mathbf{I}_{d \times d}\right]_r  \text{ } \left[\mathbf{W}, \text{ } \mathbf{A}\right]_r^{\top} \text{ } \text{ } \mathbf{h}^{(1)}}_\infty  \\&\le \mathcal{O}\left(\rho^{5} m^{-1/12} + \rho m^{-1/4} + \rho^{5} m^{-1/4}  \right)  \cdot \norm{\left[\mathbf{0}_{d \times m}, \text{ } \mathbf{I}_{d \times d}\right]_r}_{2, \infty} \\&
		= \mathcal{O}\left(\rho^{5} m^{-1/12} + \rho m^{-1/4} + \rho^{5} m^{-1/4}  \right),
	\end{align*}
	where we have used the fact that $\left[\mathbf{0}_{d \times m}, \text{ } \mathbf{I}_{d \times d}\right]_r$ doesn't depend on $\mathbf{W}$ and $\mathbf{A}$ and hence each row satisfies assumption~\ref{ass:variant_v} with $(\kappa, \zeta)=(0, 0)$.
	Now, we will show that Eq.~\ref{eq:induction_V1h1_eq2} is small. Note that $\mathbf{x}^{(1)} = \mathbf{x}^{(1)}_{(0)}$ after input normalization. Also, $\mathbf{h}^{(0)}$ is $0$ for any sequence. Thus,
	\begin{align*}
		\mathbf{D}^{(1)}_{(0)} = \mathbf{D}^{(1)}.        
	\end{align*}
	Hence,
	\begin{align*}
		\norm{\left[\mathbf{0}_{d \times m}, \text{ } \mathbf{I}_{d \times d}\right]_r  \text{ } \left[\mathbf{W}, \text{ } \mathbf{A}\right]_r^{\top} \text{ } (\mathbf{D}^{(1)}_{(0)} - \mathbf{D}^{(1)}) \text{ } \mathbf{h}^{(1)}}_\infty = 0
	\end{align*}

	Thus, continuing from Eq.~\ref{eq:induction_V1h1_eq1} and Eq.~\ref{eq:induction_V1h1_eq2}, we have
	\begin{align*}
		\norm{ \mathbf{x}^{(1)} - \bV^{(1)}  \mathbf{h}^{(1)}}_\infty &\leq \mathcal{O}\left(\rho^5 m^{-1/12} + \rho m^{-1/4} + \rho^5 m^{-1/4} \right).
	\end{align*}
	Assuming the claim is true for all $\ell \le  \ell'$, we now try to prove for $\ell = \ell' + 1$.  We have
	\begingroup \allowdisplaybreaks
	\begin{align*}
		&\norm{ \left[\text{ }\left[\bV^{(\ell')}, \text{ } \mathbf{0}_{\ell'd \times d}\right]_r,\text{ } \left[\mathbf{0}_{d \times m},\text{ } \mathbf{I}_{d \times d}\right]_r\text{ }\right]_c [\mathbf{h}^{(\ell')}, \mathbf{x}^{(\ell' + 1)}] - \bV^{(\ell' + 1)}  \mathbf{h}^{(\ell' + 1)}}_\infty \\& 
		\quad \quad \quad \quad = \norm{ \left[\bV^{(\ell')} \mathbf{h}^{(\ell')}, \mathbf{x}^{(\ell' + 1)} \right]- \bV^{(\ell' + 1)}  \mathbf{h}^{(\ell' + 1)}}_\infty \\& 
		\quad \quad \quad \quad
		\ge \norm{ \bV^{(\ell' + 1)}  \mathbf{h}^{(\ell' + 1)} - \left[[\mathbf{x}^{(1)}, \cdots, \mathbf{x}^{(\ell')}], \mathbf{x}^{(\ell' + 1)}\right]}_\infty\\& 
		\quad \quad \quad \quad\quad \quad \quad \quad- \norm{ \left[\bV^{(\ell')} \mathbf{h}^{(\ell')}, \mathbf{x}^{(\ell' + 1)} \right] - \left[[\mathbf{x}^{(1)}, \cdots, \mathbf{x}^{(\ell')}], \mathbf{x}^{(\ell' + 1)}\right]}_\infty\\& 
		\quad \quad \quad \quad
		= \norm{ \bV^{(\ell' + 1)}  \mathbf{h}^{(\ell' + 1)} - [\mathbf{x}^{(1)}, \cdots,  \mathbf{x}^{(\ell' + 1)}]}_\infty\\& 
		\quad \quad \quad \quad\quad \quad \quad \quad- \norm{ \bV^{(\ell')} \mathbf{h}^{(\ell')} - [\mathbf{x}^{(1)}, \cdots, \mathbf{x}^{(\ell')}]}_\infty.
	\end{align*}
	\endgroup
	Thus,
	\begingroup
	\allowdisplaybreaks
	\begin{align}
		&\norm{ \bV^{(\ell' + 1)}  \mathbf{h}^{(\ell' + 1)} - [\mathbf{x}^{(1)}, \cdots,  \mathbf{x}^{(\ell' + 1)}]}_\infty \nonumber\\&
		\le \norm{ \left[\text{ }\left[\bV^{(\ell')}, \text{ } \mathbf{0}_{\ell'd \times d}\right]_r,\text{ } \left[\mathbf{0}_{d \times m},\text{ } \mathbf{I}_{d \times d}\right]_r\text{ }\right]_c [\mathbf{h}^{(\ell')}, \mathbf{x}^{(\ell' + 1)}] - \bV^{(\ell' + 1)}  \mathbf{h}^{(\ell' + 1)}}_\infty \label{eq:induction_Vl_1hl_1} \\& \quad \quad \quad \quad +  \norm{ \bV^{(\ell')} \mathbf{h}^{(\ell')} - [\mathbf{x}^{(1)}, \cdots, \mathbf{x}^{(\ell')}]}_\infty \label{eq:induction_Vlhl}
	\end{align}
	\endgroup
	Using induction, we have in Eq.~\ref{eq:induction_Vlhl}, 
	\begin{align*}
		&\norm{ \bV^{(\ell')} [\mathbf{h}^{(\ell'-1)}, \mathbf{x}^{(\ell')}] - [\mathbf{x}^{(1)}, \cdots, \mathbf{x}^{(\ell')}]}_\infty \\&\le \mathcal{O}\left(\rho^{11} m^{-1/12} + \rho^{7} m^{-1/12} + \rho^{7} m^{-1/4} + \rho^{11} m^{-1/4}  \right)  \cdot \sum_{k < \ell'} \norm{\bV^{(k)}}_{2, \infty} + \mathcal{O}(\ell' \rho^2 L^{5/6} \epsilon_x^{5/3}).
	\end{align*} 
	Now, we show that Eq.~\ref{eq:induction_Vl_1hl_1} is small. 
	\begingroup\allowdisplaybreaks
	\begin{align}
		&\bV^{(\ell' + 1)}  \mathbf{h}^{(\ell' + 1)} \nonumber\\&
		= \left[\text{ }\left[\bV^{(\ell')}, \text{ } \mathbf{0}_{\ell'd \times d}\right]_r,\text{ } \left[\mathbf{0}_{d \times m},\text{ } \mathbf{I}_{d \times d}\right]_r\text{ }\right]_c  \text{ } \left[\mathbf{W}, \text{ } \mathbf{A}\right]_r^{\top} \text{ }  \mathbf{D}^{(\ell'+1)}_{(0)} \mathbf{h}^{(\ell' + 1)} \nonumber\\&
		= \left[\text{ }\left[\bV^{(\ell')}, \text{ } \mathbf{0}_{\ell'd \times d}\right]_r,\text{ } \left[\mathbf{0}_{d \times m},\text{ } \mathbf{I}_{d \times d}\right]_r\text{ }\right]_c  \text{ } \left[\mathbf{W}, \text{ } \mathbf{A}\right]_r^{\top} \text{ }  (\mathbf{D}^{(\ell'+1)}_{(0)} - \mathbf{D}^{(\ell'+1)}) \mathbf{h}^{(\ell' + 1)}\nonumber \\&
		\quad \quad \quad + \left[\text{ }\left[\bV^{(\ell')}, \text{ } \mathbf{0}_{\ell'd \times d}\right]_r,\text{ } \left[\mathbf{0}_{d \times m},\text{ } \mathbf{I}_{d \times d}\right]_r\text{ }\right]_c  \text{ } \left[\mathbf{W}, \text{ } \mathbf{A}\right]_r^{\top} \text{ }  \mathbf{D}^{(\ell'+1)} \mathbf{h}^{(\ell' + 1)} \nonumber\\&
		= \left[\text{ }\left[\bV^{(\ell')}, \text{ } \mathbf{0}_{\ell'd \times d}\right]_r,\text{ } \left[\mathbf{0}_{d \times m},\text{ } \mathbf{I}_{d \times d}\right]_r\text{ }\right]_c  \text{ } \left[\mathbf{W}, \text{ } \mathbf{A}\right]_r^{\top} \text{ }  (\mathbf{D}^{(\ell'+1)}_{(0)} - \mathbf{D}^{(\ell'+1)}) \mathbf{h}^{(\ell' + 1)} \label{eq:induction_Vl1h1_eq1}\\&
		\quad \quad \quad + \left[\text{ }\left[\bV^{(\ell')}, \text{ } \mathbf{0}_{\ell'd \times d}\right]_r,\text{ } \left[\mathbf{0}_{d \times m},\text{ } \mathbf{I}_{d \times d}\right]_r\text{ }\right]_c  \text{ } \left[\mathbf{W}, \text{ } \mathbf{A}\right]_r^{\top} \text{ }  \mathbf{h}^{(\ell' + 1)}, \label{eq:induction_Vl1h1_eq2}
	\end{align}
	\endgroup
	where in the final step, we have used the definition of $\mathbf{D}^{(\ell' + 1)}$ to get $\mathbf{h}^{(\ell' + 1)} =  \mathbf{D}^{(\ell'+1)} \mathbf{h}^{(\ell' + 1)}$. First, we will focus on Eq.~\ref{eq:induction_Vl1h1_eq2}. Using Lemma~\ref{cor:inversion_Vx}, we have
	\begin{align}
		&\norm{\left[\text{ }\left[\bV^{(\ell')}, \text{ } \mathbf{0}_{\ell'd \times d}\right]_r,\text{ } \left[\mathbf{0}_{d \times m},\text{ } \mathbf{I}_{d \times d}\right]_r\text{ }\right]_c \left[\mathbf{W},  \mathbf{A}\right]_r^{\top}   \mathbf{h}^{(\ell' + 1)} - \left[\text{ }\left[\bV^{(\ell')}, \text{ } \mathbf{0}_{\ell'd \times d}\right]_r, \left[\mathbf{0}_{d \times m},\text{ } \mathbf{I}_{d \times d}\right]_r\text{ }\right]_c \left[\mathbf{h}^{(\ell')}, \mathbf{x}^{(\ell'+1)} \right]}_{2, \infty} \nonumber \\&
		\le   \mathcal{O}\left(\rho^{11} m^{-1/12} + \rho^{7} m^{-1/12} + \rho^{7} m^{-1/4} + \rho^{11} m^{-1/4}  \right) \cdot \norm{\left[\left[\bV^{(\ell')}, \text{ } \mathbf{0}_{\ell'd \times d}\right]_r,\text{ } \left[\mathbf{0}_{d \times m},\text{ } \mathbf{I}_{d \times d}\right]_r\text{ }\right]_c}_{2, \infty} \nonumber\\&
		=  \mathcal{O}\left(\rho^{11} m^{-1/12} + \rho^{7} m^{-1/12} + \rho^{7} m^{-1/4} + \rho^{11} m^{-1/4}  \right)  \cdot \norm{\bV^{(\ell')}}_{2, \infty}. \label{eq:lemma41Vl1}
	\end{align}
	In the above steps, we have used Claim~\ref{clam:stabizable_V} to show that the rows of the matrix $\left[\left[\bV^{(\ell')}, \text{ } \mathbf{0}_{\ell'd \times d}\right]_r,\text{ } \left[\mathbf{0}_{d \times m},\text{ } \mathbf{I}_{d \times d}\right]_r\text{ }\right]_c$ satisfies assumption~\ref{ass:variant_v} with $(\kappa, \zeta)=(6, \frac{1}{6})$.
	Next, we show that Eq.~\ref{eq:induction_Vl1h1_eq1} is small, i.e.
	\begingroup \allowdisplaybreaks
	\begin{align}
		&\norm{\left[\text{ }\left[\bV^{(\ell')}, \text{ } \mathbf{0}_{\ell'd \times d}\right]_r,\text{ } \left[\mathbf{0}_{d \times m},\text{ } \mathbf{I}_{d \times d}\right]_r\text{ }\right]_c  \text{ } \left[\mathbf{W}, \text{ } \mathbf{A}\right]_r^{\top} \text{ }  (\mathbf{D}^{(\ell'+1)}_{(0)} - \mathbf{D}^{(\ell'+1)}) \mathbf{h}^{(\ell' + 1)}}_\infty \nonumber\\&= 
		\norm{\left[\text{ }\left[\bV^{(\ell')}, \text{ } \mathbf{0}_{\ell'd \times d}\right]_r,\text{ } \left[\mathbf{0}_{d \times m},\text{ } \mathbf{I}_{d \times d}\right]_r\text{ }\right]_c  \text{ } \left[\mathbf{W}, \text{ } \mathbf{A}\right]_r^{\top} \text{ } (\mathbf{D}^{(\ell'+1)}_{(0)} - \mathbf{D}^{(\ell'+1)}) \text{ } \mathbf{D}^{(\ell'+1)} \text{ } \left[\mathbf{W}, \text{ } \mathbf{A}\right]_r \left[\mathbf{h}^{(\ell')}, \mathbf{x}^{(\ell'+1)}\right] }_\infty \nonumber
		\\&= \max_{i \in [d]} \abs{ \left[\text{ }\left[\bV^{(\ell')}, \text{ } \mathbf{0}_{\ell'd \times d}\right]_r,\text{ } \left[\mathbf{0}_{d \times m},\text{ } \mathbf{I}_{d \times d}\right]_r\text{ }\right]_{1, i}^{\top}  \text{ } \left[\mathbf{W}, \text{ } \mathbf{A}\right]_r^{\top} \text{ } (\mathbf{D}^{(\ell'+1)}_{(0)} - \mathbf{I}) \text{ } \mathbf{D}^{(\ell'+1)} \text{ } \left[\mathbf{W}, \text{ } \mathbf{A}\right]_r \left[\mathbf{h}^{(\ell')}, \mathbf{x}^{(\ell'+1)}\right]} \nonumber\\&
		= \max_{i \in [d]} \abs{\sum_{k \in [m]} \left\langle \left[\mathbf{w}_k, \mathbf{a}_k\right], \left[\text{ }\left[\bV^{(\ell')}, \text{ } \mathbf{0}_{\ell'd \times d}\right]_r,\text{ } \left[\mathbf{0}_{d \times m},\text{ } \mathbf{I}_{d \times d}\right]_r\text{ }\right]_{1, i}\right\rangle d^{(\ell'+1)}_{kk} (d^{(\ell'+1)}_{(0), kk} - 1) \left\langle \left[\mathbf{w}_k, \mathbf{a}_k\right],  \left[\mathbf{h}^{(\ell')}, \mathbf{x}^{(\ell'+1)}\right] \right\rangle} \nonumber
		\\&\le \max_{i \in [d]} \norm{\left[\mathbf{W}, \text{ } \mathbf{A}\right]_r \left[\text{ }\left[\bV^{(\ell')}, \text{ } \mathbf{0}_{\ell'd \times d}\right]_r,\text{ } \left[\mathbf{0}_{d \times m},\text{ } \mathbf{I}_{d \times d}\right]_r\text{ }\right]_{1, i}}_{\infty} \nonumber \\& \quad\quad\quad\quad \cdot 
		\text{ }  \left( \sum_{k \in [m]} d^{(\ell'+1)}_{kk} (1 - d^{(\ell'+1)}_{(0), kk}) \abs{\left\langle \left[\mathbf{w}_k, \mathbf{a}_k\right],  \left[\mathbf{h}^{(\ell')}, \mathbf{x}^{(\ell'+1)}\right] \right\rangle} \right) \label{eq:eq15small1step},
	\end{align}
	\endgroup
	where we use cauchy-schwartz inequality in the final step. First note that,
	\begin{align*}
		&\max_{i \in [d]} \norm{\left[\mathbf{W}, \text{ } \mathbf{A}\right]_r \left[\text{ }\left[\bV^{(\ell')}, \text{ } \mathbf{0}_{\ell'd \times d}\right]_r,\text{ } \left[\mathbf{0}_{d \times m},\text{ } \mathbf{I}_{d \times d}\right]_r\text{ }\right]_{1, i}}_{\infty}\\&
		= \norm{\left[\mathbf{W}, \text{ } \mathbf{A}\right]_r \left[\text{ }\left[\bV^{(\ell')}, \text{ } \mathbf{0}_{\ell'd \times d}\right]_r,\text{ } \left[\mathbf{0}_{d \times m},\text{ } \mathbf{I}_{d \times d}\right]_r\text{ }\right]_{1}^{\top}}_{\infty, \infty} \\& 
		= \norm{\text{ }\left[\mathbf{W}\bV^{(\ell')\top},  \mathbf{A}\right]_{2}}_{\infty, \infty}  \\&
		\le \max\left(\norm{\mathbf{W}\bV^{(\ell')\top}}_{\infty, \infty}, \text{ } \norm{\mathbf{A}}_{\infty, \infty}\right).
	\end{align*}
	First, note that using fact~\ref{fact:max_gauss}, we have
	\begin{align*}
		\norm{\mathbf{A}}_{\infty, \infty} \le \frac{\rho}{\sqrt{m}},
	\end{align*}
	with probability at least $1 - m^2e^{\rho^2/2}$, which is equal to $1 - e^{-\Omega(\rho^2)}$, since we are using $\rho = 100L\dout \log m$. Also, from claim~\ref{claim:inftyinftyWV}, we have
	\begin{align*}
		\norm{\mathbf{W}\bV^{(\ell')\top}}_{\infty, \infty} \le \mathcal{O}(\frac{\rho}{\sqrt{m}}),
	\end{align*}
	with probability at least $1 - e^{-\Omega(\rho^2)}$.
	Hence, 
	\begin{align*}
		&\max_{i \in [d]} \norm{\left[\mathbf{W}, \text{ } \mathbf{A}\right]_r \left[\text{ }\left[\bV^{(\ell')}, \text{ } \mathbf{0}_{\ell'd \times d}\right]_r,\text{ } \left[\mathbf{0}_{d \times m},\text{ } \mathbf{I}_{d \times d}\right]_r\text{ }\right]_{1, i}}_{\infty}\\&
		\le \norm{\text{ }\left[\mathbf{W}\bV^{(\ell')\top},  \mathbf{A}\right]_{2}}_{\infty, \infty}  \\&
		\le \max\left(\norm{\mathbf{W}\bV^{(\ell')\top}}_{\infty, \infty}, \text{ } \norm{\mathbf{A}}_{\infty, \infty}\right) \\&
		\le \mathcal{O}(\frac{\rho}{\sqrt{m}}).
	\end{align*}
	By the definition of $\mathbf{D}^{(\ell' + 1)}$ and $\mathbf{D}^{(\ell'+1)}_{(0)}$, we can see that
	\begingroup \allowdisplaybreaks
	\begin{align*}
		\sum_{k \in [m]} d^{(\ell'+1)}_{kk} (1 - d^{(
			\ell'+1)}_{(0), kk}) &= \abs{\left\{k \in [m] \text{ } : \text{ }  d^{(\ell'+1)}_{kk} \text{ }  =\text{ }  1 \text{ } \And \text{ }  d^{(\ell'+1)}_{(0), kk} \text{ }  = \text{ }  0 \right\}} \\&
		\le \abs{\left\{k \in [m] \text{ } : \text{ }  d^{(\ell'+1)}_{kk} \text{ } \ne   d^{(\ell'+1)}_{(0), kk} \text{ } \right\}} \\&
		= \norm{\mathbf{D}^{(\ell'+1)} - \mathbf{D}^{(\ell'+1)}_{(0)}}_{0} \\& \le \mathcal{O}(L^{5/6} \epsilon_x^{5/3} m),
	\end{align*}
	\endgroup
	where we use Lemma~\ref{lemma:norm_ESN} in the final step. Now, we focus on 
	\begin{align*}
		\max_{k \in [m]} \abs{\left\langle \left[\mathbf{w}_k, \mathbf{a}_k\right],  \left[\mathbf{h}^{(\ell')}, \mathbf{x}^{(\ell'+1)}\right] \right\rangle} d^{(\ell'+1)}_{kk} (1 - d^{(\ell'+1)}_{(0), kk}).
	\end{align*}
	First, we note that
	\begin{align*}
		\abs{k \in [m] \text{ } : \text{ } d^{(\ell'+1)}_{kk} (1 - d^{(\ell'+1)}_{(0), kk}) = 1} = \abs{\left\{k \in [m] \text{ } : \text{ }  d^{(\ell'+1)}_{kk} \text{ }  =\text{ }  1 \text{ } \And \text{ }  d^{(\ell'+1)}_{(0), kk} \text{ }  = \text{ }  0 \right\}}.
	\end{align*}
	Hence,
	\begin{align*}
		&\abs{k \in [m] \text{ } : \text{ } d^{(\ell'+1)}_{kk} (1 - d^{(\ell'+1)}_{(0), kk}) = 1} \\&= \abs{\left\{k \in [m] \text{ } : \text{ } \left\langle \left[\mathbf{w}_k, \mathbf{a}_k\right],  \left[\mathbf{h}^{(\ell')}, \mathbf{x}^{(\ell'+1)}\right] \right\rangle \le 0 \text{ and }  \left\langle \left[\mathbf{w}_k, \mathbf{a}_k\right],  \left[\mathbf{h}_{(0)}^{(\ell')}, \mathbf{x}^{(\ell'+1)}_{(0)}\right] \right\rangle \ge 0 \right\}}.
	\end{align*}
	This implies that for the set $S_{m}= \abs{k \in [m] \text{ } : \text{ } d^{(\ell'+1)}_{kk} (1 - d^{(\ell'+1)}_{(0), kk}) = 1}$,
	\begin{align*}
		\abs{\left\langle \left[\mathbf{w}_k, \mathbf{a}_k\right],  \left[\mathbf{h}^{(\ell')}, \mathbf{x}^{(\ell'+1)}\right] \right\rangle} \le \abs{\left\langle \left[\mathbf{w}_k, \mathbf{a}_k\right],  \left[ \mathbf{h}^{(\ell')} - \mathbf{h}_{(0)}^{(\ell')}, \mathbf{x}^{(\ell'+1)} - \mathbf{x}^{(\ell'+1)}_{(0)}\right] \right\rangle}, \text{ for all } k \in S_m.
	\end{align*}
	From lemma~\ref{lemma:norm_ESN}, we have with probability $1-e^{-\Omega(\rho^2)}$,
	\begin{align*}
		\max_{k \in [m]} \abs{\left\langle \left[\mathbf{w}_k, \mathbf{a}_k\right],  \left[ \mathbf{h}^{(\ell')} - \mathbf{h}_{(0)}^{(\ell')}, \mathbf{x}^{(\ell'+1)} - \mathbf{x}^{(\ell'+1)}_{(0)}\right] \right\rangle} \le \mathcal{O}(\rho \sqrt{L} \epsilon_x m^{-1/2}).
	\end{align*}
	Thus, 
	\begin{align*}
		&\sum_{k \in [m]} d^{(\ell'+1)}_{kk} (1 - d^{(\ell'+1)}_{(0), kk}) \abs{\left\langle \left[\mathbf{w}_k, \mathbf{a}_k\right],  \left[\mathbf{h}^{(\ell')}, \mathbf{x}^{(\ell'+1)}\right] \right\rangle}  \\&
		\le \left(\sum_{k \in [m]} d^{(\ell'+1)}_{kk} (1 - d^{(\ell'+1)}_{(0), kk}) \right) \cdot  \left( \max_{k \in [m]} \abs{\left\langle \left[\mathbf{w}_k, \mathbf{a}_k\right],  \left[\mathbf{h}^{(\ell')}, \mathbf{x}^{(\ell'+1)}\right] \right\rangle} d^{(\ell'+1)}_{kk} (1 - d^{(\ell'+1)}_{(0), kk}) \right) \\&
		\le \left(\sum_{k \in [m]} d^{(\ell'+1)}_{kk} (1 - d^{(\ell'+1)}_{(0), kk}) \right) \cdot  \left( \max_{k \in S_m} \abs{\left\langle \left[\mathbf{w}_k, \mathbf{a}_k\right],  \left[\mathbf{h}^{(\ell')}, \mathbf{x}^{(\ell'+1)}\right] \right\rangle}  \right) \\&
		\le  \left(\sum_{k \in [m]} d^{(\ell'+1)}_{kk} (1 - d^{(\ell'+1)}_{(0), kk}) \right) \cdot  \left(\max_{k \in S_m}  \abs{\left\langle \left[\mathbf{w}_k, \mathbf{a}_k\right],  \left[ \mathbf{h}^{(\ell')} - \mathbf{h}_{(0)}^{(\ell')}, \mathbf{x}^{(\ell'+1)} - \mathbf{x}^{(\ell'+1)}_{(0)}\right] \right\rangle}  \right) \\&
		\le \mathcal{O}(L^{5/6} \epsilon_x^{2/3} m) \cdot \mathcal{O}(\rho \sqrt{L} \epsilon_x m^{-1/2}) = \mathcal{O}(\rho L^{5/6} \epsilon_x^{5/3} \sqrt{m}).
	\end{align*}
	Thus, finally Eq.~\ref{eq:eq15small1step} boils down to
	\begingroup \allowdisplaybreaks
	\begin{align*}
		&\norm{\left[\text{ }\left[\bV^{(\ell')}, \text{ } \mathbf{0}_{\ell'd \times d}\right]_r,\text{ } \left[\mathbf{0}_{d \times m},\text{ } \mathbf{I}_{d \times d}\right]_r\text{ }\right]_c  \text{ } \left[\mathbf{W}, \text{ } \mathbf{A}\right]_r^{\top} \text{ }  (\mathbf{D}^{(\ell'+1)}_{(0)} - \mathbf{D}^{(\ell'+1)}) \mathbf{h}^{(\ell' + 1)}}_\infty 
		\\ & \le \max_{i \in [d]} \norm{\left[\mathbf{W}, \text{ } \mathbf{A}\right]_r \left[\text{ }\left[\bV^{(\ell')}, \text{ } \mathbf{0}_{\ell'd \times d}\right]_r,\text{ } \left[\mathbf{0}_{d \times m},\text{ } \mathbf{I}_{d \times d}\right]_r\text{ }\right]_{1, i}}_{\infty} \\& \quad\quad\quad\quad \cdot \text{ }  \left( \sum_{k \in [m]} d^{(\ell'+1)}_{kk} (1 - d^{(\ell'+1)}_{(0), kk}) \abs{\left\langle \left[\mathbf{w}_k, \mathbf{a}_k\right],  \left[\mathbf{h}^{(\ell')}, \mathbf{x}^{(\ell'+1)}\right] \right\rangle} \right) \\&
		\le \mathcal{O}(\frac{\rho}{\sqrt{m}}) \cdot \mathcal{O}(\frac{\rho}{\sqrt{m}}) \cdot \mathcal{O}(\rho L^{5/6} \epsilon_x^{5/3} \sqrt{m}) \\& = \mathcal{O}(\rho^2 L^{5/6} \epsilon_x^{5/3}).
	\end{align*}
	\endgroup

	Hence, continuing from Eq.~\ref{eq:induction_Vl1h1_eq1} and Eq.~\ref{eq:induction_Vl1h1_eq2}, we have
	\begingroup\allowdisplaybreaks
	\begin{align*}
		&\bV^{(\ell' + 1)}  \mathbf{h}^{(\ell' + 1)} \nonumber\\&
		= \left[\text{ }\left[\bV^{(\ell')}, \text{ } \mathbf{0}_{\ell'd \times d}\right]_r,\text{ } \left[\mathbf{0}_{d \times m},\text{ } \mathbf{I}_{d \times d}\right]_r\text{ }\right]_c  \text{ } \left[\mathbf{W}, \text{ } \mathbf{A}\right]_r^{\top} \text{ }  (\mathbf{D}^{(\ell'+1)}_{(0)} - \mathbf{D}^{(\ell'+1)}) \mathbf{h}^{(\ell' + 1)} \\&
		\quad \quad \quad + \left[\text{ }\left[\bV^{(\ell')}, \text{ } \mathbf{0}_{\ell'd \times d}\right]_r,\text{ } \left[\mathbf{0}_{d \times m},\text{ } \mathbf{I}_{d \times d}\right]_r\text{ }\right]_c  \text{ } \left[\mathbf{W}, \text{ } \mathbf{A}\right]_r^{\top} \text{ }  \mathbf{h}^{(\ell' + 1)} \\&
		=  \left[\text{ }\left[\bV^{(\ell')}, \text{ } \mathbf{0}_{\ell'd \times d}\right]_r, \left[\mathbf{0}_{d \times m},\text{ } \mathbf{I}_{d \times d}\right]_r\text{ }\right]_c \left[\mathbf{h}^{(\ell')}, \mathbf{x}^{(\ell'+1)} \right] 
		\\& \pm \mathcal{O}\left(\rho^{11} m^{-1/12} + \rho^{7} m^{-1/12} + \rho^{7} m^{-1/4} + \rho^{11} m^{-1/4}  \right)  \cdot \norm{\bV^{(\ell')}}_{2, \infty} \\&
		\pm \mathcal{O}(\rho^2 L^{5/6} \epsilon_x^{5/3}).
	\end{align*}
	\endgroup
	Hence, from Eq.~\ref{eq:induction_Vl_1hl_1} and Eq.~\ref{eq:induction_Vlhl}, we have
	\begingroup\allowdisplaybreaks
	\begin{align*}
		&\norm{ \bV^{(\ell' + 1)}  \mathbf{h}^{(\ell' + 1)} - [\mathbf{x}^{(1)}, \cdots,  \mathbf{x}^{(\ell' + 1)}]}_\infty \\
		&\le \norm{ \left[\text{ }\left[\bV^{(\ell')}, \text{ } \mathbf{0}_{\ell'd \times d}\right]_r,\text{ } \left[\mathbf{0}_{d \times m},\text{ } \mathbf{I}_{d \times d}\right]_r\text{ }\right]_c [\mathbf{h}^{(\ell')}, \mathbf{x}^{(\ell' + 1)}] - \bV^{(\ell' + 1)}  \mathbf{h}^{(\ell' + 1)}}_\infty  \\& \quad \quad \quad \quad +  \norm{ \bV^{(\ell')} \mathbf{h}^{(\ell')} - [\mathbf{x}^{(1)}, \cdots, \mathbf{x}^{(\ell')}]}_\infty \\&
		\le \mathcal{O}\left(\rho^{11} m^{-1/12} + \rho^{7} m^{-1/12} + \rho^{7} m^{-1/4} + \rho^{11} m^{-1/4}  \right)  \cdot \sum_{k < \ell'} \norm{\bV^{(k)}}_{2, \infty} + \mathcal{O}(\ell' \rho^2 L^{5/6} \epsilon_x^{5/3})
		\\& +\mathcal{O}\left(\rho^{11} m^{-1/12} + \rho^{7} m^{-1/12} + \rho^{7} m^{-1/4} + \rho^{11} m^{-1/4}  \right)  \cdot \norm{\bV^{(\ell')}}_{2, \infty} + \mathcal{O}(L^{5/6} \rho^2 \epsilon_x^{5/3}) \\&
		= \mathcal{O}\left(\rho^{11} m^{-1/12} + \rho^{7} m^{-1/12} + \rho^{7} m^{-1/4} + \rho^{11} m^{-1/4}  \right)  \cdot \sum_{k 
			\le \ell'} \norm{\bV^{(k)}}_{2, \infty} + \mathcal{O}((\ell' + 1) \rho^2 L^{5/6} \epsilon_x^{5/3}).
	\end{align*}
	\endgroup
	Thus, the claim follows by induction.
	
\end{proof}

\begin{claim}[Restating claim~\ref{claim:induct_VW}]\label{claim:induct_VW_proof}
	\begin{equation*}
		\mathbf{V}^{(\ell)} = \obW^{[\ell]\top}, \quad \text{ for all } \ell \in [L].
	\end{equation*}
\end{claim}

\begin{proof}
	We prove the claim by induction. For $\ell = 1$,
	\begin{equation*}
		\bV^{(1)} = \left[\mathbf{0}_{d \times m}, \text{ } \mathbf{I}_{d \times d}\right]_r \text{ } \left[\mathbf{W}, \text{ } \mathbf{A}\right]_r^{\top} \mathbf{D}_{(0)}^{(1)} = \mathbf{A}^{\top} \mathbf{D}_{(0)}^{(1)} := \obW^{
			[1]\top}.
	\end{equation*}
	Assuming the claim holds true for for all $\ell \le \ell'$, we now prove for $\ell = \ell' + 1$.
	\begin{align*}
		\bV^{(\ell' + 1)} &= \left[\text{ }\left[\bV^{(\ell')}, \text{ } \mathbf{0}_{(\ell')d \times d}\right]_r,\text{ } \left[\mathbf{0}_{d \times m},\text{ } \mathbf{I}_{d \times d}\right]_r\text{ }\right]_c \text{ }\left[\mathbf{W}, \text{ } \mathbf{A}\right]_r^{\top} \mathbf{D}_{(0)}^{(\ell'+1)}\\
		&= \left[\text{ }\left[\bV^{(\ell')}, \text{ } \mathbf{0}_{(\ell')d \times d}\right]_r \left[\mathbf{W}, \text{ } \mathbf{A}\right]_r^{\top} \mathbf{D}_{(0)}^{(\ell'+1)} \text{ },  \text{ } \left[\mathbf{0}_{d \times m},\text{ } \mathbf{I}_{d \times d}\right]_r\text{ } \left[\mathbf{W}, \text{ } \mathbf{A}\right]_r^{\top} \mathbf{D}_{(0)}^{(\ell'+1)}\right]_c\\
		&= \left[\text{ }\bV^{(\ell')} \mathbf{W}^{\top}\mathbf{D}_{(0)}^{(\ell'+1)},  \text{ } \mathbf{A}^{\top}\mathbf{D}_{(0)}^{(\ell'+1)}\right]_c\\
		&= \left[\obW^{[\ell']\top}\mathbf{W}^{\top}\mathbf{D}_{(0)}^{(\ell'+1)},  \text{ } \mathbf{A}^{\top}\mathbf{D}_{(0)}^{(\ell'+1)}\right]_c := \obW^{[\ell' + 1]\top},
	\end{align*}
	where in the pre-final step, we use induction argument for $\ell = \ell'$. Hence, the claim follows from induction.
\end{proof}

\begin{claim}\label{claim:inftyinftyWV}
	With probability $1 - e^{-\Omega(\rho^2)}$, for all $\ell \in [L]$,
	\begin{align*}
		\norm{\mathbf{W} \mathbf{V}^{(\ell)\top}}_{\infty, \infty} \le \mathcal{O}(\frac{\rho}{\sqrt{m}}).
	\end{align*}
\end{claim}

\begin{proof}
	Fix an $\ell \in [L]$.
	From claim~\ref{claim:induct_VW}, we have
	\begin{align*}
		\mathbf{V}^{(\ell)\top} = \obW^{[\ell]}.
	\end{align*}
	Thus,
	\begin{align*}
		\norm{\mathbf{W} \mathbf{V}^{(\ell)\top}}_{\infty, \infty} &= \norm{\mathbf{W}  \obW^{[\ell]}}_{\infty, \infty}\\&
		\le \max_{k \le \ell} \norm{\mathbf{W}  \mathbf{W}^{(\ell, k+1)}}_{\infty, \infty} \\&
		=  \max_{k \le \ell} \norm{\mathbf{W} \mathbf{D}^{(\ell)} \mathbf{W} \cdots \mathbf{D}^{(k+1)} \mathbf{W} \mathbf{D}^{(k)} \mathbf{A}}_{\infty, \infty} \\&
		\le \mathcal{O}(\frac{\rho}{\sqrt{m}}),
	\end{align*}
	with probability at least $1 - e^{-\Omega(\rho^2)}$. Here, we use lemma~\ref{lemma:norm_ESN} in the final step that says
	\begin{align*}
		\norm{\mathbf{W} \mathbf{D}^{(j)} \mathbf{W} \cdots \mathbf{D}^{(i+1)} \mathbf{W} \mathbf{D}^{(i)} \mathbf{A}}_{\infty, \infty} \le \mathcal{O}(\frac{\rho}{\sqrt{m}}),
	\end{align*}
	with high probability for any $1 \le i \le j \le L$.
\end{proof}

\begin{claim}[Restating claim~\ref{claim:normWbracketedL}]\label{claim:normWbracketedL_proof}
	\begin{equation*}
		\norm{\obW^{[\ell]\top}}_{2, \infty} \le \mathcal{O}(L^3), \quad \text{ for all } \ell \in [L],
	\end{equation*}
	with probability at least $1 - 2 e^{-\rho^2}$.
\end{claim}

\begin{proof}
	Since $\obW^{[\ell]} = [\mathbf{W}^{(\ell)} \mathbf{A}, \mathbf{W}^{(\ell-1)} \mathbf{A}, \cdots, \mathbf{A}]_r$,
	\begingroup \allowdisplaybreaks
	\begin{align*}
		\norm{\obW^{[\ell]\top}}_{2, \infty} &= \max_{k \le \ell} \norm{\left(\mathbf{W}^{(\ell, k+1)} \mathbf{D}_{(0)}^{(k)} \mathbf{A}\right)^{\top}}_{2, \infty} \\& \le 
		\max_{k \le \ell} \norm{\mathbf{W}^{(\ell, k+1)} \mathbf{D}_{(0)}^{(k)} \mathbf{A}}_{2}
		\\& \le \max_{k \le \ell} \norm{\mathbf{W}^{(\ell, k+1)}}_2 \norm{\mathbf{D}_{(0)}^{(k)}}_2 \norm{\mathbf{A}}_{2} \\& \le  \max_{k \le \ell} \norm{\mathbf{W}^{(\ell, k+1)}} \norm{\mathbf{A}}_{2}.
	\end{align*}
	\endgroup
	From Fact~\ref{thm:norm_W}, we can show that with probability exceeding $1 - e^{-\rho^2}$,
	\begin{align*}
		\norm{\mathbf{A}} \le  \sqrt{2}(1 + \sqrt{dm^{-1}} + \sqrt{2}\rho m^{-0.5}) \le 5,
	\end{align*}
	provided $m \ge d \rho^2$.
	Also, from Lemma~\ref{lemma:norm_ESN}, we have for any $k \le L$, w.p. at least $1-e^{-\Omega(\rho^2)}$,
	\begin{align*}
		\norm{\mathbf{W}^{(\ell, k+1)}} = \norm{\mathbf{D}^{(\ell)} \mathbf{W} \mathbf{D}^{(\ell-1)} \mathbf{W} \cdots \mathbf{D}^{(k)} \mathbf{W}} \le \mathcal{O}(L^3).
	\end{align*}
	Hence,
	\begin{align*}
		\norm{\obW^{[\ell]\top}}_{2, \infty} \le&\max_{k \le \ell} \norm{\mathbf{W}^{(\ell, k+1)}} \norm{\mathbf{A}}_{2} \le 5 \cdot \mathcal{O}(L^3) = \mathcal{O}(L^3).
	\end{align*}
\end{proof}

\begin{claim}\label{clam:stabizable_V}
	Choose a random subset $\mathcal{K} \subset[m]$ of size $|\mathcal{K}|=N$. Replace the rows $\left\{\mathbf{w}_{k}, \mathbf{a}_{k}\right\}_{k \in \mathcal{K}}$ of $\mathbf{W}$ and $\mathbf{A}$ with freshly new i.i.d. samples $\widetilde{\mathbf{w}}_{k}, \widetilde{\mathbf{a}}_{k} \sim \mathcal{N}\left(0, \frac{2}{m} \mathbf{I}\right)$ to form new matrices $\widetilde{\mathbf{W}}$ and $\widetilde{\mathbf{A}}$. Define $\widetilde{\mathbf{V}}^{(\ell)}$ inductively as follows:
	\begin{align*}
		\widetilde{\mathbf{V}}^{(1)} &= \left[\mathbf{0}_{d \times m}, \text{ } \mathbf{I}_{d \times d}\right]_r \text{ }  \left[\widetilde{\mathbf{W}}, \text{ } \widetilde{\mathbf{A}}\right]_r^{\top} \text{ }  \widetilde{\mathbf{D}}^{(1)}_{(0)} \\
		\widetilde{\mathbf{V}}^{(\ell)} &= \left[\text{ }\left[\widetilde{\mathbf{V}}^{(\ell - 1)}, \text{ } \mathbf{0}_{(\ell - 1)d \times d}\right]_r,\text{ } \left[\mathbf{0}_{d \times m},\text{ } \mathbf{I}_{d \times d}\right]_r\text{ }\right]_c  \text{ } \left[\widetilde{\mathbf{W}}, \text{ } \widetilde{\mathbf{A}}\right]_r^{\top} \text{ }  \widetilde{\mathbf{D}}^{(\ell)}_{(0)}
	\end{align*}
	Then, with probability $1-e^{-\Omega(\rho^2)}$, for all $\ell \ge 2$,
	\begin{itemize}
		\item $\norm{\left[\text{ }\left[\widetilde{\mathbf{V}}^{(\ell)}, \text{ } \mathbf{0}_{\ell d \times d}\right]_r,\text{ } \left[\mathbf{0}_{d \times m},\text{ } \mathbf{I}_{d \times d}\right]_r\text{ }\right]_c  - \left[\text{ }\left[\mathbf{V}^{(\ell)}, \text{ } \mathbf{0}_{\ell d \times d}\right]_r,\text{ } \left[\mathbf{0}_{d \times m},\text{ } \mathbf{I}_{d \times d}\right]_r\text{ }\right]_c }_{2, \infty} \le \mathcal{O}(\rho^6 (N/m)^{1/6}).$
		\item $\norm{ \left( \left[\text{ }\left[\widetilde{\mathbf{V}}^{(\ell) }, \text{ } \mathbf{0}_{\ell d \times d}\right]_r,\text{ } \left[\mathbf{0}_{d \times m},\text{ } \mathbf{I}_{d \times d}\right]_r\text{ }\right]_c  - \left[\text{ }\left[\mathbf{V}^{(\ell)}, \text{ } \mathbf{0}_{\ell d \times d}\right]_r,\text{ } \left[\mathbf{0}_{d \times m},\text{ } \mathbf{I}_{d \times d}\right]_r\text{ }\right]_c  \right) \left[\mathbf{W}_{\mathcal{K}}, \mathbf{A}_{\mathcal{K}} \right]_r^{\top} }_{2, \infty} \le \mathcal{O}(\rho^6 (N/m)^{2/3}).$
	\end{itemize}
	
\end{claim}

\begin{proof}
	Let $\widetilde{\mathbf{W}}^{(k)} = \prod_{1 \le \ell \le k + 1} \widetilde{\mathbf{D}}_{(0)}^{(k - \ell + 1)} \widetilde{\mathbf{W}}$. Let $\overline{\widetilde{\mathbf{W}}}^{[L]} = [\widetilde{\mathbf{W}}^{(\ell - 1)} \widetilde{\mathbf{D}}_{(0)}^{(1)} \widetilde{\mathbf{A}}, \widetilde{\mathbf{W}}^{(\ell-2)} \widetilde{\mathbf{D}}_{(0)}^{(2)} \widetilde{\mathbf{A}}, \cdots, \widetilde{\mathbf{D}}_{(0)}^{(\ell)} \widetilde{\mathbf{A}}]_r$. Then, using the same induction technique as in Claim~\ref{claim:induct_VW}, we can show that
	\begin{align*}
		\widetilde{\mathbf{V}}^{(\ell)} = \overline{\widetilde{\mathbf{W}}}^{[\ell]\top}, \quad \text{ for all } \ell \in [L].
	\end{align*}
	Hence,
	\begingroup
	\begin{align*} \allowdisplaybreaks
		&\norm{\left[\text{ }\left[\widetilde{\mathbf{V}}^{(\ell)}, \text{ } \mathbf{0}_{(\ell - 1)d \times d}\right]_r,\text{ } \left[\mathbf{0}_{d \times m},\text{ } \mathbf{I}_{d \times d}\right]_r\text{ }\right]_c  - \left[\text{ }\left[\mathbf{V}^{(\ell)}, \text{ } \mathbf{0}_{(\ell - 1)d \times d}\right]_r,\text{ } \left[\mathbf{0}_{d \times m},\text{ } \mathbf{I}_{d \times d}\right]_r\text{ }\right]_c }_{2, \infty} \\&= \norm{\widetilde{\mathbf{V}}^{(\ell)} - \mathbf{V}^{(\ell)}}_{2, \infty} = \norm{\overline{\mathbf{W}}^{[\ell]\top} - \overline{\widetilde{\mathbf{W}}}^{[\ell]\top}}_{2, \infty}
		\\&\le \max_{k \le \ell} \norm{\left(\widetilde{\mathbf{W}}^{(\ell, k+1)} \widetilde{\mathbf{D}}_{(0)}^{(k)} \widetilde{\mathbf{A}} - \mathbf{W}^{(\ell, k+1)} \mathbf{D}_{(0)}^{(k)} \mathbf{A}\right)^{\top}}_{2, \infty} \\&
		\le \max_{k \le \ell} \norm{\left(\widetilde{\mathbf{W}}^{(\ell, k+1)} \widetilde{\mathbf{D}}_{(0)}^{(k)} \widetilde{\mathbf{A}} - \mathbf{W}^{(\ell, k+1)} \mathbf{D}_{(0)}^{(k)} \mathbf{A} \right)^{\top}}_{2, \infty} \\&
		=  \max_{k \le \ell} \norm{\left(\widetilde{\mathbf{W}}^{(\ell, k+1)} \widetilde{\mathbf{D}}_{(0)}^{(\ell, k+1)} \widetilde{\mathbf{A}} - \mathbf{W}^{(\ell, k+1)} \widetilde{\mathbf{D}}_{(0)}^{(k)} \widetilde{\mathbf{A}}\right)^{\top} + \left(\mathbf{W}^{(\ell, k+1)} \widetilde{\mathbf{D}}_{(0)}^{(k)} \widetilde{\mathbf{A}} - \mathbf{W}^{(\ell, k+1)} \mathbf{D}_{(0)}^{(k)} \mathbf{A} \right)^{\top}}_{2, \infty} \\&
		\le  \max_{k \le \ell} \norm{\left(\widetilde{\mathbf{W}}^{(\ell, k+1)} \widetilde{\mathbf{D}}_{(0)}^{(\ell, k+1)} \widetilde{\mathbf{A}} - \mathbf{W}^{(\ell, k+1)} \widetilde{\mathbf{D}}_{(0)}^{(k)} \widetilde{\mathbf{A}}\right)^{\top}}_{2, \infty} + \norm{\left(\mathbf{W}^{(\ell, k+1)} \widetilde{\mathbf{D}}_{(0)}^{(k)} \widetilde{\mathbf{A}} - \mathbf{W}^{(\ell, k+1)} \mathbf{D}_{(0)}^{(k)} \mathbf{A} \right)^{\top}}_{2, \infty} \\&
		\le \max_{k \le \ell} \norm{\left(\widetilde{\mathbf{W}}^{(\ell, k+1)} \widetilde{\mathbf{D}}_{(0)}^{(\ell, k+1)} \widetilde{\mathbf{A}} - \mathbf{W}^{(\ell, k+1)} \widetilde{\mathbf{D}}_{(0)}^{(k)} \widetilde{\mathbf{A}}\right)^{\top}}_{2, \infty} + \norm{\left(\mathbf{W}^{(\ell, k+1)} \widetilde{\mathbf{D}}_{(0)}^{(k)} \widetilde{\mathbf{A}} - \mathbf{W}^{(\ell, k+1)} \mathbf{D}_{(0)}^{(k)} \mathbf{A} \right)^{\top}}_{2}.
	\end{align*}
	\endgroup     
	We will bound the two terms separately. 
	From Fact~\ref{thm:norm_W}, we can show that with probability exceeding $1 - e^{-\rho^2}$,
	\begin{align}
		\norm{\widetilde{\mathbf{A}}} \le  \sqrt{2/m}(\sqrt{m} + \sqrt{d} + \sqrt{2}\rho) \le 5 \label{eq:normtildeA},
	\end{align}
	provided $m \ge d\rho^2$. 
	Using the above bound, we get the following for all $k \in [\ell]$:
	\begin{align*}
		\norm{\left(\widetilde{\mathbf{W}}^{(\ell, k+1)} \widetilde{\mathbf{D}}_{(0)}^{(k)} \widetilde{\mathbf{A}} - \mathbf{W}^{(\ell, k+1)} \widetilde{\mathbf{D}}_{(0)}^{(k)} \widetilde{\mathbf{A}}\right)^{\top}}_{2, \infty} &\le \max_{i \in [d]} \norm{\left(\widetilde{\mathbf{W}}^{(\ell, k+1)}  - \mathbf{W}^{(\ell, k+1)} \right) \left(\widetilde{\mathbf{D}}_{(0)}^{(k)} \widetilde{\mathbf{A}}\right)^{\top}_i}_2 \\&
		\le \max_{i \in [d]} \mathcal{O}(\rho^5 (N/m)^{1/6})  \cdot \norm{ \left(\widetilde{\mathbf{D}}_{(0)}^{(k)} \widetilde{\mathbf{A}}\right)^{\top}_i}_2 \\&
		\le \mathcal{O}(\rho^5 (N/m)^{1/6}), 
	\end{align*}
	where in the second-final step we have used Lemma~\ref{lemma:rerandESN} to have
	\begin{align*}
		\norm{\left(\widetilde{\mathbf{W}}^{(\ell, k+1)} - \mathbf{W}^{(\ell, k+1)}\right) \mathbf{v}}_2 &= \norm{ \left(\prod_{\ell \ge \ell' \ge k + 1} \widetilde{\mathbf{D}}_{(0)}^{(\ell')} \widetilde{\mathbf{W}}  - \prod_{\ell \ge \ell' \ge k + 1} \mathbf{D}_{(0)}^{(\ell')} \mathbf{W} \right) \mathbf{v} }_2 \le  \mathcal{O}(\rho^5 (N/m)^{1/6}) \norm{\mathbf{v}},
	\end{align*}
	for the vectors $\mathbf{v} \in \left\{\left(\widetilde{\mathbf{D}}_{(0)}^{(k)} \widetilde{\mathbf{A}}\right)^{\top}_i\right\}_{i \in [d]}$. 
	From Lemma~\ref{lemma:norm_ESN}, we have for any $k < \ell$, w.p. at least $1-e^{-\Omega(\rho^2)}$,
	\begin{align}
		\norm{\mathbf{W}^{(\ell, k+1)}} = \norm{\mathbf{D}^{(\ell)} \mathbf{W} \mathbf{D}^{(\ell-1)} \mathbf{W} \cdots \mathbf{D}^{(k+1)} \mathbf{W}} \le \mathcal{O}(L^3) \label{eq:normWk}.
	\end{align}
	Also, we have used the following fact:
	\begin{align*}
		\max_{i \in [d]}  \norm{ \left(\widetilde{\mathbf{D}}_{(0)}^{(k)} \widetilde{\mathbf{A}}\right)^{\top}_i}_2 &\le \norm{ \left(\widetilde{\mathbf{D}}_{(0)}^{(k)} \widetilde{\mathbf{A}}\right)^{\top}}_2 \\& \le \norm{ \left(\widetilde{\mathbf{D}}_{(0)}^{(k)} \widetilde{\mathbf{A}}\right)^{\top}}_2
	\end{align*}
	
	Again since only $N$ rows of $\mathbf{A}$ and $\widetilde{\mathbf{A}}$ are different, we can show from Fact~\ref{thm:norm_W} that with probability exceeding $1 - e^{-\Omega(\rho^2)}$,
	\begin{align}
		\norm{\mathbf{A} - \widetilde{\mathbf{A}}} \le  \sqrt{2/m}(\sqrt{N} + \sqrt{d} + \sqrt{2}\rho) \le  \mathcal{O}((\rho + \sqrt{d}) \sqrt{N/m}) \le \mathcal{O}(\rho \sqrt{N/m}) \label{eq:tildeAA},
	\end{align}
	since we are using $\rho = 100Ld \log m$. Also from Lemma~\ref{lemma:rerandESN}, we have that with probability exceeding $1 - e^{-\Omega(\rho^2)}$, 
	\begin{align*}
		\norm{\widetilde{\mathbf{D}}^{(k)}_{(0)} - \mathbf{D}^{(k)}_{(0)}}_0 \le \mathcal{O}(\rho^4 N^{1/3} m^{2/3}).
	\end{align*}
	Thus, again we can use Fact~\ref{thm:norm_W} to show that with probability exceeding $1 - 2e^{-\rho^2}$,
	\begin{align}
		\norm{\left( \widetilde{\mathbf{D}}^{(k)}_{(0)} - \mathbf{D}^{(k)}_{(0)}\right) \mathbf{A}}_2  &
		\le \sqrt{2/m} (\sqrt{d} + \sqrt{\norm{\widetilde{\mathbf{D}}^{(k)}_{(0)} - \mathbf{D}^{(k)}_{(0)}}_0} + \sqrt{2} \rho)
		\nonumber \\&
		\le \sqrt{2/m} (\sqrt{d} + \sqrt{\mathcal{O}(\rho^4 N^{1/3} m^{2/3})} + \sqrt{2} \rho) \nonumber\\&
		\le \sqrt{2/m} \mathcal{O} ((\sqrt{d} + \sqrt{2} \rho) \sqrt{\mathcal{O}(\rho^4 N^{1/3} m^{2/3})} ) \nonumber
		\\&\le \mathcal{O}(\rho^{3} N^{1/6} m^{-1/6}) \label{eq:DAtildeDA},
	\end{align}
	where again we are using $\rho = 100Ld \log m$.
	Thus,
	\begingroup \allowdisplaybreaks
	\begin{align}
		&\norm{\mathbf{W}^{(\ell, k+1)} \widetilde{\mathbf{D}}_{(0)}^{(k)} \widetilde{\mathbf{A}} - \mathbf{W}^{(\ell, k+1)} \mathbf{D}_{(0)}^{(k)} \mathbf{A}}_{2} \nonumber\\&= \norm{\mathbf{W}^{(\ell, k+1)}}_2 \norm{\widetilde{\mathbf{D}}_{(0)}^{(k)} \widetilde{\mathbf{A}} -  \mathbf{D}_{(0)}^{(k)} \mathbf{A}}_{2} \nonumber\\&
		= \norm{\mathbf{W}^{(\ell, k+1)}}_2 \norm{\widetilde{\mathbf{D}}_{(0)}^{(k)} \widetilde{\mathbf{A}} - \mathbf{D}_{(0)}^{(k)} \widetilde{\mathbf{A}} + \mathbf{D}_{(0)}^{(k)} \widetilde{\mathbf{A}} -  \mathbf{D}_{(0)}^{(k)} \mathbf{A}}_{2}  \nonumber\\&
		\le \norm{\mathbf{W}^{(\ell, k+1)}}_2 \left(\norm{\widetilde{\mathbf{D}}_{(0)}^{(k)} \widetilde{\mathbf{A}} - \mathbf{D}_{(0)}^{(k)} \widetilde{\mathbf{A}}}_2 + \norm{\mathbf{D}_{(0)}^{(k)} \widetilde{\mathbf{A}} -  \mathbf{D}_{(0)}^{(k)} \mathbf{A}}_{2} \right) \nonumber\\&
		\le \norm{\mathbf{W}^{(\ell, k+1)}}_2 \left(\norm{\widetilde{\mathbf{D}}_{(0)}^{(k)} \widetilde{\mathbf{A}} - \mathbf{D}_{(0)}^{(k)} \widetilde{\mathbf{A}}}_2 + \norm{ \widetilde{\mathbf{A}} -  \mathbf{A}}_{2} \right) \nonumber
		\\&
		\le \mathcal{O}(L^3) \cdot \left(\mathcal{O}(\rho^{3} (N/m)^{1/6}) + \mathcal{O}(\rho (N/m)^{1/2})\right) \le \mathcal{O}(\rho^{6} (N/m)^{1/6}) \label{eq:DAtilde_DA},
	\end{align}
	\endgroup
	since  we are using $\rho = 100Ld \log m$.
	Hence, we have 
	\begingroup\allowdisplaybreaks
	\begin{align*}
		\norm{\overline{\mathbf{W}}^{[\ell]\top} - \overline{\widetilde{\mathbf{W}}}^{[\ell]\top}}_{2, \infty} & \le  \max_{k \le \ell} \norm{\left(\widetilde{\mathbf{W}}^{(\ell, k+1)} \widetilde{\mathbf{D}}_{(0)}^{(k)} \widetilde{\mathbf{A}} - \mathbf{W}^{(\ell, k+1)} \widetilde{\mathbf{D}}_{(0)}^{(k)} \widetilde{\mathbf{A}}\right)^{\top}}_{2, \infty} + \norm{\mathbf{W}^{(\ell, k+1)} \widetilde{\mathbf{D}}_{(0)}^{(k)} \widetilde{\mathbf{A}} - \mathbf{W}^{(\ell, k+1)} \mathbf{D}_{(0)}^{(k)} \mathbf{A}}_{2} \\&
		\le \mathcal{O}(\rho^{6} (N/m)^{1/6}) + \mathcal{O}(\rho^{5} N^{1/6} m^{-1/6}) = \mathcal{O}(\rho^{6} (N/m)^{1/6}),
	\end{align*}
	\endgroup
	which gives the first result.
	Now, we focus on 
	\begin{align*}
		\norm{ \left( \left[\text{ }\left[\widetilde{\mathbf{V}}^{(\ell) }, \text{ } \mathbf{0}_{\ell d \times d}\right]_r,\text{ } \left[\mathbf{0}_{d \times m},\text{ } \mathbf{I}_{d \times d}\right]_r\text{ }\right]_c  - \left[\text{ }\left[\mathbf{V}^{(\ell)}, \text{ } \mathbf{0}_{\ell d \times d}\right]_r,\text{ } \left[\mathbf{0}_{d \times m},\text{ } \mathbf{I}_{d \times d}\right]_r\text{ }\right]_c  \right) \left[\mathbf{W}_{\mathcal{K}}, \mathbf{A}_{\mathcal{K}} \right]_r^{\top} }_{2, \infty}.
	\end{align*}
	Note that the above is equivalent to
	\begin{align*}
		\norm{\widetilde{\mathbf{W}}^{[\ell]\top} \mathbf{W}_{\mathcal{K}}^{\top} - \mathbf{W}^{[\ell]\top} \mathbf{W}_{\mathcal{K}}^{\top} }_{2, \infty},
	\end{align*}
	using the relation between $\widetilde{\mathbf{V}}^{(\ell) }$ and $\widetilde{\mathbf{W}}^{[\ell]\top}$, and $\mathbf{V}^{(\ell) }$ and $\mathbf{W}^{[\ell]\top}$. Continuing
	\begingroup \allowdisplaybreaks
	\begin{align}
		&\norm{\widetilde{\mathbf{W}}^{[\ell]\top} \mathbf{W}_{\mathcal{K}}^{\top} - \mathbf{W}^{[\ell]\top} \mathbf{W}_{\mathcal{K}}^{\top} }_{2, \infty}
		\nonumber\\&
		\le \max_{k \le \ell} \norm{\left(\mathbf{W}_{\mathcal{K}} \widetilde{\mathbf{W}}^{(\ell, k+1)} \widetilde{\mathbf{D}}_{(0)}^{(k)} \widetilde{\mathbf{A}} - \mathbf{W}_{\mathcal{K}} \mathbf{W}^{(\ell, k+1)} \mathbf{D}_{(0)}^{(k)} \mathbf{A}\right)^{\top}}_{2, \infty} \nonumber\\&
		= \max_{k \le \ell} \norm{\left(\mathbf{W}_{\mathcal{K}} \widetilde{\mathbf{W}}^{(\ell, k+1)} \widetilde{\mathbf{D}}_{(0)}^{(k)} \widetilde{\mathbf{A}} - \mathbf{W}_{\mathcal{K}} \mathbf{W}^{(\ell, k+1)} \widetilde{\mathbf{D}}_{(0)}^{(k)} \widetilde{\mathbf{A}} + \mathbf{W}_{\mathcal{K}} \mathbf{W}^{(\ell, k+1)} \widetilde{\mathbf{D}}_{(0)}^{(k)} \widetilde{\mathbf{A}} - \mathbf{W}_{\mathcal{K}} \mathbf{W}^{(\ell, k+1)} \mathbf{D}_{(0)}^{(k)} \mathbf{A}\right)^{\top}}_{2, \infty}  \nonumber\\&
		\le \max_{k \le \ell} \norm{\left(\mathbf{W}_{\mathcal{K}} \widetilde{\mathbf{W}}^{(\ell, k+1)} \widetilde{\mathbf{D}}_{(0)}^{(k)} \widetilde{\mathbf{A}} - \mathbf{W}_{\mathcal{K}} \mathbf{W}^{(\ell, k+1)} \widetilde{\mathbf{D}}_{(0)}^{(k)} \widetilde{\mathbf{A}}\right)^{\top}}_{2, \infty} + \norm{\left(\mathbf{W}_{\mathcal{K}} \mathbf{W}^{(\ell, k+1)} \widetilde{\mathbf{D}}_{(0)}^{(k)} \widetilde{\mathbf{A}} - \mathbf{W}_{\mathcal{K}} \mathbf{W}^{(\ell, k+1)} \mathbf{D}_{(0)}^{(k)} \mathbf{A}\right)^{\top}}_{2, \infty} \nonumber\\&
		\le\max_{k \le \ell} \norm{\left(\mathbf{W}_{\mathcal{K}} \widetilde{\mathbf{W}}^{(\ell, k+1)} \widetilde{\mathbf{D}}_{(0)}^{(k)} \widetilde{\mathbf{A}} - \mathbf{W}_{\mathcal{K}} \mathbf{W}^{(\ell, k+1)} \widetilde{\mathbf{D}}_{(0)}^{(k)} \widetilde{\mathbf{A}}\right)^{\top}}_{2, \infty} + \norm{\mathbf{W}_{\mathcal{K}} \mathbf{W}^{(\ell, k+1)} \left(\widetilde{\mathbf{D}}_{(0)}^{(k)} \widetilde{\mathbf{A}} - \mathbf{D}_{(0)}^{(k)} \mathbf{A}\right)}_2 
		\label{eq:tildeWWkappaWWkappa}
	\end{align}
	\endgroup
	Using lemma~\ref{lemma:rerandESN}, we have with probability $1-e^{-\Omega(\rho^2)}$, we have for all $k \le \ell$
	\begin{align*}
		\norm{\left(\mathbf{W}_{\mathcal{K}} \widetilde{\mathbf{W}}^{(\ell, k+1)}  - \mathbf{W}_{\mathcal{K}} \mathbf{W}^{(\ell, k+1)} \right) \mathbf{v}}_2 &= \norm{\mathbf{W}_{\mathcal{K}} \left(\prod_{\ell \ge \ell' \ge k+1} \widetilde{\mathbf{D}}^{(\ell')} \widetilde{\mathbf{W}} - \prod_{\ell \ge \ell' \ge k+1} \mathbf{D}^{(\ell')} \mathbf{W}\right) \mathbf{v}}_2 \\& \le \mathcal{O}(\rho^6 (N/m)^{2/3}) \norm{\mathbf{v}}, 
	\end{align*}
	for any fixed vector $\mathbf{v}$.We will use union bound to make sure the above property is satisfied for all vector $\mathbf{v}$ in the set $ \left\{\left(\widetilde{\mathbf{D}}_{(0)}^{(k)} \widetilde{\mathbf{A}}\right)^{\top}_i\right\}_{i \in [d]}$. From Eq.~\ref{eq:normtildeA}, with probability $1-e^{-\Omega(\rho^2)}$
	\begin{align*}
		\norm{\widetilde{\mathbf{A}}} \le  \sqrt{2}(1 + \sqrt{dm^{-1}} + \sqrt{2}\rho m^{-0.5}) \le 5,
	\end{align*}
	provided $m \ge d \rho^2$. Thus, the first term in Eq.~\ref{eq:tildeWWkappaWWkappa} can be bounded as
	\begin{align*}
		&\max_{k \le \ell} \norm{\left(\mathbf{W}_{\mathcal{K}} \widetilde{\mathbf{W}}^{(\ell, k+1)} \widetilde{\mathbf{D}}_{(0)}^{(k)} \widetilde{\mathbf{A}} - \mathbf{W}_{\mathcal{K}} \mathbf{W}^{(\ell, k+1)} \widetilde{\mathbf{D}}_{(0)}^{(k)} \widetilde{\mathbf{A}}\right)^{\top}}_{2, \infty} \\& = \max_{k \le \ell} \max_{i \in [d]} \norm{\left(\mathbf{W}_{\mathcal{K}} \widetilde{\mathbf{W}}^{(\ell, k+1)} \widetilde{\mathbf{D}}_{(0)}^{(k)}  - \mathbf{W}_{\mathcal{K}} \mathbf{W}^{(\ell, k+1)} \right) \left(\widetilde{\mathbf{D}}_{(0)}^{(k)} \widetilde{\mathbf{A}}\right)_i}_{2} \\&
		\le \max_{k \le \ell} \max_{i \in [d]} \mathcal{O}(\rho^6 (N/m)^{2/3}) \cdot \norm{ \left(\widetilde{\mathbf{D}}_{(0)}^{(k)} \widetilde{\mathbf{A}}\right)_i} \\&
		\le \max_{k \le \ell}  \mathcal{O}(\rho^6 (N/m)^{2/3}) \cdot \norm{ \widetilde{\mathbf{D}}_{(0)}^{(k)} \widetilde{\mathbf{A}}}_2 \\&
		\le \mathcal{O}(\rho^6 (N/m)^{2/3}).
	\end{align*}
	
	Now, we focus on the second term in  Eq.~\ref{eq:tildeWWkappaWWkappa}. We have for any $k \le \ell$,
	\begingroup \allowdisplaybreaks
	\begin{align*}
		&\norm{\mathbf{W}_{\mathcal{K}} \mathbf{W}^{(\ell, k+1)} \left(\widetilde{\mathbf{D}}_{(0)}^{(k)} \widetilde{\mathbf{A}} - \mathbf{D}_{(0)}^{(k)} \mathbf{A}\right)}_2 \\& =
		\norm{\mathbf{W}_{\mathcal{K}} \mathbf{W}^{(\ell, k+1)} \left(\widetilde{\mathbf{D}}_{(0)}^{(k)} \widetilde{\mathbf{A}} -  \mathbf{D}_{(0)}^{(k)} \widetilde{\mathbf{A}} + \mathbf{D}_{(0)}^{(k)} \widetilde{\mathbf{A}} - \mathbf{D}_{(0)}^{(k)} \mathbf{A}\right)}_2 \\&
		\le \norm{\mathbf{W}_{\mathcal{K}} \mathbf{W}^{(\ell, k+1)} \left(\widetilde{\mathbf{D}}_{(0)}^{(k)} \widetilde{\mathbf{A}} -  \mathbf{D}_{(0)}^{(k)} \widetilde{\mathbf{A}} \right)}_2 + \norm{\mathbf{W}_{\mathcal{K}} \mathbf{W}^{(\ell, k+1)} \left( \mathbf{D}_{(0)}^{(k)} \widetilde{\mathbf{A}} - \mathbf{D}_{(0)}^{(k)} \mathbf{A}\right)}_2 \\&
		\le \norm{\mathbf{W}_{\mathcal{K}} \mathbf{W}^{(\ell, k+1)} \left(\widetilde{\mathbf{D}}_{(0)}^{(k)}  -  \mathbf{D}_{(0)}^{(k)}  \right)}_2 \norm{\left(\widetilde{\mathbf{D}}_{(0)}^{(k)}  -  \mathbf{D}_{(0)}^{(k)}  \right)\widetilde{\mathbf{A}}}_2 + \norm{\mathbf{W}_{\mathcal{K}} \mathbf{W}^{(\ell, k+1)} \left( \mathbf{D}_{(0)}^{(k)} \widetilde{\mathbf{A}} - \mathbf{D}_{(0)}^{(k)} \mathbf{A}\right)}_2 \\&
		=  \norm{\mathbf{D}_{\mathcal{K}} \mathbf{W} \mathbf{W}^{(\ell, k+1)} \left(\widetilde{\mathbf{D}}_{(0)}^{(k)}  -  \mathbf{D}_{(0)}^{(k)}  \right)}_2 \norm{\left(\widetilde{\mathbf{D}}_{(0)}^{(k)}  -  \mathbf{D}_{(0)}^{(k)}  \right)\widetilde{\mathbf{A}}}_2 + \norm{\mathbf{D}_{\mathcal{K}} \mathbf{W} \mathbf{W}^{(\ell, k+1)} \mathbf{D}_{(0)}^{(k)} \mathbf{D}_{\mathcal{K}} \left(  \widetilde{\mathbf{A}} -  \mathbf{A}\right)}_2 \\& 
		\le \norm{\mathbf{D}_{\mathcal{K}} \mathbf{W} \mathbf{W}^{(\ell, k+1)} \left(\widetilde{\mathbf{D}}_{(0)}^{(k)}  -  \mathbf{D}_{(0)}^{(k)}  \right)}_2 \norm{\left(\widetilde{\mathbf{D}}_{(0)}^{(k)}  -  \mathbf{D}_{(0)}^{(k)}  \right)\widetilde{\mathbf{A}}}_2 + \norm{\mathbf{D}_{\mathcal{K}} \mathbf{W} \mathbf{W}^{(\ell, k+1)} \mathbf{D}_{(0)}^{(k)} \mathbf{D}_{\mathcal{K}}}_2 \norm{ \widetilde{\mathbf{A}} -  \mathbf{A} }_2
	\end{align*}
	\endgroup
	Since, $\norm{D_{\mathcal{K}}}_0 = N$ and $\norm{\widetilde{\mathbf{D}}_{(0)}^{(k)}  -  \mathbf{D}_{(0)}^{(k)} }_0 \le \mathcal{O}(\rho^4 N^{1/3} m^{2/3})$ with probability exceeding $1 - e^{-\Omega(\rho^2)}$ from Lemma~\ref{lemma:rerandESN}, we can use Lemma~\ref{lemma:norm_ESN} to get
	\begingroup \allowdisplaybreaks
	\begin{align*}
		&\norm{\mathbf{W}_{\mathcal{K}} \mathbf{W}^{(\ell, k+1)} \left(\widetilde{\mathbf{D}}_{(0)}^{(k)} \widetilde{\mathbf{A}} - \mathbf{D}_{(0)}^{(k)} \mathbf{A}\right)}_2 \\& 
		\le \norm{\mathbf{D}_{\mathcal{K}} \mathbf{W} \mathbf{W}^{(\ell, k+1)} \left(\widetilde{\mathbf{D}}_{(0)}^{(k)}  -  \mathbf{D}_{(0)}^{(k)}  \right)}_2 \norm{\left(\widetilde{\mathbf{D}}_{(0)}^{(k)}  -  \mathbf{D}_{(0)}^{(k)}  \right)\widetilde{\mathbf{A}}}_2 + \norm{\mathbf{D}_{\mathcal{K}} \mathbf{W} \mathbf{W}^{(\ell, k+1)} \mathbf{D}_{(0)}^{(k)} \mathbf{D}_{\mathcal{K}}}_2 \norm{ \widetilde{\mathbf{A}} -  \mathbf{A} }_2 \\&
		= \norm{\mathbf{D}_{\mathcal{K}} \mathbf{W} \mathbf{D}^{(\ell)} \mathbf{W} \cdots \mathbf{D}^{(k+1)} \mathbf{W} \left(\widetilde{\mathbf{D}}_{(0)}^{(k)}  -  \mathbf{D}_{(0)}^{(k)}  \right)}_2 \norm{\left(\widetilde{\mathbf{D}}_{(0)}^{(k)}  -  \mathbf{D}_{(0)}^{(k)}  \right)\widetilde{\mathbf{A}}}_2 \\& \quad \quad \quad + \norm{\mathbf{D}_{\mathcal{K}} \mathbf{W} \mathbf{W} \mathbf{D}^{(\ell)} \mathbf{W} \cdots \mathbf{D}^{(k+1)} \mathbf{W} \mathbf{D}_{(0)}^{(k)} \mathbf{D}_{\mathcal{K}}}_2 \norm{ \widetilde{\mathbf{A}} -  \mathbf{A} }_2
		\\&
		\le \mathcal{O}(\rho (N/m)^{1/2}) \cdot\left( \norm{\left(\widetilde{\mathbf{D}}_{(0)}^{(k)}  -  \mathbf{D}_{(0)}^{(k)}  \right)\widetilde{\mathbf{A}}}_2 + \norm{ \widetilde{\mathbf{A}} -  \mathbf{A} }_2\right).
	\end{align*}
	\endgroup
	Further using Eq.~\ref{eq:DAtildeDA} and Eq.~\ref{eq:tildeAA}, we have
	\begingroup \allowdisplaybreaks
	\begin{align*}
		&\norm{\mathbf{W}_{\mathcal{K}} \mathbf{W}^{(\ell, k+1)} \left(\widetilde{\mathbf{D}}_{(0)}^{(k)} \widetilde{\mathbf{A}} - \mathbf{D}_{(0)}^{(k)} \mathbf{A}\right)}_2 \\& 
		\le \mathcal{O}(\rho (N/m)^{1/2}) \cdot\left( \norm{\left(\widetilde{\mathbf{D}}_{(0)}^{(k)}  -  \mathbf{D}_{(0)}^{(k)}  \right)\widetilde{\mathbf{A}}}_2 + \norm{ \widetilde{\mathbf{A}} -  \mathbf{A} }_2\right) \\&
		\le \mathcal{O}(\rho (N/m)^{1/2}) \cdot\left( \mathcal{O}(\rho^3 (N/m)^{1/6})  +  \mathcal{O}(\rho (N/m)^{1/2})\right) = \mathcal{O}(\rho^4 (N/m)^{2/3}).
	\end{align*}
	\endgroup
	Thus, connecting all the bounds above in Eq.~\ref{eq:tildeWWkappaWWkappa}, we get 
	\begin{align}
		&\norm{\widetilde{\mathbf{W}}^{[\ell]\top} \mathbf{W}_{\mathcal{K}}^{\top} - \mathbf{W}^{[\ell]\top} \mathbf{W}_{\mathcal{K}}^{\top} }_{2, \infty} \nonumber\\&
		\le 
		\le\max_{k \le \ell} \norm{\left(\mathbf{W}_{\mathcal{K}} \widetilde{\mathbf{W}}^{(\ell, k+1)} \widetilde{\mathbf{D}}_{(0)}^{(k)} \widetilde{\mathbf{A}} - \mathbf{W}_{\mathcal{K}} \mathbf{W}^{(\ell, k+1)} \widetilde{\mathbf{D}}_{(0)}^{(k)} \widetilde{\mathbf{A}}\right)^{\top}}_{2, \infty} + \norm{\mathbf{W}_{\mathcal{K}} \mathbf{W}^{(\ell, k+1)} \left(\widetilde{\mathbf{D}}_{(0)}^{(k)} \widetilde{\mathbf{A}} - \mathbf{D}_{(0)}^{(k)} \mathbf{A}\right)}_2 \nonumber \\&
		\le \mathcal{O}(\rho^6 (N/m)^{2/3}) + \mathcal{O}(\rho^4 (N/m)^{2/3}) = \mathcal{O}(\rho^6 (N/m)^{2/3}) \label{eq:tmp_W_MATHCALwl}.
	\end{align}
\end{proof}

\section{Existence of good pseudo network: proofs}
\subsection{Proof of theorem~\ref{thm:existence_pseudo}}
\begin{definition}[Restating defintion~\ref{def:existence}] 
	Define $\mathbf{W}^{\ast}$ and $\mathbf{A}^{\ast}$ as follows.
	\begin{align*}
		\mathbf{W}^{\ast} &= 0 \\
		\mathbf{a}^{*}_{r} &= \frac{\dout }{m} \sum_{s \in [\dout ]} \sum_{r' \in [p]} b_{r, s} b_{r', s}^{\dagger} H_{r', s} \left(\theta_{r', s} \left(\langle \mathbf{w}_{r}, \overline{\mathbf{W}}^{[L]} \mathbf{w}_{r', s}^{\dagger}\rangle\right), \sqrt{m/2} a_{r, d}\right) \mathbf{e}_d, \quad \forall r \in [m],
	\end{align*}
	where
	\begin{equation*} 
		\theta_{r', s} = \frac{\sqrt{m/2}}{\norm[1]{ \overline{\mathbf{W}}^{[L]} \mathbf{w}_{r', s}^{\dagger}}},
	\end{equation*}
	and $\obW^{[L]} = [\mathbf{W}^{(L, 3)} \mathbf{D}_{(0)}^{(2)}\mathbf{A}_{[d-1]},  \cdots, \mathbf{W}^{(L, L)} 
	\mathbf{D}_{(0)}^{(L-1)}\mathbf{A}_{[d-1]}]_r$, where $\mathbf{W}^{(k_b, k_e)} = \prod_{k_b \ge \ell > k_e} \mathbf{D}_{(0)}^{(\ell)} \mathbf{W}$.
\end{definition}
Using Lemma~\ref{lemma:norm_ESN}, we have with probability at least $ 1-e^{-\Omega(\rho^2)}$, for all $\ell \in [L]$ and any vector $\mathbf{u} \in \mathbb{R}^{d}$
\begingroup \allowdisplaybreaks
\begin{align*}
	\left(1 - \frac{1}{100L}\right)^{L} \norm{\mathbf{u}} \le \norm[2]{\mathbf{W}^{(L, \ell+1)} \mathbf{D}^{(\ell)}_{(0)} \mathbf{A} \mathbf{u}} = \norm[2]{\prod_{L \ge \ell' \ge \ell+1} \mathbf{D}_{(0)}^{(\ell')} \mathbf{W} \mathbf{D}_{(0)}^{(\ell)} \mathbf{A} \mathbf{u}} \le \left(1 + \frac{1}{100L}\right)^{L} \norm{\mathbf{u}}.
\end{align*} 
\endgroup
Since, for any vector $\mathbf{u} \in \mathbb{R}^{Ld}$,
\begin{align*}
	\min_{\ell \in [L]} \frac{\norm[2]{\mathbf{W}^{(L, \ell+1)} \mathbf{D}^{(\ell)}_{(0)} \mathbf{A} \mathbf{u}_{\ell d : (\ell + 1)d}}}{\norm[2]{\mathbf{u}_{\ell d : (\ell + 1)d}}} \norm[2]{\mathbf{u}} \le \norm[2]{\mathbf{W}^{[L]} \mathbf{u}} \le \max_{\ell \in [L]} \frac{\norm[2]{\mathbf{W}^{(L, \ell+1)} \mathbf{D}^{(\ell)}_{(0)} \mathbf{A} \mathbf{u}_{\ell d : (\ell + 1)d}}}{\norm[2]{\mathbf{u}_{\ell d : (\ell + 1)d}}} \norm[2]{\mathbf{u}},
\end{align*}
where $\mathbf{u}_{\ell d : (\ell + 1)d} \in \mathbb{R}^{Ld}$ refers to a vector that is equal to the vector $\mathbf{u}$ in the dimensions from $\ell d$ to $(\ell + 1)d$ and $0$ outside, we have
\begin{align*}
	\left(1 - \frac{1}{100L}\right)^{L} \norm[2]{\mathbf{w}_{r', s}^{\dagger}} \le  \norm[2]{\overline{\mathbf{W}}^{[L]} \mathbf{w}_{r', s}^{\dagger}} \le \left(1 + \frac{1}{100L}\right)^{L} \norm[2]{\mathbf{w}_{r', s}^{\dagger}}
\end{align*}
and thus we have $\forall  r' \in [p], s \in [\dout ]$,
\begin{align}
	\left(1 + \frac{1}{100L}\right)^{-L}  \le \sqrt{2/m} \theta_{r', s} = \left(1 - \frac{1}{100L}\right)^{-L}  \label{eq:abstheta_lowerbound}.
\end{align}

\begin{theorem}[Restating theorem~\ref{thm:existence_pseudo}]\label{thm:existence_pseudo_proof}
	The construction of $\mathbf{W}^{*}$ and $\mathbf{A}^{\ast}$ in Definition~\ref{def:existence} satisfies the following. For every normalized input sequence $\mathbf{x}^{(1)}, \cdots, \mathbf{x}^{(L)}$, we have with probability at least $1-e^{-\Omega\left(\rho^{2}\right)}$ over $\mathbf{W}, \mathbf{A}, \mathbf{B},$ it holds for every $s \in [\dout ]$.
	$$
	\begin{array}{l}
		F_{s}^{(L)} \stackrel{\text { def }}{=} \sum_{i=1}^{L} \mathbf{e}_{s}^{\top} \mathbf{Back}_{i \rightarrow L} D^{(i)} \left(\mathbf{W}^{\ast} \mathbf{h}^{(i-1)} + \mathbf{A}^{\ast} \mathbf{x}^{(i)}\right) \\
		= \sum_{r \in[p]} b_{r, s}^{\dagger} \Phi_{r, s} \left(\left\langle  \mathbf{w}_{r, s}^{\dagger}, [\overline{\mathbf{x}}^{(2)}, \cdots, \overline{\mathbf{x}}^{(L-2)}]\right\rangle\right)\\ \pm \mathcal{O}(\dout Lp\rho^2 \varepsilon + \dout L^{17/6} p \rho^4 L_{\Phi} \varepsilon_x^{2/3} + \dout ^{3/2}L^5 p \rho^{11} L_{\Phi} C_{\Phi}  \mathfrak{C}_{\varepsilon}(\Phi, \mathcal{O}(\varepsilon_x^{-1}))  m^{-1/30} ).
	\end{array}
	$$
\end{theorem}

\begin{proof}
	We fix a given normalized sequence $\mathbf{x}^{(1)}, \cdots, \mathbf{x}^{(L)}$ and an index $s \in [\dout ]$.
	The pseudo network for the fixed sequence is given by
	\begingroup \allowdisplaybreaks
	\begin{align}
		F_{s}^{(L)} &= \sum_{i=1}^{L} \mathbf{e}_{s}^{\top} \mathbf{Back}_{i \rightarrow L} D^{(i)} \left(\mathbf{W}^{\ast} \mathbf{h}^{(i-1)} + \mathbf{A}^{\ast} \mathbf{x}^{(i)}\right) \nonumber\\
		&= \frac{\dout }{m} \sum_{i=1}^{L}  \sum_{s' \in [\dout ]} \sum_{r' \in [p]} \sum_{r \in [m]}  b_{r, s'} b_{r', s'}^{\dagger} \mathbf{Back}_{i \to L, r, s} \nonumber\\& \quad \quad \quad \quad H_{r', s'}\Big(\theta_{r', s'} \langle \mathbf{w}_{r}, \overline{\mathbf{W}}^{[L]} \mathbf{w}_{r', s'}^{\dagger}\rangle , \sqrt{m/2} a_{r, d}\Big) \mathbb{I}_{\mathbf{w}_r^{\top} \mathbf{h}^{(i-1)} + \mathbf{a}_r^{\top} \mathbf{x}^{(i)} \ge 0} 
	\end{align}
	\endgroup
	First of all, we can't show that the above formulation concentrates on the required signal, because of the dependencies of randomness between $\mathbf{W}$, $\mathbf{A}$, $\mathbf{Back}$, $\mathbf{\overline{\mathbf{W}}}^{[L]}$ and $\left\{\mathbf{h}^{(\ell)}\right\}_{\ell \in [L]}$.  To decouple this randomness, we use the fact that ESNs are stable to re-randomization of few rows of the weight matrices and follow the proof technique of Lemma G.3 in \cite{allen2019can}. Choose a random subset $\mathcal{K} \subset[m]$ of size $|\mathcal{K}|=N$. Define the function $F^{(L), \mathcal{K}}_s$ as
	\begin{align*}
		F_{s}^{(L), \mathcal{K}}(\mathbf{h}^{(L-1)}, \mathbf{x}^{(L)}) &\stackrel{\text { def }}{=} \frac{\dout }{m} \sum_{i=1}^{L}  \sum_{s' \in [\dout ]} \sum_{r' \in [p]} \sum_{r \in \mathcal{K}}  b_{r, s'} b_{r', s'}^{\dagger} \mathbf{Back}_{i \to L, r, s} \nonumber\\& \quad \quad \quad \quad H_{r', s'}\Big(\theta_{r', s'} \langle \mathbf{w}_{r}, \overline{\mathbf{W}}^{[L]} \mathbf{w}_{r', s'}^{\dagger}\rangle , \sqrt{m/2} a_{r, d}\Big) \mathbb{I}_{\mathbf{w}_r^{\top} \mathbf{h}^{(i-1)} + \mathbf{a}_r^{\top} \mathbf{x}^{(i)} \ge 0}.  
	\end{align*}
	
	We show the following claim.
	\begin{claim}\label{claim:singlesubset_generalization}
		With probability at least $1-e^{-\Omega(\rho^2)}$, for any $\varepsilon \in (0, \min_{r, s} \frac{1}{C_s(\Phi_{r, s}, \mathcal{O}(\varepsilon_x^{-1}) ) })$,
		\begin{align*}      &\abs{F^{(L), \mathcal{K}}_s(\mathbf{h}^{(L-1)}, \mathbf{x}^{(L)}) - \frac{\dout }{m} \sum_{i=1}^{L}  \sum_{s' \in [\dout ]} \sum_{r' \in [p]} \sum_{r \in \mathcal{K}}  b_{r, s'} b_{r', s'}^{\dagger} \mathbf{Back}_{i \to L, r, s} \Phi_{r', s} \left(\left\langle \mathbf{w}_{r', s}^{\dagger}, [\overline{\mathbf{x}}^{(1)}, \cdots, \overline{\mathbf{x}}^{(L)}]\right\rangle\right)} \\&\le  \mathcal{O}(\dout Lp\rho^8  \mathfrak{C}_{\varepsilon}(\Phi, \mathcal{O}(\varepsilon_x^{-1})) N^{5/3} m^{-7/6}) + \frac{\dout }{m} \cdot \mathcal{O}(\mathfrak{C}_{\varepsilon}(\Phi_{r' s}, \mathcal{O}(\varepsilon_x^{-1})) \rho^2 \sqrt{\dout LpN}) \\& + \frac{\dout LpN}{m} \rho^2 (\varepsilon + \mathcal{O}( L_{\Phi}\rho^5 (N/m)^{1/6}) + \mathcal{O}(\varepsilon_x^{-1} L_{\Phi} L^4 \rho^{11} m^{-1/12} +  L_{\Phi} \rho^2 L^{11/6} \varepsilon_x^{2/3})) + \mathcal{O}(\rho^8\dout Lp N^{7/6} m^{-7/6}) .
		\end{align*}
	\end{claim}
	The above claim has been restated and proven in claim~\ref{claim:singlesubset_generalization_proof}. The above claim states that the function $F^{(L), \mathcal{K}}_s(\mathbf{h}^{(L-1)}, \mathbf{x}^{(L)})$ contains information about the true function.
	
	To complete the proof, we divide the set of neurons into $m/N$ disjoint sets $\mathcal{K}_1, \cdots, \mathcal{K}_{m/N}$, each set is of size $N$. We apply the Claim~\ref{claim:singlesubset_generalization} to each subset $\mathcal{K}_i$ and then add up the errors from each subset. That is, with probability at least $1 - \frac{m}{N}e^{-\Omega(\rho^2)}$,
			\begingroup \allowdisplaybreaks
			\begin{align*}
				&F_s^{(L)}(\mathbf{h}^{(\ell-1)}, \mathbf{x}^{(\ell)}) \\&=  \sum_{j=1}^{m/N}  F_s^{(L),\mathcal{K}_i}(\mathbf{h}^{(\ell-1)}, \mathbf{x}^{(\ell)})\\
				&= \sum_{j=1}^{m/N} \frac{\dout }{m} \sum_{i=1}^{L}  \sum_{s' \in [\dout ]} \sum_{r' \in [p]} \sum_{r \in \mathcal{K}_j}  b_{r, s'} b_{r', s'}^{\dagger} \mathbf{Back}_{i \to L, r, s} \Phi_{r', s} \left(\left\langle \mathbf{w}_{r', s}^{\dagger}, [\overline{\mathbf{x}}^{(2)}, \cdots, \overline{\mathbf{x}}^{(L-1)}]\right\rangle\right) 
				+ \sum_{j=1}^{m/N} error_{\mathcal{K}_j} \\&
				=  \frac{\dout }{m} \sum_{i=1}^{L}  \sum_{s' \in [\dout ]} \sum_{r' \in [p]} \sum_{r \in [m]}  b_{r, s'} b_{r', s'}^{\dagger} \mathbf{Back}_{i \to L, r, s} \Phi_{r', s} \left(\left\langle \mathbf{w}_{r', s}^{\dagger}, [\overline{\mathbf{x}}^{(2)}, \cdots, \overline{\mathbf{x}}^{(L-1)}]\right\rangle\right) 
				+ \sum_{j=1}^{m/N} error_{\mathcal{K}_j}, 
			\end{align*}
			where by Claim~\ref{claim:singlesubset_generalization},
			\begin{align*}
				\abs{error_{\mathcal{K}_i}} &\le \mathcal{O}(\dout Lp\rho^8  \mathfrak{C}_{\varepsilon}(\Phi, \mathcal{O}(\varepsilon_x^{-1})) N^{5/3} m^{-7/6}) + \frac{\dout }{m} \cdot \mathcal{O}(\mathfrak{C}_{\varepsilon}(\Phi_{r' s}, \mathcal{O}(\varepsilon_x^{-1})) \rho^2 \sqrt{\dout LpN}) + \\& + \frac{\dout LpN}{m} \rho^2 (\varepsilon + \mathcal{O}( L_{\Phi}\rho^5 (N/m)^{1/6}) + \mathcal{O}(\varepsilon_x^{-1} L_{\Phi} L^4 \rho^{11} m^{-1/12} +  L_{\Phi} L^{11/6} \rho^2 \varepsilon_x^{2/3})) + \\& \mathcal{O}(\rho^8\dout Lp N^{7/6} m^{-7/6}).
			\end{align*}
			\endgroup
			Thus,
			\begin{align*}
				&\abs{ F_s^{(L)}(\mathbf{h}^{(\ell-1)}, \mathbf{x}^{(\ell)}) - \frac{\dout }{m} \sum_{i=1}^{L}  \sum_{s' \in [\dout ]} \sum_{r' \in [p]} \sum_{r \in [m]}  b_{r, s'} b_{r', s'}^{\dagger} \mathbf{Back}_{i \to L, r, s} \Phi_{r', s} \left(\left\langle \mathbf{w}_{r', s}^{\dagger}, [\overline{\mathbf{x}}^{(2)}, \cdots, \overline{\mathbf{x}}^{(L-1)}]\right\rangle\right) } \\& \le \mathcal{O}(\dout Lp\rho^8  \mathfrak{C}_{\varepsilon}(\Phi, \mathcal{O}(\varepsilon_x^{-1})) N^{2/3} m^{-1/6}) + \frac{\sqrt{\dout ^3Lp}}{\sqrt{N}} \cdot \mathcal{O}(\mathfrak{C}_{\varepsilon}(\Phi_{r' s}, \mathcal{O}(\varepsilon_x^{-1})) \rho^2 )  \\& + \dout Lp \rho^2 (\varepsilon + \mathcal{O}( L_{\Phi}\rho^5 (N/m)^{1/6}) + \mathcal{O}(\varepsilon_x^{-1} L_{\Phi} L^4 \rho^{11} m^{-1/12} +  L_{\Phi} L^{11/6} \rho^2 \varepsilon_x^{2/3})) + \mathcal{O}(\rho^8\dout Lp N^{1/6} m^{-1/6}),
			\end{align*}
			with probability at least $1 - \frac{m}{N}e^{-\Omega(\rho^2)}$.
			Choosing $N = m^{0.2}$, we have
			\begin{align}
				&\abs{ F_s^{(L)}(\mathbf{h}^{(\ell-1)}, \mathbf{x}^{(\ell)}) - \frac{\dout }{m} \sum_{i=1}^{L}  \sum_{s' \in [\dout ]} \sum_{r' \in [p]} \sum_{r \in [m]}  b_{r, s'} b_{r', s'}^{\dagger} \mathbf{Back}_{i \to L, r, s} \Phi_{r', s} \left(\left\langle \mathbf{w}_{r', s}^{\dagger}, [\overline{\mathbf{x}}^{(2)}, \cdots, \overline{\mathbf{x}}^{(L-1)}]\right\rangle\right) }\nonumber \\& \le \mathcal{O}(\dout Lp\rho^8  \mathfrak{C}_{\varepsilon}(\Phi, \mathcal{O}(\varepsilon_x^{-1}))  m^{-1/30}) + \mathcal{O}(\mathfrak{C}_{\varepsilon}(\Phi_{r' s}, \mathcal{O}(\varepsilon_x^{-1})) \rho^2 \sqrt{\dout ^3Lp} m^{-0.1})  \\& + \dout Lp \rho^2 (\varepsilon + \mathcal{O}( L_{\Phi}\rho^5 m^{-2/15}) + \mathcal{O}(\varepsilon_x^{-1} L_{\Phi} L^4 \rho^{11} m^{-1/12} +  L_{\Phi} L^{11/6} 
				\rho^2\varepsilon_x^{2/3})) + \mathcal{O}(\rho^8\dout Lp  m^{-2/15}) \label{eq:prefinalf},
			\end{align}
			with probability at least $1 -  m^{0.8} e^{-\Omega(\rho^2)} \ge 1 - e^{-\Omega(\rho^2)}$.  
			
			Now, in the next claim, we show that the $f$ concentrates on the desired term. 
			\begin{claim}\label{claim:simplifybig}
				With probability exceeding $1 - e^{-\Omega(\rho^2)}$,
				\begin{align*}
					&\Big|  b_{r', s}^{\dagger}  \Phi_{r', s} \left(\left\langle \mathbf{w}_{r', s}^{\dagger}, [\overline{\mathbf{x}}^{(2)}, \cdots, \overline{\mathbf{x}}^{(L-1)}]\right\rangle\right) \\& \quad \quad \quad - \frac{\dout }{m} \sum_{i=1}^{L}  \sum_{s' \in [\dout ]}  \sum_{r \in [m]}  b_{r, s'} b_{r', s'}^{\dagger} \mathbf{Back}_{i \to L, r, s} \Phi_{r', s} \left(\left\langle \mathbf{w}_{r', s}^{\dagger}, [\overline{\mathbf{x}}^{(2)}, \cdots, \overline{\mathbf{x}}^{(L-1)}]\right\rangle\right) \Big| \\&\le  \mathcal{O}(L\dout  \rho C_{\Phi} m^{-0.25}).
				\end{align*}
			\end{claim}
			The claim is restated and proven in claim~\ref{claim:simplifybig_proof}.	
				
			Thus, introducing claim~\ref{claim:simplifybig} in eq.~\ref{eq:prefinalf}, we have
			\begingroup \allowdisplaybreaks
			\begin{align*}
				&\abs{ F_s^{(L)}(\mathbf{h}^{(\ell-1)}, \mathbf{x}^{(\ell)}) - \sum_{r' \in [p]}  b_{r', s}^{\dagger} \Phi_{r', s} \left(\left\langle \mathbf{w}_{r', s}^{\dagger}, [\overline{\mathbf{x}}^{(2)}, \cdots, \overline{\mathbf{x}}^{(L-1)}]\right\rangle\right) } \\& \le \mathcal{O}(\dout Lp\rho^8  \mathfrak{C}_{\varepsilon}(\Phi, \mathcal{O}(\varepsilon_x^{-1}))  m^{-1/30}) + \mathcal{O}(\mathfrak{C}_{\varepsilon}(\Phi_{r' s}, \mathcal{O}(\varepsilon_x^{-1})) \rho^2 \sqrt{\dout ^3Lp} m^{-0.1}) \\& + \dout Lp \rho^2 (\varepsilon + \mathcal{O}( L_{\Phi}\rho^5 m^{-2/15}) + \mathcal{O}(\varepsilon_x^{-1} L_{\Phi} L^4 \rho^{11} m^{-1/12} +  L_{\Phi} L^{11/6} \rho^2 \varepsilon_x^{2/3})) \\& + \mathcal{O}(\rho^8\dout Lp  m^{-2/15})  + \mathcal{O}(Lp\dout  \rho C_{\Phi} m^{-0.25}) \\&
				\le \mathcal{O}(\dout Lp\rho^2 \varepsilon + \dout L^{17/6} p \rho^4 L_{\Phi} \varepsilon_x^{2/3} + \dout ^{3/2} L^5 p \rho^{11} L_{\Phi} C_{\Phi}  \mathfrak{C}_{\varepsilon}(\Phi, \mathcal{O}(\varepsilon_x^{-1}))  m^{-1/30} ) .
			\end{align*}
			\endgroup
		\end{proof}

\subsection{Proof of Claim~\ref{claim:singlesubset_generalization}}

\begin{claim}[Restating claim~\ref{claim:singlesubset_generalization}]\label{claim:singlesubset_generalization_proof}
	With probability at least $1-e^{-\Omega(\rho^2)}$, for any $\varepsilon \in (0, \min_{r, s} \frac{1}{C_s(\Phi_{r, s}, \mathcal{O}(\varepsilon_x^{-1}) ) })$,
	\begin{align*}      &\abs{F^{(L), \mathcal{K}}_s(\mathbf{h}^{(L-1)}, \mathbf{x}^{(L)}) - \frac{\dout }{m} \sum_{i=1}^{L}  \sum_{s' \in [\dout ]} \sum_{r' \in [p]} \sum_{r \in \mathcal{K}}  b_{r, s'} b_{r', s'}^{\dagger} \mathbf{Back}_{i \to L, r, s} \Phi_{r', s} \left(\left\langle \mathbf{w}_{r', s}^{\dagger}, [\overline{\mathbf{x}}^{(1)}, \cdots, \overline{\mathbf{x}}^{(L)}]\right\rangle\right)} \\&\le  \mathcal{O}(\dout Lp\rho^8  \mathfrak{C}_{\varepsilon}(\Phi, \mathcal{O}(\varepsilon_x^{-1})) N^{5/3} m^{-7/6}) + \frac{\dout }{m} \cdot \mathcal{O}(\mathfrak{C}_{\varepsilon}(\Phi_{r' s}, \mathcal{O}(\varepsilon_x^{-1})) \rho^2 \sqrt{\dout LpN}) \\& + \frac{\dout LpN}{m} \rho^2 (\varepsilon + \mathcal{O}( L_{\Phi}\rho^5 (N/m)^{1/6}) + \mathcal{O}(\varepsilon_x^{-1} L_{\Phi} L^4 \rho^{11} m^{-1/12} +  L_{\Phi} \rho^2 L^{11/6} \varepsilon_x^{2/3})) + \mathcal{O}(\rho^8\dout Lp N^{7/6} m^{-7/6}) .
	\end{align*}
\end{claim}

\begin{proof}
We will replace the rows $\left\{\mathbf{w}_{k}, \mathbf{a}_{k}\right\}_{k \in \mathcal{K}}$ of $\mathbf{W}$ and $\mathbf{A}$ with freshly new i.i.d. samples $\widetilde{\mathbf{w}}_{k}, \widetilde{\mathbf{a}}_{k} \sim \mathcal{N}\left(0, \frac{2}{m} \mathbf{I}\right).$ to form new matrices $\widetilde{\mathbf{W}}$ and $\widetilde{\mathbf{A}}$. For the given sequence, we follow the notation of Lemma~\ref{lemma:rerandESN} to denote the hidden states corresponding to the old and the new weight matrices. Let $\widetilde{F}^{(L), \mathcal{K}}_s$ denote the following function:
\begin{align*}
	\widetilde{F}_{s}^{(L), \mathcal{K}}(\widetilde{\mathbf{h}}^{(L-1)}, \mathbf{x}^{(L)}) &\stackrel{\text { def }}{=} \frac{\dout }{m} \sum_{i=1}^{L}  \sum_{s' \in [\dout ]} \sum_{r' \in [p]} \sum_{r \in \mathcal{K}}  b_{r, s'} b_{r', s'}^{\dagger} \widetilde{\mathbf{Back}}_{i \to L, r, s} \nonumber\\& \quad \quad \quad \quad H_{r', s'}\Big(\widetilde{\theta}_{r', s'} \langle \mathbf{w}_{r}, \overline{\widetilde{\mathbf{W}}}^{[L]} \mathbf{w}_{r', s'}^{\dagger}\rangle , \sqrt{m/2} a_{r, d}\Big) \mathbb{I}_{\mathbf{w}_r^{\top} \widetilde{\mathbf{h}}^{(i-1)} + \mathbf{a}_r^{\top} \mathbf{x}^{(i)} \ge 0}, 
\end{align*}
where
\begin{equation*} 
	\widetilde{\theta}_{r', s} = \frac{\sqrt{m/2}}{ \norm[2]{\overline{\widetilde{\mathbf{W}}}^{[L]} \mathbf{w}_{r', s}^{\dagger}}}.
\end{equation*}
Using similar technique used to find the bounds of $\theta_{r', s}$ in eq.~\ref{eq:abstheta_lowerbound}, we ca show that $\forall  r' \in [p], s \in [\dout ]$, w.p. at least $1-e^{-\Omega(\rho^2)}$ over $\widetilde{\mathbf{W}}, \widetilde{\mathbf{A}}$,
\begin{align}
	\left(1 + \frac{1}{100L}\right)^{-L}  \le \sqrt{2/m} \widetilde{\theta}_{r', s} \le  \left(1 - \frac{1}{100L}\right)^{-L}  \label{eq:abstildetheta_lowerbound}.
\end{align}
Again, there is one important relation between $\theta_{r', s}$  and $\widetilde{\theta}_{r', s}$ that we will require later on, which we prove in the next claim.
\begin{claim}\label{claim:invtildethetatheta}
	With probability at least $1-e^{-\Omega(\rho^2)}$, for all $r' \in [p], s \in [\dout ]$,
	\begin{align*}
		\abs{\widetilde{\theta}_{r', s} \theta_{r', s}^{-1} - 1} \le \mathcal{O}(\rho^5 (N/m)^{1/6}).
	\end{align*}
\end{claim}
The claim has been restated and proven in claim~\ref{claim:invtildethetatheta_proof}. A simple corollary of the above claim is given below.

\begin{corollary}
	\label{cor:tildetheta_diff_theta}
	With probability at least $1-e^{-\Omega(\rho^2)}$, for all $r' \in [p], s \in [\dout ]$,
	\begin{align*}
		\abs{\widetilde{\theta}_{r', s} - \theta_{r', s}} \le \mathcal{O}(\rho^5 (N/m)^{1/6}).
	\end{align*}
\end{corollary}
The above corollary follows from the bounds on $\theta_{r',s}$ from eq.~\ref{eq:abstheta_lowerbound}.

We will first show that $\widetilde{F}_{s}^{(L), \mathcal{K}}(\widetilde{\mathbf{h}}^{(L-1)}, \mathbf{x}^{(L)})$ and $F_{s}^{(L), \mathcal{K}}(\mathbf{h}^{(L-1)}, \mathbf{x}^{(L)})$ are close. The claim has been restated and proven in claim~\ref{claim:diffftildef_proof}.

\begin{claim}\label{claim:diffftildef}
	With probability at least $1-e^{-\Omega(\rho^2)}$,
	\begingroup\allowdisplaybreaks
	\begin{align*}
		\abs{\widetilde{F}_{s}^{(L), \mathcal{K}}(\widetilde{\mathbf{h}}^{(L-1)}, \mathbf{x}^{(L)}) - F_{s}^{(L), \mathcal{K}}(\mathbf{h}^{(L-1)}, \mathbf{x}^{(L)})} \le \mathcal{O}(\dout Lp\rho^8  \mathfrak{C}_{\varepsilon}(\Phi, \mathcal{O}(\varepsilon_x^{-1})) N^{5/3} m^{-7/6}).
	\end{align*}
	\endgroup
\end{claim}

Now, we show that $\widetilde{F}$ is close to the desired signal in the two claims below. 
	\begin{claim}\label{claim:difftildefphi}
	With probability at least $1-e^{-\Omega(\rho^2)}$,
	\begin{align*}
		\Big|&\widetilde{F}_{s}^{(L), \mathcal{K}}(\widetilde{\mathbf{h}}^{(L-1)}, \mathbf{x}^{(L)}) - \frac{\dout }{m} \sum_{i=1}^{L}  \sum_{s' \in [\dout ]} \sum_{r' \in [p]} \sum_{r \in \mathcal{K}}  b_{r, s'} b_{r', s'}^{\dagger} \widetilde{\mathbf{Back}}_{i \to L, r, s} \Phi_{r', s} \left(\left\langle \mathbf{w}_{r', s}^{\dagger}, [\overline{\mathbf{x}}^{(1)}, \cdots, \overline{\mathbf{x}}^{(L)}]\right\rangle\right)\Big| \\&\le \frac{\dout }{m} \cdot \mathcal{O}(\mathfrak{C}_{\varepsilon}(\Phi_{r' s}, \mathcal{O}(\varepsilon_x^{-1})) \rho^2 \sqrt{\dout LpN}) \\& + \frac{\dout LpN}{m} \rho^2 (\varepsilon + \mathcal{O}( L_{\Phi}\rho^5 (N/m)^{1/6}) + \mathcal{O}(L_{\Phi}\varepsilon_x^{-1}  L^4 \rho^{11} m^{-1/12} +  L_{\Phi} \rho^2 L^{11/6} \varepsilon_x^{2/3})),
	\end{align*}
	for any $\varepsilon \in (0, \min_{r, s} \frac{\sqrt{3}}{C_s(\Phi_{r, s}, \varepsilon_x^{-1})})$.
\end{claim}

\begin{claim}\label{claim:fbacktildeback}
	with probability at least $1 - e^{-\Omega(\rho^2)}$,
	\begin{align*}
		&\Big| \frac{\dout }{m} \sum_{i=1}^{L}  \sum_{s' \in [\dout ]} \sum_{r' \in [p]} \sum_{r \in \mathcal{K}}  b_{r, s'} b_{r', s'}^{\dagger} \widetilde{\mathbf{Back}}_{i \to L, r, s} \Phi_{r', s} \left(\left\langle \mathbf{w}_{r', s}^{\dagger}, [\overline{\mathbf{x}}^{(1)}, \cdots, \overline{\mathbf{x}}^{(L)}]\right\rangle\right) \\& - \frac{\dout }{m} \sum_{i=1}^{L}  \sum_{s' \in [\dout ]} \sum_{r' \in [p]} \sum_{r \in \mathcal{K}}  b_{r, s'} b_{r', s'}^{\dagger} \mathbf{Back}_{i \to L, r, s} \Phi_{r', s} \left(\left\langle \mathbf{w}_{r', s}^{\dagger}, [\overline{\mathbf{x}}^{(1)}, \cdots, \overline{\mathbf{x}}^{(L)}]\right\rangle\right) \Big| \\& \le \mathcal{O}(\rho^8 C_{\Phi} \dout Lp N^{7/6} m^{-7/6}). 
	\end{align*}
\end{claim}
The above two claims have been restated and proven in claim~\ref{claim:difftildefphi_proof} and~\ref{claim:fbacktildeback_proof} respectively.

	Thus, from Claim~\ref{claim:diffftildef}, Claim~\ref{claim:difftildefphi} and Claim~\ref{claim:fbacktildeback}, we have with probability at least $1-e^{-\Omega(\rho^2)}$, for any $\varepsilon \in (0, \min_{r, s} \frac{1}{C_s(\Phi_{r, s}, \mathcal{O}(\varepsilon_x^{-1}) ) })$,
\begingroup \allowdisplaybreaks
\begin{align*}      
	&\abs{F^{(L), \mathcal{K}}_s(\mathbf{h}^{(L-1)}, \mathbf{x}^{(L)}) - \frac{\dout }{m} \sum_{i=1}^{L}  \sum_{s' \in [\dout ]} \sum_{r' \in [p]} \sum_{r \in \mathcal{K}}  b_{r, s'} b_{r', s'}^{\dagger} \mathbf{Back}_{i \to L, r, s} \Phi_{r', s} \left(\left\langle \mathbf{w}_{r', s}^{\dagger}, [\overline{\mathbf{x}}^{(2)}, \cdots, \overline{\mathbf{x}}^{(L-1)}]\right\rangle\right)} \\&\le 
	\abs{F^{(L), \mathcal{K}}_s(\mathbf{h}^{(L-1)}, \mathbf{x}^{(L)}) - \widetilde{F}^{(L), \mathcal{K}}_s(\widetilde{\mathbf{h}}^{(L-1)}, \mathbf{x}^{(L)})} \\& +\abs{\widetilde{F}^{(L), \mathcal{K}}_s(\widetilde{\mathbf{h}}^{(L-1)}, \mathbf{x}^{(L)}) - \frac{\dout }{m} \sum_{i=1}^{L}  \sum_{s' \in [\dout ]} \sum_{r' \in [p]} \sum_{r \in \mathcal{K}}  b_{r, s'} b_{r', s'}^{\dagger} \widetilde{\mathbf{Back}}_{i \to L, r, s} \Phi_{r', s} \left(\left\langle \mathbf{w}_{r', s}^{\dagger}, [\overline{\mathbf{x}}^{(2)}, \cdots, \overline{\mathbf{x}}^{(L-1)}]\right\rangle\right)} \\& + \Big| \frac{\dout }{m} \sum_{i=1}^{L}  \sum_{s' \in [\dout ]} \sum_{r' \in [p]} \sum_{r \in \mathcal{K}}  b_{r, s'} b_{r', s'}^{\dagger} \widetilde{\mathbf{Back}}_{i \to L, r, s} \Phi_{r', s} \left(\left\langle \mathbf{w}_{r', s}^{\dagger}, [\overline{\mathbf{x}}^{(2)}, \cdots, \overline{\mathbf{x}}^{(L-1)}]\right\rangle\right) \\& -  \frac{\dout }{m} \sum_{i=1}^{L}  \sum_{s' \in [\dout ]} \sum_{r' \in [p]} \sum_{r \in \mathcal{K}}  b_{r, s'} b_{r', s'}^{\dagger} \mathbf{Back}_{i \to L, r, s} \Phi_{r', s} \left(\left\langle \mathbf{w}_{r', s}^{\dagger}, [\overline{\mathbf{x}}^{(2)}, \cdots, \overline{\mathbf{x}}^{(L-1)}]\right\rangle\right) \Big|
	\\&
	\le \mathcal{O}(\dout Lp\rho^8  \mathfrak{C}_{\varepsilon}(\Phi, \mathcal{O}(\varepsilon_x^{-1})) N^{5/3} m^{-7/6}) + \frac{\dout }{m} \cdot \mathcal{O}(\mathfrak{C}_{\varepsilon}(\Phi_{r' s}, \mathcal{O}(\varepsilon_x^{-1})) \rho^2 \sqrt{\dout Lp N})  \\& + \frac{\dout LpN}{m} \rho^2 (\varepsilon + \mathcal{O}( L_{\Phi}\rho^5 (N/m)^{1/6}) + \mathcal{O}(\varepsilon_x^{-1} L_{\Phi} L^4 \rho^{11} m^{-1/12} +  L_{\Phi} L^{11/6} \rho^2 \varepsilon_x^{2/3})) + \mathcal{O}(\rho^8\dout Lp N^{7/6} m^{-7/6}) .
\end{align*}
\endgroup

\end{proof}


\subsection{Helping lemmas}
\subsubsection{Function approximation using hermite polynomials}
The following theorem on approximating a smooth function using hermite polynomials has been taken from \cite{allen2019learning} and we will use this theorem to show that pseudo RNNs can approximate the target concept class.
\begin{theorem}[Lemma 6.2 in \cite{allen2019learning}]\label{Thm:Smooth_H}
	For every smooth function $\phi$, every $\varepsilon \in \left(0, \frac{1}{\mathfrak{C}_{s}\left(\phi, 1\right)}\right)$ there exists a $H:\mathbb{R}^2 \to \left(-\mathfrak{C}_\varepsilon\left(\phi, 1\right), \mathfrak{C}_\varepsilon\left(\phi, 1\right)\right)$, satisfying $\abs{H} \le \mathfrak{C}_\varepsilon\left(\phi, 1\right)$, and is $\mathfrak{C}_\varepsilon\left(\phi, 1\right)$-lipschitz continuous in the first variable and for all $x_1 \in (-1, 1)$
	\begin{equation*}
		\left|\mathbb{E}_{\alpha_1, \beta_1, b_0}\left[\mathbb{I}_{\alpha_{1} x_{1}+\beta_{1} \sqrt{1-x_{1}^{2}} + b_0   \geq 0} {H\left(\alpha_{1}, b_0\right)}\right]-\phi\left(x_{1}\right)\right| \leq \varepsilon
	\end{equation*}
	where $\alpha_1, \beta_1\text{ and } b_0 \sim \mathcal{N}\left(0, 1\right)$ are independent random variables.
\end{theorem}

In \cite{allen2019learning}, the function $H$ is shown to be lipschitz continuous in expectation w.r.t. the first variable $\alpha_1$ which follows a normal distribution. However, one can also show that the function $H$ is lipschitz continuous w.r.t. the first variable, even when the variable is perturbed by bounded noise to a variable that does not necessarily follow a gaussian distribution i.e. one can show that
\begin{align*}
	\left|\mathbb{E}_{\alpha_1, \beta_1, b_0 \sim \mathcal{N}(0, 1)} \mathbb{E}_{\theta: \abs{\theta} \le \gamma} \left[H\left(\alpha_{1}, b_0\right) - H\left(\alpha_{1} + \theta, b_0\right)\right]\right| \le \gamma \mathfrak{C}_\varepsilon(\phi, 1). 
\end{align*}
The proof will follow along the similar lines of Claim C.2 in \cite{allen2019learning}. We give a brief overview here. The function $H$ was shown to be a weighted combination of different hermite polynomials. Using the following property of hermite polynomials,
\begin{align*}
	h_i(x+y) = \sum_{k=0}^{i} {i \choose k} x^{i-k} h_k(y),
\end{align*}
we expand the function $H\left(\alpha_{1} + \theta, b_0\right)$ and then, bound each term using the procedure in Claim C.2 of \cite{allen2019learning}.
\todo{Ask Navin, if I have to add more details.}

\begin{corollary}\label{Cor:Smooth_H}
	For any $\sigma > 0$, $r_x > 0$ s.t. $\sigma \ge r_x/10$, $k_0 \ge 0$, and for every smooth function $\phi$, any $\varepsilon \in \left(0, \frac{r_x}{\sigma \mathfrak{C}_{s}\left(\phi, k_0 r_x\right)}\right)$ there exists a $H:\mathbb{R}^2 \to \left(- 
	\frac{\sigma}{r_x}\mathfrak{C}_\varepsilon\left(\phi,  k_0 r_x\right), \frac{\sigma}{r_x} \mathfrak{C}_\varepsilon\left(\phi,  k_0 r_x\right)\right)$, which is $ \frac{\sigma}{r_x}\mathfrak{C}_\varepsilon\left(\phi, k_0 r_x\right)$-lipschitz continuous and for all $x_1 \in (-r_x, r_x)$
	\begin{equation*}
		\left|\mathbb{E}_{\alpha_1, \beta_1, b_0}\left[\mathbb{I}_{\alpha_{1} x_{1}+\beta_{1} \sqrt{r_x^2-x_{1}^{2}} + b_0   \geq 0} {H\left(\alpha_{1}, b_0\right)}\right]-\phi\left(k_0 x_{1}\right)\right| \leq \varepsilon
	\end{equation*}
	where $\alpha_1, \beta_1 \sim \mathcal{N}\left(0, 1\right) \text{ and } b_0 \sim \mathcal{N}\left(0, \sigma^2\right)$ are independent random variables.
\end{corollary}

\begin{lemma}[Function Approximators]\label{Def:Function_approx}
	Let $r_x = \sqrt{2 + (L-2)\varepsilon_x^2}$. For each $\Phi_{r, s}$ and a constant $k_{0, r, s} = \Theta(\frac{1}{\varepsilon_x})$, there exists a function $H_{r, s}$ such that for any $\varepsilon \in (0, \min_{r, s} \frac{r_x}{ C_s(\Phi_{r, s}, k_{0, r, s} r_x)})$,  $H_{r, s}:\mathbb{R}^2 \to \left(-\frac{1}{r_x} \mathfrak{C}_\varepsilon\left(\Phi_{r, s}, k_{0, r, s} r_x\right),\frac{1}{r_x} \mathfrak{C}_\varepsilon\left(\Phi_{r, s}, k_{0, r, s} r_x\right)\right)$, is $\frac{1}{r_x} \mathfrak{C}_\varepsilon\left(\Phi_{r, s},  k_{0, r, s} r_x\right)$-lipschitz continuous, and for all $x_1 \in (-r_x, r_x)$
	\begin{equation*}
		\left|\mathbb{E}_{\alpha_1, \beta_1, b_0}\left[\mathbb{I}_{\alpha_{1} x_{1}+\beta_{1} \sqrt{r_x^2 - x_{1}^{2}} + b_0   \geq 0} {H_{r, s}\left(\alpha_{1}, b_0\right)}\right]-\Phi_{r, s}\left(k_{0, r, s} x_{1} \right)\right| \leq \varepsilon
	\end{equation*}
	where $\alpha_1, \beta_1 \sim \mathcal{N}\left(0, 1\right) \text{ and } b_0 \sim \mathcal{N}\left(0, 1\right)$ are independent random variables.
\end{lemma}
For any $\varepsilon_x \le \frac{1}{L}$, we can see that for all $\Phi_{r, s}$,
$\abs{H_{r, s}} \le \frac{1}{\sqrt{2}} \mathfrak{C}_\varepsilon\left(\Phi_{r, s},  \mathcal{O}(\varepsilon_x^{-1})\right)$ and $H_{r, s}$ is $\frac{1}{\sqrt{2}} \mathfrak{C}_\varepsilon\left(\Phi_{r, s},  \mathcal{O}(\varepsilon_x^{-1})\right)$ lipschitz, for any $\varepsilon \le \frac{\sqrt{3}}{C_s(\Phi_{r, s}, \mathcal{O}(\varepsilon_x^{-1}))}$.

\subsubsection{Proofs of the helping lemmas}
First, we mention one of the properties on correlations of $\mathbf{Back}_{i \to j}$ matrices, which will be heavily used later on.
\begin{lemma}[Lemma C.1 in \cite{allen2019can}]\label{lemma:backward_correlation}
	For every $\varepsilon_x < 1/L$ and every normalized input sequence, $\mathbf{x}_1, \mathbf{x}_2, ..., \mathbf{x}_{L}$, with probability at least 1 - $e^{-\Omega(\rho^2)}$ over $\mathbf{W}$, $\mathbf{A}$ and $\mathbf{B}$: for every $1 \le i \le j < j' \le L$, 
	\begin{equation*}
		\abs{\langle \mathbf{u}^{\top} \mathbf{Back}_{i \to j}, \mathbf{v}^{\top} \mathbf{Back}_{i \to j'} \rangle} \le \mathcal{O}\left(m^{0.75} \rho\right) \norm{\mathbf{u}} \norm{\mathbf{v}},
	\end{equation*}
	for any two vectors $\mathbf{u}$ and $\mathbf{v}$ in $\mathbb{R}^{\dout }$.
\end{lemma}





		\begin{claim}[Restating claim~\ref{claim:invtildethetatheta}]\label{claim:invtildethetatheta_proof}
			With probability at least $1-e^{-\Omega(\rho^2)}$, for all $r' \in [p], s \in [\dout ]$,
			\begin{align*}
				\abs{\widetilde{\theta}_{r', s} \theta_{r', s}^{-1} - 1} \le \mathcal{O}(\rho^5 (N/m)^{1/6}).
			\end{align*}
		\end{claim}
		
		\begin{proof}
			First of all,
			\begin{align*}
				\sqrt{2/m} \abs{\widetilde{\theta}_{r', s}^{-1} - \theta_{r', s}^{-1} } &= 
				\abs{\norm[2]{\overline{\widetilde{\mathbf{W}}}^{[L]} \mathbf{w}_{r', s}^{\dagger}} - \norm[2]{\overline{\mathbf{W}}^{[L]} \mathbf{w}_{r', s}^{\dagger}}} \\& \le 
				\norm[2]{\overline{\widetilde{\mathbf{W}}}^{[L]} \mathbf{w}_{r', s}^{\dagger}- \overline{\mathbf{W}}^{[L]} \mathbf{w}_{r', s}^{\dagger}}  \\&
				\le \max_{\ell \le L} \norm{\left(\widetilde{\mathbf{W}}^{(L, 
						\ell)} - \mathbf{W}^{(L, \ell)}\right) \mathbf{w}_{r', s}^{\dagger}}_2  \\&
				\le \mathcal{O}(\rho^5 (N/m)^{1/6})
			\end{align*}
			where in the pre-final step, we have used Lemma~\ref{lemma:rerandESN} to have w.p. exceeding $1-e^{-\Omega(\rho^2)}$ for any $\ell \le L$,
			\begin{align*}
				\norm{\left(\widetilde{\mathbf{W}}^{(L,
						\ell)} - \mathbf{W}^{(L, \ell)}\right) \mathbf{w}_{r', s}^{\dagger}}_2 &= \norm{\left( \prod_{L \ge \ell' \ge \ell} \widetilde{\mathbf{D}}_{(0)}^{(\ell)} \widetilde{\mathbf{W}}  - \prod_{L \ge \ell' \ge \ell} \mathbf{D}_{(0)}^{(\ell)} \mathbf{W} \right)  \mathbf{w}_{r', s}^{\dagger}}_2\\& \le  \mathcal{O}(\rho^5 (N/m)^{1/6}) \cdot \norm[2]{\mathbf{w}_{r', s}^{\dagger}} = \mathcal{O}(\rho^5 (N/m)^{1/6}).
			\end{align*}
			Hence,
			\begin{align*}
				\abs{ \widetilde{\theta}_{r', s} \theta_{r', s}^{-1} - 1} \le \sqrt{m/2} \abs{\widetilde{\theta}_{r', s} }  \mathcal{O}(\rho^5 (N/m)^{1/6}) \le
				\mathcal{O}(\rho^5 (N/m)^{1/6}),
			\end{align*}
			where we have used the upper bound on $\sqrt{m/2} \abs{\widetilde{\theta}_{r', s} }$ from eq.~\ref{eq:abstildetheta_lowerbound} in the final step. 
		\end{proof}

		
		\begin{claim}[Restating claim~\ref{claim:diffftildef}]\label{claim:diffftildef_proof}
			With probability at least $1-e^{-\Omega(\rho^2)}$,
			\begingroup\allowdisplaybreaks
			\begin{align*}
				\abs{\widetilde{F}_{s}^{(L), \mathcal{K}}(\widetilde{\mathbf{h}}^{(L-1)}, \mathbf{x}^{(L)}) - F_{s}^{(L), \mathcal{K}}(\mathbf{h}^{(L-1)}, \mathbf{x}^{(L)})} \le \mathcal{O}(\dout Lp\rho^8  \mathfrak{C}_{\varepsilon}(\Phi, \mathcal{O}(\varepsilon_x^{-1})) N^{5/3} m^{-7/6}).
			\end{align*}
			\endgroup
		\end{claim}
		
		\begin{proof}
			We break the required term into three different terms.
			\begingroup\allowdisplaybreaks
			\begin{align}
				&\abs{\widetilde{F}_{s}^{(L), \mathcal{K}}(\widetilde{\mathbf{h}}^{(L-1)}, \mathbf{x}^{(L)}) - F_{s}^{(L), \mathcal{K}}(\mathbf{h}^{(L-1)}, \mathbf{x}^{(L)})}\nonumber\\&=
				\Big| \frac{\dout }{m} \sum_{i=1}^{L}  \sum_{s' \in [\dout ]} \sum_{r' \in [p]} \sum_{r \in \mathcal{K}}  b_{r, s'} b_{r', s'}^{\dagger} \widetilde{\mathbf{Back}}_{i \to L, r, s} H_{r', s'}\Big(\widetilde{\theta}_{r', s'} \langle \mathbf{w}_{r}, \overline{\widetilde{\mathbf{W}}}^{[L]} \mathbf{w}_{r', s'}^{\dagger}\rangle , \sqrt{m/2} a_{r, d}\Big) \mathbb{I}_{\mathbf{w}_r^{\top} \widetilde{\mathbf{h}}^{(i-1)} + \mathbf{a}_r^{\top} \mathbf{x}^{(i)} \ge 0} \nonumber\\&
				\quad \quad -  \frac{\dout }{m} \sum_{i=1}^{L}  \sum_{s' \in [\dout ]} \sum_{r' \in [p]} \sum_{r \in \mathcal{K}}  b_{r, s'} b_{r', s'}^{\dagger} \mathbf{Back}_{i \to L, r, s}  H_{r', s'}\Big(\theta_{r', s'} \langle \mathbf{w}_{r}, \overline{\mathbf{W}}^{[L]} \mathbf{w}_{r', s'}^{\dagger}\rangle , \sqrt{m/2} a_{r, d}\Big) \mathbb{I}_{\mathbf{w}_r^{\top} \mathbf{h}^{(i-1)} + \mathbf{a}_r^{\top} \mathbf{x}^{(i)} \ge 0} \Big| \nonumber \\&
				\le  \Big| \frac{\dout }{m} \sum_{i=1}^{L}  \sum_{s' \in [\dout ]} \sum_{r' \in [p]} \sum_{r \in \mathcal{K}}  b_{r, s'} b_{r', s'}^{\dagger} \widetilde{\mathbf{Back}}_{i \to L, r, s}  H_{r', s'}\Big(\widetilde{\theta}_{r', s'} \langle \mathbf{w}_{r}, \overline{\widetilde{\mathbf{W}}}^{[L]} \mathbf{w}_{r', s'}^{\dagger}\rangle , \sqrt{m/2} a_{r, d}\Big) \mathbb{I}_{\mathbf{w}_r^{\top} \widetilde{\mathbf{h}}^{(i-1)} + \mathbf{a}_r^{\top} \mathbf{x}^{(i)} \ge 0} \nonumber\\&
				\quad \quad -  \frac{\dout }{m} \sum_{i=1}^{L}  \sum_{s' \in [\dout ]} \sum_{r' \in [p]} \sum_{r \in \mathcal{K}}  b_{r, s'} b_{r', s'}^{\dagger} \mathbf{Back}_{i \to L, r, s}  H_{r', s'}\Big(\widetilde{\theta}_{r', s'} \langle \mathbf{w}_{r}, \overline{\widetilde{\mathbf{W}}}^{[L]} \mathbf{w}_{r', s'}^{\dagger}\rangle , \sqrt{m/2} a_{r, d}\Big) \mathbb{I}_{\mathbf{w}_r^{\top} \widetilde{\mathbf{h}}^{(i-1)} + \mathbf{a}_r^{\top} \mathbf{x}^{(i)} \ge 0} \Big| \label{eq:rerandRNN_1}\\&
				\quad \quad + \Big| \frac{\dout }{m} \sum_{i=1}^{L}  \sum_{s' \in [\dout ]} \sum_{r' \in [p]} \sum_{r \in \mathcal{K}}  b_{r, s'} b_{r', s'}^{\dagger} \mathbf{Back}_{i \to L, r, s}  H_{r', s'}\Big(\widetilde{\theta}_{r', s'} \langle \mathbf{w}_{r}, \overline{\widetilde{\mathbf{W}}}^{[L]} \mathbf{w}_{r', s'}^{\dagger}\rangle , \sqrt{m/2} a_{r, d}\Big) \mathbb{I}_{\mathbf{w}_r^{\top} \widetilde{\mathbf{h}}^{(i-1)} + \mathbf{a}_r^{\top} \mathbf{x}^{(i)} \ge 0} \nonumber\\&
				\quad \quad -  \frac{\dout }{m} \sum_{i=1}^{L}  \sum_{s' \in [\dout ]} \sum_{r' \in [p]} \sum_{r \in \mathcal{K}}  b_{r, s'} b_{r', s'}^{\dagger} \mathbf{Back}_{i \to L, r, s}  H_{r', s'}\Big(\widetilde{\theta}_{r', s'} \langle \mathbf{w}_{r}, \overline{\widetilde{\mathbf{W}}}^{[L]} \mathbf{w}_{r', s'}^{\dagger}\rangle , \sqrt{m/2} a_{r, d}\Big) \mathbb{I}_{\mathbf{w}_r^{\top} \mathbf{h}^{(i-1)} + \mathbf{a}_r^{\top} \mathbf{x}^{(i)} \ge 0} \Big| \label{eq:rerandRNN_2}\\&
				\quad\quad + \Big| \frac{\dout }{m} \sum_{i=1}^{L}  \sum_{s' \in [\dout ]} \sum_{r' \in [p]} \sum_{r \in \mathcal{K}}  b_{r, s'} b_{r', s'}^{\dagger} \mathbf{Back}_{i \to L, r, s}  H_{r', s'}\Big(\widetilde{\theta}_{r', s'} \langle \mathbf{w}_{r}, \overline{\widetilde{\mathbf{W}}}^{[L]} \mathbf{w}_{r', s'}^{\dagger}\rangle , \sqrt{m/2} a_{r, d}\Big) \mathbb{I}_{\mathbf{w}_r^{\top} \mathbf{h}^{(i-1)} + \mathbf{a}_r^{\top} \mathbf{x}^{(i)} \ge 0} \nonumber\\&
				\quad \quad -  \frac{\dout }{m} \sum_{i=1}^{L}  \sum_{s' \in [\dout ]} \sum_{r' \in [p]} \sum_{r \in \mathcal{K}}  b_{r, s'} b_{r', s'}^{\dagger} \mathbf{Back}_{i \to L, r, s}  H_{r', s'}\Big(\theta_{r', s'} \langle \mathbf{w}_{r}, \overline{\mathbf{W}}^{[L]} \mathbf{w}_{r', s'}^{\dagger}\rangle , \sqrt{m/2} a_{r, d}\Big) \mathbb{I}_{\mathbf{w}_r^{\top} \mathbf{h}^{(i-1)} + \mathbf{a}_r^{\top} \mathbf{x}^{(i)} \ge 0} \Big|. \label{eq:rerandRNN_3}
			\end{align}
			\endgroup
			
			We now show that each of the three equations, eq.~\ref{eq:rerandRNN_1}, eq.~\ref{eq:rerandRNN_2} and eq.~\ref{eq:rerandRNN_3} are small. First, we will need a couple of bounds on the terms that appear in the equations.
			\begin{itemize}
				\item Since $b_{r, s'} \sim \mathcal{N}(0, \frac{1}{\dout })$, using the fact~\ref{fact:max_gauss}, we can show that with probability $1-e^{-\Omega(\rho^2)}$,
				$\max_{r, s'} | b_{r, s'} | \le \frac{\rho}{\sqrt{\dout }}.$ 
				\item From the definition of concept class,  $\max_{r',s'} | b_{r', s'}^{\dagger} | \le 1$. 
				\item By the definition of $H$ from def~\ref{Def:Function_approx}, we have $\max_{r',s'} |  H_{r', s'}\Big(\widetilde{\theta}_{r', s'} \langle \mathbf{w}_{r}, \overline{\widetilde{\mathbf{W}}}^{[L]} \mathbf{w}_{r', s'}^{\dagger}\rangle , \sqrt{m/2} a_{r, d}\Big) | \le \mathfrak{C}_{\varepsilon}(\Phi, \mathcal{O}(\varepsilon_x^{-1}))$.
				\item Since $\mathbf{b}_s \sim \mathcal{N}(0, \frac{1}{\dout } \mathbf{I})$, using fact~\ref{lem:chi-squared}, we can show that $\norm{\mathbf{b}_s} \le \mathcal{O}(\frac{\rho}{\sqrt{\dout }})$, w.p. $1-e^{-\rho^2}$. Hence, from lemma~\ref{lemma:norm_ESN}, we have w.p. atleast $1 - e^{-\Omega(\rho^2)}$, for any $1 \le i \le j \le L, s \in [\dout ], r \in [m]$, $\abs{\mathbf{e}_s^{\top} \mathbf{Back}_{i \to j} \mathbf{e}_r} = 
				\abs{\mathbf{b}_s^{\top} \mathbf{D}^{(\ell)} \mathbf{W} \cdots \mathbf{D}^{(i+1)} \mathbf{W}\mathbf{e}_r} \le \norm[2]{\mathbf{b}_s} \norm[2]{\mathbf{D}^{(\ell)} \mathbf{W} \cdots \mathbf{D}^{(i+1)} \mathbf{W}\mathbf{e}_r} \le \mathcal{O}(\frac{\rho}{\sqrt{\dout }})$.
			\end{itemize}
			
			First, let's focus on eq.~\ref{eq:rerandRNN_1}. From Lemma~\ref{lemma:rerandESN}, we have with probability at least $1 - e^{-\Omega(\rho^2)}$,
			\begin{align*}
				\abs{\mathbf{e}_s^{\top} \left(\mathbf{Back_{i \to j}} - \widetilde{\mathbf{Back}}_{i \to j} \right) \mathbf{e}_r} 
				&= \abs{\mathbf{b}_s^{\top} \left( \mathbf{D}^{(j)} \mathbf{W} \cdots \mathbf{D}^{(i)} \mathbf{W} - \widetilde{\mathbf{D}}^{(j)} \widetilde{\mathbf{W}} \cdots \widetilde{\mathbf{D}}^{(i)} \widetilde{\mathbf{W}} \right) \mathbf{e}_r}
				\\&
				\le \norm[2]{\mathbf{b}_s} \norm[2]{\left( \mathbf{D}^{(j)} \mathbf{W} \cdots \mathbf{D}^{(i)} \mathbf{W} - \widetilde{\mathbf{D}}^{(j)} \widetilde{\mathbf{W}} \cdots \widetilde{\mathbf{D}}^{(i)} \widetilde{\mathbf{W}} \right) \mathbf{e}_r}
				\\&\le \mathcal{O}(\rho^7 \dout ^{-1/2} N^{1/6} m^{-1/6}) , \text{ for all }  r \in [m], s \in [\dout ] \text{ and } 1 \le i \le j \le L.
			\end{align*}
			Thus, 
			\begingroup \allowdisplaybreaks
			\begin{align}
				& \Big| \frac{\dout }{m} \sum_{i=1}^{L}  \sum_{s' \in [\dout ]} \sum_{r' \in [p]} \sum_{r \in \mathcal{K}}  b_{r, s'} b_{r', s'}^{\dagger} \widetilde{\mathbf{Back}}_{i \to L, r, s}  H_{r', s'}\Big(\widetilde{\theta}_{r', s'} \langle \mathbf{w}_{r}, \overline{\widetilde{\mathbf{W}}}^{[L]} \mathbf{w}_{r', s'}^{\dagger}\rangle , \sqrt{m/2} a_{r, d}\Big) \mathbb{I}_{\mathbf{w}_r^{\top} \widetilde{\mathbf{h}}^{(i-1)} + \mathbf{a}_r^{\top} \mathbf{x}^{(i)} \ge 0} \nonumber\\&
				-  \frac{\dout }{m} \sum_{i=1}^{L}  \sum_{s' \in [\dout ]} \sum_{r' \in [p]} \sum_{r \in \mathcal{K}}  b_{r, s'} b_{r', s'}^{\dagger} \mathbf{Back}_{i \to L, r, s}  H_{r', s'}\Big(\widetilde{\theta}_{r', s'} \langle \mathbf{w}_{r}, \overline{\widetilde{\mathbf{W}}}^{[L]} \mathbf{w}_{r', s'}^{\dagger}\rangle , \sqrt{m/2} a_{r, d}\Big) \mathbb{I}_{\mathbf{w}_r^{\top} \widetilde{\mathbf{h}}^{(i-1)} + \mathbf{a}_r^{\top} \mathbf{x}^{(i)} \ge 0} \Big|\nonumber\\&
				\le \sum_{i=1}^{L} \sum_{r' \in [p]} \sum_{s' \in [\dout ]} \sum_{r \in \mathcal{K}} \frac{\dout }{m} \abs{\mathbf{e}_s^{\top} \left(\mathbf{Back_{i \to j}} - \widetilde{\mathbf{Back}}_{i \to j} \right) \mathbf{e}_r}  \nonumber\\& \quad \quad \quad \quad \cdot \abs{b_{r, s'} b_{r', s'}^{\dagger}  H_{r', s'}\Big(\widetilde{\theta}_{r', s'} \langle \mathbf{w}_{r}, \overline{\widetilde{\mathbf{W}}}^{[L]} \mathbf{w}_{r', s'}^{\dagger}\rangle , \sqrt{m/2} a_{r, d}\Big) \mathbb{I}_{\mathbf{w}_r^{\top} \widetilde{\mathbf{h}}^{(i-1)} + \mathbf{a}_r^{\top} \mathbf{x}^{(i)} \ge 0}} \nonumber\\&
				\le \sum_{i=1}^{L} \sum_{r' \in [p]} \sum_{s' \in [\dout ]} \sum_{r \in \mathcal{K}} \frac{\dout }{m} \abs{\mathbf{e}_s^{\top} \left(\mathbf{Back_{i \to j}} - \widetilde{\mathbf{Back}}_{i \to j} \right) \mathbf{e}_r}  \nonumber\\& \quad \quad \quad \quad \cdot \abs{b_{r, s'}} \abs{b_{r', s'}^{\dagger}} \abs{  H_{r', s'}\Big(\widetilde{\theta}_{r', s'} \langle \mathbf{w}_{r}, \overline{\widetilde{\mathbf{W}}}^{[L]} \mathbf{w}_{r', s'}^{\dagger}\rangle , \sqrt{m/2} a_{r, d}\Big)} \abs{\mathbb{I}_{\mathbf{w}_r^{\top} \widetilde{\mathbf{h}}^{(i-1)} + \mathbf{a}_r^{\top} \mathbf{x}^{(i)} \ge 0}} \nonumber\\&
				\le \sum_{i=1}^{L} \sum_{r' \in [p]} \sum_{s' \in [\dout ]} \sum_{r \in \mathcal{K}} \frac{\dout }{m} \cdot \mathcal{O}(\dout ^{-1/2} \rho^7 (N/m)^{1/6}) \cdot \frac{\rho}{\sqrt{\dout }} \cdot 1 \cdot \mathfrak{C}_{\varepsilon}(\Phi, \mathcal{O}(\varepsilon_x^{-1})) \cdot 1 \\&
				\le \mathcal{O}( \dout  pL \rho^8 \mathfrak{C}_{\varepsilon}(\Phi_{r', s'}, \mathcal{O}(\varepsilon_x^{-1})) (N/m)^{7/6}) \label{eq:rerandRNN_1SOLVE}.
			\end{align}
			\endgroup

			Now, we focus on eq.~\ref{eq:rerandRNN_2}. Lemma~\ref{lemma:rerandESN} shows that with probability at least $1 - e^{-\Omega(\rho^2)}$,
			\begin{equation*}
				\abs[0]{\mathbf{w}_r^{\top} (\widetilde{\mathbf{h}}^{(L-1)} -  \mathbf{h}^{(L-1)})} \leq \mathcal{O}\left(\rho^{5} N^{2 / 3} m^{-2 / 3}\right) \quad \text { for every } r \in [m], \ell \in[L].
			\end{equation*}
			
			From lemma~\ref{lemma:norm_ESN}, we have w.p. at least $1 - e^{-\Omega(\rho^2)}$ for any $s \le \frac{\rho^2}{m}$,
			\begin{align*}
				\abs{\left\{r \in [m] \Big| \abs{\mathbf{w}_r^{\top} \mathbf{h}^{(L-1)} + \mathbf{a}_r^{\top} \mathbf{x}^{(L-1)}} \le \frac{s}{\sqrt{m}} \right\}} \le \mathcal{O}(sm).
			\end{align*}
			This can be modified for the subset $\mathcal{K}$,  w.p. at least $1 - e^{-\Omega(\rho^2)}$ for any $s \le \frac{\rho^2}{m}$,
			\begin{align*}
				\abs{\left\{r \in \mathcal{K} \Big| \abs{\mathbf{w}_r^{\top} \mathbf{h}^{(L-1)} + \mathbf{a}_r^{\top} \mathbf{x}^{(L-1)}} \le  \frac{s}{\sqrt{m}} \right\}} \le \mathcal{O}(sN).
			\end{align*}
			
			Thus, w.p. at least $1 - e^{-\Omega(\rho^2)}$,
			\begin{equation*}
				\sum_{r \in \mathcal{K}} \mathbb{I}\left[\abs{\mathbf{w}_r^{\top} \mathbf{h}^{(L-1)} + \mathbf{a}_r^{\top} \mathbf{x}^{(L-1)}} \le \rho^{5} N^{2 / 3} m^{-2 / 3}\right] \le \mathcal{O}(\rho^{5} N^{5 / 3} m^{-1/6}).
			\end{equation*}
			
			Hence, that implies  w.p. at least $1 - e^{-\Omega(\rho^2)}$,
			\begingroup \allowdisplaybreaks
			\begin{align*}
				&\sum_{r \in \mathcal{K}} \abs{\mathbb{I}\left[\mathbf{w}_r^{\top} \mathbf{h}^{(L-1)} + \mathbf{a}_r^{\top} \mathbf{x}^{(L-1)} \right] - \mathbb{I}\left[\mathbf{w}_r^{\top} \widetilde{\mathbf{h}}^{(L-1)} + \mathbf{a}_r^{\top} \mathbf{x}^{(L-1)} \right]}\\&
				\le \sum_{r \in \mathcal{K}}\mathbb{I}\left[\abs[0]{\mathbf{w}_r^{\top} \mathbf{h}^{(L-1)} + \mathbf{a}_r^{\top} \mathbf{x}^{(L-1)}} \le \abs[0]{\mathbf{w}_r^{\top} \widetilde{\mathbf{h}}^{(L-1)} + \mathbf{a}_r^{\top} \mathbf{x}^{(L-1)} - \mathbf{w}_r^{\top} \mathbf{h}^{(L-1)} - \mathbf{a}_r^{\top} \mathbf{x}^{(L-1)}} \right] \\&
				= \sum_{r \in \mathcal{K}}\mathbb{I}\left[\abs[0]{\mathbf{w}_r^{\top} \mathbf{h}^{(L-1)} + \mathbf{a}_r^{\top} \mathbf{x}^{(L-1)}} \le \abs[0]{\mathbf{w}_r^{\top} (\widetilde{\mathbf{h}}^{(L-1)} -  \mathbf{h}^{(L-1)}}) \right]
				\\&\le \sum_{r \in \mathcal{K}}\mathbb{I}\left[\abs[0]{\mathbf{w}_r^{\top} \mathbf{h}^{(L-1)} + \mathbf{a}_r^{\top} \mathbf{x}^{(L-1)}} \le \mathcal{O}\left(\rho^{5} N^{2 / 3} m^{-2 / 3}\right) \right] \\&
				\le \mathcal{O}(\rho^{5} N^{5 / 3} m^{-1/6}).
			\end{align*}  
			\endgroup
			Thus, we have in eq.~\ref{eq:rerandRNN_2},
			\begingroup \allowdisplaybreaks
			\begin{align}
				&\Big| \frac{\dout }{m} \sum_{i=1}^{L}  \sum_{s' \in [\dout ]} \sum_{r' \in [p]} \sum_{r \in \mathcal{K}}  b_{r, s'} b_{r', s'}^{\dagger} \mathbf{Back}_{i \to L, r, s}  H_{r', s'}\Big(\widetilde{\theta}_{r', s'} \langle \mathbf{w}_{r}, \overline{\widetilde{\mathbf{W}}}^{[L]} \mathbf{w}_{r', s'}^{\dagger}\rangle , \sqrt{m/2} a_{r, d}\Big) \mathbb{I}_{\mathbf{w}_r^{\top} \widetilde{\mathbf{h}}^{(i-1)} + \mathbf{a}_r^{\top} \mathbf{x}^{(i)} \ge 0} \nonumber\\&
				\quad \quad -  \frac{\dout }{m} \sum_{i=1}^{L}  \sum_{s' \in [\dout ]} \sum_{r' \in [p]} \sum_{r \in \mathcal{K}}  b_{r, s'} b_{r', s'}^{\dagger} \mathbf{Back}_{i \to L, r, s}  H_{r', s'}\Big(\widetilde{\theta}_{r', s'} \langle \mathbf{w}_{r}, \overline{\widetilde{\mathbf{W}}}^{[L]} \mathbf{w}_{r', s'}^{\dagger}\rangle , \sqrt{m/2} a_{r, d}\Big) \mathbb{I}_{\mathbf{w}_r^{\top} \mathbf{h}^{(i-1)} + \mathbf{a}_r^{\top} \mathbf{x}^{(i)} \ge 0} \Big| \nonumber\\&
				\le \max_{r, s} \abs{b_{r, s}} \cdot \max_{r', s'} \abs{b^{\dagger}_{r',s'}} \cdot \max_{i,r,s} \abs{\mathbf{Back}_{i \to L, r, s}} \cdot \max_{r', s', r}\abs{H_{r', s'}\Big(\widetilde{\theta}_{r', s'} \langle \mathbf{w}_{r}, \overline{\widetilde{\mathbf{W}}}^{[L]} \mathbf{w}_{r', s'}^{\dagger}\rangle , \sqrt{m/2} a_{r, d}\Big)} \nonumber\\& \quad\quad\quad\quad
				\cdot \frac{\dout }{m} \sum_{i=1}^{L} \sum_{s' \in [\dout ]} \sum_{r' \in [p]} \sum_{r \in \mathcal{K}} \abs{\mathbb{I}_{\mathbf{w}_r^{\top} \mathbf{h}^{(i-1)} + \mathbf{a}_r^{\top} \mathbf{x}^{(i)} \ge 0} - \mathbb{I}_{\mathbf{w}_r^{\top} \widetilde{\mathbf{h}}^{(i-1)} + \mathbf{a}_r^{\top} \mathbf{x}^{(i)} \ge 0}} \nonumber\\&
				\le \frac{\rho}{\sqrt{\dout }} \cdot 1 \cdot \mathcal{O}(\frac{\rho}{\sqrt{\dout }}) \cdot \mathfrak{C}_{\varepsilon}(\Phi, \mathcal{O}(\varepsilon_x^{-1})) \cdot \frac{\dout ^2Lp }{m} \cdot \mathcal{O}(\rho^4 N^{5/3} m^{1/6}) \le \mathcal{O}(\dout Lp\rho^6 N^{5/3} m^{-7/6} \mathfrak{C}_{\varepsilon}(\Phi, \mathcal{O}(\varepsilon_x^{-1}))) \label{eq:rerandRNN_2SOLVE}. 
			\end{align}
			\endgroup
			
			Now, we focus on eq.~\ref{eq:rerandRNN_3}. We have
			\begingroup \allowdisplaybreaks
			\begin{align}
				&\Big| \frac{\dout }{m} \sum_{i=1}^{L}  \sum_{s' \in [\dout ]} \sum_{r' \in [p]} \sum_{r \in \mathcal{K}}  b_{r, s'} b_{r', s'}^{\dagger} \mathbf{Back}_{i \to L, r, s}  H_{r', s'}\Big(\widetilde{\theta}_{r', s'} \langle \mathbf{w}_{r}, \overline{\widetilde{\mathbf{W}}}^{[L]} \mathbf{w}_{r', s'}^{\dagger}\rangle , \sqrt{m/2} a_{r, d}\Big) \mathbb{I}_{\mathbf{w}_r^{\top} \mathbf{h}^{(i-1)} + \mathbf{a}_r^{\top} \mathbf{x}^{(i)} \ge 0} \nonumber\\&
				\quad \quad -  \frac{\dout }{m} \sum_{i=1}^{L}  \sum_{s' \in [\dout ]} \sum_{r' \in [p]} \sum_{r \in \mathcal{K}}  b_{r, s'} b_{r', s'}^{\dagger} \mathbf{Back}_{i \to L, r, s}  H_{r', s'}\Big(\theta_{r', s'} \langle \mathbf{w}_{r}, \overline{\mathbf{W}}^{[L]} \mathbf{w}_{r', s'}^{\dagger}\rangle , \sqrt{m/2} a_{r, d}\Big) \mathbb{I}_{\mathbf{w}_r^{\top} \mathbf{h}^{(i-1)} + \mathbf{a}_r^{\top} \mathbf{x}^{(i)} \ge 0} \Big| \nonumber\\&
				\le \max_{r,s'} \abs{b_{r, s'}} \cdot \max_{r',s'} \abs{b_{r', s'}^{\dagger}} \cdot \max_{i, r, s} \abs{\mathbf{Back}_{i \to L, r, s}} \cdot \max_{i, r} \abs{\mathbb{I}_{\mathbf{w}_r^{\top} \mathbf{h}^{(i-1)} + \mathbf{a}_r^{\top} \mathbf{x}^{(i)} \ge 0}} \nonumber\\&
				\cdot \frac{\dout }{m} \sum_{i=1}^{L}  \sum_{s' \in [\dout ]} \sum_{r' \in [p]} \sum_{r \in \mathcal{K}} \abs{H_{r', s'}\Big(\theta_{r', s'} \langle \mathbf{w}_{r}, \overline{\mathbf{W}}^{[L]} \mathbf{w}_{r', s'}^{\dagger}\rangle , \sqrt{m/2} a_{r, d}\Big) - H_{r', s'}\Big(\widetilde{\theta}_{r', s'} \langle \mathbf{w}_{r}, \overline{\widetilde{\mathbf{W}}}^{[L]} \mathbf{w}_{r', s'}^{\dagger}\rangle , \sqrt{m/2} a_{r, d}\Big)} \nonumber\\&
				\le \frac{\rho}{\sqrt{\dout }} \cdot 1 \cdot \mathcal{O}(\frac{\rho}{\sqrt{\dout }}) \cdot 1 \\& \cdot \frac{\dout }{m} \sum_{i=1}^{L}  \sum_{s' \in [\dout ]} \sum_{r' \in [p]} \sum_{r \in \mathcal{K}} \cdot \mathfrak{C}_{\varepsilon}(\Phi_{r',s'}, \mathcal{O}(\varepsilon_x^{-1}))  \cdot \abs{\theta_{r', s'} \langle \mathbf{w}_{r}, \overline{\mathbf{W}}^{[L]} \mathbf{w}_{r', s'}^{\dagger}\rangle -  \widetilde{\theta}_{r', s'} \langle \mathbf{w}_{r}, \overline{\widetilde{\mathbf{W}}}^{[L]} \mathbf{w}_{r', s'}^{\dagger}\rangle} \label{eq:rerandRNN_3_preprefinal}\\&
				\le \frac{\rho}{\sqrt{\dout }} \cdot 1 \cdot \mathcal{O}(\frac{\rho}{\sqrt{\dout }}) \cdot 1 \cdot \frac{\dout ^2pL N}{m} \cdot \mathfrak{C}_{\varepsilon}(\Phi, \mathcal{O}(\varepsilon_x^{-1})) \cdot \mathcal{O}(\rho^6 N^{2/3} m^{-2/3} ) \label{eq:rerandRNN_3_prefinal} \\& 
				\le \mathcal{O}(\dout Lp\rho^8  \mathfrak{C}_{\varepsilon}(\Phi, \mathcal{O}(\varepsilon_x^{-1})) N^{5/3} m^{-5/3}) \nonumber,
			\end{align}
			\endgroup
			where we get eq.~\ref{eq:rerandRNN_3_preprefinal} by using the lipschitz continuity of the function $H_{r', s'}$ from def.~\ref{Def:Function_approx}. We get eq.~\ref{eq:rerandRNN_3_prefinal} by bounding the following term:
			\begin{align*}
				&\abs{\theta_{r', s'} \langle \mathbf{w}_{r}, \overline{\mathbf{W}}^{[L]} \mathbf{w}_{r', s'}^{\dagger}\rangle -  \widetilde{\theta}_{r', s'} \langle \mathbf{w}_{r}, \overline{\widetilde{\mathbf{W}}}^{[L]} \mathbf{w}_{r', s'}^{\dagger}\rangle} \\&
				\le \abs{\theta_{r', s'} \left(\langle \mathbf{w}_{r}, \overline{\mathbf{W}}^{[L]} \mathbf{w}_{r', s'}^{\dagger}\rangle -  \langle \mathbf{w}_{r}, \overline{\widetilde{\mathbf{W}}}^{[L]} \mathbf{w}_{r', s'}^{\dagger}\rangle\right)} + \abs{\left(\theta_{r', s'} - \widetilde{\theta}_{r', s'} \right) \langle \mathbf{w}_{r}, \overline{\widetilde{\mathbf{W}}}^{[L]} \mathbf{w}_{r', s'}^{\dagger}\rangle} \\&
				\le \abs{\theta_{r', s'}} \cdot \abs{\left\langle \mathbf{w}_{r}, \left(\overline{\mathbf{W}}^{[L]} -  \overline{\widetilde{\mathbf{W}}}^{[L]} \right) \mathbf{w}_{r', s'}^{\dagger}\right\rangle} + \abs{\theta_{r', s'} - \widetilde{\theta}_{r', s'} } \cdot \abs{ \langle \mathbf{w}_{r}, \overline{\widetilde{\mathbf{W}}}^{[L]} \mathbf{w}_{r', s'}^{\dagger}\rangle} \\&
				\le \mathcal{O}(\rho^6 (N/m)^{2/3}),
			\end{align*}
			where we use the following bounds that are true for all $r \in \mathcal{K}, r' \in [p], s' \in [\dout ]$ w.p. at least $1-e^{-\Omega(\rho^2)}$:
			\begin{itemize}
				\item Eq.~\ref{eq:abstheta_lowerbound} gives an upper bound of $O(1)$ on $|\theta_{r',s'}|$.
				\item Eq.~\ref{eq:tmp_W_MATHCALwl} can be easily modified to get a similar upper bound on $|\langle \mathbf{w}_{r}, (\overline{\mathbf{W}}^{[L]} -  \overline{\widetilde{\mathbf{W}}}^{[L]} ) \mathbf{w}_{r', s'}^{\dagger}\rangle |$.
				\item Cor.~\ref{cor:tildetheta_diff_theta} gives an upper bound on $|\theta_{r',s'} - \widetilde{\theta}_{r',s'}|$.
				\item Since $\norm[2]{\overline{\widetilde{\mathbf{W}}}^{[L]} \mathbf{w}_{r', s'}^{\dagger}} := \sqrt{m/2} \widetilde{\theta}_{r', s'}^{-1}$, we can use Eq.~\ref{eq:abstildetheta_lowerbound} to give an upper bound on the norm. Then, we can use Fact~\ref{fact:max_gauss} to bound $\max_{r \in \mathcal{K}} |\langle \mathbf{w}_r, \overline{\widetilde{\mathbf{W}}}^{[L]} \mathbf{w}_{r', s'}^{\dagger}\rangle| = \frac{\rho}{\sqrt{m}} \cdot \norm[2]{\overline{\widetilde{\mathbf{W}}}^{[L]} \mathbf{w}_{r', s'}^{\dagger}}$.
			\end{itemize}
		\end{proof}
				\begin{claim}[Restating claim~\ref{claim:difftildefphi}]\label{claim:difftildefphi_proof}
					With probability at least $1-e^{-\Omega(\rho^2)}$,
					\begin{align*}
						\Big|&\widetilde{F}_{s}^{(L), \mathcal{K}}(\widetilde{\mathbf{h}}^{(L-1)}, \mathbf{x}^{(L)}) - \frac{\dout }{m} \sum_{i=1}^{L}  \sum_{s' \in [\dout ]} \sum_{r' \in [p]} \sum_{r \in \mathcal{K}}  b_{r, s'} b_{r', s'}^{\dagger} \widetilde{\mathbf{Back}}_{i \to L, r, s} \Phi_{r', s} \left(\left\langle \mathbf{w}_{r', s}^{\dagger}, [\overline{\mathbf{x}}^{(1)}, \cdots, \overline{\mathbf{x}}^{(L)}]\right\rangle\right)\Big| \\&\le \frac{\dout }{m} \cdot \mathcal{O}(\mathfrak{C}_{\varepsilon}(\Phi_{r' s}, \mathcal{O}(\varepsilon_x^{-1})) \rho^2 \sqrt{\dout LpN}) \\& + \frac{\dout LpN}{m} \rho^2 (\varepsilon + \mathcal{O}( L_{\Phi}\rho^5 (N/m)^{1/6}) + \mathcal{O}(L_{\Phi}\varepsilon_x^{-1}  L^4 \rho^{11} m^{-1/12} +  L_{\Phi} \rho^2 L^{11/6} \varepsilon_x^{2/3})),
					\end{align*}
					for any $\varepsilon \in (0, \min_{r, s} \frac{\sqrt{3}}{C_s(\Phi_{r, s}, \varepsilon_x^{-1})})$.
				\end{claim}
				
				\begin{proof}
					We will take the expectation w.r.t. the weights $\left\{ \mathbf{w}_r, \mathbf{a}_r \right\}_{r \in \mathcal{K}}$. The difference between $\widetilde{F}$ and the expected value is given by
					\begingroup \allowdisplaybreaks
					\begin{align}
						&\abs{\widetilde{F}_{s}^{(L), \mathcal{K}}(\widetilde{\mathbf{h}}^{(L-1)}, \mathbf{x}^{(L)}) - \mathbb{E}_{\left\{ \mathbf{w}_r, \mathbf{a}_r \right\}_{r \in \mathcal{K}}} \widetilde{F}_{s}^{(L), \mathcal{K}}(\widetilde{\mathbf{h}}^{(L-1)}, \mathbf{x}^{(L)})} \nonumber\\
						&=\Big|\frac{\dout }{m} \sum_{i=1}^{L}  \sum_{s' \in [\dout ]} \sum_{r' \in [p]} \sum_{r \in \mathcal{K}}  b_{r, s'} b_{r', s'}^{\dagger} \widetilde{\mathbf{Back}}_{i \to L, r, s} H_{r', s}\Big(\widetilde{\theta}_{r', s} \langle \mathbf{w}_{r}, \overline{\widetilde{\mathbf{W}}}^{[L]} \mathbf{w}_{r', s}^{\dagger}\rangle, \sqrt{m/2} a_{r, d}\Big) \mathbb{I}_{\mathbf{w}_r^{\top} \widetilde{\mathbf{h}}^{(L-1)} + \mathbf{a}_r^{\top} \mathbf{x}^{(L)} \ge 0} \nonumber\\&
						\quad\quad\quad\quad -  \mathbb{E}_{[\mathbf{w}, \mathbf{a}] \sim \mathcal{N}(0, \frac{2}{m}\mathbf{I})} \frac{\dout }{m} \sum_{i=1}^{L}  \sum_{s' \in [\dout ]} \sum_{r' \in [p]} \sum_{r \in \mathcal{K}}  b_{r, s'} b_{r', s'}^{\dagger} \widetilde{\mathbf{Back}}_{i \to L, r, s} \nonumber\\& \quad\quad\quad\quad \quad\quad\quad\quad \cdot H_{r', s}\Big(\widetilde{\theta}_{r', s} \langle \mathbf{w}, \overline{\widetilde{\mathbf{W}}}^{[L]} \mathbf{w}_{r', s}^{\dagger}\rangle, \sqrt{m/2} a_{d}\Big) \mathbb{I}_{\mathbf{w}^{\top} \widetilde{\mathbf{h}}^{(L-1)} + \mathbf{a}^{\top} \mathbf{x}^{(L)} \ge 0}\Big| \nonumber\\&
						=\Big|\frac{\dout }{m} \sum_{i=1}^{L}  \sum_{s' \in [\dout ]} \sum_{r' \in [p]} \sum_{r \in \mathcal{K}}  b_{r, s'} b_{r', s'}^{\dagger} \widetilde{\mathbf{Back}}_{i \to L, r, s} H_{r', s}\Big(\widetilde{\theta}_{r', s} \langle \mathbf{w}_{r}, \overline{\widetilde{\mathbf{W}}}^{[L]} \mathbf{w}_{r', s}^{\dagger}\rangle, \sqrt{m/2} a_{r, d}\Big) \mathbb{I}_{\mathbf{w}_r^{\top} \widetilde{\mathbf{h}}^{(L-1)} + \mathbf{a}_r^{\top} \mathbf{x}^{(L)} \ge 0} \nonumber\\&
						\quad\quad\quad\quad -  \frac{\dout }{m} \sum_{i=1}^{L}  \sum_{s' \in [\dout ]} \sum_{r' \in [p]} \sum_{r \in \mathcal{K}}  b_{r, s'} b_{r', s'}^{\dagger} \widetilde{\mathbf{Back}}_{i \to L, r, s} \nonumber\\& \quad\quad\quad\quad \quad\quad\quad\quad \cdot  \mathbb{E}_{\mathbf{w}, \mathbf{a} \sim \mathcal{N}(0, \frac{2}{m}\mathbf{I})} H_{r', s}\Big(\widetilde{\theta}_{r', s} \langle \mathbf{w}, \overline{\widetilde{\mathbf{W}}}^{[L]} \mathbf{w}_{r', s}^{\dagger}\rangle, \sqrt{m/2} a_{d}\Big) \mathbb{I}_{\mathbf{w}^{\top} \widetilde{\mathbf{h}}^{(L-1)} + \mathbf{a}^{\top} \mathbf{x}^{(L)} \ge 0}\Big|, \label{Eqn:conc_H}
					\end{align}
					\endgroup
					where in the final step, we have used the fact that $\widetilde{\mathbf{Back}}$ and $\mathbf{B}$ are independent of  the variables $\left\{ \mathbf{w}_r, \mathbf{a}_r \right\}_{r \in \mathcal{K}}$ w.r.t. which we are taking the expectation.
					
					Note that, the random variable under consideration is a bounded random variable, because: using the fact that $b_{r, s'} \sim \mathcal{N}(0, 1)$, it is bounded by $\rho$ with high probability, $\widetilde{Back}_{i \to L, r, s} = \mathbf{b}_s^{\top} (\prod_{i \le \ell \le L} \mathbf{D}^{(\ell)} \mathbf{W}) \mathbf{e}_r$ is bounded by $\mathcal{O}(\rho)$ using bound on norm of $\mathbf{b}_s$ and Claim~\ref{lemma:norm_ESN}, and the function $H$ is bounded by def.~\ref{Def:Function_approx}.
					Denoting the inequality in eq.~\ref{Eqn:conc_H} as 
					$P(\left\{ \mathbf{w}_r, \mathbf{a}_r \right\}_{r \in \mathcal{K}})$, we get using hoeffding's inequality for bounded variables (fact~\ref{fact:hoeffding})
					\begin{equation*}
						\Pr \left[ P(\left\{ \mathbf{w}_r, \mathbf{a}_r \right\}_{r \in \mathcal{K}}) > \frac{\dout }{m} \cdot  \mathcal{O}(\mathfrak{C}_{\varepsilon}(\Phi_{r' s}, \mathcal{O}(\varepsilon_x^{-1})) \rho^2 \sqrt{\dout LpN}) \right] \le e^{-\rho^2/8}.
					\end{equation*}
					
					Now, we focus on the expected value in eq.~\ref{Eqn:conc_H}.
					For typographical simplicity in the next few steps, we denote the vector $\mathbf{v} =  \widetilde{\mathbf{h}}^{(L-1)}$, vector $\mathbf{q} = \sqrt{m/2} \cdot \mathbf{w}$ and vector $\mathbf{t}=\widetilde{\theta}_{r', s} \cdot \sqrt{2/m} \cdot \overline{\widetilde{\mathbf{W}}}^{[L]} \mathbf{w}_{r', s}^{\dagger}$. Also, let $\mathbf{t}^{\perp}$ denote a vector in the subspace orthogonal to $\mathbf{t}$ that is closest to the vector $\mathbf{v}$.
					
					By the definition of the function $H_{r' ,s}$ from 
					Def.~\ref{Def:Function_approx}, where we use $k_{0, r, s} = \varepsilon_x^{-1} \sqrt{m/2} \theta_{r', s}^{-1}$  for each $r' \in [p], s \in [\dout ]$ in def.~\ref{Def:Function_approx} ($\theta_{r' ,s}$ is defined in def.~\ref{def:existence}),
					we have
					\begingroup
					\allowdisplaybreaks
					\begin{align*}
						&\mathbb{E}_{\mathbf{w}, \mathbf{a} \sim \mathcal{N}(0, \frac{2}{m}\mathbf{I})}  H_{r', s}\Big(\widetilde{\theta}_{r', s} \langle \mathbf{w}, \overline{\widetilde{\mathbf{W}}}^{[L]} \mathbf{w}_{r', s}^{\dagger}\rangle , \sqrt{m/2} a_{d}\Big)  \cdot\mathbb{I} \left[ \langle \mathbf{w},  \widetilde{\mathbf{h}}^{(L-1)}\rangle + a_{d}  \ge 0 \right] \\&
						= \mathbb{E}_{\mathbf{w}, \mathbf{a} \sim \mathcal{N}(0, \frac{2}{m}\mathbf{I})}  H_{r', s}\Big(\widetilde{\theta}_{r', s} \langle \mathbf{w}, \overline{\widetilde{\mathbf{W}}}^{[L]} \mathbf{w}_{r', s}^{\dagger}\rangle , \sqrt{m/2} a_{d}\Big) \cdot\mathbb{I} \left[ \sqrt{m/2}\langle \mathbf{w},  \widetilde{\mathbf{h}}^{(L-1)}\rangle + \sqrt{m/2} a_{d}  \ge 0 \right] \\&
						= \mathbb{E}_{\mathbf{w}, \mathbf{a} \sim \mathcal{N}(0, \frac{2}{m}\mathbf{I})}  H_{r', s}\Big(  \langle \mathbf{q}, \mathbf{t}  \rangle, \sqrt{m/2} a_{d}\Big) \cdot \mathbb{I} \left[\langle \mathbf{v}, \mathbf{q} \rangle + \sqrt{m/2} a_{d}  \ge 0 \right] \\&
						= \mathbb{E}_{\mathbf{w}, \mathbf{a} \sim \mathcal{N}(0, \frac{2}{m}\mathbf{I})}  H_{r', s}\Big(  \langle \mathbf{q}, \mathbf{t}  \rangle, \sqrt{m/2} a_{d}\Big) \cdot  \mathbb{I} \left[\langle \mathbf{v}, \mathbf{t} \rangle \langle \mathbf{q}, \mathbf{t} \rangle +  \sqrt{\norm{\mathbf{v}}^2 - \langle \mathbf{v}, \mathbf{t} \rangle^2}  \langle \mathbf{q}, \mathbf{t}^{\perp} \rangle+ \sqrt{m/2} a_{d}  \ge 0 \right] \\&
						= \Phi_{r', s} \left(\varepsilon_x^{-1} (\sqrt{m/2}\theta_{r', s}^{-1}) \langle\mathbf{t}, \mathbf{v}\rangle\right) \pm \varepsilon 
						\\&
						= \Phi_{r', s} \left( \varepsilon_x^{-1} \theta_{r', s}^{-1} \widetilde{\theta}_{r', s} \langle \overline{\widetilde{\mathbf{W}}}^{[L]} \mathbf{w}_{r', s}^{\dagger}, \mathbf{\widetilde{h}}^{(L-1)} \rangle  \right) \pm \varepsilon \\&
						= \Phi_{r', s} \left( \varepsilon_x^{-1}  \langle \overline{\widetilde{\mathbf{W}}}^{[L]} \mathbf{w}_{r', s}^{\dagger}, \mathbf{\widetilde{h}}^{(L-1)}\rangle  \pm \mathcal{O}(\rho^5 (N/m)^{1/6})  \right) \pm \varepsilon \\&
						= \Phi_{r', s} \left( \varepsilon_x^{-1}  \langle \overline{\widetilde{\mathbf{W}}}^{[L]} \mathbf{w}_{r', s}^{\dagger}, \mathbf{\widetilde{h}}^{(L-1)}  \rangle  \right) \pm \varepsilon \pm \mathcal{O}(L_{\Phi_{r',s}} \rho^5 (N/m)^{1/6}),
					\end{align*}
					\endgroup
					where in the pre-final step, we have used claim~\ref{claim:invtildethetatheta} to bound the value of $\theta_{r', s}^{-1} \widetilde{\theta}_{r', s}$ and in the final step, we have used the lipschitz constant of $\Phi_{r', s}$ in the desired range. Corollary~\ref{cor:Invertibility_ESN} shows that with probability at least $1 - e^{-\Omega(\rho^2)}$ w.r.t. the weights $\widetilde{\mathbf{W}}$ and $\widetilde{\mathbf{A}}$, 
					\begin{align*}
						&\abs{\overline{\widetilde{\mathbf{W}}}^{[L]\top} \widetilde{h}^{(L-1)} - \varepsilon_x[\overline{\mathbf{x}}^{(2)}, 
							\cdots, \overline{\mathbf{x}}^{(L-1)}]}  \\&\le \mathcal{O}\left(L^4 \cdot (\rho^{11} m^{-1/12} + \rho^{7} m^{-1/12} + \rho^{7} m^{-1/4} + \rho^{11} m^{-1/4})  \right) + \mathcal{O}(\rho^2 L^{11/6} \varepsilon_x^{2/3}) \\&
						\le  \mathcal{O}(L^4 \rho^{11} m^{-1/12} + \rho^2 L^{11/6} \varepsilon_x^{5/3}). 
					\end{align*}
					Thus,
					\begingroup \allowdisplaybreaks
					\begin{align*}
						&\mathbb{E}_{\mathbf{w}, \mathbf{a} \sim \mathcal{N}(0, \frac{2}{m}\mathbf{I})}  H_{r', s}\Big(\widetilde{\theta}_{r', s} \langle \mathbf{w}, \overline{\widetilde{\mathbf{W}}}^{[L]} \mathbf{w}_{r', s}^{\dagger}\rangle , \sqrt{m/2} a_{d}\Big)  \cdot\mathbb{I} \left[ \langle \mathbf{w},  \widetilde{\mathbf{h}}^{(L-1)}\rangle + a_{d}  \ge 0 \right] \\&
						= \Phi_{r', s} \left( \varepsilon_x^{-1} \langle \overline{\widetilde{\mathbf{W}}}^{[L]} \mathbf{w}_{r', s}^{\dagger}, \mathbf{\widetilde{h}}^{(L-1)} \rangle  \right) \pm \varepsilon \pm \mathcal{O}(L_{\Phi_{r', s}} \rho^5 (N/m)^{1/6})
						\\&
						= \Phi_{r', s} \left( \varepsilon_x^{-1} \langle     \mathbf{w}_{r', s}^{\dagger}, \overline{\widetilde{\mathbf{W}}}^{[L]\top} \mathbf{\widetilde{h}}^{(L-1)} \rangle \right) \pm \varepsilon \pm \mathcal{O}(L_{\Phi_{r', s}} \rho^5 (N/m)^{1/6}) \\&
						= \Phi_{r', s} \left(\varepsilon_x^{-1} \langle     \mathbf{w}_{r', s}^{\dagger}, \varepsilon_x[\overline{\mathbf{x}}^{(2)}, \cdots, \overline{\mathbf{x}}^{(L-1)}] \rangle  \pm \varepsilon' \right) \pm \varepsilon \pm \mathcal{O}(L_{\Phi_{r', s}} \rho^5 (N/m)^{1/6})  \\&
						= \Phi_{r', s} \left( \langle     \mathbf{w}_{r', s}^{\dagger}, [\overline{\mathbf{x}}^{(2)}, \cdots, \overline{\mathbf{x}}^{(L-1)}] \rangle \right) \pm \varepsilon \pm \mathcal{O}(L_{\Phi_{r', s}} \rho^5 (N/m)^{1/6}) \pm L_{\Phi_{r', s}}\varepsilon',
					\end{align*}
					\endgroup
					where $\varepsilon \in (0, \min_{r, s} \frac{\sqrt{3}}{C_s(\Phi_{r, s}, \varepsilon_x^{-1})})$ and  $\varepsilon' =  \mathcal{O}(\varepsilon_x^{-1} L^4 \rho^{11} m^{-1/12} + \rho^2 L^{11/6} \varepsilon_x^{2/3}).$ Thus, we have
					\begingroup \allowdisplaybreaks
					\begin{align}
						&\Big|\frac{\dout }{m} \sum_{i=1}^{L}  \sum_{s' \in [\dout ]} \sum_{r' \in [p]} \sum_{r \in \mathcal{K}}  b_{r, s'} b_{r', s'}^{\dagger} \widetilde{\mathbf{Back}}_{i \to L, r, s} \nonumber\\& \quad\quad\quad\quad \quad\quad\quad\quad \cdot  \mathbb{E}_{\mathbf{w}, \mathbf{a} \sim \mathcal{N}(0, \frac{2}{m}\mathbf{I})} H_{r', s}\Big(\widetilde{\theta}_{r', s} \langle \mathbf{w}, \overline{\widetilde{\mathbf{W}}}^{[L]} \mathbf{w}_{r', s}^{\dagger}\rangle, \sqrt{m/2} a_{d}\Big) \mathbb{I}_{\mathbf{w}^{\top} \widetilde{\mathbf{h}}^{(L-1)} + \mathbf{a}^{\top} \mathbf{x}^{(L)} \ge 0} \nonumber\\& - 
						\frac{\dout }{m} \sum_{i=1}^{L}  \sum_{s' \in [\dout ]} \sum_{r' \in [p]} \sum_{r \in \mathcal{K}}  b_{r, s'} b_{r', s'}^{\dagger} \widetilde{\mathbf{Back}}_{i \to L, r, s} \Phi_{r', s} \left( \langle     \mathbf{w}_{r', s}^{\dagger}, [\overline{\mathbf{x}}^{(2)}, \cdots, \overline{\mathbf{x}}^{(L-1)}] \rangle \right) \Big| \\& \le \frac{\dout }{m} \sum_{i=1}^{L}  \sum_{s' \in [\dout ]} \sum_{r' \in [p]} \sum_{r \in \mathcal{K}}  b_{r, s'} b_{r', s'}^{\dagger} \widetilde{\mathbf{Back}}_{i \to L, r, s} \cdot (\varepsilon + \mathcal{O}(L_{\Phi_{r', s}} \rho^5 (N/m)^{1/6}) + L_{\Phi_{r', s}}\varepsilon') \nonumber\\&
						\le \frac{\dout }{m} \sum_{i=1}^{L}  \sum_{s' \in [\dout ]} \sum_{r' \in [p]} \sum_{r \in \mathcal{K}}  \abs{b_{r, s'}} \abs{b_{r', s'}^{\dagger}} \abs{\widetilde{\mathbf{Back}}_{i \to L, r, s}} \cdot (\varepsilon + \max_{r', s} L_{\Phi_{r', s}}\varepsilon' + \mathcal{O}(L_{\Phi_{r', s}} \rho^5 (N/m)^{1/6})) \nonumber\\&
						\le \frac{\dout LpN}{m} \rho^2 (\varepsilon + \mathcal{O}(L_{\Phi} \rho^5 (N/m)^{1/6}) + L_{\Phi}\varepsilon') \label{eq:expectedtheta},
					\end{align}
					\endgroup
					where $\varepsilon \in (0, \min_{r, s} \frac{\sqrt{3}}{C_s(\Phi_{r, s}, \varepsilon_x^{-1})})$ and  $\varepsilon' =  \mathcal{O}(\varepsilon_x^{-1} L^4 \rho^{11} m^{-1/12} + \rho^2 L^{11/6} \varepsilon_x^{2/3}).$ 
					Hence, using eq.~\ref{eq:expectedtheta} and eq.~\ref{Eqn:conc_H}, we have w.p. at least $1 - e^{-\Omega(\rho^2)}$,
					\begingroup \allowdisplaybreaks
					\begin{align*}
						\Big|&\widetilde{F}_{s}^{(L), \mathcal{K}}(\widetilde{\mathbf{h}}^{(L-1)}, \mathbf{x}^{(L)}) - \frac{\dout }{m} \sum_{i=1}^{L}  \sum_{s' \in [\dout ]} \sum_{r' \in [p]} \sum_{r \in \mathcal{K}}  b_{r, s'} b_{r', s'}^{\dagger} \widetilde{\mathbf{Back}}_{i \to L, r, s} \Phi_{r', s} \left(\left\langle \mathbf{w}_{r', s}^{\dagger}, [\overline{\mathbf{x}}^{(1)}, \cdots, \overline{\mathbf{x}}^{(L)}]\right\rangle\right)\Big| \\&
						\le \abs{\widetilde{F}_{s}^{(L), \mathcal{K}}(\widetilde{\mathbf{h}}^{(L-1)}, \mathbf{x}^{(L)}) - \mathbb{E}_{\left\{ \mathbf{w}_r, \mathbf{a}_r \right\}_{r \in \mathcal{K}}} \widetilde{F}_{s}^{(L), \mathcal{K}}(\widetilde{\mathbf{h}}^{(L-1)}, \mathbf{x}^{(L)})}  \\&
						+ \Big|\mathbb{E}_{\left\{ \mathbf{w}_r, \mathbf{a}_r \right\}_{r \in \mathcal{K}}} \widetilde{F}_{s}^{(L), \mathcal{K}}(\widetilde{\mathbf{h}}^{(L-1)}, \mathbf{x}^{(L)}) \\& \quad \quad \quad \quad - \frac{\dout }{m} \sum_{i=1}^{L}  \sum_{s' \in [\dout ]} \sum_{r' \in [p]} \sum_{r \in \mathcal{K}}  b_{r, s'} b_{r', s'}^{\dagger} \widetilde{\mathbf{Back}}_{i \to L, r, s} \Phi_{r', s} \left(\left\langle \mathbf{w}_{r', s}^{\dagger}, [\overline{\mathbf{x}}^{(1)}, \cdots, \overline{\mathbf{x}}^{(L)}]\right\rangle\right)\Big| \\&
						\le  \frac{\dout LpN}{m} \rho^2 (\varepsilon + \mathcal{O}(L_{\Phi} \rho^5 (N/m)^{1/6}) + L_{\Phi}\varepsilon') + \frac{\dout }{m} \cdot \mathcal{O}(\mathfrak{C}_{\varepsilon}(\Phi_{r' s}, \mathcal{O}(\varepsilon_x^{-1})) \rho^2 \sqrt{\dout LpN}),
					\end{align*}
					where $\varepsilon \in (0, \min_{r, s} \frac{\sqrt{3}}{C_s(\Phi_{r, s}, \varepsilon_x^{-1})})$ and  $\varepsilon' =  \mathcal{O}(\varepsilon_x^{-1} L^4 \rho^{11} m^{-1/12} + \rho^2 L^{11/6} \varepsilon_x^{2/3}).$ 
					\endgroup
					
				\end{proof}
				
				\begin{claim}[Restating claim~\ref{claim:fbacktildeback}]\label{claim:fbacktildeback_proof}
					With probability at least $1 - e^{-\Omega(\rho^2)}$,
					\begin{align*}
						&\Big| \frac{\dout }{m} \sum_{i=1}^{L}  \sum_{s' \in [\dout ]} \sum_{r' \in [p]} \sum_{r \in \mathcal{K}}  b_{r, s'} b_{r', s'}^{\dagger} \widetilde{\mathbf{Back}}_{i \to L, r, s} \Phi_{r', s} \left(\left\langle \mathbf{w}_{r', s}^{\dagger}, [\overline{\mathbf{x}}^{(1)}, \cdots, \overline{\mathbf{x}}^{(L)}]\right\rangle\right) \\& - \frac{\dout }{m} \sum_{i=1}^{L}  \sum_{s' \in [\dout ]} \sum_{r' \in [p]} \sum_{r \in \mathcal{K}}  b_{r, s'} b_{r', s'}^{\dagger} \mathbf{Back}_{i \to L, r, s} \Phi_{r', s} \left(\left\langle \mathbf{w}_{r', s}^{\dagger}, [\overline{\mathbf{x}}^{(1)}, \cdots, \overline{\mathbf{x}}^{(L)}]\right\rangle\right) \Big| \\& \le \mathcal{O}(\rho^8 C_{\Phi} \dout Lp N^{7/6} m^{-7/6}). 
					\end{align*}
				\end{claim}
				
				\begin{proof}
					From Lemma~\ref{lemma:rerandESN}, we have with probability at least $1 - e^{-\Omega(\rho^2)}$,
					\begin{align*}
						\abs{\mathbf{e}_s^{\top} \left(\mathbf{Back_{i \to j}} - \widetilde{\mathbf{Back}}_{i \to j} \right) \mathbf{e}_r} 
						&= \abs{\mathbf{b}_s^{\top} \left( \mathbf{D}^{(j)} \mathbf{W} \cdots \mathbf{D}^{(i)} \mathbf{W} - \widetilde{\mathbf{D}}^{(j)} \widetilde{\mathbf{W}} \cdots \widetilde{\mathbf{D}}^{(i)} \widetilde{\mathbf{W}} \right) \mathbf{e}_r}
						\\&
						\le \norm[2]{\mathbf{b}_s} \norm[2]{\left( \mathbf{D}^{(j)} \mathbf{W} \cdots \mathbf{D}^{(i)} \mathbf{W} - \widetilde{\mathbf{D}}^{(j)} \widetilde{\mathbf{W}} \cdots \widetilde{\mathbf{D}}^{(i)} \widetilde{\mathbf{W}} \right) \mathbf{e}_r}
						\\&\le \mathcal{O}(\dout ^{-1/2} \rho^7 N^{1/6} m^{-1/6}) , \text{ for all }  r \in [m], s \in [\dout ] \text{ and } 1 \le i \le j \le L.
					\end{align*}
					Hence,
					\begingroup \allowdisplaybreaks
					\begin{align*}
						&\Big| \frac{\dout }{m} \sum_{i=1}^{L}  \sum_{s' \in [\dout ]} \sum_{r' \in [p]} \sum_{r \in \mathcal{K}}  b_{r, s'} b_{r', s'}^{\dagger} \widetilde{\mathbf{Back}}_{i \to L, r, s} \Phi_{r', s} \left(\left\langle \mathbf{w}_{r', s}^{\dagger}, [\overline{\mathbf{x}}^{(1)}, \cdots, \overline{\mathbf{x}}^{(L)}]\right\rangle\right) \\& - \frac{\dout }{m} \sum_{i=1}^{L}  \sum_{s' \in [\dout ]} \sum_{r' \in [p]} \sum_{r \in \mathcal{K}}  b_{r, s'} b_{r', s'}^{\dagger} \mathbf{Back}_{i \to L, r, s} \Phi_{r', s} \left(\left\langle \mathbf{w}_{r', s}^{\dagger}, [\overline{\mathbf{x}}^{(1)}, \cdots, \overline{\mathbf{x}}^{(L)}]\right\rangle\right) \Big|
						\\&
						= \Big| \frac{\dout }{m} \sum_{i=1}^{L}  \sum_{s' \in [\dout ]} \sum_{r' \in [p]} \sum_{r \in \mathcal{K}}  b_{r, s'} b_{r', s'}^{\dagger} \mathbf{e}_r^{\top} \left( \widetilde{\mathbf{Back}}_{i \to L} \Phi_{r', s} - \mathbf{Back}_{i \to L}\right) \mathbf{e}_s \Phi_{r', s} \left(\left\langle \mathbf{w}_{r', s}^{\dagger}, [\overline{\mathbf{x}}^{(1)}, \cdots, \overline{\mathbf{x}}^{(L)}]\right\rangle\right) \Big| \\&
						\le \frac{\dout }{m} \sum_{i=1}^{L}  \sum_{s' \in [\dout ]} \sum_{r' \in [p]} \sum_{r \in \mathcal{K}}  \abs{b_{r, s'}} \cdot \abs{b_{r', s'}^{\dagger}} \cdot  \abs{\Phi_{r', s} \left(\left\langle \mathbf{w}_{r', s}^{\dagger}, [\overline{\mathbf{x}}^{(1)}, \cdots, \overline{\mathbf{x}}^{(L)}]\right\rangle\right)} \cdot \abs{ \mathbf{e}_r^{\top} \left( \widetilde{\mathbf{Back}}_{i \to L} \Phi_{r', s} - \mathbf{Back}_{i \to L}\right) \mathbf{e}_s } \\&
						\le \frac{\dout ^2LpN}{m} \cdot \frac{\rho}{\sqrt{\dout }} \cdot 1 \cdot C_{\Phi} \cdot \mathcal{O}(\dout ^{-1/2} \rho^7 N^{1/6} m^{-1/6}) \\&
						\le \mathcal{O}(\rho^8 C_{\Phi} \dout Lp N^{7/6} m^{-7/6}).
					\end{align*}
					\endgroup
					In the final step, we have used the bounds of different terms as follows.
					we will need a couple of bounds on the terms that appear in the equations.
					\begin{itemize}
						\item Using the fact~\ref{fact:max_gauss}, we can show that with probability $1-e^{-\Omega(\rho^2)}$,
						$\max_{r, s'} | b_{r, s'} | \le \frac{\rho}{\sqrt{\dout }}.$ 
						\item From the definition of concept class,  $\max_{r',s'} | b_{r', s'}^{\dagger} | \le 1$ and $\max_{r', s} \abs{\Phi_{r', s}} \le C_{\Phi}$ in the desired range.
					\end{itemize}
				\end{proof}


			\begin{claim}[Restating claim~\ref{claim:simplifybig}]\label{claim:simplifybig_proof}
				With probability exceeding $1 - e^{-\Omega(\rho^2)}$,
				\begin{align*}
					&\Big|  b_{r', s}^{\dagger}  \Phi_{r', s} \left(\left\langle \mathbf{w}_{r', s}^{\dagger}, [\overline{\mathbf{x}}^{(2)}, \cdots, \overline{\mathbf{x}}^{(L-1)}]\right\rangle\right) \\& - \frac{\dout }{m} \sum_{i=1}^{L}  \sum_{s' \in [\dout ]}  \sum_{r \in [m]}  b_{r, s'} b_{r', s'}^{\dagger} \mathbf{Back}_{i \to L, r, s} \Phi_{r', s} \left(\left\langle \mathbf{w}_{r', s}^{\dagger}, [\overline{\mathbf{x}}^{(2)}, \cdots, \overline{\mathbf{x}}^{(L-1)}]\right\rangle\right) \Big| \\&\le  \mathcal{O}(L\dout  \rho C_{\Phi} m^{-0.25}).
				\end{align*}
			\end{claim}
			
			\begin{proof}
				\begingroup \allowdisplaybreaks
				\begin{align*}
					&\Big| \Phi_{r', s} \left(\left\langle \mathbf{w}_{r', s}^{\dagger}, [\overline{\mathbf{x}}^{(2)}, \cdots, \overline{\mathbf{x}}^{(L-1)}]\right\rangle\right) \\& \quad \quad  - \frac{\dout }{m} \sum_{i=1}^{L}  \sum_{s' \in [\dout ]}  \sum_{r \in [m]}  b_{r, s'} b_{r', s'}^{\dagger} \mathbf{Back}_{i \to L, r, s} \Phi_{r', s} \left(\left\langle \mathbf{w}_{r', s}^{\dagger}, [\overline{\mathbf{x}}^{(2)}, \cdots, \overline{\mathbf{x}}^{(L-1)}]\right\rangle\right) \Big| \nonumber\\&
					\le 
					\Big| \Phi_{r', s} \left(\left\langle \mathbf{w}_{r', s}^{\dagger}, [\overline{\mathbf{x}}^{(2)}, \cdots, \overline{\mathbf{x}}^{(L-1)}]\right\rangle\right)  - \frac{\dout }{m} \sum_{r \in [m]}  b_{r, s} b_{r', s}^{\dagger} \mathbf{Back}_{L \to L, r, s} \Phi_{r', s} \left(\left\langle \mathbf{w}_{r', s}^{\dagger}, [\overline{\mathbf{x}}^{(2)}, \cdots, \overline{\mathbf{x}}^{(L-1)}]\right\rangle\right) \Big| \\&
					+ \Big| \ \frac{\dout }{m} \sum_{s'\in [\dout ]: s' \ne s} \sum_{r \in [m]}  b_{r, s'} b_{r', s'}^{\dagger} \mathbf{Back}_{L \to L, r, s} \Phi_{r', s} \left(\left\langle \mathbf{w}_{r', s}^{\dagger}, [\overline{\mathbf{x}}^{(2)}, \cdots, \overline{\mathbf{x}}^{(L-1)}]\right\rangle\right) \Big| \\&
					+ \Big| \ \frac{\dout }{m}  \sum_{i=1}^{L-1} \sum_{s'\in [\dout ]: s' \ne s} \sum_{r \in [m]}  b_{r, s'} b_{r', s'}^{\dagger} \mathbf{Back}_{i \to L, r, s} \Phi_{r', s} \left(\left\langle \mathbf{w}_{r', s}^{\dagger}, [\overline{\mathbf{x}}^{(2)}, \cdots, \overline{\mathbf{x}}^{(L-1)}]\right\rangle\right) \Big| \\&
					+ \Big|  \frac{\dout }{m} \sum_{i=1}^{L-1} \sum_{r \in [m]}  b_{r, s} b_{r', s}^{\dagger} \mathbf{Back}_{i \to L, r, s} \Phi_{r', s} \left(\left\langle \mathbf{w}_{r', s}^{\dagger}, [\overline{\mathbf{x}}^{(2)}, \cdots, \overline{\mathbf{x}}^{(L-1)}]\right\rangle\right) \Big|.
				\end{align*}
				\endgroup
				
				Since, $\mathbf{B} = \mathbf{Back}_{L \to L}$ by definition, we can simplify the above 4 terms as
				\begingroup \allowdisplaybreaks
				\begin{align}
					&\Big| \Phi_{r', s} \left(\left\langle \mathbf{w}_{r', s}^{\dagger}, [\overline{\mathbf{x}}^{(2)}, \cdots, \overline{\mathbf{x}}^{(L-1)}]\right\rangle\right) \\& - \frac{\dout }{m} \sum_{i=1}^{L}  \sum_{s' \in [\dout ]}  \sum_{r \in [m]}  b_{r, s'} b_{r', s'}^{\dagger} \mathbf{Back}_{i \to L, r, s} \Phi_{r', s} \left(\left\langle \mathbf{w}_{r', s}^{\dagger}, [\overline{\mathbf{x}}^{(2)}, \cdots, \overline{\mathbf{x}}^{(L-1)}]\right\rangle\right) \Big| \nonumber\\& 
					\le 
					\Big| \Phi_{r', s} \left(\left\langle \mathbf{w}_{r', s}^{\dagger}, [\overline{\mathbf{x}}^{(2)}, \cdots, \overline{\mathbf{x}}^{(L-1)}]\right\rangle\right)  - \frac{\dout }{m} \sum_{r \in [m]}  b^2_{r, s} b_{r', s}^{\dagger} \Phi_{r', s} \left(\left\langle \mathbf{w}_{r', s}^{\dagger}, [\overline{\mathbf{x}}^{(2)}, \cdots, \overline{\mathbf{x}}^{(L-1)}]\right\rangle\right) \Big| \nonumber\\&
					+ \Big| \ \frac{\dout }{m} \sum_{s'\in [\dout ]: s' \ne s} \sum_{r \in [m]}  b_{r, s'} b_{r', s'}^{\dagger} b_{r, s} \Phi_{r', s} \left(\left\langle \mathbf{w}_{r', s}^{\dagger}, [\overline{\mathbf{x}}^{(2)}, \cdots, \overline{\mathbf{x}}^{(L-1)}]\right\rangle\right) \Big| \nonumber\\&
					+ \Big| \ \frac{\dout }{m}  \sum_{i=1}^{L} \sum_{s'\in [\dout ]: s' \ne s} \sum_{r \in [m]}  \mathbf{Back}_{L \to L, r, s'} b_{r', s'}^{\dagger} \mathbf{Back}_{i \to L, r, s} \Phi_{r', s} \left(\left\langle \mathbf{w}_{r', s}^{\dagger}, [\overline{\mathbf{x}}^{(2)}, \cdots, \overline{\mathbf{x}}^{(L-1)}]\right\rangle\right) \Big| \nonumber\\&
					+ \Big|  \frac{\dout }{m} \sum_{i=1}^{L-1} \sum_{r \in [m]}  \mathbf{Back}_{L \to L, r, s}  b_{r', s}^{\dagger} \mathbf{Back}_{i \to L, r, s} \Phi_{r', s} \left(\left\langle \mathbf{w}_{r', s}^{\dagger}, [\overline{\mathbf{x}}^{(2)}, \cdots, \overline{\mathbf{x}}^{(L-1)}]\right\rangle\right) \Big| \nonumber\\&
					= \Big| \Phi_{r', s} \left(\left\langle \mathbf{w}_{r', s}^{\dagger}, [\overline{\mathbf{x}}^{(2)}, \cdots, \overline{\mathbf{x}}^{(L-1)}]\right\rangle\right)  - \frac{\dout }{m} \sum_{r \in [m]}  b^2_{r, s} b_{r', s}^{\dagger} \Phi_{r', s} \left(\left\langle \mathbf{w}_{r', s}^{\dagger}, [\overline{\mathbf{x}}^{(2)}, \cdots, \overline{\mathbf{x}}^{(L-1)}]\right\rangle\right) \Big| \label{eq:Comptosimple1}\\&
					+ \Big| \ \frac{\dout }{m} \sum_{s'\in [\dout ]: s' \ne s} \sum_{r \in [m]}  b_{r, s'} b_{r', s'}^{\dagger} b_{r, s} \Phi_{r', s} \left(\left\langle \mathbf{w}_{r', s}^{\dagger}, [\overline{\mathbf{x}}^{(2)}, \cdots, \overline{\mathbf{x}}^{(L-1)}]\right\rangle\right) \Big| \label{eq:Comptosimple2}\\&
					+ \Big| \ \frac{\dout }{m}  \sum_{i=1}^{L-1} \sum_{s'\in [\dout ]: s' \ne s}  b_{r', s'}^{\dagger} \Phi_{r', s} \left(\left\langle \mathbf{w}_{r', s}^{\dagger}, [\overline{\mathbf{x}}^{(2)}, \cdots, \overline{\mathbf{x}}^{(L-1)}]\right\rangle\right) \left\langle \mathbf{e}_s^{\top} \mathbf{Back}_{L \to L}, \mathbf{e}_{s'}^{\top} \mathbf{Back}_{i \to L} \right\rangle  \Big| \label{eq:Comptosimple3}\\&
					+ \Big|  \frac{\dout }{m} \sum_{i=1}^{L-1}  b_{r', s}^{\dagger} \Phi_{r', s} \left(\left\langle \mathbf{w}_{r', s}^{\dagger}, [\overline{\mathbf{x}}^{(2)}, \cdots, \overline{\mathbf{x}}^{(L-1)}]\right\rangle\right) \left\langle \mathbf{e}_s^{\top} \mathbf{Back}_{L \to L}, \mathbf{e}_s^{\top} \mathbf{Back}_{i \to L} \right\rangle  \Big| \label{eq:Comptosimple4}.
				\end{align}
				\endgroup
				First, we can use Lemma~\ref{lemma:backward_correlation} to show that both eq.~\ref{eq:Comptosimple3} and eq.~\ref{eq:Comptosimple4} are small.
				\begin{align*}
					&\Big| \ \frac{\dout }{m}  \sum_{i=1}^{L-1} \sum_{s'\in [\dout ]: s' \ne s}  b_{r', s'}^{\dagger} \Phi_{r', s} \left(\left\langle \mathbf{w}_{r', s}^{\dagger}, [\overline{\mathbf{x}}^{(2)}, \cdots, \overline{\mathbf{x}}^{(L-1)}]\right\rangle\right) \left\langle \mathbf{e}_s^{\top} \mathbf{Back}_{L \to L}, \mathbf{e}_{s'}^{\top} \mathbf{Back}_{i \to L} \right\rangle  \Big| \\&
					\le \sum_{i=1}^{L-1} \sum_{s'\in [\dout ]: s' \ne s} \frac{\dout }{m} \cdot \abs{b_{r', s'}^{\dagger}} \cdot \abs{\Phi_{r', s} \left(\left\langle \mathbf{w}_{r', s}^{\dagger}, [\overline{\mathbf{x}}^{(2)}, \cdots, \overline{\mathbf{x}}^{(L-1)}]\right\rangle\right)} \cdot \abs{\left\langle \mathbf{e}_s^{\top} \mathbf{Back}_{L \to L}, \mathbf{e}_{s'}^{\top} \mathbf{Back}_{i \to L} \right\rangle } \\& \le  \mathcal{O}(L\dout \rho C_{\Phi} m^{-0.25}).
				\end{align*}
				
				Also, 
				\begin{align*}
					&\Big|  \frac{\dout }{m} \sum_{i=1}^{L-1}  b_{r', s}^{\dagger} \Phi_{r', s} \left(\left\langle \mathbf{w}_{r', s}^{\dagger}, [\overline{\mathbf{x}}^{(2)}, \cdots, \overline{\mathbf{x}}^{(L-1)}]\right\rangle\right) \left\langle \mathbf{e}_s^{\top} \mathbf{Back}_{L \to L}, \mathbf{e}_s^{\top} \mathbf{Back}_{i \to L} \right\rangle  \Big| \\&
					\le \sum_{i=1}^{L-1}  \frac{\dout }{m} \cdot \abs{b_{r', s}^{\dagger}} \cdot \abs{\Phi_{r', s} \left(\left\langle \mathbf{w}_{r', s}^{\dagger}, [\overline{\mathbf{x}}^{(2)}, \cdots, \overline{\mathbf{x}}^{(L-1)}]\right\rangle\right)} \cdot \abs{\left\langle \mathbf{e}_s^{\top} \mathbf{Back}_{L \to L}, \mathbf{e}_{s}^{\top} \mathbf{Back}_{i \to L} \right\rangle } \\& \le  \mathcal{O}(L \rho C_{\Phi} m^{-0.25}).
				\end{align*}
				
				Since, $\mathbf{b}_{s} \sim \mathcal{O}(0, \frac{1}{\dout }\mathbb{I})$, we can show using using fact~\ref{lem:chi-squared} that with probability at least $1 - e^{-\Omega(\rho^2)}$,
				\begin{align*}
					\abs{\frac{\dout }{m} \sum_{r \in [m]} b_{r, s}^2 - 1} \le \mathcal{O}(\frac{\rho}{\sqrt{m}}).
				\end{align*}   
				Hence, eq.~\ref{eq:Comptosimple1} can be simplified as 
				\begin{align*}
					&\Big| \Phi_{r', s} \left(\left\langle \mathbf{w}_{r', s}^{\dagger}, [\overline{\mathbf{x}}^{(2)}, \cdots, \overline{\mathbf{x}}^{(L-1)}]\right\rangle\right)  - \frac{\dout }{m} \sum_{r \in [m]}  b^2_{r, s} b_{r', s}^{\dagger} \Phi_{r', s} \left(\left\langle \mathbf{w}_{r', s}^{\dagger}, [\overline{\mathbf{x}}^{(2)}, \cdots, \overline{\mathbf{x}}^{(L-1)}]\right\rangle\right) \Big| \\&
					\le \mathcal{O}(\frac{\rho}{\sqrt{m}}) \cdot \abs{b_{r', s}^{\dagger}} \abs{\Phi_{r', s} \left(\left\langle \mathbf{w}_{r', s}^{\dagger}, [\overline{\mathbf{x}}^{(2)}, \cdots, \overline{\mathbf{x}}^{(L-1)}]\right\rangle\right)} 
					\le \mathcal{O}(C_{\Phi}\frac{\rho}{\sqrt{m}}). 
				\end{align*}
				Also, 
				\begin{align*}
					\frac{\dout }{m} \sum_{r \in [m]}  b_{r, s} b_{r, s'} = \frac{1}{2m} \left(\norm{b_{r, s} + b_{r, s'}}^2 - \norm{b_{r, s} - b_{r, s'}}^2\right).    
				\end{align*}
				Since, both $\mathbf{b}_s$ and $\mathbf{b}_{s'}$ are independent gaussian vectors, $\mathbf{b}_{s} + \mathbf{b}_{s'} \sim \mathcal{N}(0, \frac{2}{\dout }\mathbb{I})$ and $\mathbf{b}_{s} - \mathbf{b}_{s'} \sim \mathcal{N}(0, \frac{2}{\dout }\mathbb{I})$. Hence, using fact~\ref{lem:chi-squared} we have with probability at least $1 - e^{-\Omega(\rho^2)}$, for all $s' \in [\dout ]$,
				\begin{align*}
					&\abs{\frac{\dout }{m} \sum_{r \in [m]} (b_{r, s'} + b_{r, s})^2 - 2} \le \mathcal{O}(\frac{\rho}{\sqrt{m}}) \\&
					\abs{\frac{\dout }{m} \sum_{r \in [m]} (b_{r, s'} - b_{r, s})^2 - 2} \le \mathcal{O}(\frac{\rho}{\sqrt{m}}),
				\end{align*}
				and thus
				\begin{align*}
					\abs{\frac{\dout }{m} \sum_{r \in [m]}  b_{r, s} b_{r, s'}} \le \mathcal{O}(\frac{\rho}{\sqrt{m}}).
				\end{align*}
				This can be used to simplify eq.~\ref{eq:Comptosimple4}.
				\begin{align*}
					&\Big|  \frac{\dout }{m} \sum_{s'\in [\dout ]: s' \ne s} \sum_{r \in [m]}  b_{r, s'} b_{r', s'}^{\dagger} b_{r, s} \Phi_{r', s} \left(\left\langle \mathbf{w}_{r', s}^{\dagger}, [\overline{\mathbf{x}}^{(2)}, \cdots, \overline{\mathbf{x}}^{(L-1)}]\right\rangle\right) \Big| \\&
					\le  \sum_{s'\in [\dout ]: s' \ne s} \Big|  \frac{\dout }{m} \sum_{r \in [m]}  b_{r, s'} b_{r', s'}^{\dagger} b_{r, s} \Phi_{r', s} \left(\left\langle \mathbf{w}_{r', s}^{\dagger}, [\overline{\mathbf{x}}^{(2)}, \cdots, \overline{\mathbf{x}}^{(L-1)}]\right\rangle\right) \Big|\\&
					\le \mathcal{O}(\frac{\rho}{\sqrt{m}})  \sum_{s'\in [\dout ]: s' \ne s} \cdot \abs{b_{r', s'}^{\dagger}} \abs{\Phi_{r', s} \left(\left\langle \mathbf{w}_{r', s}^{\dagger}, [\overline{\mathbf{x}}^{(2)}, \cdots, \overline{\mathbf{x}}^{(L-1)}]\right\rangle\right)}\\&
					\le \mathcal{O}(C_{\Phi} \dout  \frac{\rho}{\sqrt{m}}).
				\end{align*}
				Hence, adding everything up, we have with probability exceeding $1 - e^{-\Omega(\rho^2)}$,
				\begin{align*}
					&\Big| \Phi_{r', s} \left(\left\langle \mathbf{w}_{r', s}^{\dagger}, [\overline{\mathbf{x}}^{(2)}, \cdots, \overline{\mathbf{x}}^{(L-1)}]\right\rangle\right)  \\&- \frac{\dout }{m} \sum_{i=1}^{L}  \sum_{s' \in [\dout ]}  \sum_{r \in [m]}  b_{r, s'} b_{r', s'}^{\dagger} \mathbf{Back}_{i \to L, r, s} \Phi_{r', s} \left(\left\langle \mathbf{w}_{r', s}^{\dagger}, [\overline{\mathbf{x}}^{(2)}, \cdots, \overline{\mathbf{x}}^{(L-1)}]\right\rangle\right) \Big| \\&
					\le  \mathcal{O}(C_{\Phi}\frac{\rho}{\sqrt{m}}) + \mathcal{O}(C_{\Phi} \dout  \frac{\rho}{\sqrt{m}}) + \mathcal{O}(L\dout \rho C_{\Phi} m^{-0.25}) +  \mathcal{O}(L \rho C_{\Phi} m^{-0.25})  \\&
					\le \mathcal{O}(L\dout  \rho C_{\Phi} m^{-0.25}).
				\end{align*}
			\end{proof}
			Thus, introducing claim~\ref{claim:simplifybig} in eq.~\ref{eq:prefinalf}, we have
			\begingroup \allowdisplaybreaks
			\begin{align*}
				&\abs{ F_s^{(L)}(\mathbf{h}^{(\ell-1)}, \mathbf{x}^{(\ell)}) - \sum_{r' \in [p]}  b_{r', s}^{\dagger} \Phi_{r', s} \left(\left\langle \mathbf{w}_{r', s}^{\dagger}, [\overline{\mathbf{x}}^{(2)}, \cdots, \overline{\mathbf{x}}^{(L-1)}]\right\rangle\right) } \\& \le \mathcal{O}(\dout Lp\rho^8  \mathfrak{C}_{\varepsilon}(\Phi, \mathcal{O}(\varepsilon_x^{-1}))  m^{-1/30}) + \mathcal{O}(\mathfrak{C}_{\varepsilon}(\Phi_{r' s}, \mathcal{O}(\varepsilon_x^{-1})) \rho^2 \sqrt{\dout ^3Lp} m^{-0.1}) \\& + \dout Lp \rho^2 (\varepsilon + \mathcal{O}( L_{\Phi}\rho^5 m^{-2/15}) + \mathcal{O}(\varepsilon_x^{-1} L_{\Phi} L^4 \rho^{11} m^{-1/12} +  L_{\Phi} L^{11/6} \rho^2 \varepsilon_x^{2/3})) \\& + \mathcal{O}(\rho^8\dout Lp  m^{-2/15})  + \mathcal{O}(Lp\dout  \rho C_{\Phi} m^{-0.25}) \\&
				\le \mathcal{O}(\dout Lp\rho^2 \varepsilon + \dout L^{17/6} p \rho^4 L_{\Phi} \varepsilon_x^{2/3} + \dout ^{3/2} L^5 p \rho^{11} L_{\Phi} C_{\Phi}  \mathfrak{C}_{\varepsilon}(\Phi, \mathcal{O}(\varepsilon_x^{-1}))  m^{-1/30} ) .
			\end{align*}
			\endgroup
		
		\begin{lemma}\label{lemma:norm_WA}
			With probability at least $1-e^{-\Omega(\rho^2)}$,
			\begin{align*}
				& \norm{\mathbf{W}^{\ast}}_F = 0,
				\\
				&\norm{\mathbf{A}^{\ast}}_F \le \mathcal{O}\left(\rho \dout ^{1/2} \frac{\mathfrak{C}_{\varepsilon}(\Phi, \mathcal{O}(\varepsilon_x^{-1}))}{\sqrt{m}}\right).
			\end{align*}
			\begin{proof}
				The norm of $\mathbf{W}^{\ast}$ follows from the fact that it is a zero matrix. From def.~\ref{def:existence}, we have that
				\begin{align*}
					\mathbf{a}^{*}_{r} = \frac{\dout }{m} \sum_{s \in [\dout ]} \sum_{r' \in [p]} b_{r, s} b_{r', s}^{\dagger} H_{r', s} \left(\theta_{r', s} \left(\langle \mathbf{w}_{r}, \overline{\mathbf{W}}^{[L]} \mathbf{w}_{r', s}^{\dagger}\rangle\right), \sqrt{m/2} a_{r, d}\right) \mathbf{e}_d, \quad \forall r \in [m],
				\end{align*}
				where
				\begin{equation*} 
					\theta_{r', s} = \frac{\sqrt{m/2}}{\norm[1]{ \overline{\mathbf{W}}^{[L]} \mathbf{w}_{r', s}^{\dagger}}}.
				\end{equation*}
				Since, there is a dependence between $\mathbf{w}_r$ and $\overline{\mathbf{W}}^{[L]}$, we again need to re-randomize some rows of $\overline{\mathbf{W}}^{[L]}$ as has been done in thm.~\ref{thm:existence_pseudo}. Following the steps as has been done to bound eq.~\ref{eq:rerandRNN_3}, we can get
				\begin{align*}
					\mathbf{A}^{\ast} = \widetilde{\mathbf{A}}^{\ast} + \overline{\mathbf{A}}^{\ast},
				\end{align*}
				where $\norm{\overline{\mathbf{A}}^{\ast}}_F \le \mathcal{O}(\mathfrak{C}_{\varepsilon}(\Phi, \mathcal{O}(\varepsilon_x^{-1})) \rho^6 m^{-5/6})$ with probability at least $1-e^{-\Omega(\rho^2)}$ and for each $r \in [m]$,
				\begin{align*}
					\widetilde{\mathbf{a}}^{*}_{r} = \frac{\dout }{m} \sum_{s \in [\dout ]} \sum_{r' \in [p]} b_{r, s} b_{r', s}^{\dagger} H_{r', s} \left(\theta_{r', s} \left(\langle \mathbf{w}_{r}, \overline{\widetilde{\mathbf{W}}}^{[L]} \mathbf{w}_{r', s}^{\dagger}\rangle\right), \sqrt{m/2} a_{r, d}\right) \mathbf{e}_d, 
				\end{align*}
				where $\overline{\widetilde{\mathbf{W}}}^{[L]}$ doesn't depend on the weight vector $\mathbf{w}_r$. Using the properties of the function $H_{r', s}$ from def.~\ref{Def:Function_approx}, we can show that with probability at least $1-e^{-\Omega(\rho^2)}$,
				\begin{align*}
					\norm{\widetilde{\mathbf{A}}^{\ast}}_F \le \mathcal{O}(\dout ^{1/2} \mathfrak{C}_{\varepsilon}(\Phi, \mathcal{O}(\varepsilon_x^{-1})) \rho m^{-1/2}).
				\end{align*}
				\todo{Add few more details if you have time.}
			\end{proof}
		\end{lemma}
\section{Optimization and Generalization: proofs}\label{sec:optim_general_proofs}
\subsection{Proof of lemma~\ref{lem:trainloss}}
\begin{lemma}[Restating lemma~\ref{lem:trainloss}]\label{lem:trainloss_proof}
	For a constant $\varepsilon_x = \frac{1}{\operatorname{poly}(\rho)}$ and for every constant $\varepsilon \in \left(0, \frac{1}{p \cdot \operatorname{poly}(\rho) \cdot \mathfrak{C}_{\mathfrak{s}}(\Phi, \mathcal{O}(\varepsilon_x^{-1}))}\right),$ there exists $C^{\prime}=\mathfrak{C}_{\varepsilon}(\Phi, \mathcal{O}(\varepsilon_x^{-1}))$ and a parameter $\lambda=\Theta\left(\frac{\varepsilon}{L \rho}\right)$
	so that, as long as $m \geq \operatorname{poly}\left(C', p, L, \dout , \varepsilon^{-1}\right)$ and $N \geq \Omega\left(\frac{\rho^{3} p C_{\Phi}^2}{\varepsilon^2}\right),$ setting learning rate $\eta=\Theta\left(\frac{1}{\varepsilon \rho^{2} m}\right)$ and
	$T=\Theta\left(\frac{p^{2}  C'^2 \mathrm{poly}(\rho)}{\varepsilon^{2}}\right),$ we have
	\begin{align*}
		\underset{\mathrm{sgd}}{\mathbb{E}}\Big[\frac{1}{T} \sum_{t=0}^{T-1}  \underset{(\obx, \mathbf{y}^{\ast}) \sim \mathcal{Z}}{\mathbb{E}} \mathrm{Obj}(\obx, \mathbf{y}^{\ast};  \mathbf{W}+\mathbf{W}_{t}, \mathbf{A} + \mathbf{A}_t) \Big] \leq \mathrm{OPT} + \frac{\varepsilon}{2} + \frac{1}{\mathrm{poly}(\rho)},
	\end{align*}
	and $\left\|W_{t}\right\|_{F} \leq \frac{\Delta}{\sqrt{m}}$ for $\Delta=\frac{C'^{2} p^{2} \mathrm { poly }(\rho)}{\varepsilon^{2}}$.
\end{lemma}

\begin{proof}
	The proof will follow exactly the same routine as lemma 7.1 in \cite{allen2019can}. We allow $\mathbf{A}$ to change, which leads to changes in the proof. We outline the major differences here for completeness. For simplicity, we outline the proof for Gradient Descent.
	
	The training objective is given by
	\begin{align*}
		\mathrm{Obj}( \mathbf{W}_{t},  \mathbf{A}_t) &= \underset{\left(\obx, y^{\ast}\right) \sim \mathcal{Z}}{\mathbb{E}} \mathrm{Obj}(\obx, y^{\ast};  \mathbf{W}_{t},  \mathbf{A}_t),\text{ where } \\
		\mathrm{Obj}(\obx, y^{\ast};  \mathbf{W}_{t},  \mathbf{A}_t) &= G(\lambda F^{(\ell)}_{\mathrm{rnn}}(\bx, y^{\ast};  \mathbf{W} + \mathbf{W}_{t},  \mathbf{A} + \mathbf{A}_t)).
	\end{align*}
	Let $\obx$ be a true sequence and $\bx$ be its normalized version. Let's consider the matrices $\mathbf{W}+\mathbf{W}_t$, $\mathbf{A}+\mathbf{A}_t$ after SGD iteration $t$. Let
	\begin{itemize}
		\item  at RNN cell $i$,  $\mathbf{h}^{(i)}$,  $\mathbf{Back}_{i \to L}$ and  $\mathbf{D}^{(i)}$ are defined w.r.t. $\mathbf{A}, \mathbf{W}, \mathbf{B}, \bx$.
		\item at RNN cell $i$,  $\mathbf{h}^{(i)} + \mathbf{h}^{(i)}_t$,  $\mathbf{Back}_{i \to L} + \mathbf{Back}_{i \to L, t}$ and  $\mathbf{D}^{(i)} + \mathbf{D}^{(i)}_t$ are defined w.r.t. $\mathbf{A} + \mathbf{A}_t, \mathbf{W} + \mathbf{W}_t, \mathbf{B}, \bx$.
	\end{itemize}
	
	Define the following regularization term:
	\begin{align*}
		R(\bx; \mathbf{W}', \mathbf{A}') &= \sum_{i = 2}^{L} (\mathbf{Back}_{i \to L} + \mathbf{Back}_{i \to L, t}) (\mathbf{D}^{(i)} + \mathbf{D}^{(i)}_t) (\mathbf{W}'(\mathbf{h}^{(i-1)} +  \mathbf{h}^{(i-1)}_t)  + \mathbf{A}' \bx^{(i)}), \\&
		= \sum_{i = 2}^{L} (\mathbf{Back}_{i \to L} + \mathbf{Back}_{i \to L, t}) (\mathbf{D}^{(i)} + \mathbf{D}^{(i)}_t) \left(\left[\mathbf{W}', \mathbf{A}'\right]_r \left[\mathbf{h}^{(i-1)} +  \mathbf{h}^{(i-1)}_t, \bx^{(i)}\right]\right)
	\end{align*}
	which is a linear function over $[\mathbf{W}', \mathbf{A}']_r$. Define the following regularized loss function:
	\begin{align*}
		\widetilde{G}(\mathbf{W}', \mathbf{A}') &= \underset{\left(\obx, y^{\ast}\right) \sim \mathcal{Z}}{\mathbb{E}} \widetilde{G}(\obx, y^{\ast};  \mathbf{W}',  \mathbf{A}'),\text{ where } \\ 
		\widetilde{G}(\obx, y^{\ast};  \mathbf{W}',  \mathbf{A}') &= G(\lambda F^{(L)}_{\mathrm{rnn}}(\obx;  \mathbf{W} + \mathbf{W}_t,  \mathbf{A} + \mathbf{A}_t) + \lambda  R(\bx; \mathbf{W}', \mathbf{A}'))
	\end{align*}
	Note that,
	$ \widetilde{G}(\mathbf{0}, \mathbf{0}) = \mathrm{Obj}(\mathbf{W}_t, \mathbf{A}_t)$ and $\nabla_{[\mathbf{W}', \mathbf{A}']} \widetilde{G}(\mathbf{0}, \mathbf{0}) = \nabla_{[\mathbf{W}_t, \mathbf{A}_t]} \mathrm{Obj}(\mathbf{W}_t, \mathbf{A}_t)$. First of all, 
	we have
	\begin{align*}
		&F^{(\ell)}_{\mathrm{rnn}}(\bx;  \mathbf{W} + \mathbf{W}_t,  \mathbf{A} + \mathbf{A}_t) - F^{(\ell)}_{\mathrm{rnn}}(\bx;  \mathbf{W},  \mathbf{A}) \tag{difference between RNN output}\\&=  \sum_{\ell \in [L]} \mathbf{Back}_{\ell \rightarrow L}  \mathbf{D}^{(\ell)} \left(\mathbf{W}_t \mathbf{h}^{(\ell-1)} + \mathbf{A}_t \mathbf{x}^{(\ell)} \right) + \varepsilon' \tag{using lemma~\ref{lemma:perturb_NTK_small_output}}\\&
		= \sum_{\ell \in [L]}\left(\mathbf{Back}_{\ell \rightarrow L}+\mathbf{Back}_{\ell \rightarrow L, t}\right)\left(\mathbf{D}^{(\ell)}+\mathbf{D}^{(\ell)}_t\right) \left(\mathbf{W}_t\left(\mathbf{h}^{(\ell-1)}+h^{(\ell-1)}_t\right) + \mathbf{A}_t \mathbf{x}^{(\ell)} \right) + \varepsilon' + \varepsilon'' \tag{using lemma~\ref{lemma:perturb_NTK_small}}\\&
		=  R(\bx; \mathbf{W}_t, \mathbf{A}_t) + \varepsilon' + \varepsilon'' \tag{using the definition of $R$},
	\end{align*}
	where $0 \le \varepsilon', \varepsilon'' \le \mathcal{O}(\rho^7 \Delta^{4/3} m^{-1/6})$. Also, from lemma~\ref{lemma:perturb_small_target}, we have
	\begin{align*}
	    R(\bx; \mathbf{W}^{\ast}, \mathbf{A}^{\ast}) = F^{\ast}(\obx) \pm \varepsilon''', \label{eq:RFast}
	\end{align*}
	with $\varepsilon''' \le \varepsilon/2 + poly(\rho)^{-1}$, when $\varepsilon_x \le poly(\rho)^{-1}$, $\varepsilon < (p \cdot poly(\rho) \cdot \mathfrak{C}_s(\Phi, \mathcal{O}(\varepsilon_x^{-1})))^{-1}$ and $m \ge poly(\varrho)$.
	Hence, 
	\begin{align}
		& \widetilde{G}\left(\frac{1}{\lambda} \mathbf{W}^{\ast} - \mathbf{W}_t, \frac{1}{\lambda}\mathbf{A}^{\ast} - \mathbf{A}_t\right) \nonumber\\&= G\left(\lambda F_{\mathrm{rnn}}^{(L)}\left(\bx;  \mathbf{W} + \mathbf{W}_t,  \mathbf{A} + \mathbf{A}_t\right) + \lambda  R\left(\bx; \frac{1}{\lambda} \mathbf{W}^{\ast} - \mathbf{W}_t, \frac{1}{\lambda}\mathbf{A}^{\ast} - \mathbf{A}_t \right)\right) \nonumber\\&
		= G\left(\lambda F_{\mathrm{rnn}}^{(L)}\left(\bx;  \mathbf{W},  \mathbf{A}\right) + \lambda R\left(\bx;  \mathbf{W}_t, \mathbf{A}_t \right) + \lambda  R\left(\bx; \frac{1}{\lambda} \mathbf{W}^{\ast} - \mathbf{W}_t, \frac{1}{\lambda}\mathbf{A}^{\ast} - \mathbf{A}_t \right)\right) \pm \varepsilon' \pm \varepsilon'' \tag{using the difference of  $F_{\mathrm{rnn}}^{(L)}$ derived above} \nonumber\\&
		=  G\left(\lambda F_{\mathrm{rnn}}^{(L)}\left(\obx;  \mathbf{W},  \mathbf{A}\right) +  R\left(\bx;  \mathbf{W}^{\ast}, \mathbf{A}^{\ast} \right)\right) \pm \varepsilon' \pm \varepsilon'' \nonumber\\&
		= G\left(R\left(\bx;  \mathbf{W}^{\ast}, \mathbf{A}^{\ast} \right)\right) \pm \varepsilon' \pm \varepsilon'' \pm \varepsilon \label{eq:optimality_step3}
		\\&= G(F^{\ast}(\bx)) \pm \varepsilon' \pm \varepsilon'' \pm \varepsilon \pm \varepsilon''' \tag{using eq.~\ref{eq:RFast}}\\&= OPT + \mathcal{O}(\varepsilon) + \frac{1}{poly(\rho)},\label{eq:optimality}
	\end{align}
	after setting everything properly. Here $\lambda$ is chosen above such that with high probability $\abs{\lambda F_{\mathrm{rnn}}^{(L)}\left(\obx;  \mathbf{W},  \mathbf{A}\right)} \le \varepsilon$ to get eq.~\ref{eq:optimality_step3}.
	
	Now, at each step of gradient descent, we have
	\begin{align*}
		[\mathbf{W}_{t+1}, \mathbf{A}_{t+1}] &= [\mathbf{W}_{t}, \mathbf{A}_{t}] - \eta \nabla_{[\mathbf{W}_{t}, \mathbf{A}_{t}]} \mathrm{Obj}(\mathbf{W}_{t}, \mathbf{A}_{t}) \\&
		= [\mathbf{W}_{t}, \mathbf{A}_{t}] - \eta \nabla_{[\mathbf{W}', \mathbf{A}']} \widetilde{G}(\mathbf{0}, \mathbf{0}) \tag{using the equivalence between gradient derived above}. 
	\end{align*}
	Hence, we have
	\begin{align*}
		\norm{[\mathbf{W}_{t+1}, \mathbf{A}_{t+1}] - \frac{1}{\lambda}[\mathbf{W}^{\ast}, \mathbf{A}^{\ast}]}^2 &= \norm{[\mathbf{W}_{t}, \mathbf{A}_{t}] - \frac{1}{\lambda}[\mathbf{W}^{\ast}, \mathbf{A}^{\ast}]}^2 + \norm{[\mathbf{W}_{t+1}, \mathbf{A}_{t+1}] - [\mathbf{W}_t, \mathbf{A}_t]}^2 \\&+ 2 \left\langle [\mathbf{W}_{t+1}, \mathbf{A}_{t+1}] - [\mathbf{W}_t, \mathbf{A}_t],  [\mathbf{W}_{t}, \mathbf{A}_{t}] - \frac{1}{\lambda}[\mathbf{W}^{\ast}, \mathbf{A}^{\ast}] \right\rangle \\&
		=  \norm{[\mathbf{W}_{t}, \mathbf{A}_{t}] - \frac{1}{\lambda}[\mathbf{W}^{\ast}, \mathbf{A}^{\ast}]}^2 + \eta^2 \norm{\nabla_{[\mathbf{W}', \mathbf{A}']} \widetilde{G}(\mathbf{0}, \mathbf{0})}^2 \tag{from descent update}
		\\& - 2\eta \left\langle\nabla_{[\mathbf{W}', \mathbf{A}']} \widetilde{G}(\mathbf{0}, \mathbf{0}),  [\mathbf{W}_{t}, \mathbf{A}_{t}] - \frac{1}{\lambda}[\mathbf{W}^{\ast}, \mathbf{A}^{\ast}] \right\rangle \\&
		\ge \norm{[\mathbf{W}_{t}, \mathbf{A}_{t}] - \frac{1}{\lambda}[\mathbf{W}^{\ast}, \mathbf{A}^{\ast}]}^2 + \eta^2 \norm{\nabla_{[\mathbf{W}', \mathbf{A}']} \widetilde{G}(\mathbf{0}, \mathbf{0})}^2 \\& - 2\eta \left(\widetilde{G}(\mathbf{W}_t - \frac{1}{\lambda} \mathbf{W}^{\ast}, \mathbf{A}_t - \frac{1}{\lambda} \mathbf{A}^{\ast}) - \widetilde{G}(0, 0)\right) 
		\tag{using the convexity of $\widetilde{G}$}\\&
		= \norm{[\mathbf{W}_{t}, \mathbf{A}_{t}] - \frac{1}{\lambda}[\mathbf{W}^{\ast}, \mathbf{A}^{\ast}]}^2 + \eta^2 \norm{\nabla_{[\mathbf{W}_t, \mathbf{A}_t]} \mathrm{Obj}(\mathbf{W}_t, \mathbf{A}_t)}^2 \\& - 2\eta \left(OPT +\mathcal{O}(\varepsilon) + \frac{1}{poly(\rho)} - \mathrm{Obj}(\mathbf{W}_t, \mathbf{A}_t)\right) \tag{using eq.~\ref{eq:optimality}}.
	\end{align*}
 Thus, 
	\begin{align*}
		\frac{1}{T} \sum_{t \in [T]} \mathrm{Obj}(\mathbf{W}_t, \mathbf{A}_t) &\le  \frac{1}{2\eta T}  \left(\norm{[\mathbf{W}_T, \mathbf{A}_T] - \frac{1}{\lambda}[\mathbf{W}^{\ast}, \mathbf{A}^{\ast}]}^2 - \norm{ \frac{1}{\lambda}[\mathbf{W}^{\ast}, \mathbf{A}^{\ast}]}^2 \right)  \\& + \frac{\eta}{2T} \sum_{t \in [T]} \norm{\nabla_{[\mathbf{W}_t, \mathbf{A}_t]} \mathrm{Obj}(\mathbf{W}_t, \mathbf{A}_t)}^2 + OPT + \frac{1}{poly(\rho)} +  \mathcal{O}(\varepsilon).
	\end{align*}
	We can then finish the proof by bounding $ \norm[1]{\nabla_{[\mathbf{W}_t, \mathbf{A}_t]} \mathrm{Obj}(\mathbf{W}_t, \mathbf{A}_t)} \approx \norm[1]{\nabla \mathrm{Obj}(\mathbf{0}, \mathbf{0})} \le \mathcal{O}(\lambda \rho^2 \sqrt{m})$ for $\mathbf{W}_t, \mathbf{A}_t \le \frac{\Delta}{\sqrt{m}}$. We then show that $\Delta =  C'^2 poly(\rho) \varepsilon^{-2}$, since it can be bounded by the term $\eta T \sqrt{m} \sup_{t \in [T]} \cdot \norm[1]{\nabla_{[\mathbf{W}_t, \mathbf{A}_t]} \mathrm{Obj}(\mathbf{W}_t, \mathbf{A}_t)}$.
\end{proof}

\subsection{Helping lemmas}

\begin{lemma} \label{lemma:perturb_NTK_small} [first order coupling]
	Let $\mathbf{W}, \mathbf{A}, \mathbf{B}$ be at random initialization, $\mathbf{x}^{(1)}, \cdots, \mathbf{x}^{(L)}$ be a fixed normalized input sequence, and $\Delta \in\left[\varrho^{-100}, \varrho^{100}\right] .$ With probability at least $1-e^{-\Omega(\rho)}$ over $\mathbf{W}, \mathbf{A}, \mathbf{B}$ the following holds. Given any matrices $W^{\prime}$ with $\left\|\mathbf{W}^{\prime}\right\|_{2} \leq \frac{\Delta}{\sqrt{m}},$ $\mathbf{A}^{\prime}$ with $\left\|\mathbf{A}^{\prime}\right\|_{2} \leq \frac{\Delta}{\sqrt{m}},$ and any $\widetilde{\mathbf{W}}$ with $\|\widetilde{\mathbf{W}}\|_{2} \leq \frac{\omega}{\sqrt{m}},$ $\widetilde{\mathbf{A}}$ with $\|\widetilde{\mathbf{A}}\|_{2} \leq \frac{\omega}{\sqrt{m}},$ letting
	$\mathbf{h}^{(\ell)}, \mathbf{D}^{(\ell)},$  $\mathbf{Back}_{i \rightarrow L}$ be defined with respect to $\mathbf{W}, \mathbf{A}, \mathbf{B}, \obx,$ and $\mathbf{h}^{(\ell)} + \mathbf{h}^{(\ell)\prime}, \mathbf{D}^{(\ell)} + \mathbf{D}^{(\ell)\prime},$  $\mathbf{Back}_{i \rightarrow L} + \mathbf{Back}^{\prime}_{\ell \rightarrow j}$ be defined with respect to $\mathbf{W}+\mathbf{W}^{\prime}, \mathbf{A} + \mathbf{A}',  \mathbf{B}, \obx,$
	then
	$$
	\begin{array}{l}
		\| \sum_{\ell \in [L]}\left(\mathbf{Back}_{\ell \rightarrow L}+\mathbf{Back}_{\ell \rightarrow L}^{\prime}\right)\left(\mathbf{D}^{(\ell)}+\mathbf{D}^{(\ell)\prime}\right) \left(\widetilde{\mathbf{W}}\left(\mathbf{h}^{(\ell-1)}+h^{(\ell-1)\prime}\right) + \widetilde{\mathbf{A}} \mathbf{x}^{(\ell)} \right) \\
		\quad\quad\quad\quad -\sum_{\ell \in [L]} \mathbf{Back}_{\ell \rightarrow L}  \mathbf{D}^{(\ell)} \left(\widetilde{\mathbf{W}} \mathbf{h}^{(\ell-1)} + \widetilde{\mathbf{A}} \mathbf{x}^{(\ell)} \right) \| \leq O\left(\frac{\omega \rho^{6} \Delta^{1 / 3}}{m^{1 / 6}}\right).
	\end{array}
	$$
\end{lemma}

\begin{proof}
	The proof will follow the same technique as has been used in Lemma 6.2 in \cite{allen2019can}. We give a brief overview here. 
	
	We allow a change in $\mathbf{A}$ by $\mathbf{A}^{\prime}$, which wasn't allowed in their lemma. However, we show now that the primary 3 properties (specified in Lemma F.1 in \cite{allen2019can}) used to prove the lemma change only by a constant factor, with the introduction of perturbation in $\mathbf{A}$. With probability at least $1-e^{-\Omega(\rho^2)}$,
	\begin{enumerate}
		\item $\norm{\mathbf{h}^{(\ell)\prime}}_2 \leq \mathcal{O}\left(\rho^{6} \Delta / \sqrt{m}\right).$ 
		\item $\norm{\mathbf{D}^{(\ell)\prime}}_{0} \leq \mathcal{O}\left(\rho^{4} \Delta^{2 / 3} m^{2 / 3}\right).$
		\item $\norm{ \mathbf{Back}_{\ell \rightarrow L}^{\prime} }_{2} \leq \mathcal{O}\left(\Delta^{1 / 3} \rho^{6} m^{1 / 3}\right).$
	\end{enumerate}
	Property 1 and property 2 will follow from Claim C.2 and property 3 will follow from Claim C.9 in \cite{allen2019convergence_rnn} with the following change. Due to the introduction of perturbation in $\mathbf{A}$, eq. C.2 in \cite{allen2019convergence_rnn} changes to
	\begin{align*}
		\mathbf{g}^{(\ell)\prime} = \mathbf{W}^{\prime} \mathbf{D}^{(\ell)} \mathbf{g}^{(\ell)} + (\mathbf{W} + \mathbf{W}^{\prime})\cdot \mathbf{D}^{(\ell)\prime}\cdot \mathbf{g}^{(\ell)} + (\mathbf{W} + \mathbf{W}^{\prime}) \cdot  (\mathbf{D}^{(\ell)} + \mathbf{D}^{(\ell)\prime}) \cdot  \mathbf{g}^{(\ell)\prime} + \mathbf{A}^{\prime} \mathbf{x}^{(\ell)},
	\end{align*}
	where we introduce an extra last term. Thus, since $\norm{\mathbf{A}^{\prime} \mathbf{x}^{(\ell)}} \le \norm{\mathbf{A}^{\prime}} \norm{\mathbf{x}^{(\ell)}} \le \mathcal{O}(\frac{\Delta}{\sqrt{m}})$, $\mathbf{g}^{(\ell)\prime}$ can be similarly written as $\mathbf{g}^{(\ell)\prime}_1 + \mathbf{g}^{(\ell)\prime}_2$, where $\norm{\mathbf{g}^{(\ell)\prime}_1} \le \tau_1$ and $\norm{\mathbf{g}^{(\ell)\prime}_2}_0 \le \tau_2$, with $\tau_1$ just changing by a factor 2 in eq. C.1\cite{allen2019convergence_rnn}. This minor change percolates to minor changes in the constant factors in $\norm{\mathbf{h}^{(\ell)\prime}}$, $\norm{\mathbf{D}^{(\ell)\prime}}_{0}$ and $\norm{ \mathbf{Back}_{\ell \rightarrow L}^{\prime} }_{2}$.
	
	Now, the proof follows from the following set of equations.
	\begin{align*}
		&\| \sum_{\ell \in [L]}\left(\mathbf{Back}_{\ell \rightarrow L}+\mathbf{Back}_{\ell \rightarrow L}^{\prime}\right)\left(\mathbf{D}^{(\ell)}+\mathbf{D}^{(\ell)\prime}\right) \left(\widetilde{\mathbf{W}}\left(\mathbf{h}^{(\ell-1)}+\mathbf{h}^{(\ell-1)\prime}\right) + \widetilde{\mathbf{A}} \mathbf{x}^{(\ell)} \right) \\
		&\quad\quad\quad\quad -\sum_{\ell \in [L]} \mathbf{Back}_{\ell \rightarrow L}  \mathbf{D}^{(\ell)} \left(\widetilde{\mathbf{W}} \mathbf{h}^{(\ell-1)} + \widetilde{\mathbf{A}} \mathbf{x}^{(\ell)} \right) \| \\&
		\le \| \sum_{\ell \in [L]}\mathbf{Back}_{\ell \rightarrow L}^{\prime} \left(\mathbf{D}^{(\ell)}+\mathbf{D}^{(\ell)\prime}\right) \left(\widetilde{\mathbf{W}}\left(\mathbf{h}^{(\ell-1)}+\mathbf{h}^{(\ell-1)\prime}\right) + \widetilde{\mathbf{A}} \mathbf{x}^{(\ell)} \right) \| \\&
		\le \underbrace{\|\sum_{\ell \in [L]} \mathbf{Back}_{\ell \rightarrow L}^{\prime} \left(\mathbf{D}^{(\ell)}+\mathbf{D}^{(\ell)\prime}\right) \widetilde{\mathbf{W}} \left(\mathbf{h}^{(\ell-1)}+\mathbf{h}^{(\ell-1)\prime}\right)\|}_{\text{Term 1}} \\& + \underbrace{ \|\sum_{\ell \in [L]} \mathbf{Back}_{\ell \rightarrow L} \left(\mathbf{D}^{(\ell)}+\mathbf{D}^{(\ell)\prime}\right) \widetilde{\mathbf{W}} \mathbf{h}^{(\ell-1)\prime}\|}_{\text{Term 2}}  + 
		\underbrace{\|\sum_{\ell \in [L]} \mathbf{Back}_{\ell \rightarrow L} \mathbf{D}^{(\ell)\prime} \widetilde{\mathbf{W}} \mathbf{h}^{(\ell-1)}\|}_{\text{Term 3}} \\&
		+ \underbrace{\|\sum_{\ell \in [L]} \mathbf{Back}_{\ell \rightarrow L}^{\prime} \left(\mathbf{D}^{(\ell)} +\mathbf{D}^{(\ell)\prime}\right) \widetilde{\mathbf{A}} \mathbf{x}^{(\ell)}\|}_{\text{Term 4}} + \underbrace{\|\sum_{\ell \in [L]} \mathbf{Back}_{\ell \rightarrow L} \mathbf{D}^{(\ell)\prime} \widetilde{\mathbf{A}} \mathbf{x}^{(\ell)}\|}_{\text{Term 5}} 
	\end{align*}
	Term 1, 2 and 3 appear in the proof of Claim 6.2\cite{allen2019can}. Terms 4 and 5 can be bounded using similar technique by using the bound on $\norm{\mathbf{Back}^{\prime}_{\ell \to L}}$ and $\norm{\mathbf{D}^{(\ell)\prime}}_0$ respectively.
\end{proof}

\begin{lemma} \label{lemma:perturb_NTK_small_output} [first order approximation]
	Let $\mathbf{W}, \mathbf{A}, \mathbf{B}$ be at random initialization, $\mathbf{x}^{(1)}, \cdots, \mathbf{x}^{(L)}$ be a fixed normalized input sequence, and $\Delta \in\left[\varrho^{-100}, \varrho^{100}\right] .$ With probability at least $1-e^{-\Omega(\rho)}$ over $\mathbf{W}, \mathbf{A}, \mathbf{B}$ the following holds. Given any matrices $W^{\prime}$ with $\left\|\mathbf{W}^{\prime}\right\|_{2} \leq \frac{\Delta}{\sqrt{m}},$ $\mathbf{A}^{\prime}$ with $\left\|\mathbf{A}^{\prime}\right\|_{2} \leq \frac{\Delta}{\sqrt{m}},$, letting
	$\mathbf{h}^{(\ell)}, \mathbf{D}^{(\ell)},$  $\mathbf{Back}_{i \rightarrow L}$ be defined with respect to $\mathbf{W}, \mathbf{A}, \mathbf{B}, \obx,$ and $\mathbf{h}^{(\ell)} + \mathbf{h}^{(\ell)\prime}, \mathbf{D}^{(\ell)} + \mathbf{D}^{(\ell)\prime},$  $\mathbf{Back}_{i \rightarrow L} + \mathbf{Back}^{\prime}_{\ell \rightarrow j}$ be defined with respect to $\mathbf{W}+\mathbf{W}^{\prime}, \mathbf{A} + \mathbf{A}',  \mathbf{B}, \obx,$
	then
	\begin{align*}
	    &\norm{F^{(L)}_{\mathrm{rnn}}(\bx; \mathbf{W} + \mathbf{W}', \mathbf{A} + \mathbf{A}') - F^{(L)}_{\mathrm{rnn}}(\bx; \mathbf{W}, \mathbf{A}) - F^{(L)}(\mathbf{x}, \mathbf{W}', \mathbf{A}')} \\&
		= \norm{\mathbf{B} \mathbf{h}^{(L)\prime} - \sum_{\ell \in [L]} \mathbf{Back}_{\ell \rightarrow L}  \mathbf{D}^{(\ell)} \left(\mathbf{W}' \mathbf{h}^{(\ell-1)} + \mathbf{A}' \mathbf{x}^{(\ell)} \right)} \le \mathcal{O}(\frac{\rho^7 \Delta^{4/3}}{m^{1/6}}).
	\end{align*}
\end{lemma}

\begin{proof}
	The proof will follow the same technique as has been used in Lemma 6.1 in \cite{allen2019can}. We give a brief outline here. 
	
	We allow a change in $\mathbf{A}$ by $\mathbf{A}^{\prime}$, which wasn't allowed in their lemma. This leads to an introduction of an additional term in eq. H.1 in \cite{allen2019can}. That is, there exist diagonal matrices $\mathbf{D}^{(\ell)\prime\prime}$, where $d^{(\ell)\prime\prime}_{rr} \in [-1, 1]$ and is non zero only when $d^{(\ell)\prime}_{rr} \ne d^{(\ell)}_{rr}$, 
	\begin{align*}
		\mathbf{B} (\mathbf{h}^{(L)} + \mathbf{h}^{(L)\prime}) -  \mathbf{B} \mathbf{h}^{(L)} &= \underbrace{\sum_{i=1}^{L-1} \mathbf{B} (\mathbf{D}^{(L)} + \mathbf{D}^{(L)\prime\prime}) \mathbf{W} \cdots \mathbf{W} (\mathbf{D}^{(i+1)} + \mathbf{D}^{(i+1)\prime\prime}) \mathbf{W}' (\mathbf{h}^{(i)} + \mathbf{h}^{(i)\prime})}_{\text{Term 1}} \\&
		+ \underbrace{\sum_{i=1}^{L-1} \mathbf{B} (\mathbf{D}^{(L)} + \mathbf{D}^{(L)\prime\prime}) \mathbf{W} \cdots \mathbf{W} (\mathbf{D}^{(i+1)} + \mathbf{D}^{(i+1)\prime\prime}) \mathbf{A}' \bx^{(i)}}_{\text{Term 2}}.
	\end{align*}
	In lemma 6.2 of \cite{allen2019can}, Term 1 was shown to be close to \\ $\sum_{i=1}^{L-1} \mathbf{B} \mathbf{D}^{(L)} \mathbf{W} \cdots \mathbf{W} \mathbf{D}^{(i+1)} \mathbf{W}' \mathbf{h}^{(i)}$  by $\mathcal{O}(\frac{\rho^7 \Delta^{4/3}}{m^{1/6}})$. The bound will stay the same, since we have shown similar bounds for $\norm{\mathbf{D}^{(\ell)\prime}}_0$ and $\norm{\mathbf{h}^{(\ell)\prime}}_2$ in the proof of lemma~\ref{lemma:perturb_NTK_small}. 
	
	Using the same technique, we can show that Term 2 is close to $\sum_{i=1}^{L-1} \mathbf{B} \mathbf{D}^{(L)}  \mathbf{W} \cdots \mathbf{W} \mathbf{D}^{(i+1)} \mathbf{A}' \bx^{(i)}$, since Term 2 can be similarly broken down into at most $2^L$ terms of the form $$(\mathbf{B}\mathbf{D}\mathbf{W}\cdots\mathbf{D}\mathbf{W})\mathbf{D}'' (\mathbf{W}\cdots\mathbf{D}\mathbf{W})\mathbf{D}'' \cdots \mathbf{D}'' (\mathbf{W}\cdots\mathbf{D}\mathbf{W}) \mathbf{A}'\mathbf{x}^{(i)}$$ and each term can then be similarly bounded to give an extra error bound $\mathcal{O}(\frac{\rho^7 \Delta^{4/3}}{m^{1/6}})$.
\end{proof}


\begin{lemma}\label{lemma:perturb_small_target}
	Let $\mathbf{W}^{\ast}$ and $\mathbf{A}^{\ast}$ be as defined in def. \ref{def:existence}. Let $\mathbf{W}, \mathbf{A}, \mathbf{B}$ be at random initialization, $\mathbf{x}^{(1)}, \cdots, \mathbf{x}^{(L)}$ be a fixed normalized input sequence, and $\Delta \in\left[\varrho^{-100}, \varrho^{100}\right] .$ With probability at least $1-e^{-\Omega(\rho)}$ over $\mathbf{W}, \mathbf{A}, \mathbf{B}$ the following holds. Given any matrices $W^{\prime}$ with $\left\|\mathbf{W}^{\prime}\right\|_{2} \leq \frac{\Delta}{\sqrt{m}},$ $\mathbf{A}^{\prime}$ with $\left\|\mathbf{A}^{\prime}\right\|_{2} \leq \frac{\Delta}{\sqrt{m}},$. Letting
	$\mathbf{h}^{(\ell)}, \mathbf{D}^{(\ell)},$  $\mathbf{Back}_{i \rightarrow L}$ be defined with respect to $\mathbf{W}, \mathbf{A}, \mathbf{B}, \obx,$ and $\mathbf{h}^{(\ell)} + \mathbf{h}^{(\ell)\prime}, \mathbf{D}^{(\ell)} + \mathbf{D}^{(\ell)\prime},$  $\mathbf{Back}_{i \rightarrow L} + \mathbf{Back}^{\prime}_{\ell \rightarrow j}$ be defined with respect to $\mathbf{W}+\mathbf{W}^{\prime}, \mathbf{A} + \mathbf{A}',  \mathbf{B}, \obx,$
	then for all $s \in [k]$
	\begin{align*}
		& \sum_{\ell \in [L]} \mathbf{e}_s^{\top} \left(\mathbf{Back}_{\ell \rightarrow L}+\mathbf{Back}_{\ell \rightarrow L}^{\prime}\right)\left(\mathbf{D}^{(\ell)}+\mathbf{D}^{(\ell)\prime}\right) \left(\mathbf{W}^{\ast}\left(\mathbf{h}^{(\ell-1)}+h^{(\ell-1)\prime}\right) + \mathbf{A}^{\ast} \mathbf{x}^{(\ell)} \right) \\& 
		= \sum_{r' \in [p]}  b_{r', s}^{\dagger} \Phi_{r', s} \left(\left\langle \mathbf{w}_{r', s}^{\dagger}, [\overline{\mathbf{x}}^{(2)}, \cdots, \overline{\mathbf{x}}^{(L-1)}]\right\rangle\right)  \\& \pm \mathcal{O}(\dout Lp\rho^2 \varepsilon + \dout L^{7/3} p \rho^2 L_{\Phi} \epsilon_x^{2/3} + \dout  L^5 p \rho^{11} L_{\Phi} C_{\Phi}  C_{\varepsilon}(\Phi, \mathcal{O}(\epsilon_x^{-1}))  m^{-1/30} ) \\& \pm  O\left(\frac{ C_{\varepsilon}(\Phi, \mathcal{O}(\epsilon_x^{-1})) \dout ^{1/2} \rho^{8} \Delta^{1 / 3}}{m^{1 / 6}}\right).
	\end{align*}
\end{lemma}

\begin{proof}
	The proof follows from Lemma~\ref{lemma:perturb_NTK_small}, using the bound on $\norm{\mathbf{W}^{\ast}}$ and $\norm{\mathbf{A}^{\ast}}$ from lemma~\ref{lemma:norm_WA}.
\end{proof}

\section{On Concept Classes}\label{sec:diff}
    The concept class in \cite{allen2019can} matched the output to a true label at each step using loss function $G$, i.e. $F^{\ast}$ belongs to $\mathbb{R}^{ L \times d} \to \Reals^{\dout}$, given by
    \begin{equation} \label{eqn:AL_concept_class}
    F^{\ast (j)}_s\left(\mathbf{x}\right) = \sum_{i : i < j} \sum_{r \in [p]} \phi_{i \to j, r, s} (\mathbf{w}_{i \to j, r, s}^T \mathbf{x}^{(i)}), 
    \end{equation}
    for all $j \in [2, L]$ and $s \in [\dout ]$. Here $\phi_{i \to j, r, s}: \Reals \to \Reals$ are smooth functions and $\mathbf{w}_{i \to j, r, s}$ unit vectors. We can rewrite \eqref{eqn:AL_concept_class} in a more compact vector form
    \begin{equation} \label{eqn:AL_concept_class_simplified}
        F^{\ast (j)}\left(\mathbf{x}\right) =  \sum_{i : i < j} \psi_{i \to j}(\mathbf{x}^{(i)}),
    \end{equation}
where $\psi_{i \to j}(\mathbf{x}^{(i)})$ is defined in the obvious way: it is the vector of inner sums in \eqref{eqn:AL_concept_class}.
 From the previous equation it is clear that $F^{\ast (j)}(\mathbf{x}^{(i)})$ is a \emph{sum} of 
functions of individual tokens $\mathbf{x}^{(i)}$. This suggests that this concept class can represent only a limited set of concepts. A clean framework for illustrating these issues is afforded by the task of recognizing membership in a given formal language. Fix a finite alphabet $\Sigma$ with each letter also encoded by a vector so that it can be processed by RNNs. We say that the RNN recognizes a language $\Lambda$ over $\Sigma$ if after processing the sequences $(\bx^{(1)}, \ldots, \bx^{(j)})$ encoding a string $w = (w_1, \ldots, w_j) \in \Sigma^j$, the output $\mathbf{y}^{(j)}$ satisfies $\abs[0]{\mathbf{y}^{(j)}-1}<1/3$, if $w \in \Lambda$ and $\abs[0]{\mathbf{y}^{(j)}}<1/3$, otherwise. For simplicity, in the following, we will require the more stringent conditions $\mathbf{y}^{(j)}=1$ and  $\mathbf{y}^{(j)}=0$; these can be easily relaxed with some extra work. Since our output is binary, the output dimension $\dout $ is set to $1$; this is the setting in which our experiments are also done.

    Below, we give examples of some simple regular languages that the above concept class can't recognize but can be recognized by functions in our concept class with small complexity. 

We first consider a simple regular language $L_1$ over the alphabet $\{0,1\}$ given by the regular expression $0^\ast 10^\ast$. In words, a string is in $L_1$ iff it contains a single $1$. This language can be thought of as modeling the occurrence of an event (a single blip) in a time series.

Consider the set $S$ of strings $\{0^q 00 0^{L-q-2}, 0^q 11 0^{L-q-2}, 0^q 01 0^{L-q-2}, 0^q 10 0^{L-q-2}\}$ where $0 \leq q \leq L-2$. 
Clearly, $0^q 00 0^{L-q-2} \notin L_1$ and $0^q 11 0^{L-q-2} \notin L_1$ whereas $0^q 01 0^{L-q-2} \in L_1$ and $0^q 10 0^{L-q-2} \in L_1$.
We choose uniform distribution on $S$ as the data distribution $D_{L_1}$. 
\begin{theorem}\label{thm:allencantdl1}
Any concept class of type \eqref{eqn:AL_concept_class} must err with probability at least $1/4$ on $D_{L_1}$.
\end{theorem}
\begin{proof}\emph{(sketch)} 
Fix a $q$. Let $w=0^q a b 0^{L-q-2}$ where $a, b \in \{0, 1\}$, we can rewrite \eqref{eqn:AL_concept_class_simplified} as 
\begin{align*}
    F^{\ast (L)}(w) &= 
    \sum_{i : i \leq q } \alpha_i(0) +  \alpha_{q+1}(a) + \alpha_{q+2}(b) + \sum_{i : q+3 \leq i \leq L } \alpha_i(0) \\
    &= A + \alpha_{q+1}(a) + \alpha_{q+2}(b),
\end{align*}
where each $(\alpha_i(0), \alpha_i(1)) \in \Reals^2$ is any two-dimensional vector.
Now, we must have $A + \alpha_{q+1}(1) + \alpha_{q+2}(0) = 1$ and $A + \alpha_{q+1}(0) + \alpha_{q+2}(1) = 1$. And also,
$A + \alpha_{q+1}(0) + \alpha_{q+2}(0) = 0$ and $A + \alpha_{q+1}(1) + \alpha_{q+2}(1) = 0$. Summing the first two equations gives
$2A +  \alpha_{q+1}(0)  + \alpha_{q+1}(1) + \alpha_{q+2}(0) + \alpha_{q+2}(1) = 2$ and summing the next two equations gives
$2A +  \alpha_{q+1}(0)  + \alpha_{q+1}(1) + \alpha_{q+2}(0) + \alpha_{q+2}(1) = 0$. Thus at least one of the four equations above must fail.
The concept class thus incurs an error with probability at least 1/4.
\end{proof}

We can show that our concept class (Eq.~\eqref{eq:concept_class}) can recognize the language $D_{L_1}$. Assume that we get an length-$L$ string as a length-$L$ input sequence $\bx$, with `$0$' represented by one-dimensional vector $0$ and `$1$' represented by one-dimensional vector $1$. E.g. `$0010$' will be represented as a sequence $0,0,1,0$. Then, one can count the number of $1$'s in the given input and claim that if the number of $1$'s is exactly $1$, the string belongs to the language $D_{L_1}$. The required condition can be checked using a single neuron with activation $\phi(x) = 2x - x^2$, which is a quadratic activation, and weight vector containing all ones ($\mathbf{1}$).  Hence, one can show that acceptance condition is satisfied iff $\phi(\langle\mathbf{1}, \bx\rangle + 1/2)$ is positive. Thus overall, we have shown that the language $D_{L_1}$ can be computed by a one-hidden layer neural network with a quadratic activation and $1$ neuron, implying that our concept class can approximate the language $D_{L_1}$.

\paragraph{Other pattern matching languages.} $D_{L_1}$ can be thought of as a very simple pattern matching problem.  In fact, we can show a more general class of languages that can be learned by our concept class efficiently. Consider the following language: a string (of length at most $L$) belongs to the language iff it contains a particular substring (of some constant length $k$). We will denote this substring by $\bar{s}$. Assume that we get an $L$-length string as $L$-dimensional input $\bx$, with `$0$' represented by a one-dimensional vector $-1$ and `$1$' represented by one-dimensional vector $1$. E.g. `$0010$' will be represented by the sequence $-1,-1,1,-1$. Let $\mathbf{v}_{\bar{s}}$ denote the vector representation of the sequence for the substring $\bar{s}$. Then, we can enumerate all the consecutive substrings in the input and check if the required substring occurs in at least one of them. Mathematically, this translates to creating a one layer neural network with $(L-k+1)$ neurons and activation function $\phi(t) = e^{ct}$, for some constant $c = \Omega(\log L)$. The $i$-th neuron will contain the weight vector $\mathbf{v}_i$, where the substring between position $i$ and $i+k-1$ contains $\mathbf{v}_{\bar{s}}$ and the rest of the positions contain $0$. One can check that if the input string contains the desired substring $\bar{s}$, then
$\sum_{i=1}^{L} \phi(\langle \mathbf{v}_i, \bx\rangle - k) \ge 1$, otherwise it is less than $\frac{1}{L}$. Thus, overall we have shown that the language can be recognized by a one-layer network with exponential activations. Since, we have discussed before that exponential activations have $O(1)$ complexity (see Def.~\ref{def:complexity}), we have shown that our concept class can efficiently solve the pattern matching problem.

We can generalize the above ideas to address some other related problems where we need to find multiple substrings, problems where we need to make sure that the number of times a particular substring occurs is at most a certain limit, etc.

\paragraph{General regular languages.} More generally, our concept class can express all regular languages. However, the complexity of the concept class can be super-polynomial in the sequence length $L$ depending on the regular language. Here is a sketch of a general construction.
As previously mentioned, RNNs with ReLU activations and finite precision are known to be equivalent to deterministic finite automata (DFA) and thus capture regular languages \cite{korsky2019computational}. 
    The ReLU can be approximated by polynomials \cite{lorentz} so that the resulting RNN still approximates the DFA up to some required length (the larger the length, the better the approximation needs to be---and the higher the degree of the approximating polynomial). In turn, such an RNN using polynomial activations can be easily represented by our concept class. The complexity (Def.~\ref{def:complexity}) of the concept class is small as polynomials have small complexity. We omit the routine but technical details of this construction. 
    
    Many regular languages allow special treatment though. For example, consider the language $\mathsf{PARITY}$. $\mathsf{PARITY}$ is the language over alphabet $\{0, 1\}$ with a string $w = (w_1, \ldots, w_j) \in \mathsf{PARITY}$ iff
$w_1+\ldots +w_j = 1 \,\mathrm{mod}\, 2$, for $j \geq 1$. We can show that $\mathsf{PARITY}$ is hard for the above concept class for the uniform distribution on $\{0,1\}^L$. A simple proof of this can be obtained via Boolean Fourier analysis (e.g., \cite{odonnell}) which we now sketch. In this setting, 
we note that $\mathsf{PARITY}$ of $L$ bits corresponds to a degree-$L$ polynomial $(2 w_1 -1 )(2w_2 -1)\ldots (2w_L-1)$; the output now takes values in $\{-1, 1\}$ instead of $\{0, 1\}$. On the other hand, the functions in \eqref{eqn:AL_concept_class} with $\dout=1$ correspond to linear functions of the form $\sum_i \alpha_i w_i + \beta_i$ for some constants $\alpha_i, \beta_i \in \Reals$ for all $i$. Using these facts, the correlation between the two can be easily shown to be $0$ via the Plancherel--Parseval theorem, which implies that all functions of type \eqref{eqn:AL_concept_class} make significant error on $\mathsf{PARITY}$. 

However, we can show that $\mathsf{PARITY}$ is easily expressible by our concept class with small complexity. Assume that we get an length-$L$ string as a length-$L$ input sequence $\bx$, with `$0$' represented by one-dimensional vector $0$ and `$1$' represented by one-dimensional vector $1$. E.g. `$0010$' will be represented as a sequence $0,0,1,0$. Then, one can count the number of $1$'s in the given input and claim that if the number of $1$'s is even, the string belongs to the language $\mathsf{PARITY}$. The required condition can be checked using a single neuron with activation $\phi(x) = \cos(\pi x)$ and weight vector containing all ones ($\mathbf{1}$). Hence, one can show that acceptance condition is satisfied iff $\phi( \langle\mathbf{1}, \bx\rangle - 1)$ is positive. Thus overall, we have shown that the language $\mathsf{PARITY}$ can be computed by a one-hidden layer neural network with a $\cos$ activation and $1$ neuron. Since, we have discussed before that $\cos$ activations have $O(1)$ complexity (see Def.~\ref{def:complexity}), we have shown that our concept class can efficiently recognize $\mathsf{PARITY}$.


We performed experiments on the ability of RNNs to learn various regular languages (see sec.~\ref{sec:expts} for details). In almost all of the regular languages that we tested on, RNNs can achieve near perfect test accuracies (table~\ref{table:regular}).

\section{Experiments}\label{sec:expts}

\begin{figure}[!ht]
\centering
\begin{subfigure}
  \centering
  \includegraphics[width=0.5\linewidth]{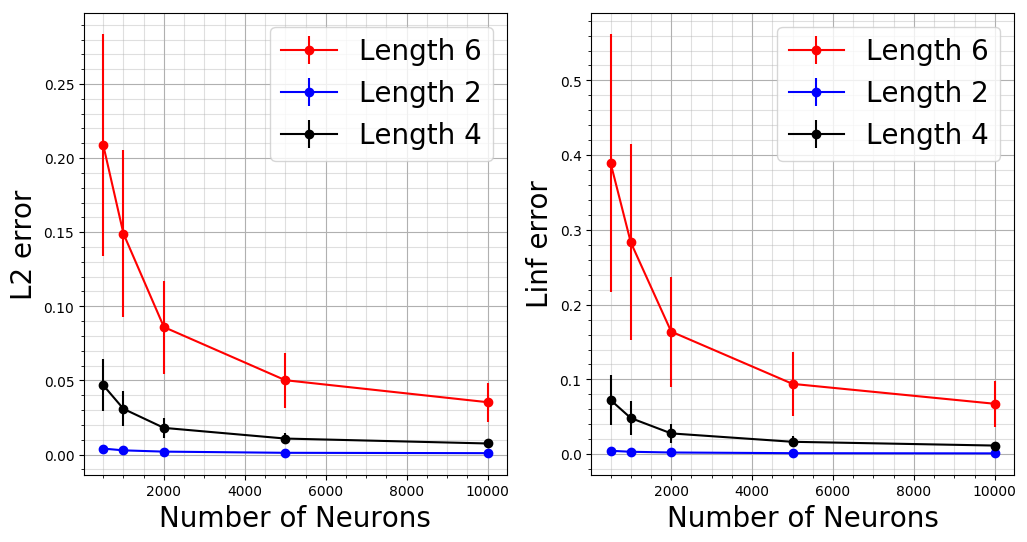}
  \caption{Data dimension: 2}
\end{subfigure}
\begin{subfigure}
  \centering
  \includegraphics[width=0.5\linewidth]{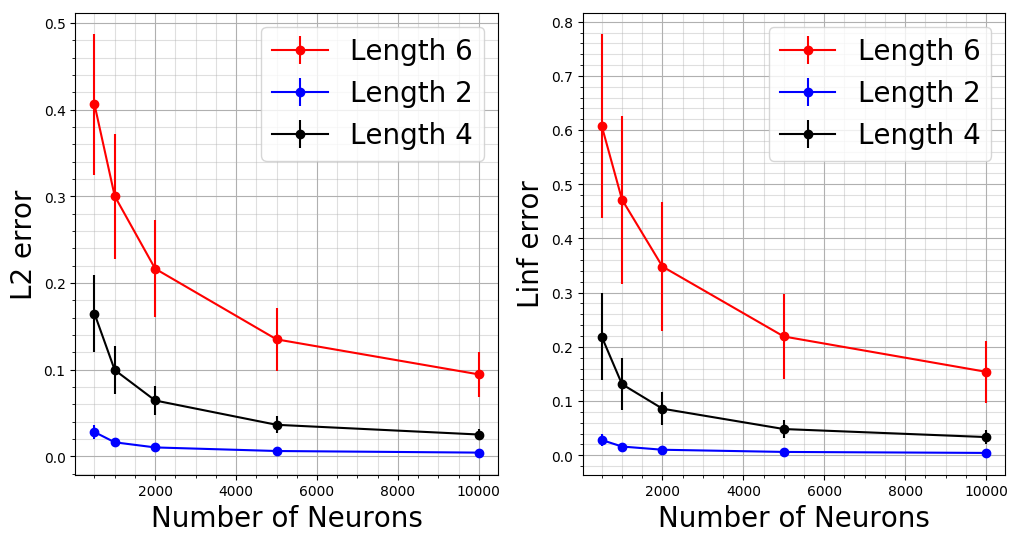}
  \caption{Data dimension: 4}
\end{subfigure}
\begin{subfigure}
  \centering
  \includegraphics[width=0.5\linewidth]{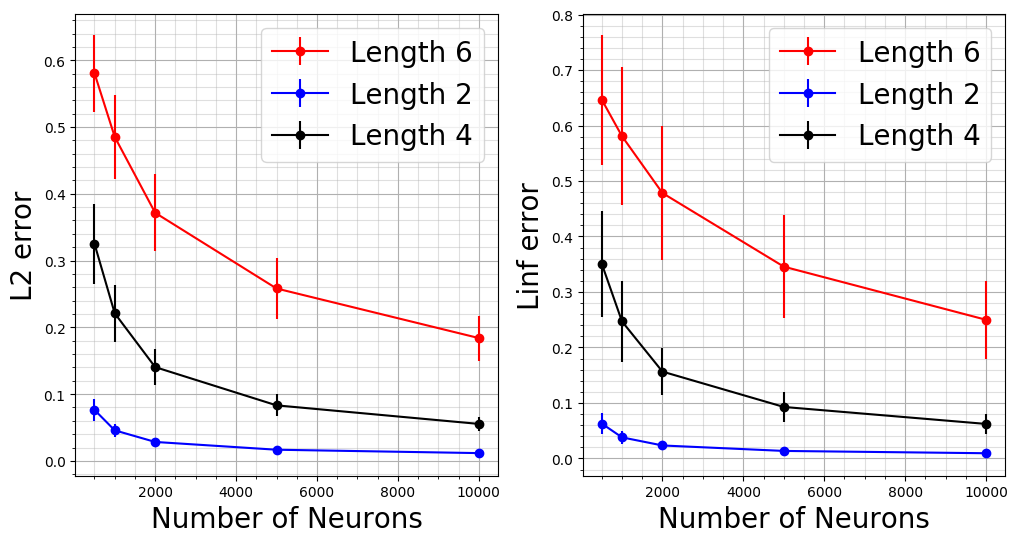}
  \caption{Data dimension: 8}
\end{subfigure}%
\caption{Invertibitiliy of RNNs at random initialization: Checking behavior of inversion error with number of neurons and the sequence length at different data dimensions.}
\label{fig:RNN_inver}
\end{figure}
\textbf{RNN inversion at random initialization.} We consider a randomly initialized RNN, with the entries of the weights $\mathbf{W}$ and $\mathbf{A}$ randomly picked from the distribution $\mathcal{N}(0, 1)$. Sequences are generated i.i.d. from normal distribution i.e. for each sequence, $\bx^{(i)} \sim N(0, \mathbf{I})$ for each $i \in [L]$. We use SGD with batch size 128, momentum $0.9$ and learning rate $0.1$ to compute the linear matrix $\obW^{[L]}$ so that $\norm[0]{\obW^{[L]} \mathbf{h}^{(L)} - [\bx^{(1)}, \ldots, \bx^{(L)}]}^2$ is minimized. We compute the following two quantities on the test dataset, containing $1000$ sequences: average $L_2$ error given by $\mathbb{E}_{\bx} \frac{\norm[0]{\obW^{[L]} \mathbf{h}^{(L)} - [\bx^{(1)}, \ldots, \bx^{(L)}]}}{\norm[0]{[\bx^{(1)}, \ldots, \bx^{(L)}]}}$ and average $L_\infty$ error given by $\mathbb{E}_{\bx} \norm[0]{\obW^{[L]} \mathbf{h}^{(L)} - [\bx^{(1)}, \ldots, \bx^{(L)}]}_{\infty}$. We plot both the quantities for different settings of data dimension $d$, sequence length $L$ and the number of neurons $m$. $L$ takes values from the set $\{2, 4, 6\}$, $d$ takes from $\{2, 4, 8\}$ and $m$ takes from $\{500, 1000, 2000, 5000, 10000\}$ (Figure~\ref{fig:RNN_inver}). The trends support our bounds in Theorem~\ref{thm:Invertibility_ESN_outline}, i.e. the error increases with increasing $L$ and decreases with increasing $m$. Note that the data distribution is different from the one assumed in normalized sequence Def.~\ref{def:normalized_seq}. It was easier to conduct experiments in the current data setting and a similar statement as Thm.~\ref{thm:Invertibility_ESN_outline} can be given.

\textbf{Performance of RNNs on different regular languages. } We check the performance of RNNs on the formal language recognition task for a wide variety of regular languages. We follow the set-up in \cite{BhattamishraAG20} who conducted experiments on LSTMs etc. but not on RNNs.

We consider the regular languages as considered in \cite{BhattamishraAG20}.
Tomita grammars \cite{tomita:cogsci82} contain 7
regular languages representable by DFAs of small
sizes, a popular benchmark for evaluating recurrent models (see references in \cite{BhattamishraAG20}). 
We reproduce the definitions of the Tomita grammars from there verbatim:
Tomita Grammars are 7 regular langauges defined on the alphabet $\Sigma = \{0, 1\}$.
Tomita-1 has the regular expression $1^\ast$.
Tomita-2 is defined by the regular expression $(10)^\ast$.
Tomita-3 accepts the strings where odd number
of consecutive 1s are always followed by an even
number of $0$'s. Tomita-4 accepts the strings that
do not contain three consecutive $0$'s. In Tomita-5 only
the strings containing an even number of $0$'s and
even number of $1$'s are allowed. In Tomita-6 the
difference in the number of $1$'s and $0$'s should be
divisible by 3 and finally, Tomita-7 has the regular
expression $0^\ast 1^\ast 0^\ast 1^\ast$. 

We also check the performance of RNNs on $\mathrm{Parity}$, which contains all languages with strings of the form $(w_1, \ldots, w_L)$ s.t. $w_1 + \ldots + w_L = 1 \mod 2$. Languages $\mathcal{D}_n$ are recursively defined as the set of all strings of the form $(0w1)^{\ast}$, where $w \in \mathcal{D}_{n-1}$, with $\mathcal{D}_0$ containing only $\epsilon$, the empty word. Other languages considered are $(00)^{\ast}$, $(0101)^{\ast}$ and $(00)^{\ast}(11)^{\ast}$. Table~\ref{table:regular} shows the number of examples in train and test data, the range of the length of the strings in the language, and the test accuracy of the RNNs with activation functions $\relu$ and $\tanh$ on the regular languages mentioned above. 

\begin{center}
\begin{table}[!ht]
\centering
\begin{tabular}{|| c | c | c | c | c ||} 
 \hline
 Task & No. of Training/Test examples  & Range of length of strings & RNN(Relu) & RNN(Tanh)  \\ [0.5ex] 
 \hline\hline
 Tomita 1 & 50/100 & [2, 50] & 1.0 & 1.0 \\
 Tomita 2 & 25/50 & [2, 50] & 1.0 & 1.0 \\
 Tomita 3 & 10000/2000 & [2, 50] & 1.0 & 1.0 \\
 Tomita 4 & 10000/2000 & [2, 50] & 1.0 & 1.0 \\
 Tomita 5 & 10000/2000 & [2, 50] & 1.0 & 1.0\\
 Tomita 6 & 10000/2000 & [2, 50] & 1.0 & 1.0\\
 Tomita 7 & 10000/2000 & [2, 50] & 0.259 & 0.99\\
 Parity & 10000/2000 & [2, 50] & 1.0 & 1.0\\
 $\mathcal{D}_2$ & 10000/2000 & [2, 100] & 1.0 & 1.0 \\
 $\mathcal{D}_3$ & 10000/2000 & [2, 100] & 0.99 & 1.0\\
 $\mathcal{D}_4$ & 10000/2000 & [2, 100] & 1.0 & 0.99\\
 $(00)^{\ast}$ & 250/50 & [2, 500] & 1.0 & 1.0\\
 $(0101)^{\ast}$ & 125/25 & [4, 500] & 0.99 & 1.0 \\
 $(00)^{\ast}(11)^{\star}$ & 10000/2000 & [2, 200] & 0.99 & 1.0 
 \\[1ex]
 \hline
\end{tabular}
\caption{Performance of RNNs on different regular languages.}
\label{table:regular}
\end{table}
\end{center}

We vary $m$, the dimension of the hidden state, in the range $[3, 32]$, used RMSProp optimizer~\cite{hinton2014coursera} with the smoothing constant $\alpha = 0.99$ and varied the learning rate in the range $[10^{-2}, 10^{-3}]$. For each language
we train models corresponding to each language
for $100$ epochs and a batch size of $32$. We experimented with two different activations $\relu$ and $\tanh$. 
In all but one case (Tomita 7 with ReLU) the test accuracies with near-perfect. This was the case across runs. Tomita 7 results could perhaps be improved by more extensive hyperparameter tuning. 
We train and test on strings of length up to 50, and in a few cases strings of larger lengths (when the number of strings in the language is small). 

\end{document}